\documentclass[10pt,twocolumn,letterpaper]{article}

\usepackage{iccv}
\usepackage{times}
\usepackage{epsfig}
\usepackage{graphicx}
\usepackage{amsmath}
\usepackage{amssymb}
\usepackage{authblk}

\usepackage{array}
\usepackage{capt-of}
\usepackage{comment}
\usepackage{booktabs}
\usepackage{multirow}
\usepackage{multicol}
\usepackage[dvipsnames]{xcolor}

\usepackage{appendix}

\usepackage[pagebackref=true,breaklinks=true,letterpaper=true,colorlinks,bookmarks=false]{hyperref}

\iccvfinalcopy %

\def\iccvPaperID{3929} %

\begin{document}

\title{ReStyle: A Residual-Based StyleGAN Encoder via Iterative Refinement}

\author{\vspace{-0.4cm}Yuval Alaluf $\ \hspace{0.05cm}$~~~~Or Patashnik~~~~~~~~Daniel Cohen-Or \\
Blavatnik School of Computer Science, Tel Aviv University \vspace{-0.3cm}
}

\newcommand{\w}{$\mathcal{W}$\xspace}
\newcommand*\concat{\mathbin{\|}}
\newcommand{\topic}[1]{\vspace{1mm}\noindent\textbf{#1}}
\newcommand{\norm}[1]{\left\lVert#1\right\rVert}

\maketitle
\ificcvfinal\thispagestyle{empty}\fi

\begin{abstract}
Recently, the power of unconditional image synthesis has significantly advanced through the use of Generative Adversarial Networks (GANs). The task of inverting an image into its corresponding latent code of the trained GAN is of utmost importance as it allows for the manipulation of real images, leveraging the rich semantics learned by the network. Recognizing the limitations of current inversion approaches, in this work we present a novel inversion scheme that extends current encoder-based inversion methods by introducing an iterative refinement mechanism. Instead of directly predicting the latent code of a given real image using a single pass, the encoder is tasked with predicting a residual with respect to the current estimate of the inverted latent code in a self-correcting manner. Our residual-based encoder, named ReStyle, attains improved accuracy compared to current state-of-the-art encoder-based methods with a negligible increase in inference time. We analyze the behavior of ReStyle to gain valuable insights into its iterative nature. We then evaluate the performance of our residual encoder and analyze its robustness compared to optimization-based inversion and state-of-the-art encoders. Code is available via our project page: \small{\url{https://yuval-alaluf.github.io/restyle-encoder/}}
\end{abstract}

\begin{figure}
    \centering
    \includegraphics[width=\linewidth]{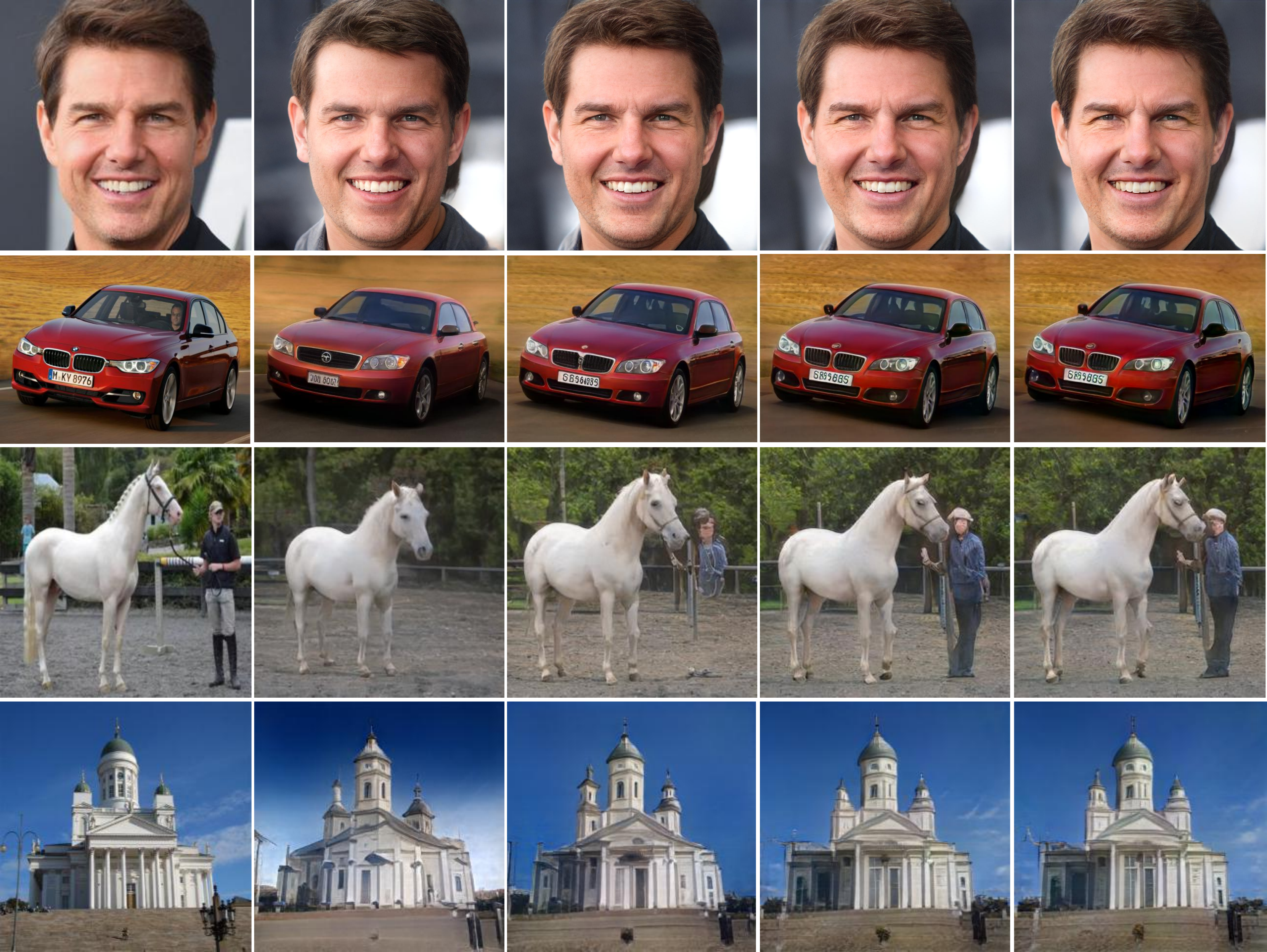} \\
    {\small
    {}\quad Input \qquad Iterative Outputs $\longrightarrow$ \quad \qquad \qquad \qquad \qquad \qquad 
    }
    \vspace{0.1cm}
    \caption{
    Different from conventional encoder-based inversion techniques, our residual-based ReStyle scheme incorporates an iterative refinement mechanism to progressively converge to an accurate inversion of real images.
    For each domain, we show the input image on the left followed by intermediate inversion outputs.}
    \label{fig:teaser}
\end{figure}

\section{Introduction}
Recently, Generative Adversarial Networks (GANs) have grown in popularity thanks to their ability to synthesize images of high visual quality and diversity. Beyond their phenomenal realism and fidelity on numerous domains, recent works have shown that GANs, e.g., StyleGAN~\cite{karras2019style, karras2020analyzing, karras2020training}, effectively encode semantic information in their latent spaces~\cite{harkonen2020ganspace,shen2020interpreting,jahanian2020steerability}. Notably, it has been shown that StyleGAN's learnt latent space \w has disentanglement properties~\cite{collins2020editing, shen2020interpreting, yang2020semantic} which allow one to perform extensive image manipulations by leveraging a well-trained StyleGAN generator. Such manipulations, however, have often been applied to synthetic images generated by the GAN itself. 
To apply such edits on \textit{real} images, one must first \textit{invert} the given image into StyleGAN's latent space. That is, retrieve the latent code $\textbf{w}$ such that passing $\textbf{w}$ to the pre-trained StyleGAN generator returns the original image. To do so, it has become common practice to invert real images into an extension of \w, denoted $\mathcal{W}+$~\cite{abdal2019image2stylegan}.

Previous works have explored learning-based inversion approaches and train encoders to map a given real image into its corresponding latent code~\cite{creswell2018inverting, pidhorskyi2020adversarial, zhu2020domain, guan2020collaborative, richardson2020encoding, tov2021designing}. Compared to per-image latent vector optimization~\cite{lipton2017precise,creswell2018inverting,abdal2019image2stylegan,abdal2020image2stylegan++,karras2020analyzing}, encoders are significantly faster, as they invert using a single forward pass, and converge to areas of the latent space which are more suitable for editing~\cite{zhu2020domain,tov2021designing}. However, in terms of reconstruction accuracy, there remains a significant gap between learning-based and optimization-based inversion methods. Hence, while significant progress has been made in learning-based inversions, designing a proper encoder and training scheme remains a challenge with many works still resorting to using a per-image optimization.

Recognizing that obtaining an accurate inversion in a single shot is difficult, we introduce a novel encoder-based inversion scheme tasked with encoding real images into the extended $\mathcal{W}+$ StyleGAN latent space. 
Unlike typical encoder-based inversion methods that infer the input's inverted latent code using a single forward pass, our scheme introduces an iterative feedback mechanism. Specifically, the inversion is performed using several forward passes by feeding the encoder with the output of the previous iteration along with the original input image. This allows the encoder to leverage knowledge learned in previous iterations to focus on the relevant regions needed for achieving an accurate reconstruction of the input image. Viewing this formulation in terms of the latent space, our \textit{residual encoder} is trained to predict the residual, or an offset, between the current latent code and the new latent code at each step. Doing so allows the encoder to progressively converge its inversion toward the target code and reconstruction, see Figure~\ref{fig:teaser}.  Note also that the inversion is predicted solely using the encoder with \textit{no} per-image optimization performed thereafter. 

In a sense, our inversion scheme, named \emph{ReStyle}, can be viewed as \textit{learning} to perform a \textit{small} number of steps (e.g., $10$) in a residual-based manner within the latent space of a pre-trained unconditional generator. ReStyle is generic in the sense that it can be applied to various encoder architectures and loss objectives for the StyleGAN inversion task.

We perform extensive experiments to show that ReStyle achieves a significant improvement in reconstruction quality compared to standard feed-forward encoders. This is achieved with a negligible increase in inference time, which is still an order of magnitude faster than the time-costly optimization-based inversion. 
We also analyze the iterative nature of our approach. Specifically, we first demonstrate which image regions are refined at each iterative feedback step demonstrating that our scheme operates in a coarse-to-fine manner.
Second, we show that the absolute magnitude of change at each step decreases, with the predicted residuals converging after only a small number of steps. 

To demonstrate the generalization of ReStyle beyond the StyleGAN inversion task and its appealing properties compared to current inversion techniques, we continue our analysis by exploring the robustness of our scheme on downstream tasks and special use-cases. To this end, we perform latent space manipulations~\cite{harkonen2020ganspace, shen2020interpreting, shen2020closedform} on the inverted latent codes to see if the embeddings are semantically meaningful. We then explore an \textit{encoder bootstrapping} technique allowing one to leverage two well-trained encoders to obtain a more faithful translation of a given real image. 

\section{Background and Related Works}

The idea of employing an iterative refinement scheme is not new. Carreira \etal~\cite{carreira2016human} introduced an iterative feedback mechanism for human pose estimation. Other works have proposed using iterative refinement for optical flow~\cite{hur2019iterative}, object pose estimation~\cite{wang2019densefusion, huang2020iterative}, object detection~\cite{rajaram2016refinenet}, and semantic segmentation~\cite{zhang2019canet} among other tasks. 
To the best of our knowledge, we are the first to adopt an iterative refinement approach for a learned inversion of real images.

\begin{figure*}[tb]
    \centering
    \setlength{\belowcaptionskip}{-15pt}
    \includegraphics[width=0.95\linewidth]{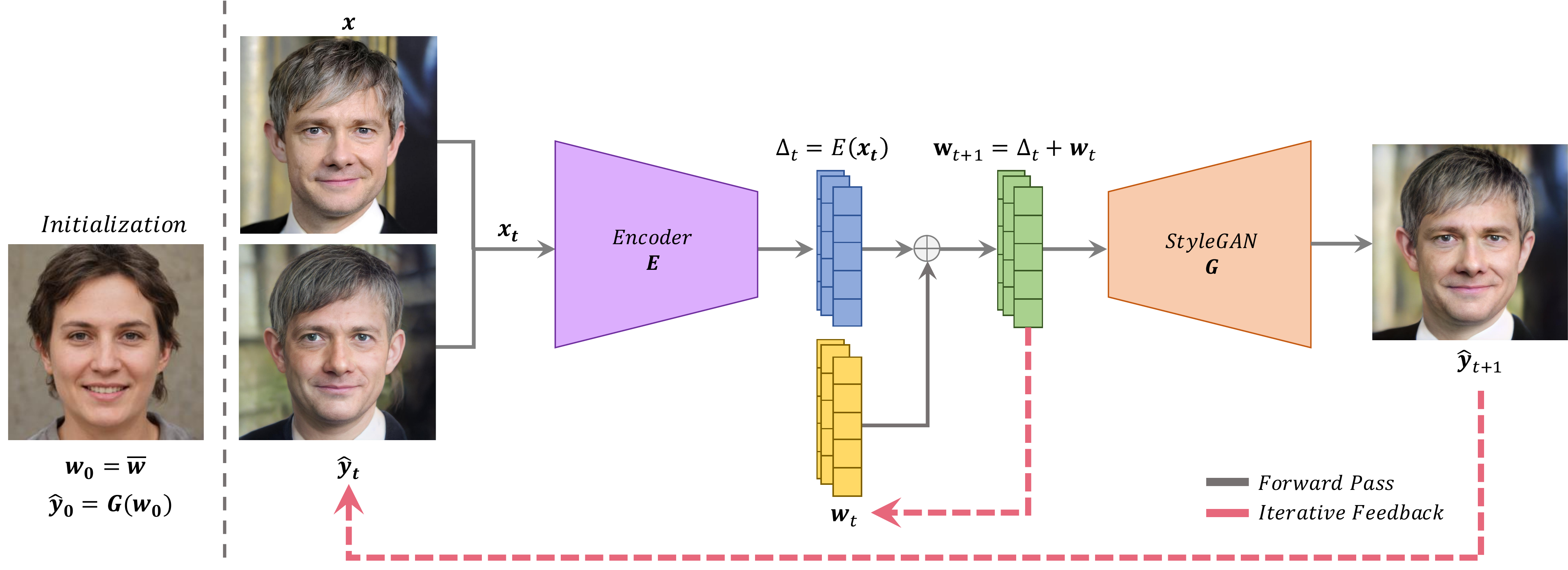}
    \vspace{0.05cm}
    \caption{
    \textit{\textbf{Our ReStyle iterative inversion scheme}}. Given an input image $\textbf{x}$, the scheme is initialized with the average latent code $\textbf{w}_0$ and its corresponding image $\hat{\textbf{y}}_0$.
    Consider step $t$. 
    ReStyle operates on an extended input obtained by concatenating $\textbf{x}$ with the image $\hat{\textbf{y}}_{t}$ corresponding to the current inversion prediction $\textbf{w}_{t}\in\mathcal{W}+$ (shown in yellow).
    The encoder $E$ is then tasked with predicting a \textit{residual} latent code, $\Delta_t\in\mathcal{W}+$ (shown in blue). 
    The predicted residual is then added to the previous latent code $\textbf{w}_{t}$ to obtain the updated latent code prediction $\textbf{w}_{t+1}$ (shown in green).
    Finally, passing the newly computed latent code to the generator $G$ results in an updated reconstruction $\hat{\textbf{y}}_{t+1}$, which is then passed as input in the following step.
    During training, the loss objectives are computed at each forward pass with back-propagation performed accordingly. A similar multi-step process is performed during inference.
    }
    \label{fig:architecture}
\end{figure*}

\subsection{GAN Inversion}
The task of \textit{GAN Inversion} was first introduced by Zhu \etal~\cite{zhu2016generative} for projecting real images into their latent representations. 
In their pioneering work, the authors demonstrate how performing such an inversion enables one to leverage the semantics of the GAN's latent space for performing various image manipulation tasks.
Some works~\cite{zhu2016generative,lipton2017precise,creswell2018inverting,abdal2019image2stylegan,abdal2020image2stylegan++,karras2020analyzing,tewari2020pie} approach this task by directly optimizing the latent vector to minimize the reconstruction error for a given image. These works typically achieve high reconstruction quality but require several minutes per image. Other approaches design an encoder~\cite{zhu2016generative,pidhorskyi2020adversarial,zhu2020domain,guan2020collaborative,richardson2020encoding,tov2021designing}
to learn a direct mapping from a given image to its corresponding latent vector. While these methods are substantially more efficient than pure optimization, they typically achieve inferior reconstruction quality. Attempting to balance this trade-off, some works have additionally proposed a hybrid approach and combine the two by using an encoder for initializing the optimization~\cite{zhu2016generative,pbayliesstyleganencoder,guan2020collaborative,zhu2020domain}. We refer the reader to Xia \etal ~\cite{xia2021gan} for a comprehensive survey on GAN inversion.

\subsection{Latent Space Embedding via Learned Encoders}
To perform image manipulations on real images, methods typically follow an ``invert first, edit later" approach. There, an image is first embedded into its corresponding latent code, which is then edited in a semantically meaningful manner. Diverging from the above, recent works~\cite{nitzan2020disentangling, richardson2020encoding,alaluf2021matter,chai2021latent} have proposed end-to-end methods for leveraging the high-quality images generated by GANs for various image-to-image translation and image editing tasks. In these works, a real input image is directly encoded into the transformed latent code which is then fed into the generator to obtain the desired transformed image. By training an encoder with some additional constraint, these works are able to directly solve various tasks without the need for inverting the images beforehand. Other works~\cite{xu2021generative} have explored utilizing features produced by a learned StyleGAN encoder for solving various down-stream tasks such as face verification and layout prediction. These works further emphasize the advantage of training a powerful encoder into the latent space of a pre-trained unconditional generator.

\subsection{Latent Space Manipulation}
With the recent advancements in image synthesis through GANs~\cite{Goodfellow2014GenerativeAN}, many works have proposed diverse methods for understanding and controlling their latent representations for performing extensive image manipulations. 

Various works~\cite{denton2019detecting, goetschalckx2019ganalyze, shen2020interpreting} use fully-supervised approaches to find latent directions corresponding to various attributes such as age, gender, and expression. On the other end of the supervision spectrum, several methods ~\cite{harkonen2020ganspace, voynov2020unsupervised, wang2021a} find directions in a completely unsupervised manner.
Others have explored techniques that go beyond a linear traversal of the latent space. Tewari \etal \cite{tewari2020stylerig} employ a pre-trained 3DMM to learn semantic face edits. Shen \etal \cite{shen2020closedform} learn versatile edit directions through the eigenvector decomposition of the generator weights. Abdal \etal \cite{abdal2020styleflow} learn non-linear paths via normalizing flows conditioned on a target attribute. Finally, Patashnik \etal \cite{patashnik2021styleclip} utilize CLIP to manipulate images using an input text-prompt.
By designing an efficient and accurate inversion method, one is able to leverage these works for manipulating real images.
\section{Preliminaries}
\subsection{Encoder-Based Inversion Methods}
Recall that our goal is to train an encoder tasked with inverting real images into the latent space of a pre-trained StyleGAN generator. Let $E$ and $G$ denote our encoder and StyleGAN generator, respectively. Given a source image $\textbf{x}$, our goal is to generate an image $\hat{\textbf{y}} = G(E(\textbf{x}))$ such that $\hat{\textbf{y}} \approx \textbf{x}$. 
Observe that in conventional encoder-based inversion methods, the reconstructed image $\hat{\textbf{y}}$ is simply computed using a single forward pass through $E$ and $G$ via StyleGAN's latent space representation. 

For learning to perform the inversion, these methods introduce a set of losses used to train the encoder network $E$ on the reconstruction task.
For training the encoder, most encoder-based methods employ a weighted combination of a pixel-wise L2 loss and a perceptual loss (e.g., LPIPS~\cite{zhang2018unreasonable}) to guide the training process. 
Recently, Richardson \etal~\cite{richardson2020encoding} extend these losses and introduce a dedicated identity loss to achieve improved reconstruction on the human facial domain. To attain improved editability over the inverted latent codes, Tov \etal~\cite{tov2021designing} additionally introduce two regularization losses during training.
Observe that during training, the pre-trained generator network $G$ typically remains fixed. 

\section{Method}~\label{method}
We now turn to describe our ReStyle scheme and build on the conventional, single-pass encoding approach introduced above.
Given an input image $\textbf{x}$, ReStyle performs $N>1$ steps to predict the image inversion $\textbf{w} = E(\textbf{x})$ and corresponding reconstruction $\hat{\textbf{y}}$. 
Here, we define a \textit{step} to be a single forward pass through $E$ and $G$. As such, observe that the conventional encoding process, being performed with a single step, is a special case of ReStyle where $N=1$.

For training the encoder network $E$, we define a single training \textit{iteration} to be a set of $N$ steps performed on a batch of images.
As with conventional encoding schemes, ReStyle uses a curated set of loss objectives for training $E$ on the inversion task while the pre-trained generator $G$ remains fixed.
Observe that the loss objectives are computed at each forward pass (i.e., step) with the encoder weights updated accordingly via back-propagation (i.e., back-propagation occurs $N$ times per batch). 

During inference, the same multi-step process (without the loss computation) is performed to compute the image inversion and reconstruction. 
Notably, for a given batch of images, we find that a small number of steps are needed for convergence (e.g., $N<10$), resulting in fast inference time.

We now turn to more formally describe ReStyle's inversion process, illustrated in Figure~\ref{fig:architecture}. 
At each step $t$, ReStyle operates on an expanded input by concatenating $\textbf{x}$ with the current prediction for the reconstructed image $\hat{\textbf{y}}_t$:
\begin{equation}~\label{eq:input}
    \textbf{x}_t := \textbf{x} \concat \hat{\textbf{y}}_t.
\end{equation}
Given the extended $6$-channel input $\textbf{x}_t$, the encoder $E$ is tasked with computing a residual code $\Delta_t$, with respect to the latent code predicted in the previous step. 
That is,
\begin{equation}~\label{eq:delta}
    \Delta_t := E(\textbf{x}_t).
\end{equation}
The new prediction for the latent code corresponding to the inversion of the input image $\textbf{x}$ is then updated as:
\begin{equation}~\label{eq:latent}
    \textbf{w}_{t+1} \gets \Delta_t + \textbf{w}_{t}.
\end{equation}
This new latent $\textbf{w}_{t+1}$ is passed through the generator $G$ to obtain the updated prediction for the reconstructed image: 
\begin{equation}~\label{eq:output}
    \hat{\textbf{y}}_{t+1} := G(\textbf{w}_{t+1}).
\end{equation}
Finally, the updated prediction $\hat{\textbf{y}}_{t+1}$ is set as the additional input channels in the next step, as defined by Equation~\ref{eq:input}.

This procedure is initialized with an initial guess $\textbf{w}_{0}$ and corresponding image $\hat{\textbf{y}}_{0}$. In our experiments, these are set to be the generator's average style vector and its corresponding synthesized image, respectively. 

Observe that constraining the encoder to invert the given image in a single step, as is typically done, imposes a hard constraint on the training process. Conversely, our training scheme can, in a sense, be viewed as relaxing this constraint. 
In the above formulation, the encoder learns how to best take \textit{several} steps in the latent space with respect to an initial guess $\textbf{w}_0$ guided by the output obtained in the previous step. This relaxed constraint allows the encoder to iteratively narrow down its inversion to the desired target latent code in a self-correcting manner. One may also view the ReStyle steps in a similar manner to the steps of optimization, with the key difference that here the steps are \textit{learned} by the encoder for efficiently performing the inversion.

\subsection{Encoder Architecture}~\label{sec:encoder_arch}
To show that the presented training scheme can be applied to different encoder architectures and loss objectives, we apply the ReStyle scheme on the state-of-the-art encoders from Richardson \etal~\cite{richardson2020encoding} (pSp) and Tov \etal~\cite{tov2021designing} (e4e). These two encoders employ a Feature Pyramid Network~\cite{lin2017feature} over a ResNet~\cite{he2015deep} backbone and extract the style features from three intermediate levels. Such a hierarchical encoder is well-motivated for well-structured domains such as the facial domain in which the style inputs can be roughly divided into three levels of detail. With that, we find such a design to have a negligible impact on less-structured, multi-modal domains while introducing an increased overhead. Moreover, we find that the multi-step nature of ReStyle alleviates the need for such a complex encoder architecture.

We therefore choose to design simpler variants of the pSp and e4e encoders. Rather than extracting the style features from three intermediate levels along the encoder, all style vectors are extracted from the final $16\times16$ feature map. Given a StyleGAN generator with $k$ style inputs, $k$ different \textit{map2style} blocks introduced in pSp are then used to down-sample the feature map to obtain the corresponding $512$-dimensional style input. An overview of the architecture is presented in Figure~\ref{fig:simplified_encoder_architecture} with additional details and ablations provided in Appendices~\ref{sec:fpn_arch} and ~\ref{ablation_study}, respectively.

\begin{figure}
    \centering
    \setlength{\belowcaptionskip}{-7pt}
    \includegraphics[width=0.915\linewidth]{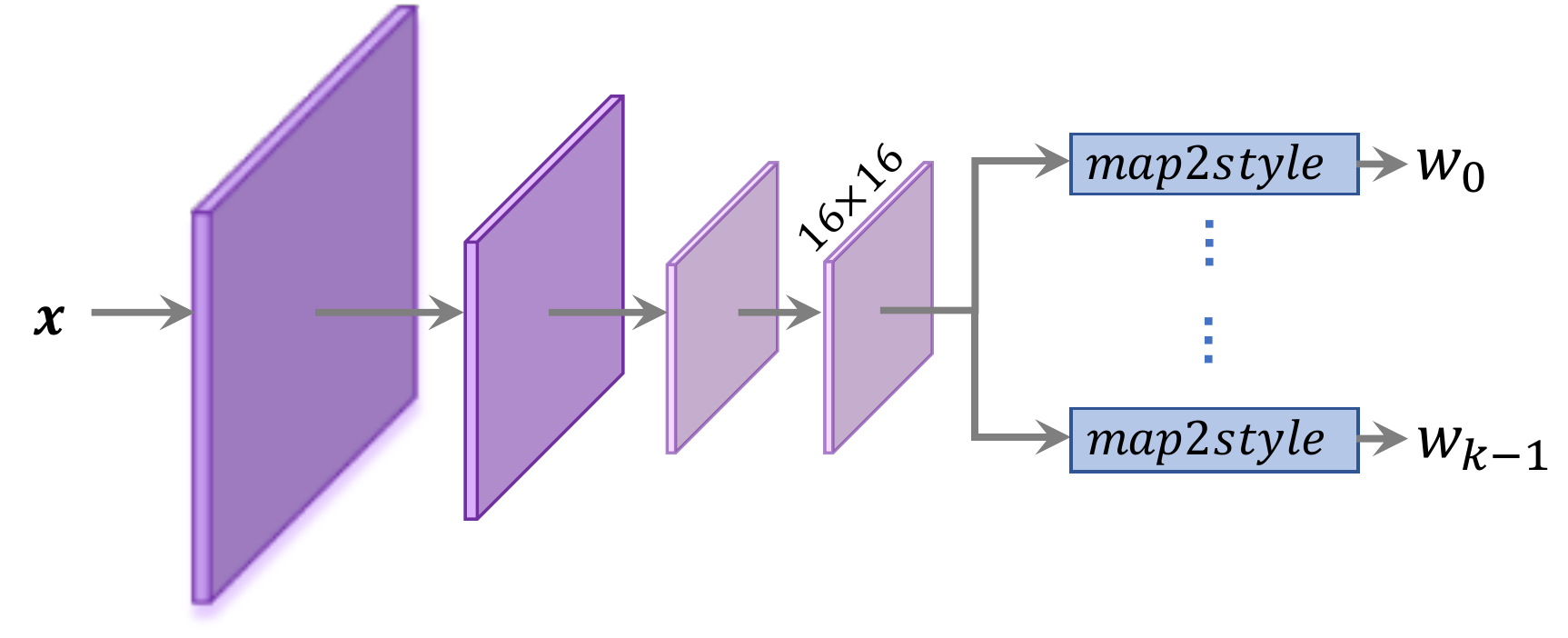}
    \caption{\textit{Our simplified encoder architecture.} All $k$ input style vectors of the generator are extracted from the encoder's final $16\times16$ feature map which is passed through $k$ \textit{map2style} blocks~\cite{richardson2020encoding}.}
    \label{fig:simplified_encoder_architecture}
\end{figure}
\begin{figure*}
    \centering
    \setlength{\belowcaptionskip}{-6pt}
    \setlength{\tabcolsep}{1pt}
    {\small
    \begin{tabular}{c c c c c c c c c c}

        \includegraphics[width=0.095\textwidth]{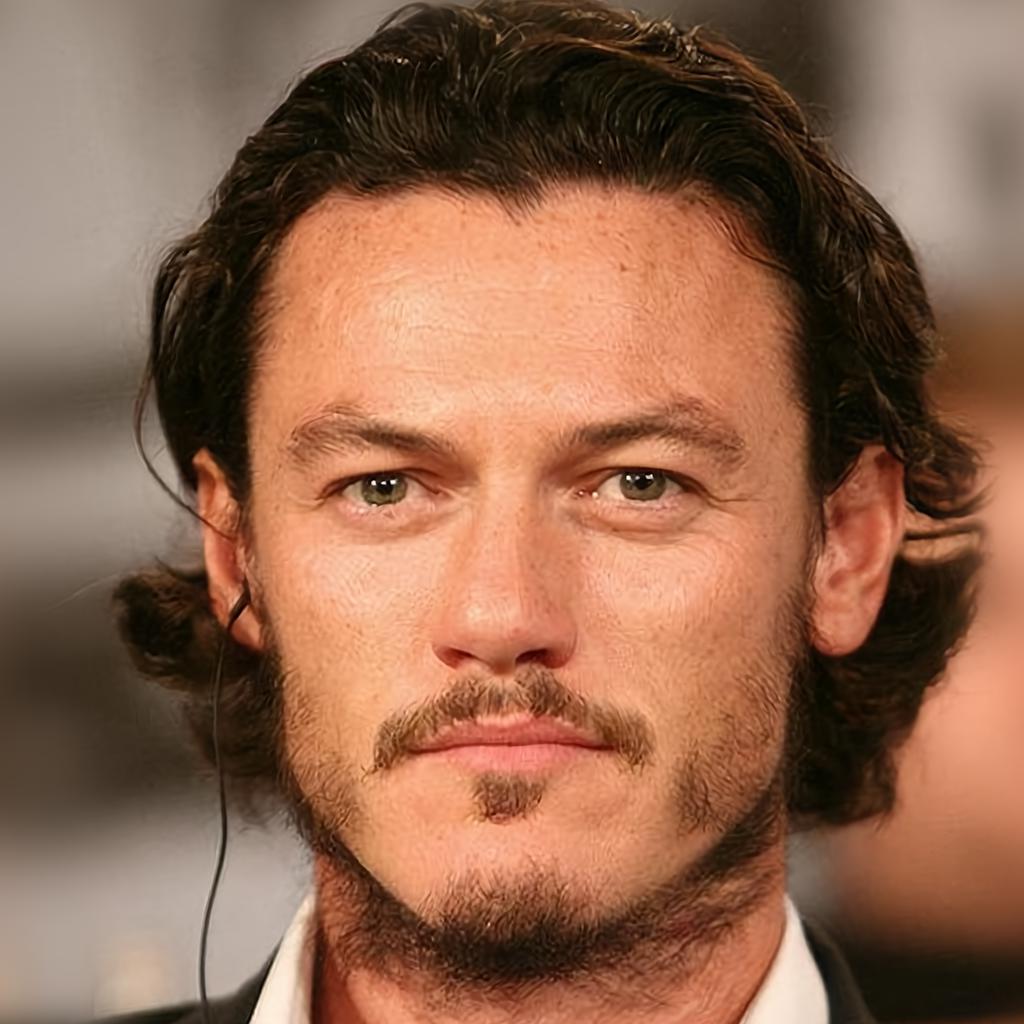} &
        \includegraphics[width=0.095\textwidth]{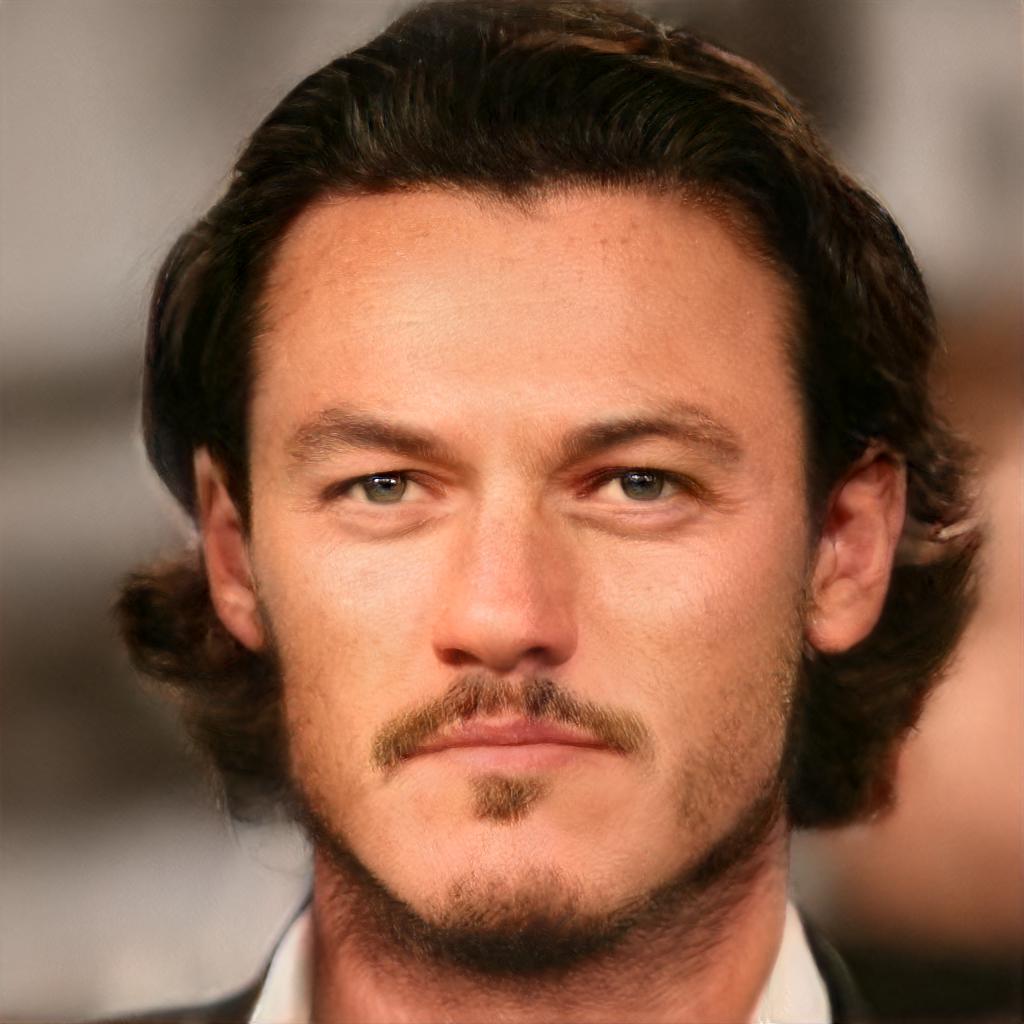} &
        \includegraphics[width=0.095\textwidth]{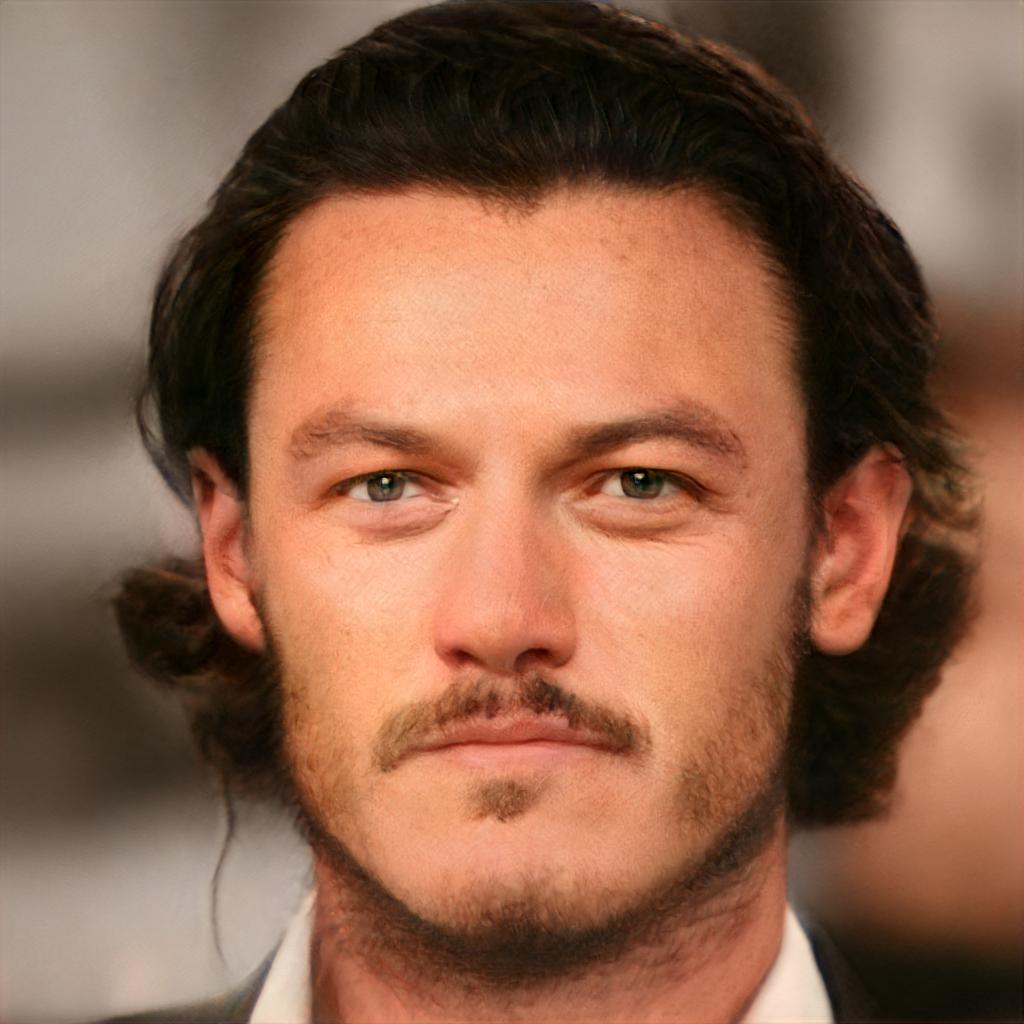} &
        \includegraphics[width=0.095\textwidth]{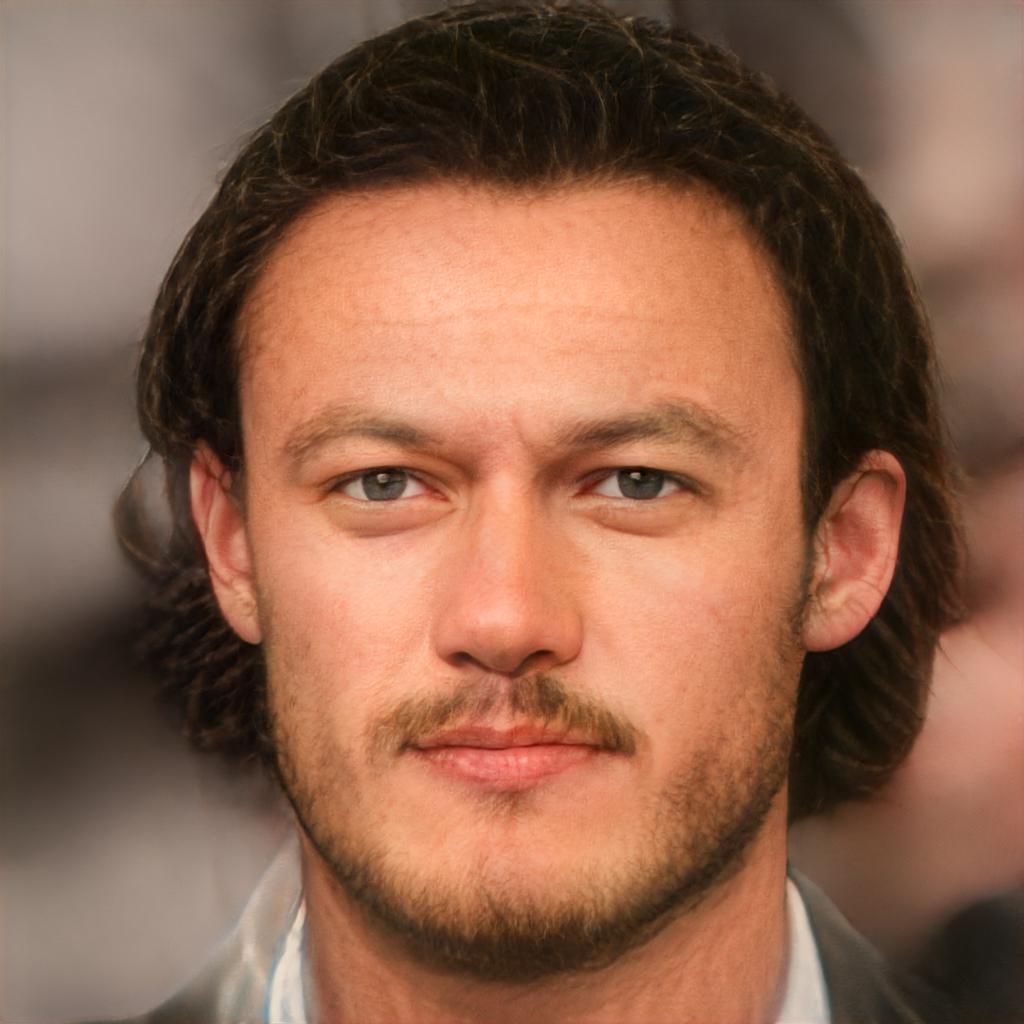} &
        \includegraphics[width=0.095\textwidth]{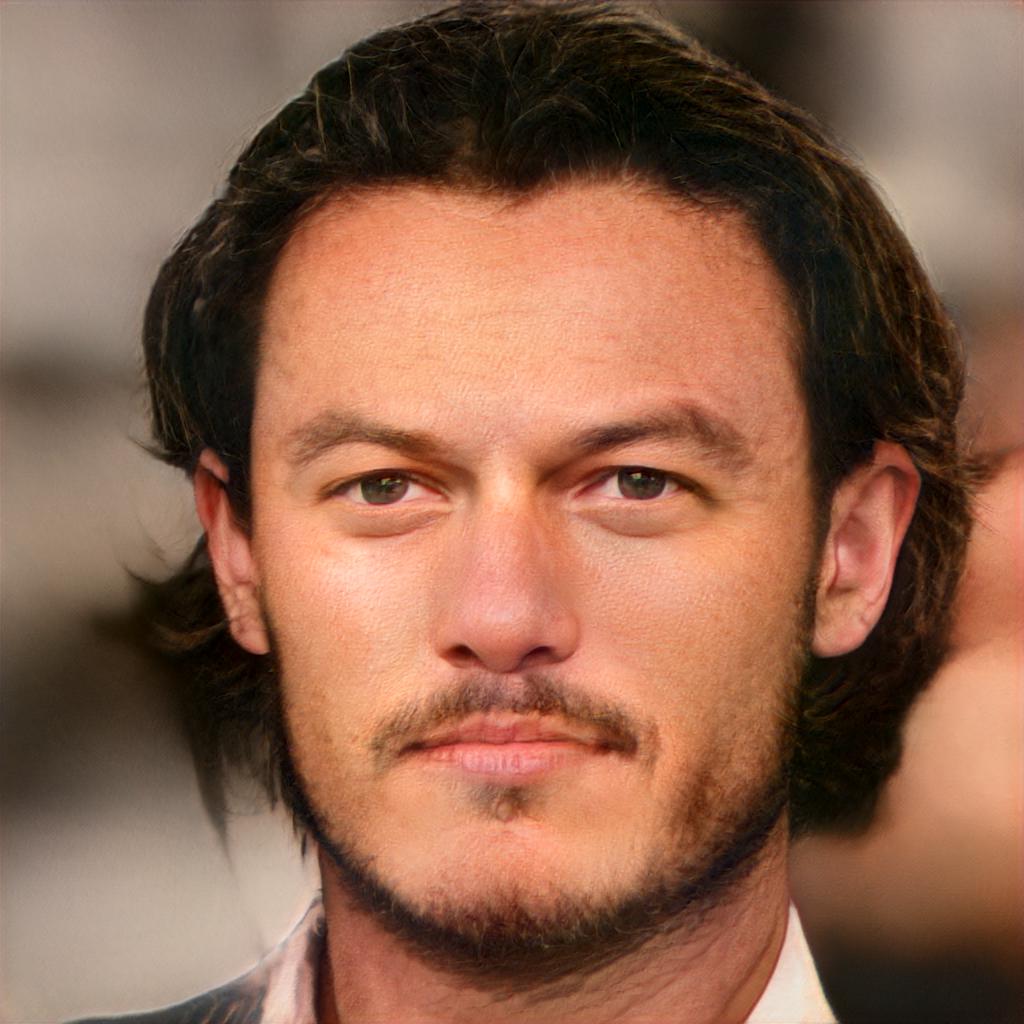} &
        \hspace{0.05cm}
        \includegraphics[width=0.095\textwidth]{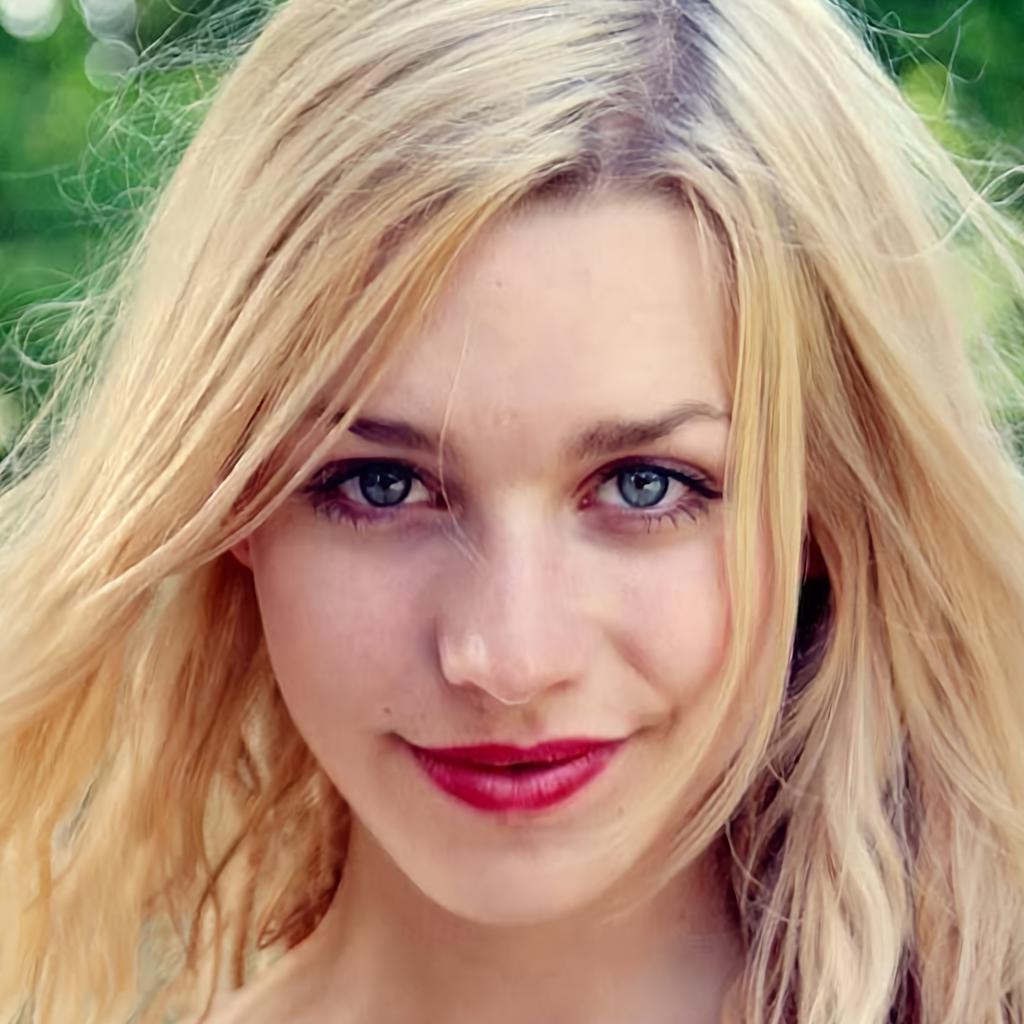} &
        \includegraphics[width=0.095\textwidth]{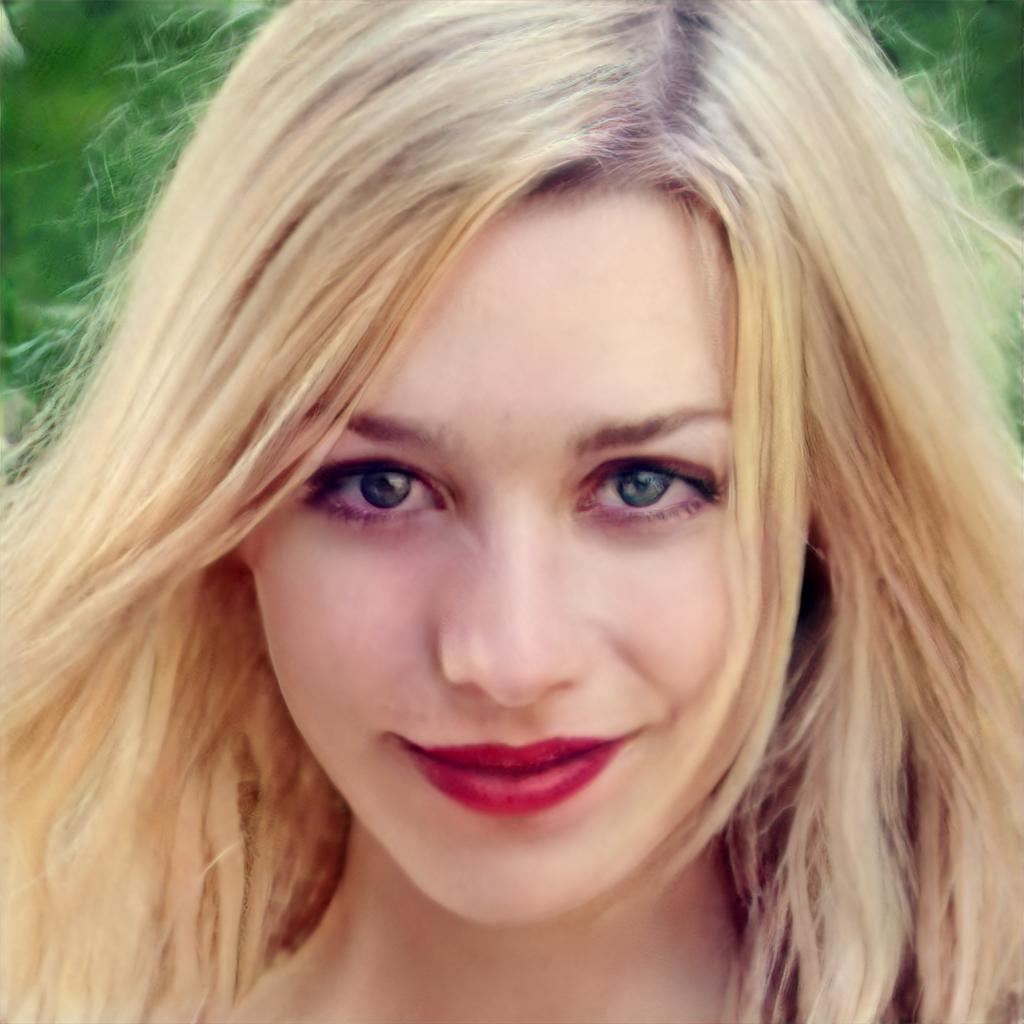} &
        \includegraphics[width=0.095\textwidth]{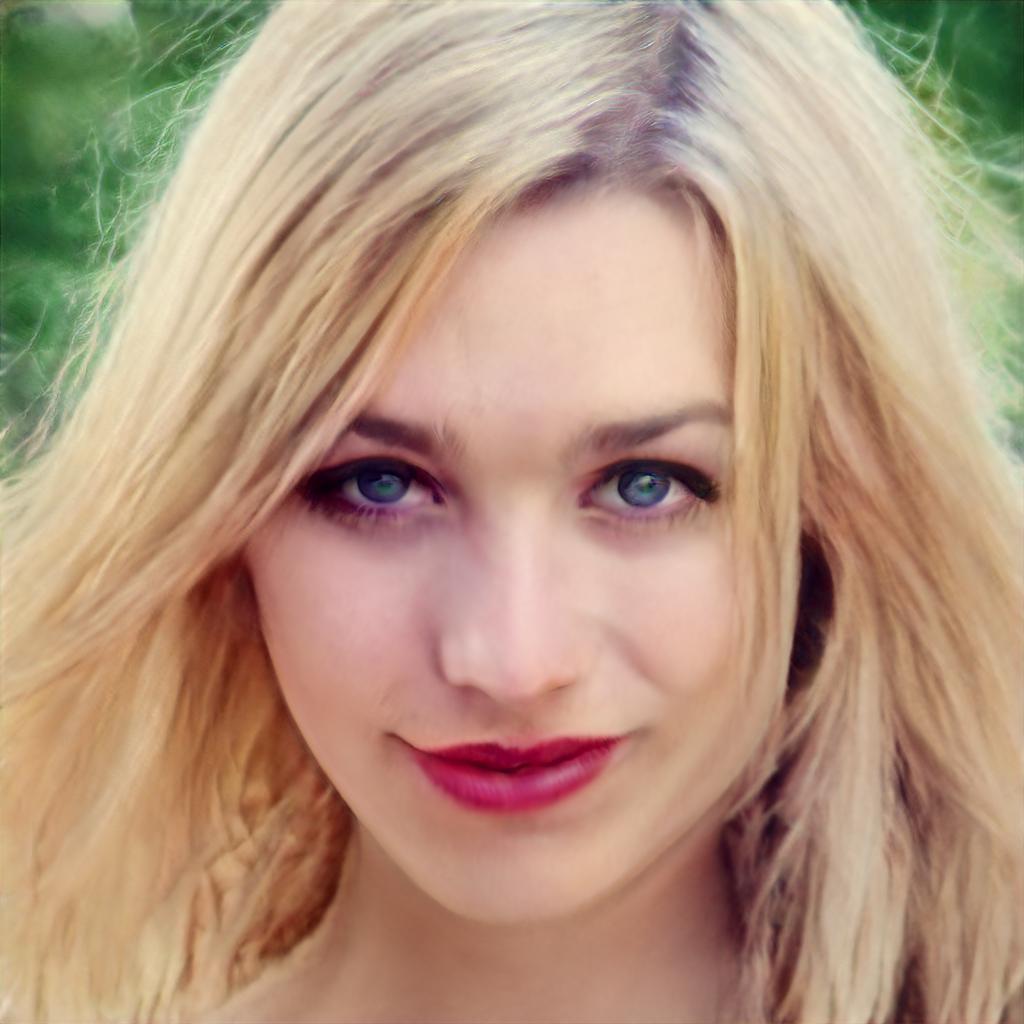} &
        \includegraphics[width=0.095\textwidth]{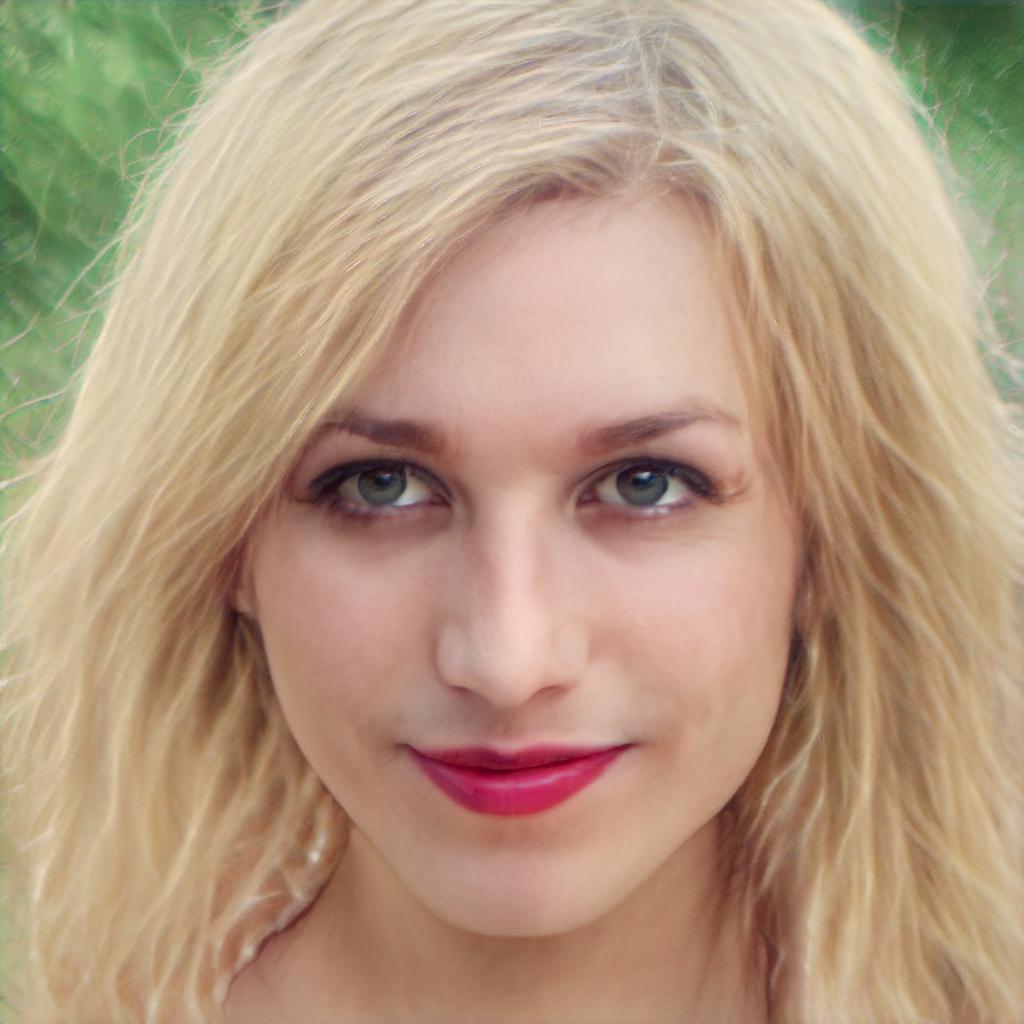} &
        \includegraphics[width=0.095\textwidth]{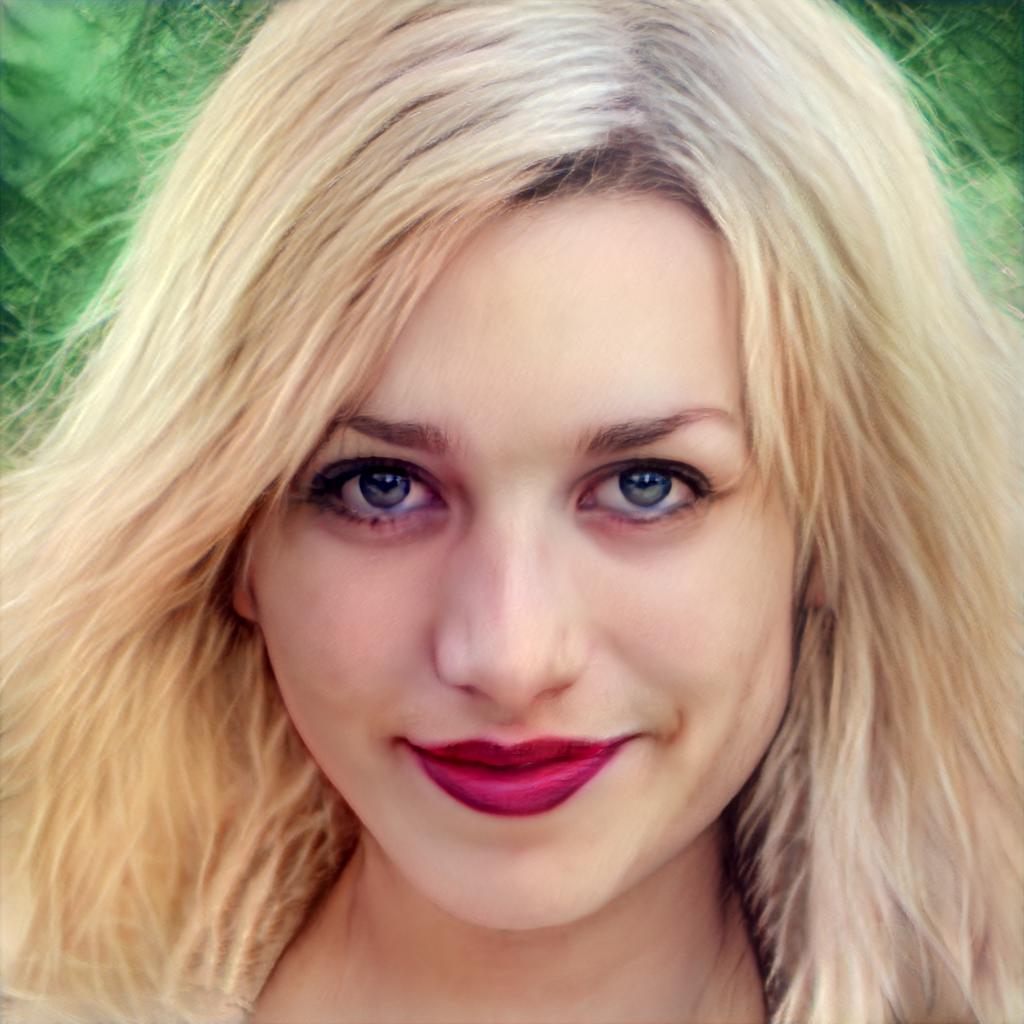} \\

        \includegraphics[width=0.095\textwidth]{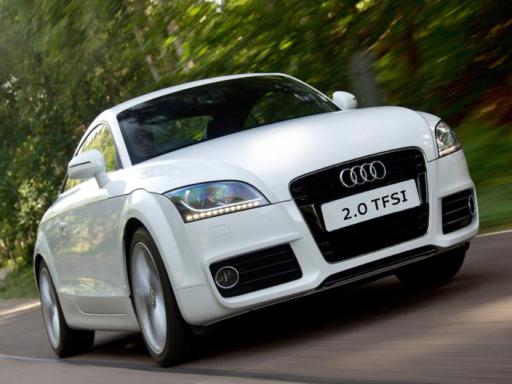} &
        \includegraphics[width=0.095\textwidth]{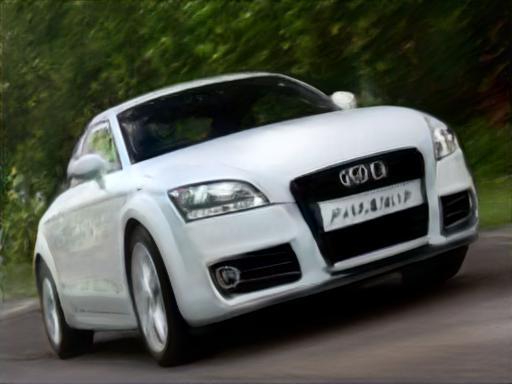} &
        \includegraphics[width=0.095\textwidth]{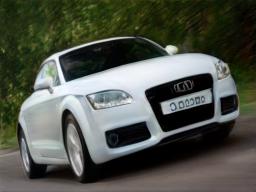} &
        \includegraphics[width=0.095\textwidth]{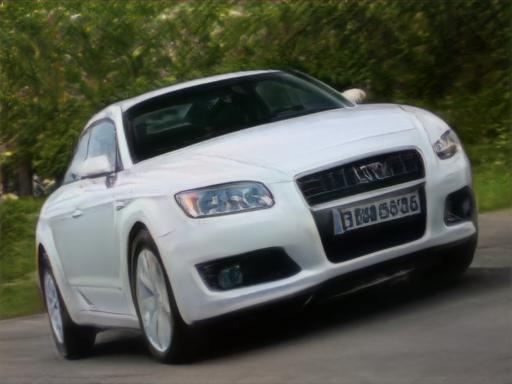} &
        \includegraphics[width=0.095\textwidth]{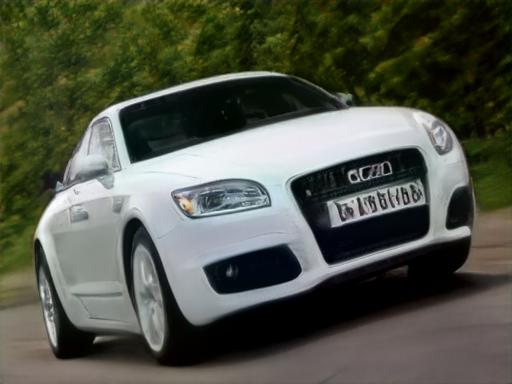} &
        \hspace{0.05cm}
        \includegraphics[width=0.095\textwidth]{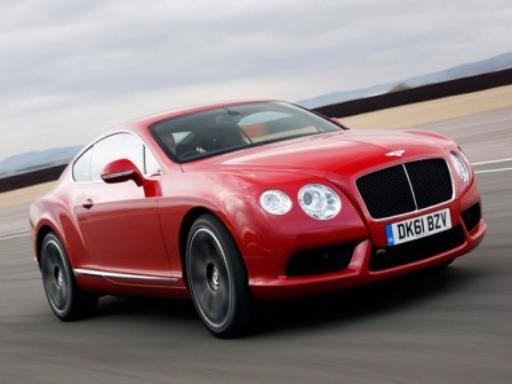} &
        \includegraphics[width=0.095\textwidth]{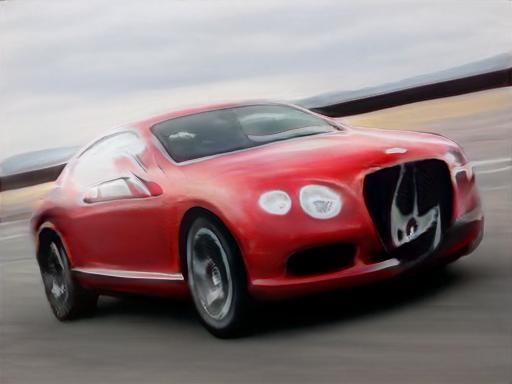} &
        \includegraphics[width=0.095\textwidth]{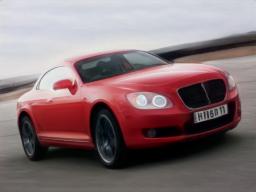} &
        \includegraphics[width=0.095\textwidth]{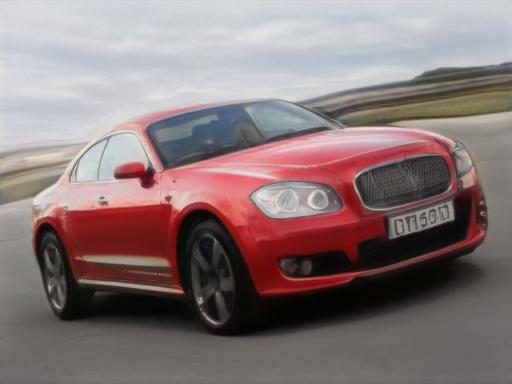} &
        \includegraphics[width=0.095\textwidth]{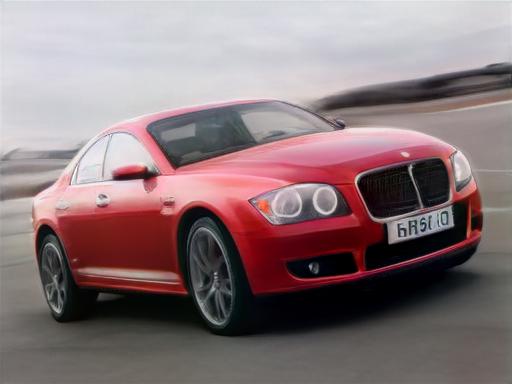} \\
        
        \includegraphics[width=0.095\textwidth]{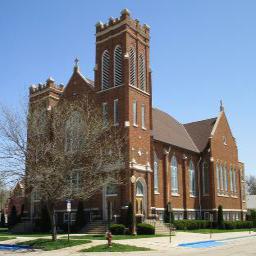} &
        \includegraphics[width=0.095\textwidth]{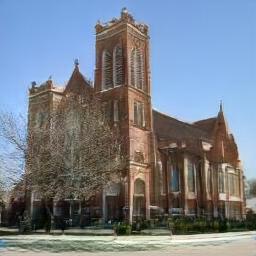} &
        \includegraphics[width=0.095\textwidth]{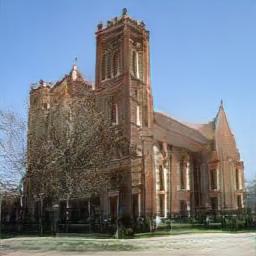} &
        \includegraphics[width=0.095\textwidth]{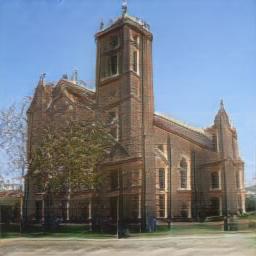} &
        \includegraphics[width=0.095\textwidth]{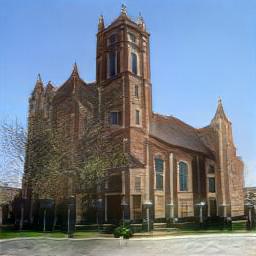} &
        \hspace{0.05cm}
        \includegraphics[width=0.095\textwidth]{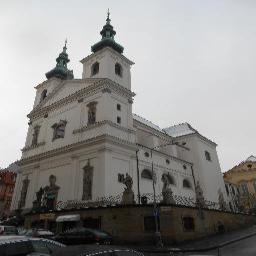} &
        \includegraphics[width=0.095\textwidth]{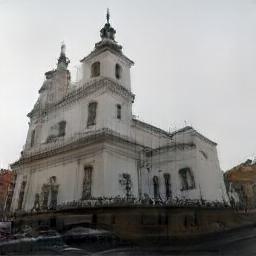} &
        \includegraphics[width=0.095\textwidth]{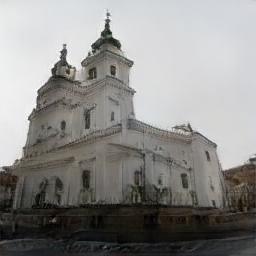} &
        \includegraphics[width=0.095\textwidth]{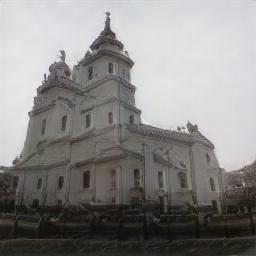} &
        \includegraphics[width=0.095\textwidth]{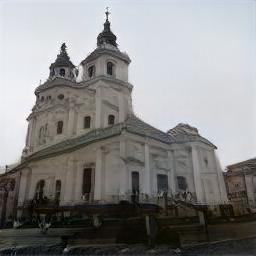} \\

        Input & Optimization & Hybrid & pSp & $\text{ReStyle}_{pSp}$ & Input & Optimization & Hybrid & pSp & $\text{ReStyle}_{pSp}$ \\
        
        \includegraphics[width=0.095\textwidth]{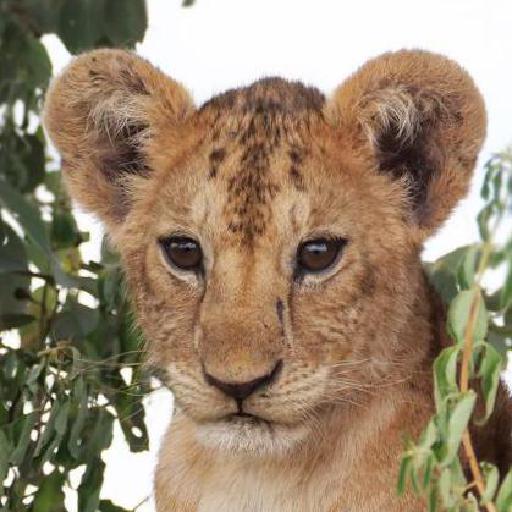} &
        \includegraphics[width=0.095\textwidth]{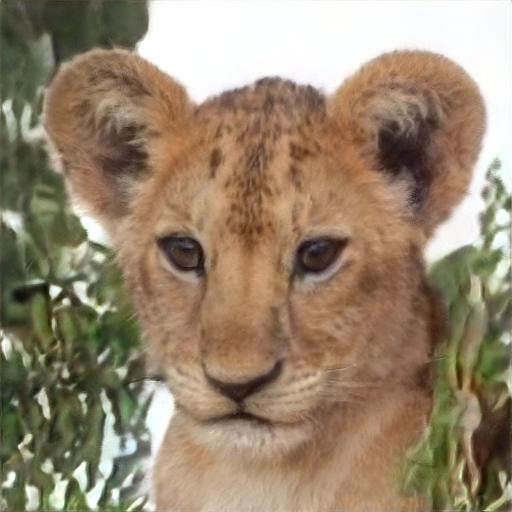} &
        \includegraphics[width=0.095\textwidth]{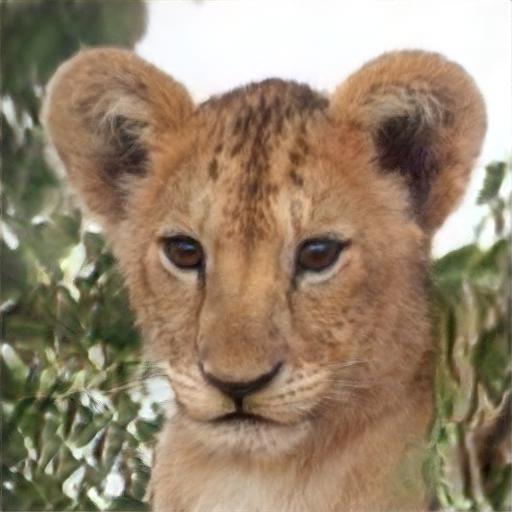} &
        \includegraphics[width=0.095\textwidth]{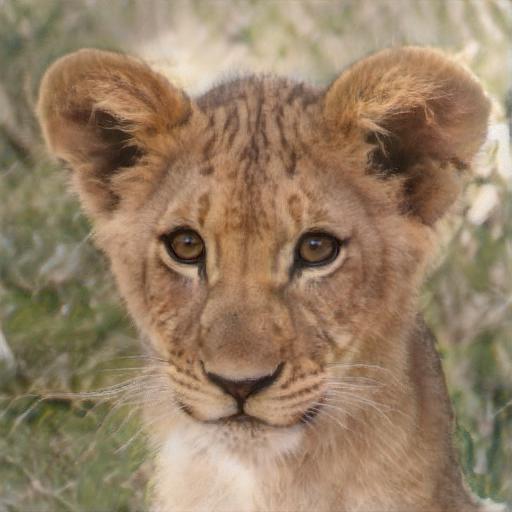} &
        \includegraphics[width=0.095\textwidth]{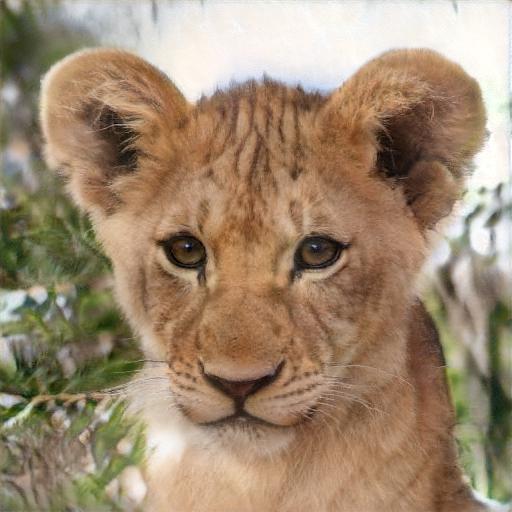} &
        \hspace{0.05cm}
        \includegraphics[width=0.095\textwidth]{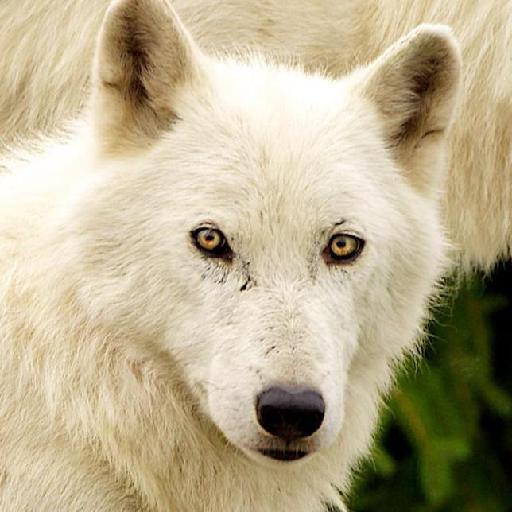} &
        \includegraphics[width=0.095\textwidth]{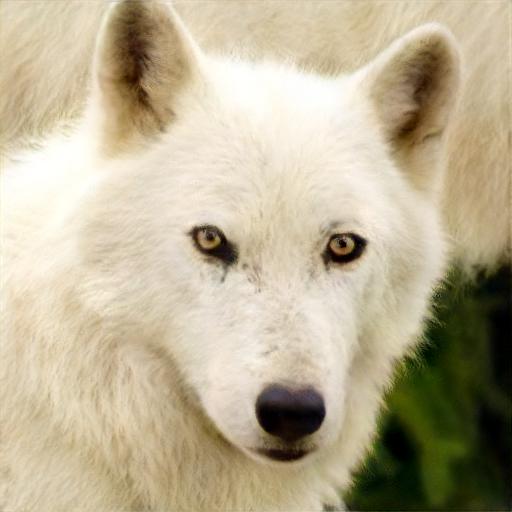} &
        \includegraphics[width=0.095\textwidth]{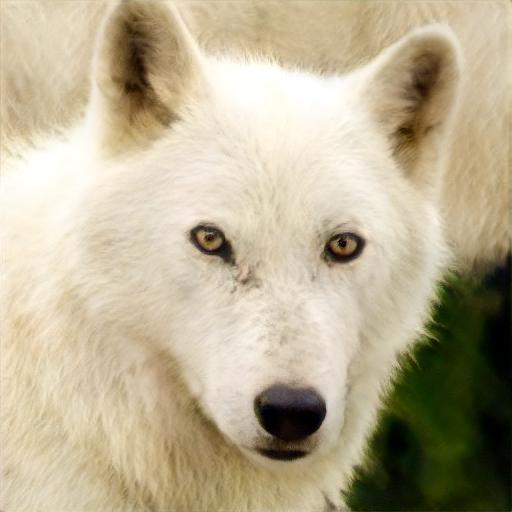} &
        \includegraphics[width=0.095\textwidth]{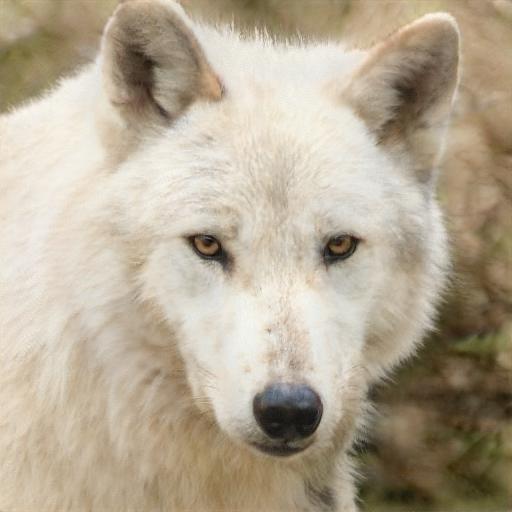} &
        \includegraphics[width=0.095\textwidth]{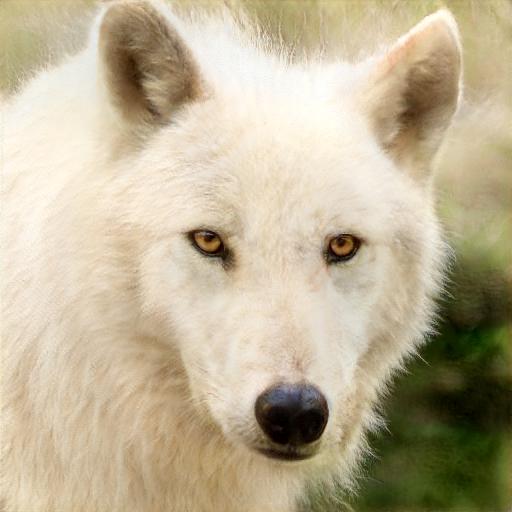} \\

        \includegraphics[width=0.095\textwidth]{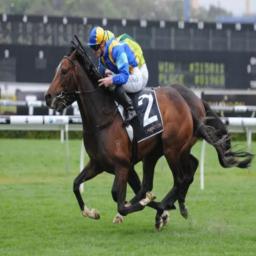} &
        \includegraphics[width=0.095\textwidth]{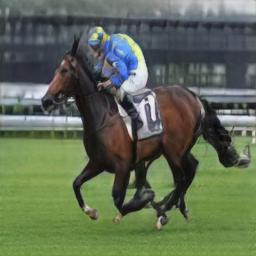} &
        \includegraphics[width=0.095\textwidth]{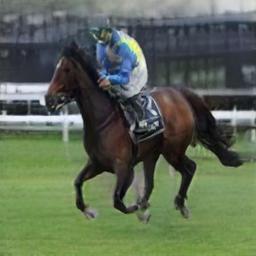} &
        \includegraphics[width=0.095\textwidth]{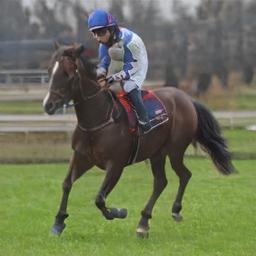} &
        \includegraphics[width=0.095\textwidth]{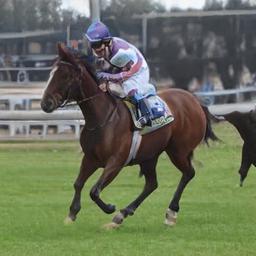} &
        \hspace{0.05cm}
        \includegraphics[width=0.095\textwidth]{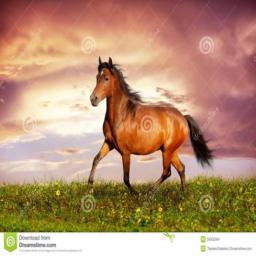} &
        \includegraphics[width=0.095\textwidth]{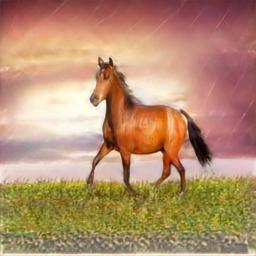} &
        \includegraphics[width=0.095\textwidth]{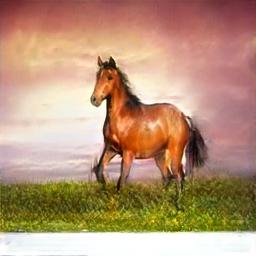} &
        \includegraphics[width=0.095\textwidth]{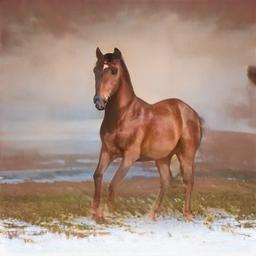} &
        \includegraphics[width=0.095\textwidth]{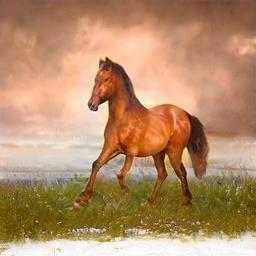} \\
        
        Input & Optimization & Hybrid & e4e & $\text{ReStyle}_{e4e}$ & Input & Optimization & Hybrid & e4e & $\text{ReStyle}_{e4e}$

    \end{tabular}
    
    }
    \vspace{0.1cm}
    \caption{\textit{Qualitative Comparison.} We compare various encoder-based and optimization-based inversion methods with our ReStyle scheme applied over pSp~\cite{richardson2020encoding} and e4e~\cite{tov2021designing} (denoted by $\text{ReStyle}_{pSp}$ and $\text{ReStyle}_{e4e}$). Hybrid results are obtained by performing optimization on the latent codes obtained by the adjacent encoder. Additional comparisons are provided in Appendix~\ref{sec:additional_results}. Best viewed zoomed-in.}
    \label{fig:qualitative_comparison}
\end{figure*}

\section{Experiments}

\subsection{Settings}~\label{settings}

\vspace{-0.3cm}
\topic{\textit{Datasets.}}
We conduct extensive evaluations on a diverse set of domains to illustrate the generalization of our approach. For the human facial domain we use the FFHQ~\cite{karras2019style} dataset for training and the CelebA-HQ~\cite{liu2015deep, karras2017progressive} test set for evaluation. For the cars domains, we use the Stanford Cars~\cite{KrauseStarkDengFei-Fei_3DRR2013} dataset for training and evaluation. Additional evaluations are performed on the LSUN~\cite{yu2016lsun} Horse and Church datasets as well as the AFHQ Wild~\cite{choi2020stargan} dataset.

\topic{\textit{Baselines.}}
Throughout this section, we explore and analyze encoder-based, optimization-based, and hybrid inversion techniques. 
For encoder-based methods, we compare our ReStyle approach with the IDInvert encoder from Zhu \etal~\cite{zhu2020domain}, pSp from Richardson \etal~\cite{richardson2020encoding}, and e4e from Tov \etal~\cite{tov2021designing}.
For optimization-based methods, we compare our results with the inversion technique from Karras \etal~\cite{karras2020analyzing}. 
For each of the above encoder-based inversion methods we also perform optimization on the resulting latents for a comparison with hybrid approaches. Additional details can be found in Appendix~\ref{sec:baselines}.

\vspace{0.03cm}
\topic{\textit{Architecture and Training Details.}}
For the facial domain, we employ the ResNet-IRSE50 architecture from Deng \etal~\cite{deng2019arcface} pre-trained for facial recognition. For all other domains, we use a ResNet34 network pre-trained on ImageNet. These networks have a modified input layer to accommodate the $6$-channel input used by ReStyle.
All results were obtained using StyleGAN2~\cite{karras2020analyzing} generators. 

Throughout this section, we apply ReStyle on pSp~\cite{richardson2020encoding} and e4e~\cite{tov2021designing} using the loss objectives and training details (e.g., batch size, loss weights) as originally defined in their respective works. Note that when applying ReStyle, we utilize the simplified encoder architecture presented in Section~\ref{sec:encoder_arch} for extracting the image inversion.
All ReStyle encoders are trained using $N=5$ steps per batch.

\subsection{Comparison with Inversion Methods}~\label{sec:comparison}
We first compare ReStyle with current state-of-the-art StyleGAN inversion techniques. While per-image optimization techniques have achieved superior image reconstruction compared to learning-based approaches, they come with a significantly higher computational cost. Therefore, when analyzing the inversion approaches, it is essential to measure reconstruction quality with respect to inference time, resulting in a so-called \textit{quality-time trade-off}. 

\vspace{0.05cm}
\topic{\textit{Qualitative Evaluation.}}
We begin by showing a qualitative comparison of ReStyle and the alternative inversion approaches across various domains in Figure~\ref{fig:qualitative_comparison}. 
It is important to emphasize that we do not claim to achieve superior reconstruction quality over optimization. The comparison instead serves to show that ReStyle is visually comparable to the latter. Attaining comparable reconstruction quality with a significantly lower inference time places ReStyle at an appealing point on the quality-time trade-off curve. 

With that, we do note the improved reconstruction obtained by ReStyle in comparison with the pSp and e4e encoders, especially in the preservation of fine details. For example in the comparison with pSp (the top three rows), observe the collar of the man in the top left and the hair of the woman in the top right. Similarly, observe the Audi symbol and the license plate in the car comparison on the left-hand side. In the comparison with e4e (the bottom two rows), observe how ReStyle better captures the background of the wild animals and the pose of the horse.
A large-scale gallery of results and comparisons is provided in Appendix~\ref{sec:additional_results}.

\begin{figure*}
    \centering
    \setlength{\belowcaptionskip}{-2.5pt}
    \setlength{\tabcolsep}{1pt}
    \begin{tabular}{c c c}

        \includegraphics[width=0.33\textwidth]{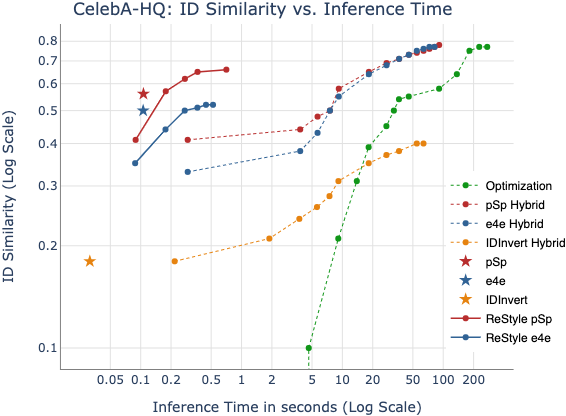} &
        \hspace{0.05cm}
        \includegraphics[width=0.33\textwidth]{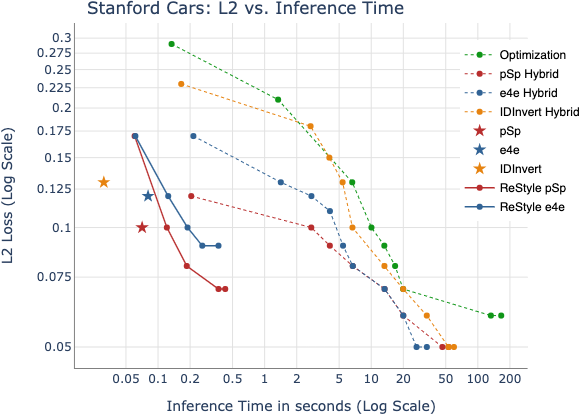} &       
        \hspace{0.05cm}
        \includegraphics[width=0.33\textwidth]{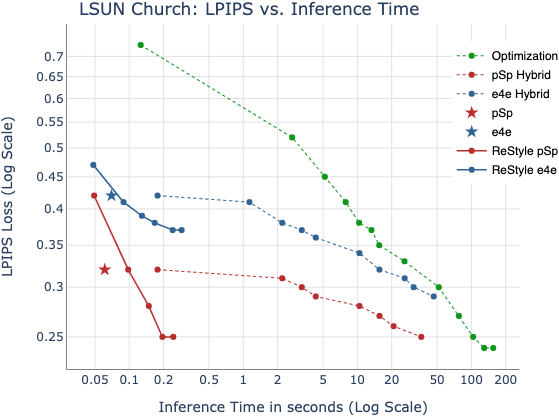}          

    \end{tabular}
    \vspace{0.1cm}
    \caption{\textit{Quantitative comparison.} We compare ReStyle with current state-of-the-art optimization-based and encoder-based methods 
    by analyzing reconstruction via three evaluation metrics --- ID similarity for faces, L2 loss for cars, and LPIPS loss for churches --- while measuring each method's inference time.
    Each encoder-based method is represented using a $\star$ symbol. The corresponding hybrid method is marked using a dashed line of the same color with the ReStyle applied over the base method shown using a solid line of the same color.
    Optimization results are shown using a dashed \textcolor{ForestGreen}{green} line.
    Methods based on pSp are shown in \textcolor{red}{red} with methods based on e4e shown in \textcolor{blue}{blue}. Finally, results obtained using IDInvert~\cite{zhu2020domain} are shown in \textcolor{orange}{orange}.  
    Note that both axes are shown in log-scale. 
    }
    \label{fig:quantitative_comparison}
\end{figure*}

\vspace{0.15cm}
\topic{\textit{Quantitative Evaluation.}}
We now perform a quantitative comparison of the inversion approaches across various domains. To measure both pixel-wise and perceptual similarities we apply the commonly-used $L_2$ and LPIPS~\cite{zhang2018unreasonable} metrics.
In addition, for the facial domain, to measure each method's ability to preserve input identity, we measure the identity similarity between the reconstructions and their sources using the CurricularFace~\cite{huang2020curricularface} recognition method.

To illustrate the trade-off between the different methods we additionally measure each method's inference time per image.
As mentioned, both optimization and ReStyle can be viewed as a continuous curve on a quality-time graph --- with each additional step, we attain improved reconstruction quality at the cost of additional inference time.

To provide a complete comparison of all inversion methods, we construct a quality-time graph for each domain. Such graphs can be visualized in Figure~\ref{fig:quantitative_comparison}. 
To form each graph, we performed the following evaluations for each inversion technique. For each encoder-based inversion, we ran a single forward pass to obtain the reconstruction image, resulting in a single point on the graph.
For measuring the optimization technique from~\cite{karras2020analyzing}, we invert the input image using a varying number of steps from $1$ optimization step up to $1,500$ steps. 
For hybrid approaches, given the computed latent codes obtained from the corresponding encoder, we performed optimization with an increasing number of steps between $1$ to $500$ steps. 
Finally, for our two ReStyle encoders, we performed up to $10$ feedback loops. 

We begin by analyzing the facial domain.
Compared to the conventional pSp and e4e encoders, our ReStyle variants match or surpass their counterparts. More notably, while optimization techniques achieve improved identity similarity compared to ReStyle, they require $\approx20\times$ more time to match the similarity attained by ReStyle.
A similar trade-off can be observed in the cars domain where now the advantage of ReStyle over typical encoders is more pronounced when evaluating the $L_2$ loss of the reconstructions. 
In the unstructured churches domain, ReStyle applied over pSp nearly matches both optimization and hybrid techniques in reconstruction quality with a significantly lower inference time. 
Observe that the first output of ReStyle may be \textit{worse} than that of a conventional encoder due to the more relaxed training formulation of ReStyle as it is trained to perform multiple steps at inference. With that, ReStyle quickly matches or surpasses the quality of single-shot encoders.

These comparisons point to the appealing nature of ReStyle: although optimization typically achieves superior reconstruction, ReStyle offers an excellent balance between reconstruction quality and inference time. 
See Appendix~\ref{sec:quant_comparison} for results on all domains and metrics. 

\begin{figure}
    \centering
    \setlength{\belowcaptionskip}{-5pt}
    \setlength{\tabcolsep}{1pt}
    {\small 
    \begin{tabular}{c c c c}

        \includegraphics[width=0.1\textwidth]{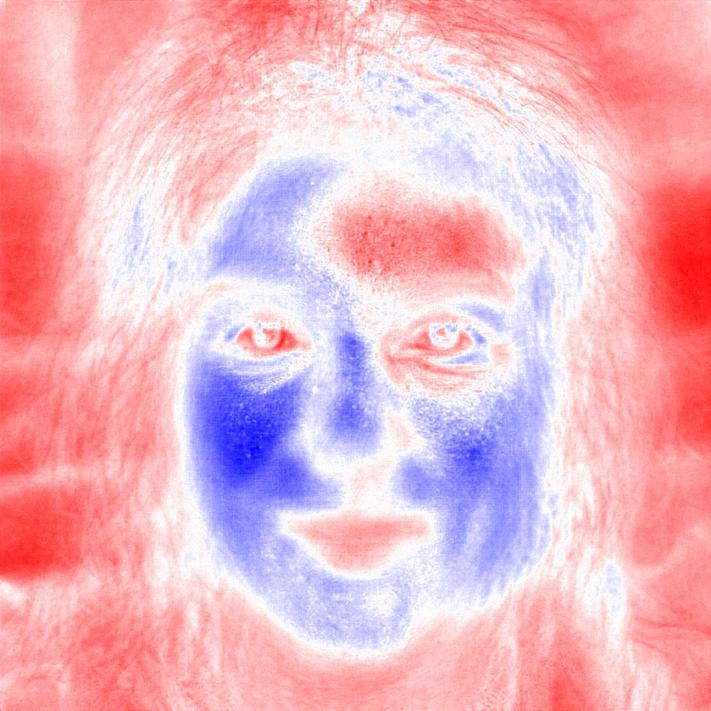} &
        \includegraphics[width=0.1\textwidth]{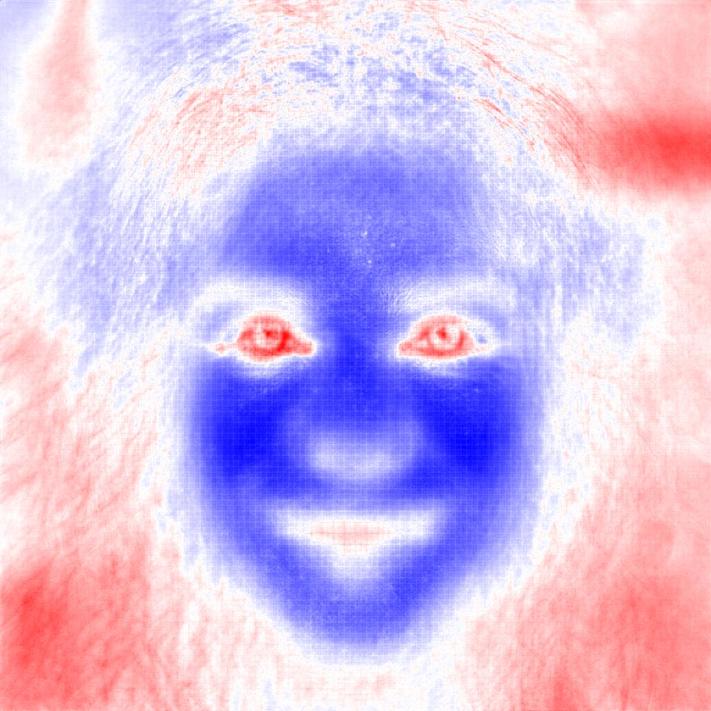} &
        \includegraphics[width=0.1\textwidth]{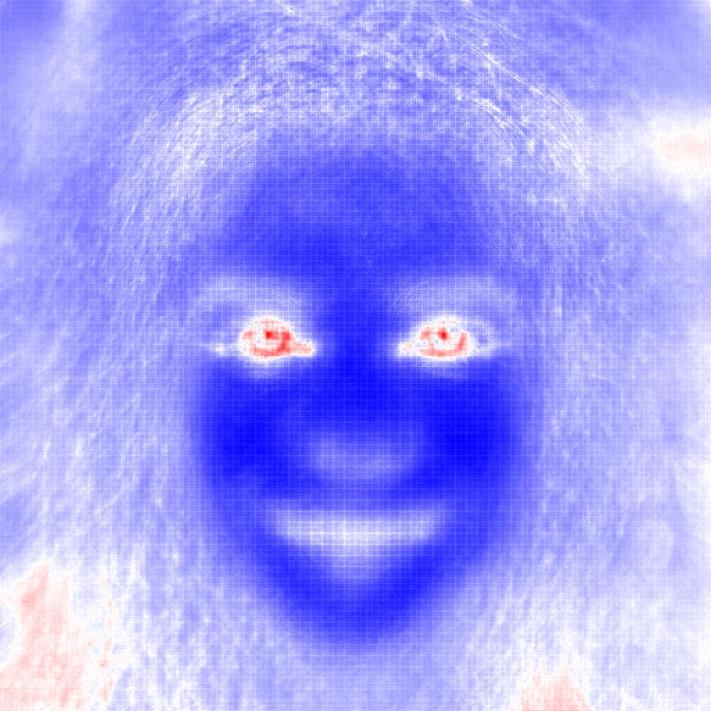} &
        \includegraphics[width=0.1\textwidth]{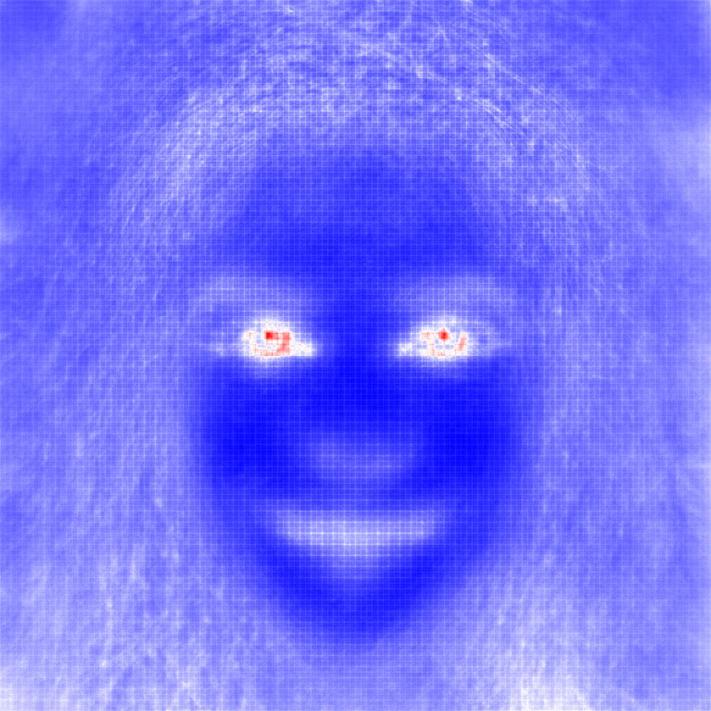}
        \tabularnewline
        $1\to2$ & $2\to3$ & $3\to4$ & $4\to5$
        
    \end{tabular}
    }
    \vspace{0.1cm}
    \caption{In each sub-image, we display a heatmap showing which image regions changed the most (in \textcolor{red}{red}) and which regions changed the least (in \textcolor{blue}{blue}) between the specified iterations.}
    \label{fig:image_diffs_facial_domain}
\end{figure}

\begin{figure}
    \centering
    \setlength{\belowcaptionskip}{-5pt}
    \setlength{\tabcolsep}{1pt}
    { \small
    \begin{tabular}{c c c c c}
        \includegraphics[width=0.1\textwidth]{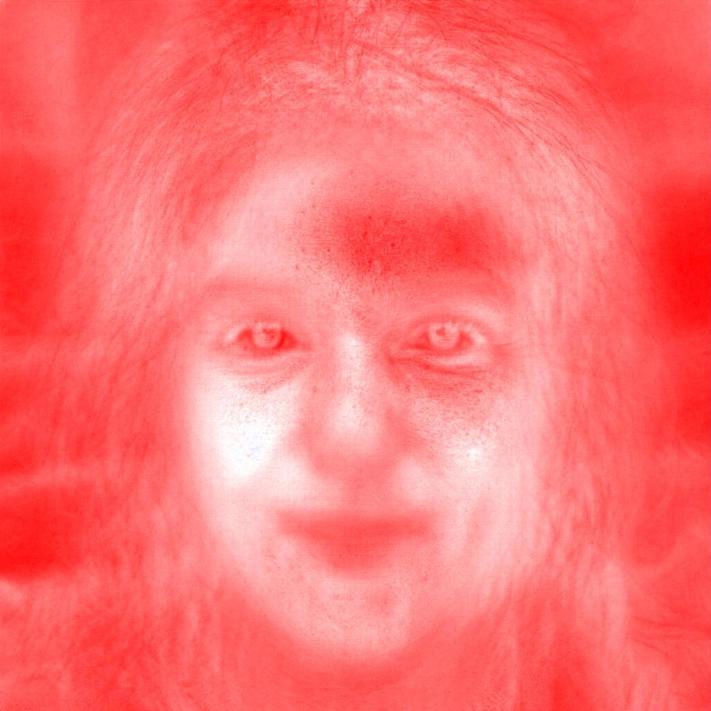} &
        \includegraphics[width=0.1\textwidth]{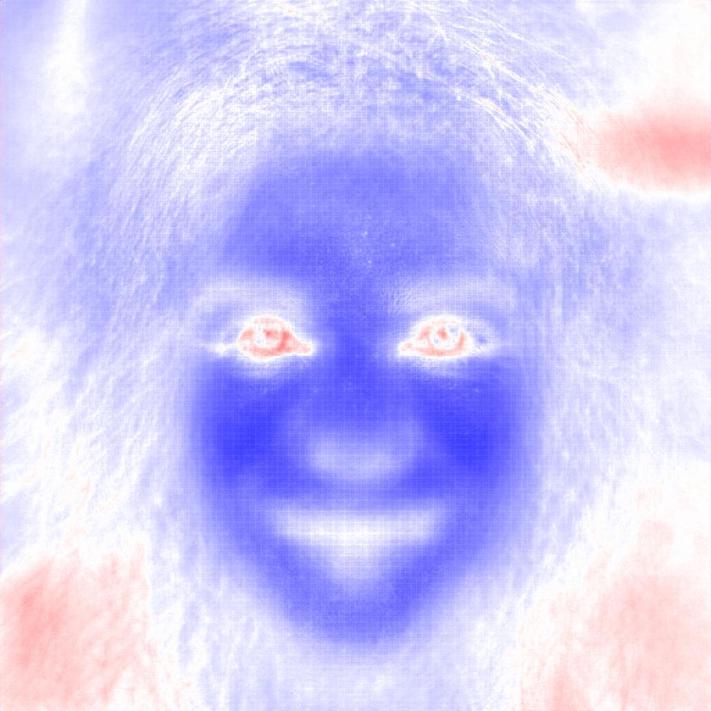} &
        \includegraphics[width=0.1\textwidth]{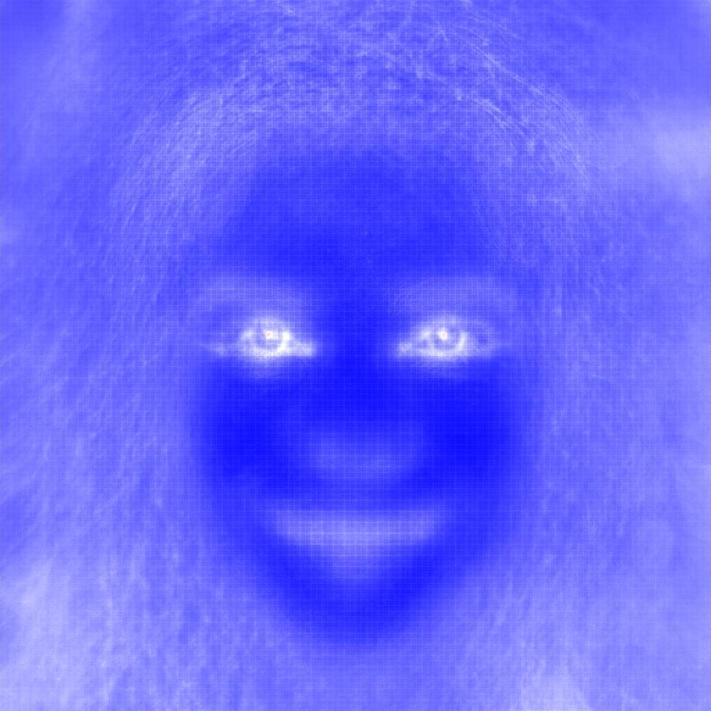} &
        \includegraphics[width=0.1\textwidth]{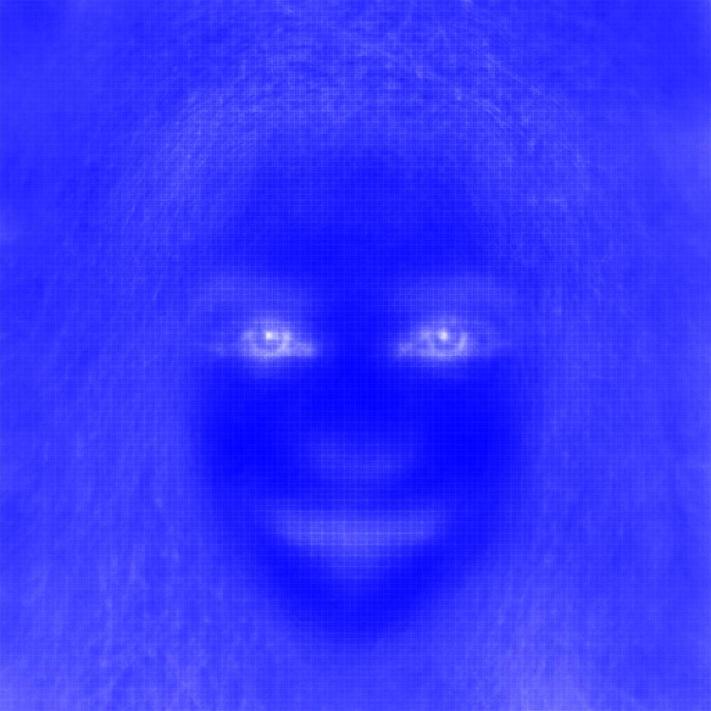}
        \tabularnewline
        $1\to2$ & $2\to3$ & $3\to4$ & $4\to5$ 
    \end{tabular}
    }
    \vspace{0.1cm}
    \caption{Similar to Figure~\ref{fig:image_diffs_facial_domain}, with the difference that here all images are normalized with respect to each other. As shown, the magnitude of change decreases with each step.}
    \label{fig:image_diffs_facial_domain_global}
\end{figure}

\begin{figure*}
    \centering
    \setlength{\belowcaptionskip}{-5pt}
    \setlength{\tabcolsep}{1pt}
    {\small
    \begin{tabular}{c c c c c c c c c c}

        \raisebox{0.225in}{\rotatebox[origin=t]{90}{Age}} &
        \includegraphics[width=0.1\textwidth]{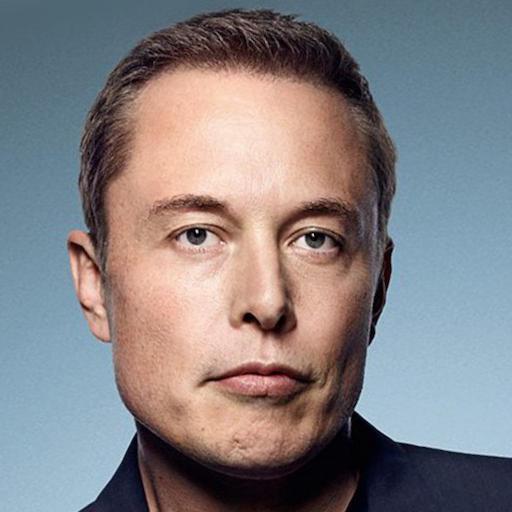} &
        \includegraphics[width=0.1\textwidth]{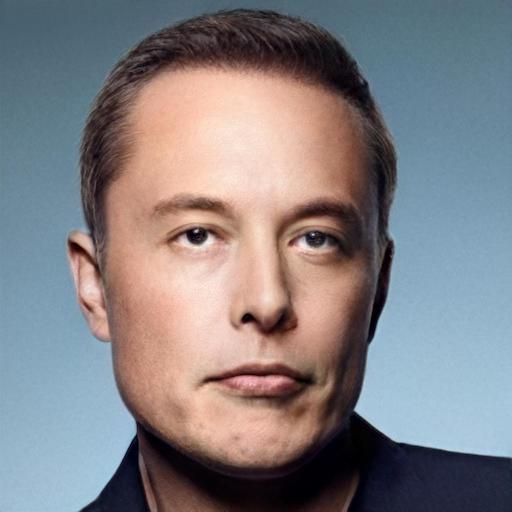} &
        \includegraphics[width=0.1\textwidth]{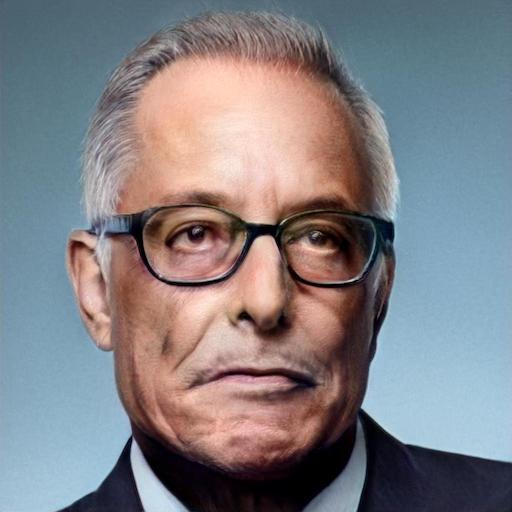} &
        \includegraphics[width=0.1\textwidth]{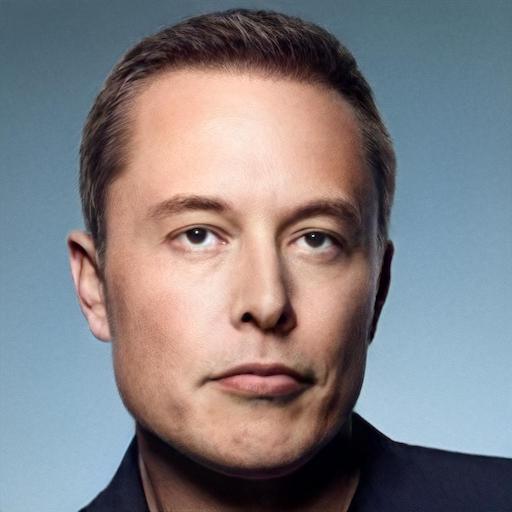} &
        \includegraphics[width=0.1\textwidth]{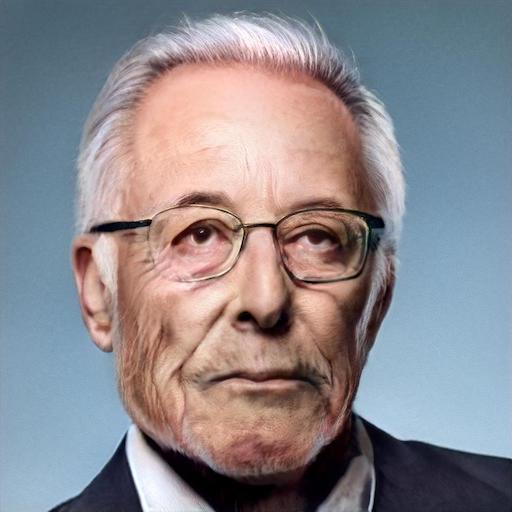} &
        \includegraphics[width=0.1\textwidth]{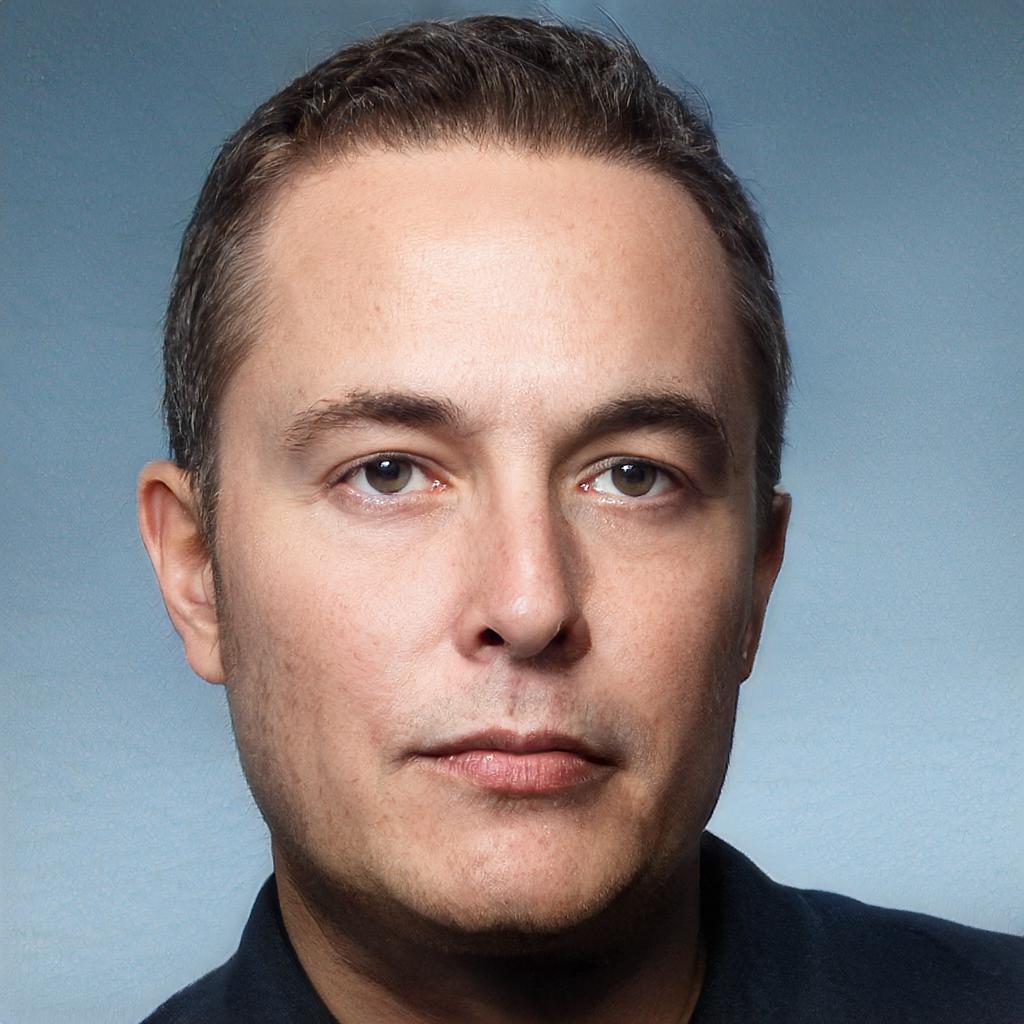} &
        \includegraphics[width=0.1\textwidth]{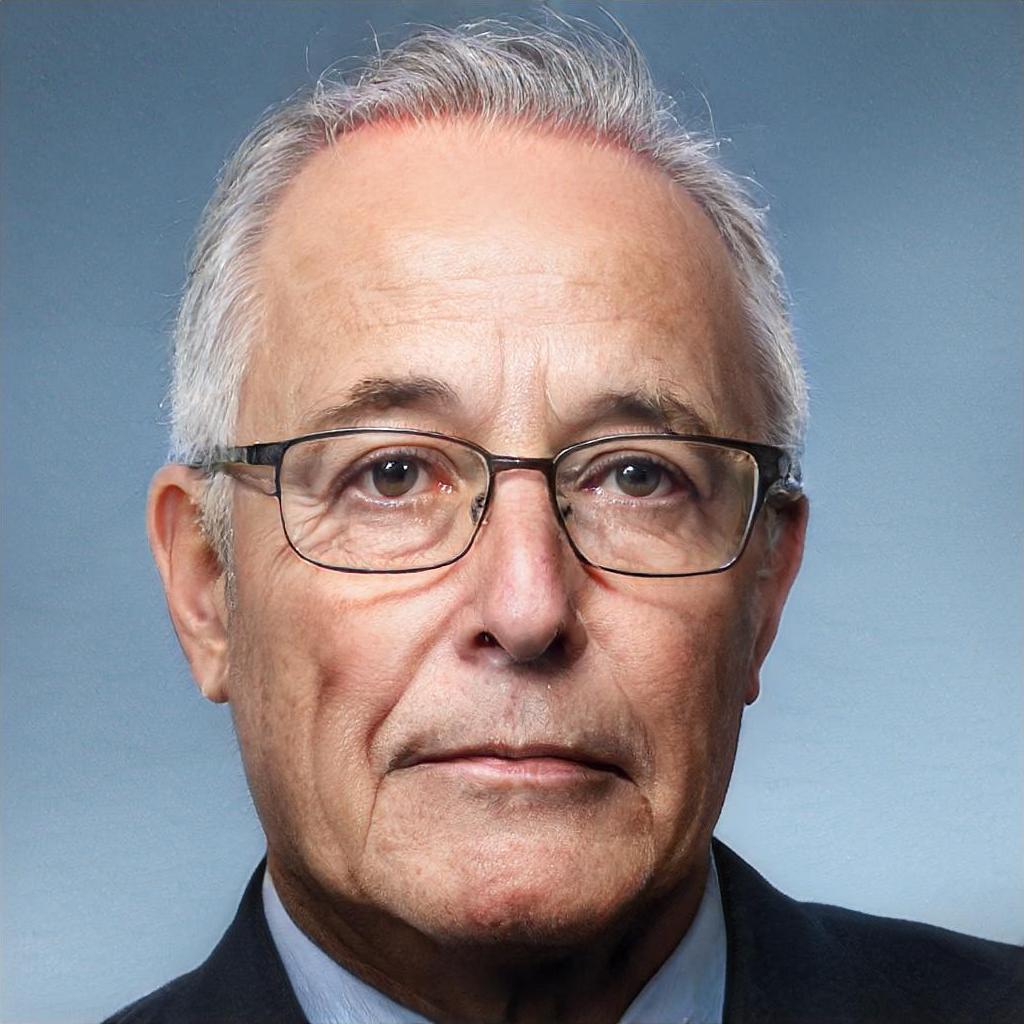} &
        \includegraphics[width=0.1\textwidth]{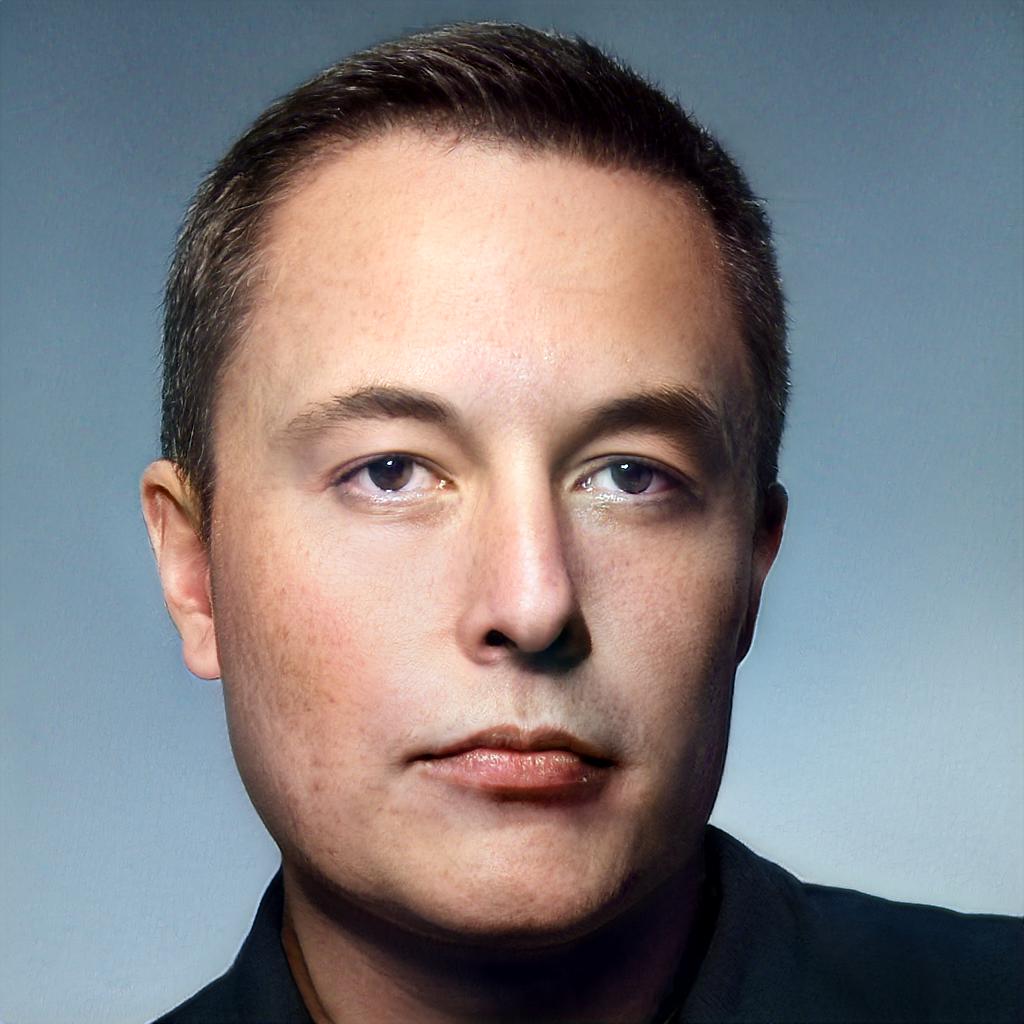} &
        \includegraphics[width=0.1\textwidth]{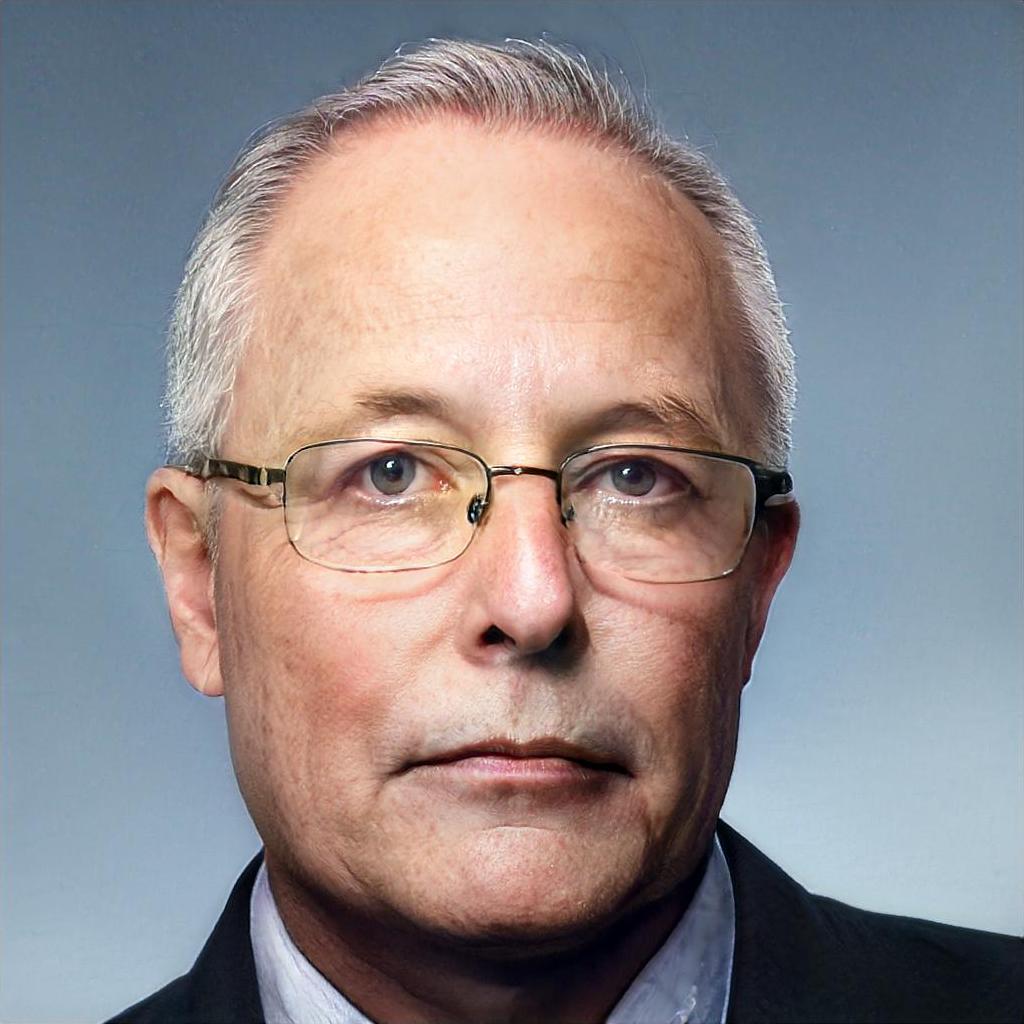} \\
        
        \raisebox{0.25in}{\rotatebox[origin=t]{90}{Smile}} &
        \includegraphics[width=0.1\textwidth]{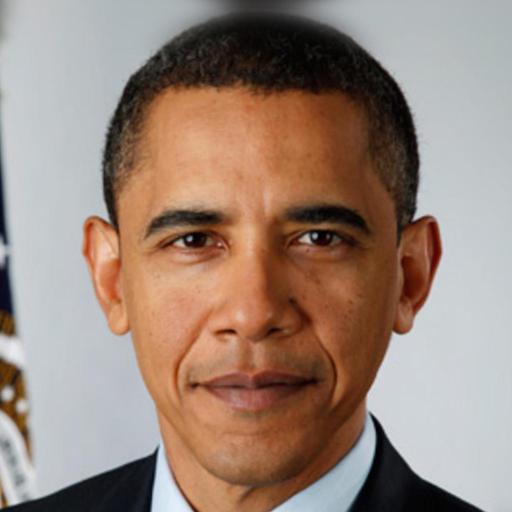} &
        \includegraphics[width=0.1\textwidth]{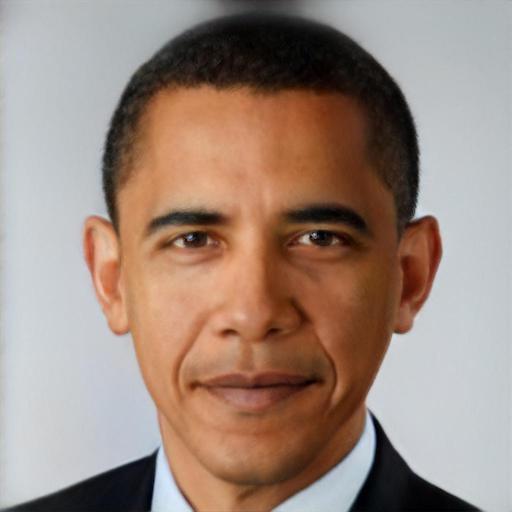} &
        \includegraphics[width=0.1\textwidth]{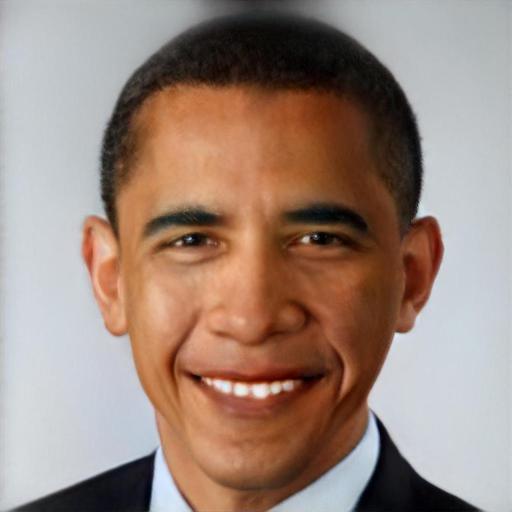} &
        \includegraphics[width=0.1\textwidth]{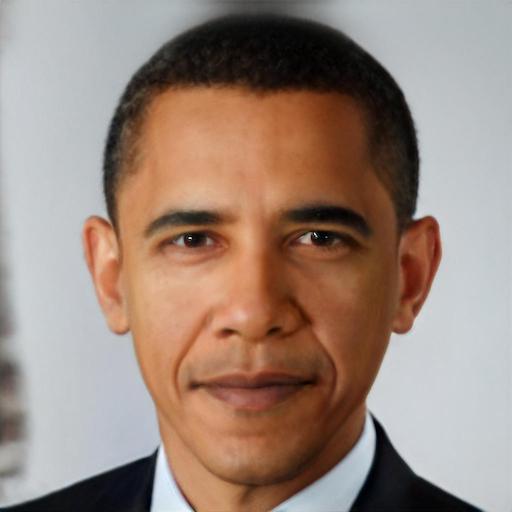} &
        \includegraphics[width=0.1\textwidth]{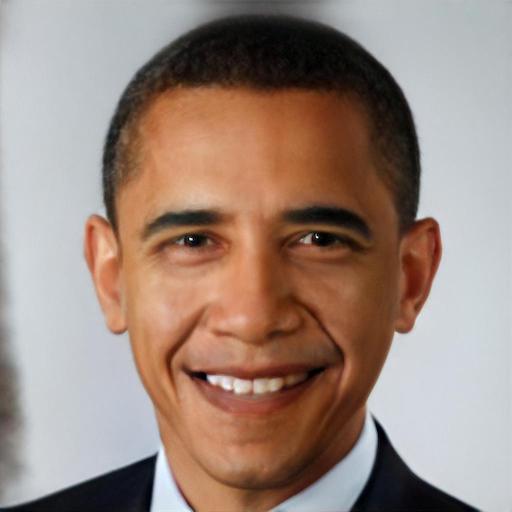} &
        \includegraphics[width=0.1\textwidth]{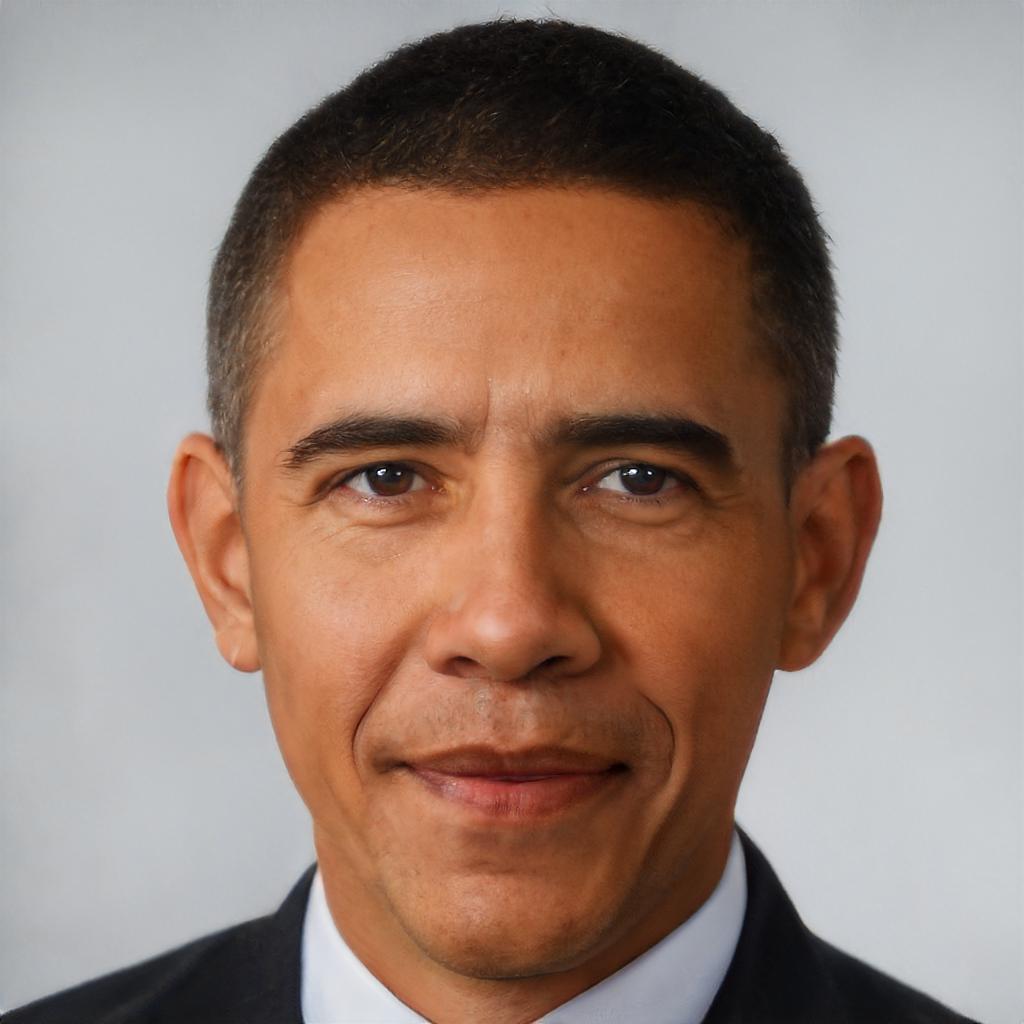} &
        \includegraphics[width=0.1\textwidth]{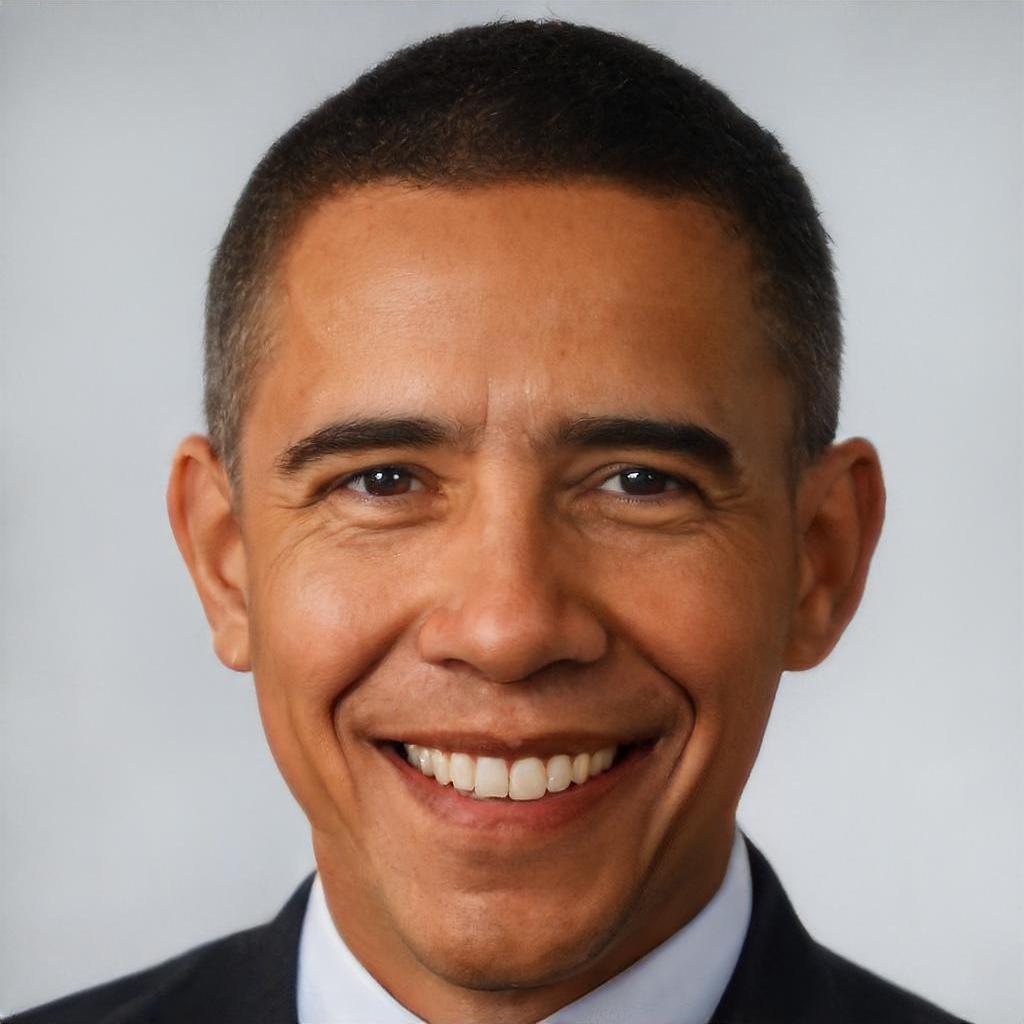} &
        \includegraphics[width=0.1\textwidth]{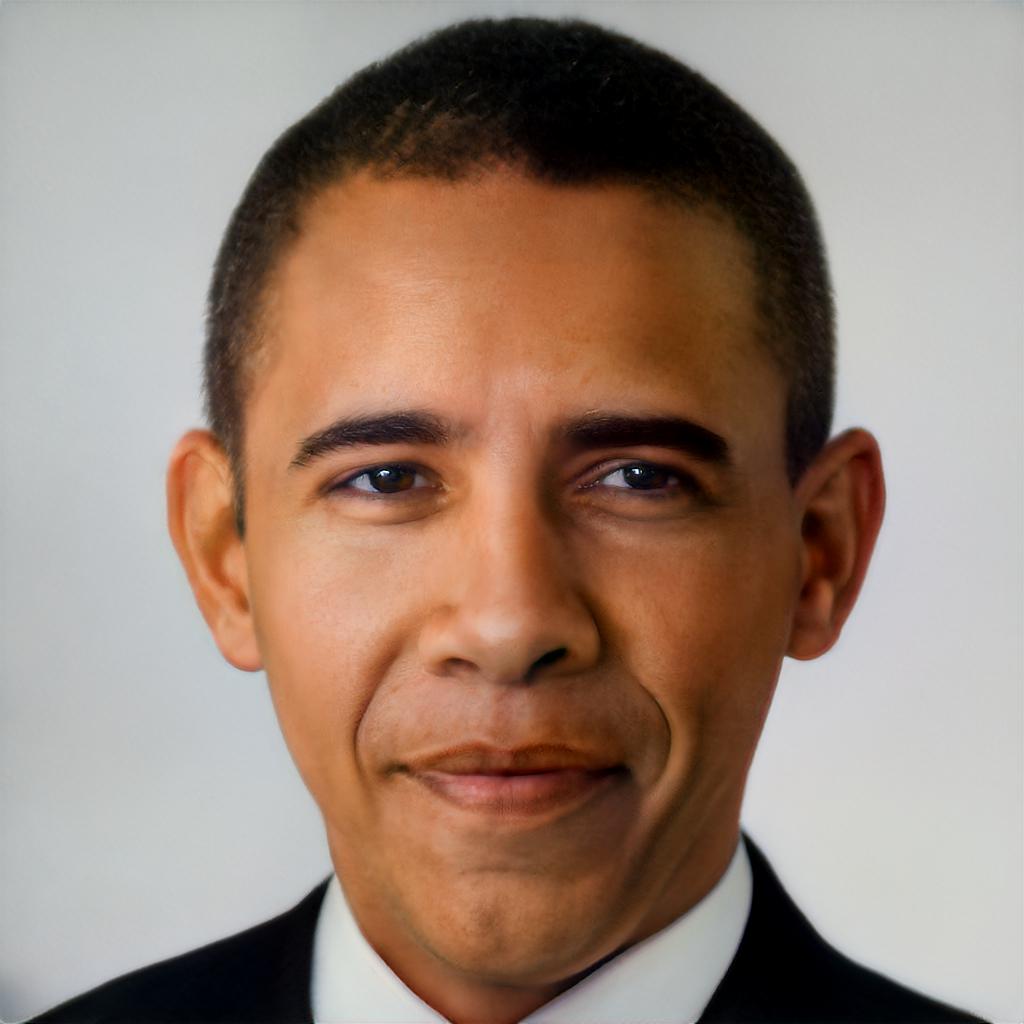} &
        \includegraphics[width=0.1\textwidth]{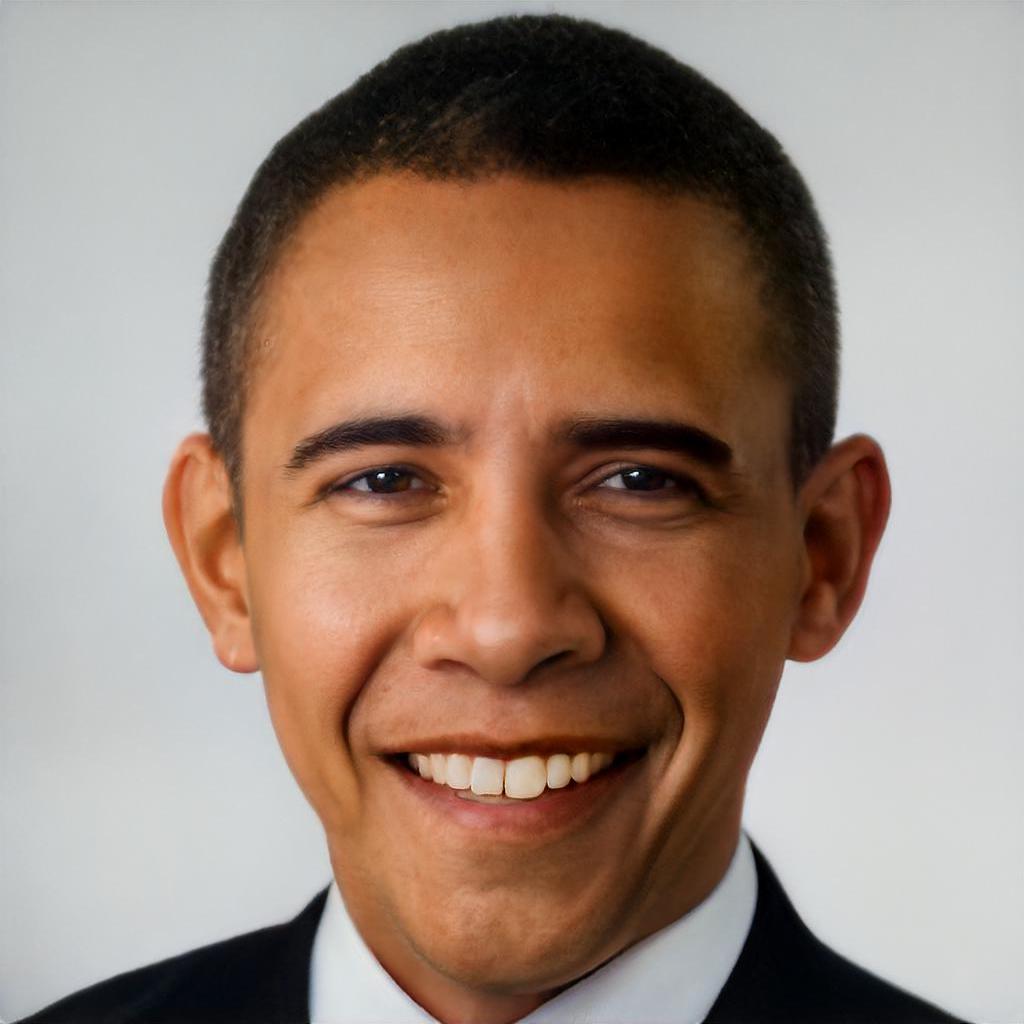} \\
        
        \raisebox{0.175in}{\rotatebox[origin=t]{90}{View}} &
        \includegraphics[width=0.1\textwidth]{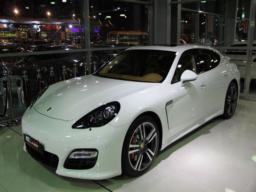} &
        \includegraphics[width=0.1\textwidth]{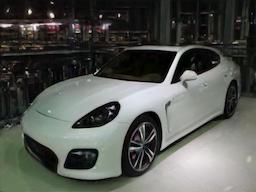} &
        \includegraphics[width=0.1\textwidth]{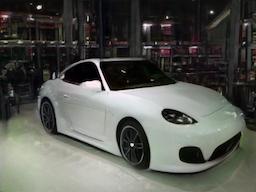} &
        \includegraphics[width=0.1\textwidth]{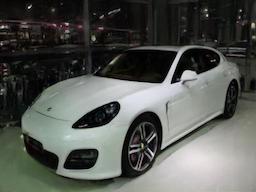} &
        \includegraphics[width=0.1\textwidth]{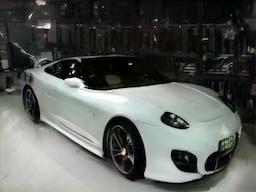} &
        \includegraphics[width=0.1\textwidth]{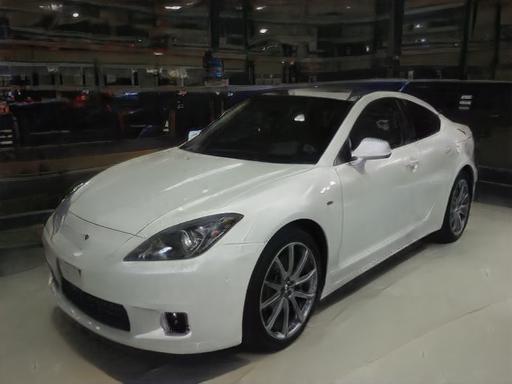} &
        \includegraphics[width=0.1\textwidth]{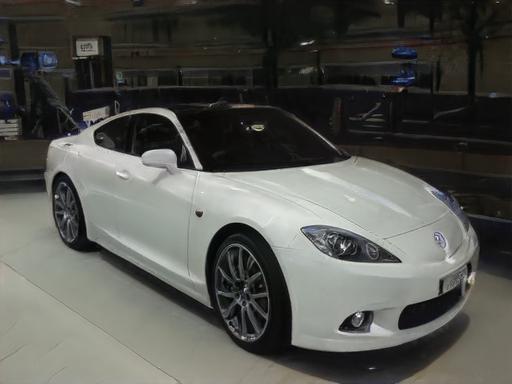} &
        \includegraphics[width=0.1\textwidth]{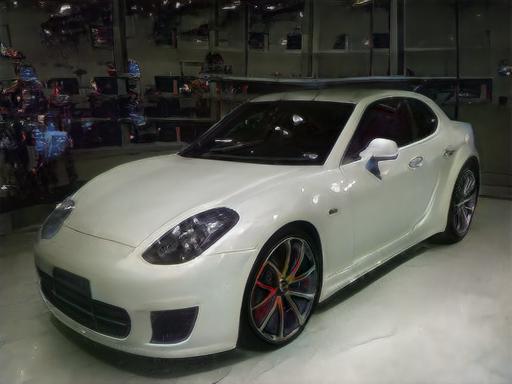} &
        \includegraphics[width=0.1\textwidth]{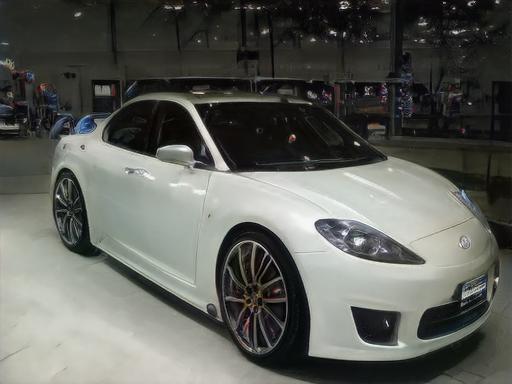} \\
        
        \raisebox{0.175in}{\rotatebox[origin=t]{90}{Cube}} &
        \includegraphics[width=0.1\textwidth]{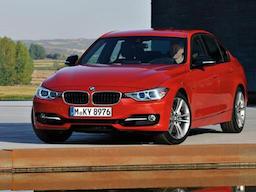} &
        \includegraphics[width=0.1\textwidth]{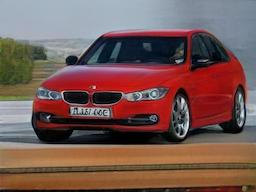} &
        \includegraphics[width=0.1\textwidth]{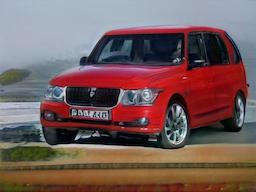} &
        \includegraphics[width=0.1\textwidth]{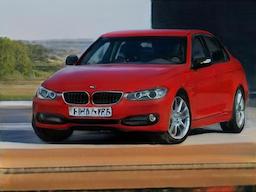} &
        \includegraphics[width=0.1\textwidth]{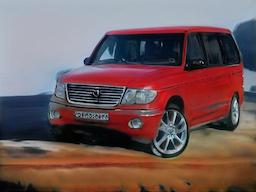} &
        \includegraphics[width=0.1\textwidth]{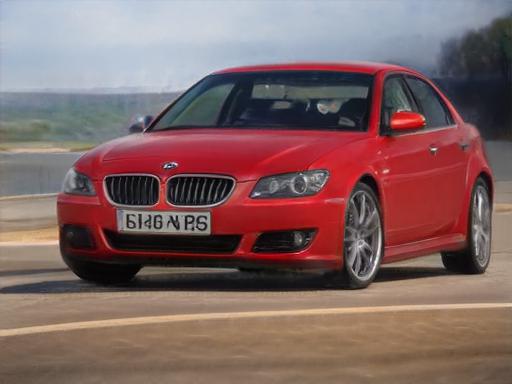} &
        \includegraphics[width=0.1\textwidth]{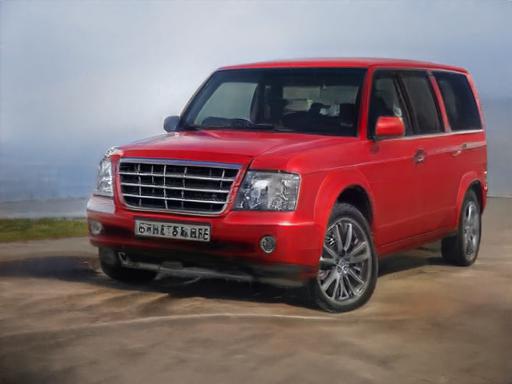} &
        \includegraphics[width=0.1\textwidth]{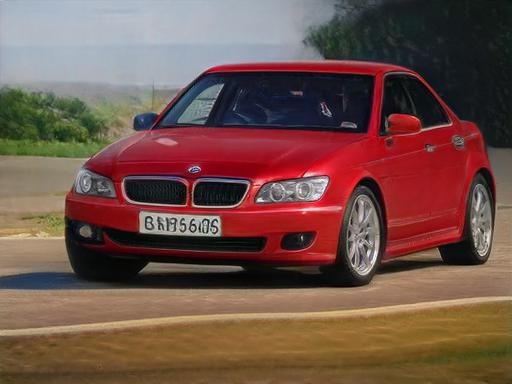} &
        \includegraphics[width=0.1\textwidth]{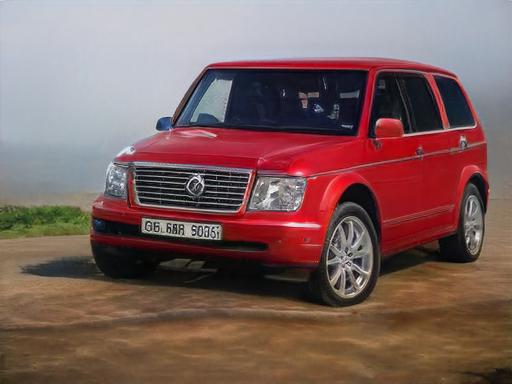} \\
        
        \raisebox{0.25in}{\rotatebox[origin=t]{90}{Pose + Zoom}} &
        \includegraphics[width=0.1\textwidth]{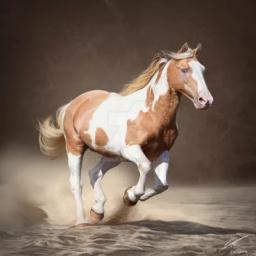} &
        \includegraphics[width=0.1\textwidth]{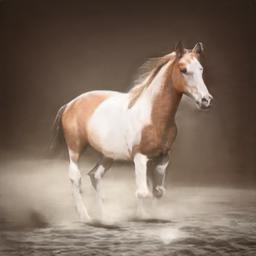} &
        \includegraphics[width=0.1\textwidth]{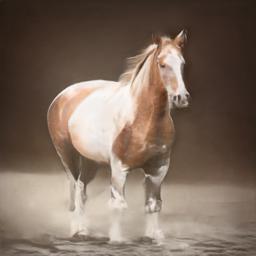} &
        \includegraphics[width=0.1\textwidth]{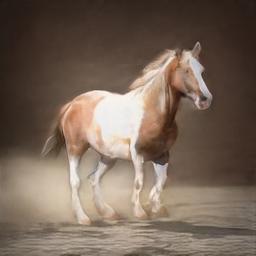} &
        \includegraphics[width=0.1\textwidth]{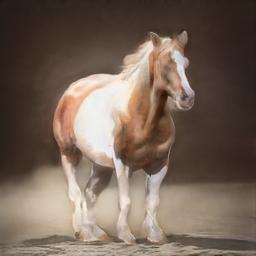} &
        \includegraphics[width=0.1\textwidth]{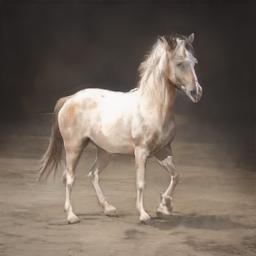} &
        \includegraphics[width=0.1\textwidth]{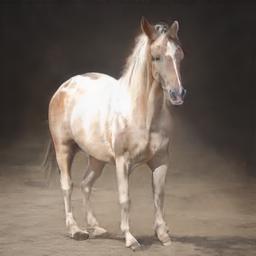} &
        \includegraphics[width=0.1\textwidth]{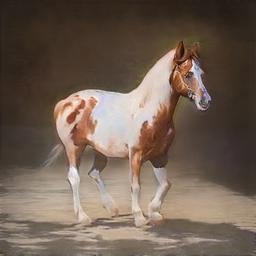} &
        \includegraphics[width=0.1\textwidth]{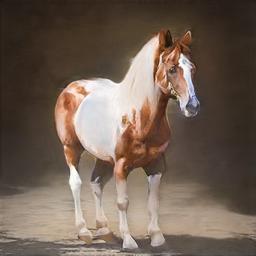} \\

        &
        \begin{tabular}{@{}c@{}}Input\end{tabular} &
        \begin{tabular}{@{}c@{}}Optimization \\ Inversion\end{tabular} &
        \begin{tabular}{@{}c@{}}Optimization \\ Edit\end{tabular} &
        \begin{tabular}{@{}c@{}}Hybrid \\ Inversion\end{tabular} &
        \begin{tabular}{@{}c@{}}Hybrid \\ Edit\end{tabular} &
        \begin{tabular}{@{}c@{}}e4e \\ Inversion\end{tabular} &
        \begin{tabular}{@{}c@{}}e4e \\ Edit\end{tabular} &
        \begin{tabular}{@{}c@{}}$\text{ReStyle}_{e4e}$ \\ Inversion\end{tabular} &
        \begin{tabular}{@{}c@{}}$\text{ReStyle}_{e4e}$ \\ Edit\end{tabular}

    \end{tabular}
    }
    \vspace{0.1cm}

    \caption{\textit{Editing comparison.} Applying edits on inversions of several methods. For performing the edits in the human facial domain we use InterFaceGAN~\cite{shen2020interpreting}, for the cars domain we use GANSpace~\cite{harkonen2020ganspace}, and for the horse domain we use SeFa~\cite{shen2020closedform}. Image credits: ~\cite{obama},\cite{musk}.}
    \label{fig:editing_comparison}
\end{figure*}

\subsection{ReStyle Analysis}~\label{sec:analysis}
In this section, we explore various aspects of ReStyle to gain a stronger understanding of its behavior and attain key insights into its efficiency. Specifically, we analyze the main details focused on by the encoder at each step and analyze the number of steps needed for convergence during inference. Additional analyses in both the image space and latent space can be found in Appendix~\ref{sec:additional_analysis}. 

\vspace{0.15cm}
\topic{\textit{Where's the focus?}}
We begin by exploring which regions of the image are focused on by the encoder at each step during inference. To do so, we consider the human facial domain. For each step $t$ and each input image $\textbf{x}$, we compute the squared difference in the image space between the generated images at steps $t$ and $t-1$. That is, we compute $d=\norm{\textbf{y}_t - \textbf{y}_{t-1}}_2$
where $\textbf{y}_t$ is defined in Equation~\ref{eq:output}.

Averaging over all test samples we obtain the average image difference between the two steps. Finally, we normalize the average image to the range $[0,1]$ and visualize the regions of the image that incur the most change at the current step $t$.
We visualize this process in Figure~\ref{fig:image_diffs_facial_domain} showing ReStyle's incremental refinements. As can be seen, in the early steps the encoder focuses on refining the background and pose while in subsequent steps the encoder moves its focus to adjusting finer details along the eyes and hair. 

In Figure~\ref{fig:image_diffs_facial_domain} we show only the magnitude of change \textit{within} each step. That is, the absolute magnitude of change may vary between the different steps. To show that the overall amount of change decreases with each step, we refer the reader to Figure~\ref{fig:image_diffs_facial_domain_global}. There, all images are normalized with respect to each other allowing one to see how the largest changes occur in the first step and decrease thereafter.

In a sense, the encoder operates in a coarse-to-fine manner, beginning by concentrating on low frequency details which are then gradually complemented by adjusting high frequency, fine-level details. 

\begin{figure}
    \centering
    \setlength{\belowcaptionskip}{-8pt}
    \setlength{\tabcolsep}{1pt}
\renewcommand{\arraystretch}{0.5}
    {\small
    \begin{tabular}{c c c c c}
    
        \includegraphics[width=0.09\textwidth]{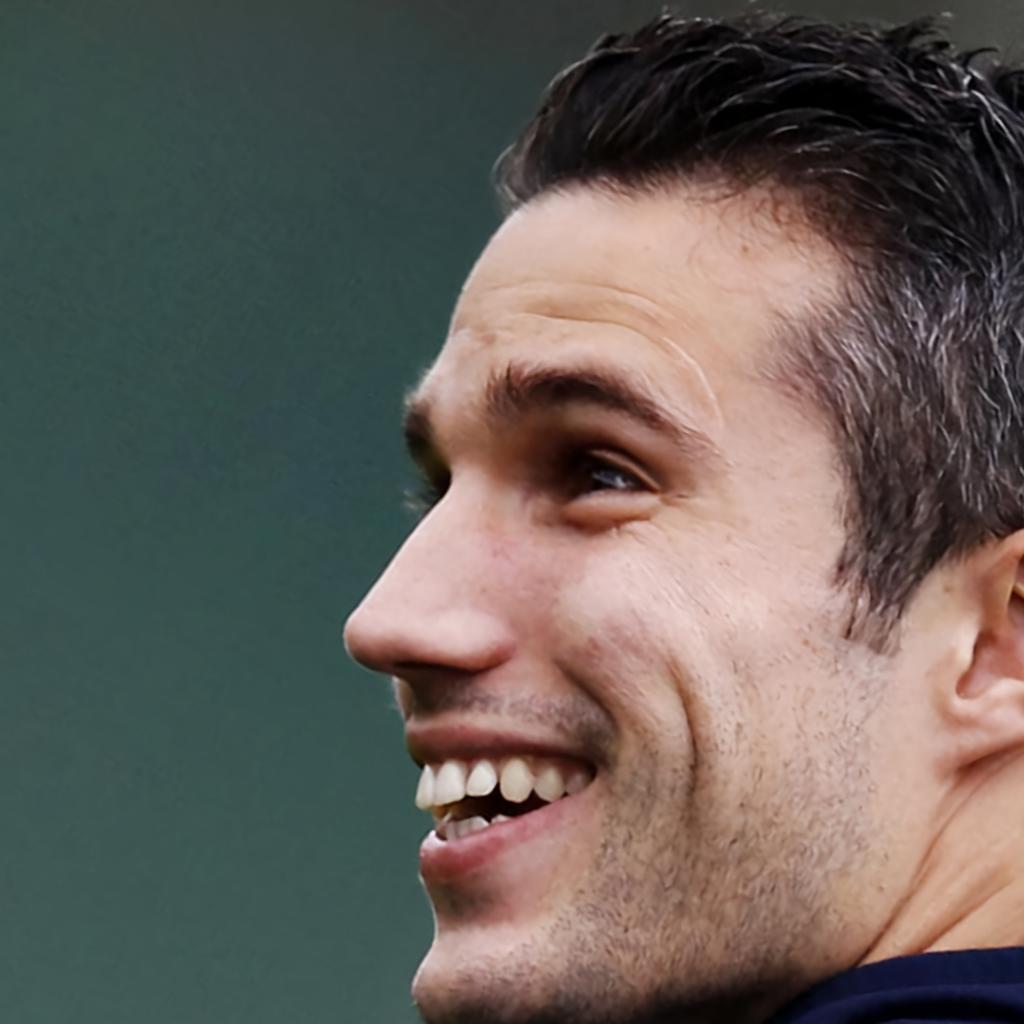} &
        \includegraphics[width=0.09\textwidth]{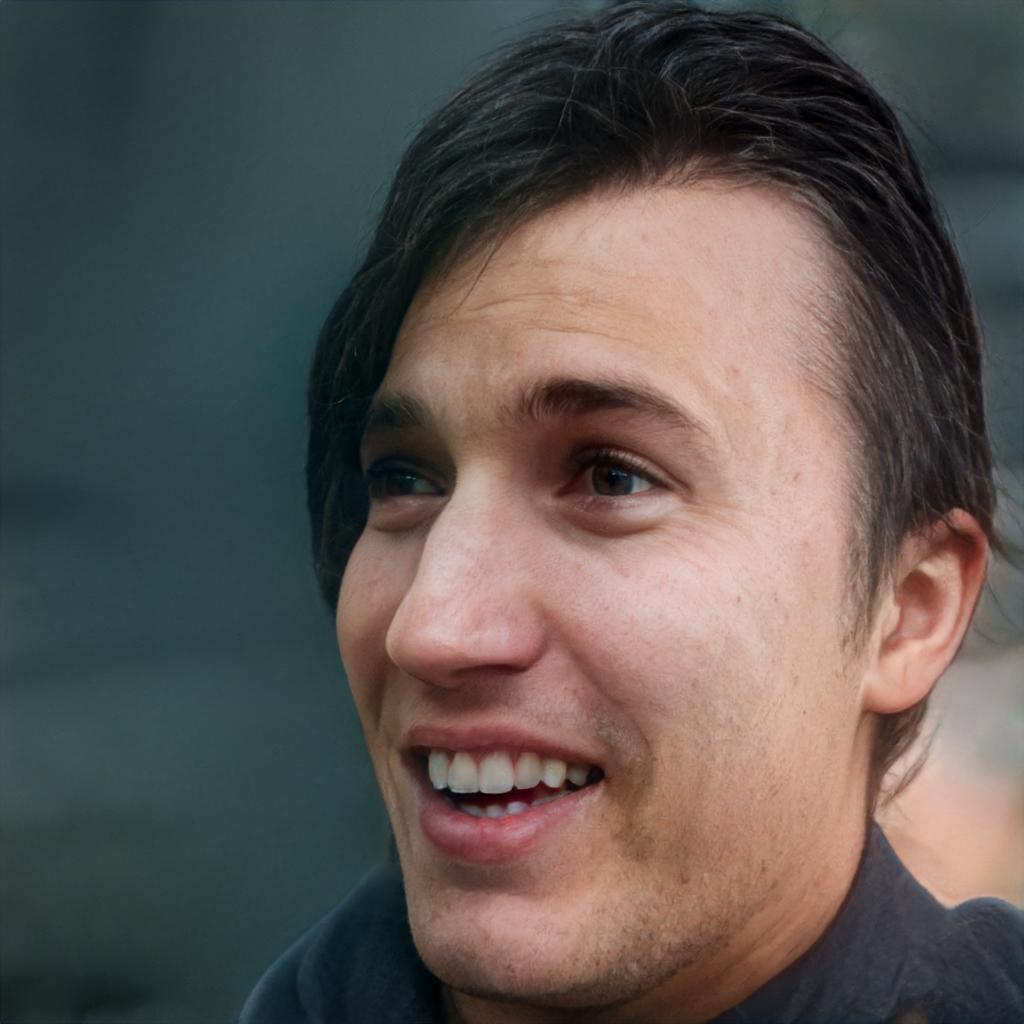} &
        \includegraphics[width=0.09\textwidth]{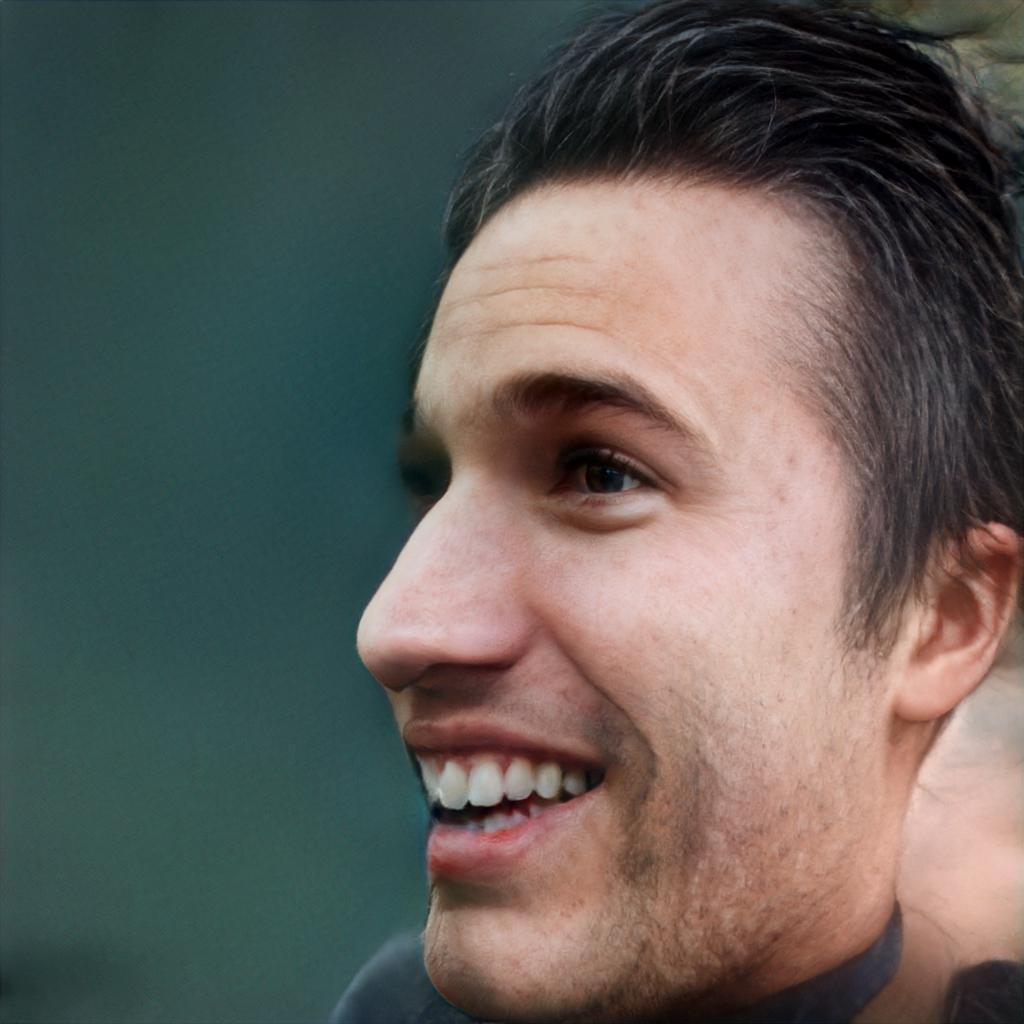} &
        \includegraphics[width=0.09\textwidth]{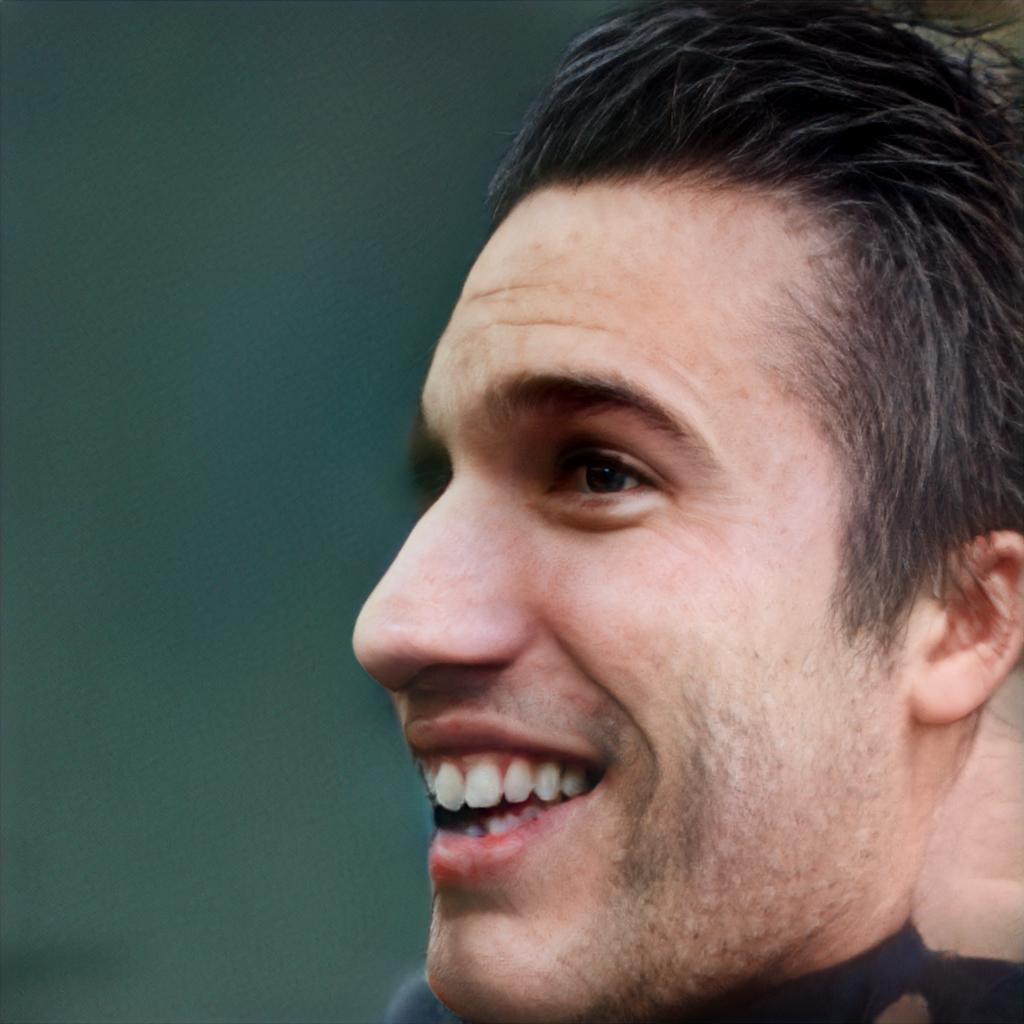} &
        \includegraphics[width=0.09\textwidth]{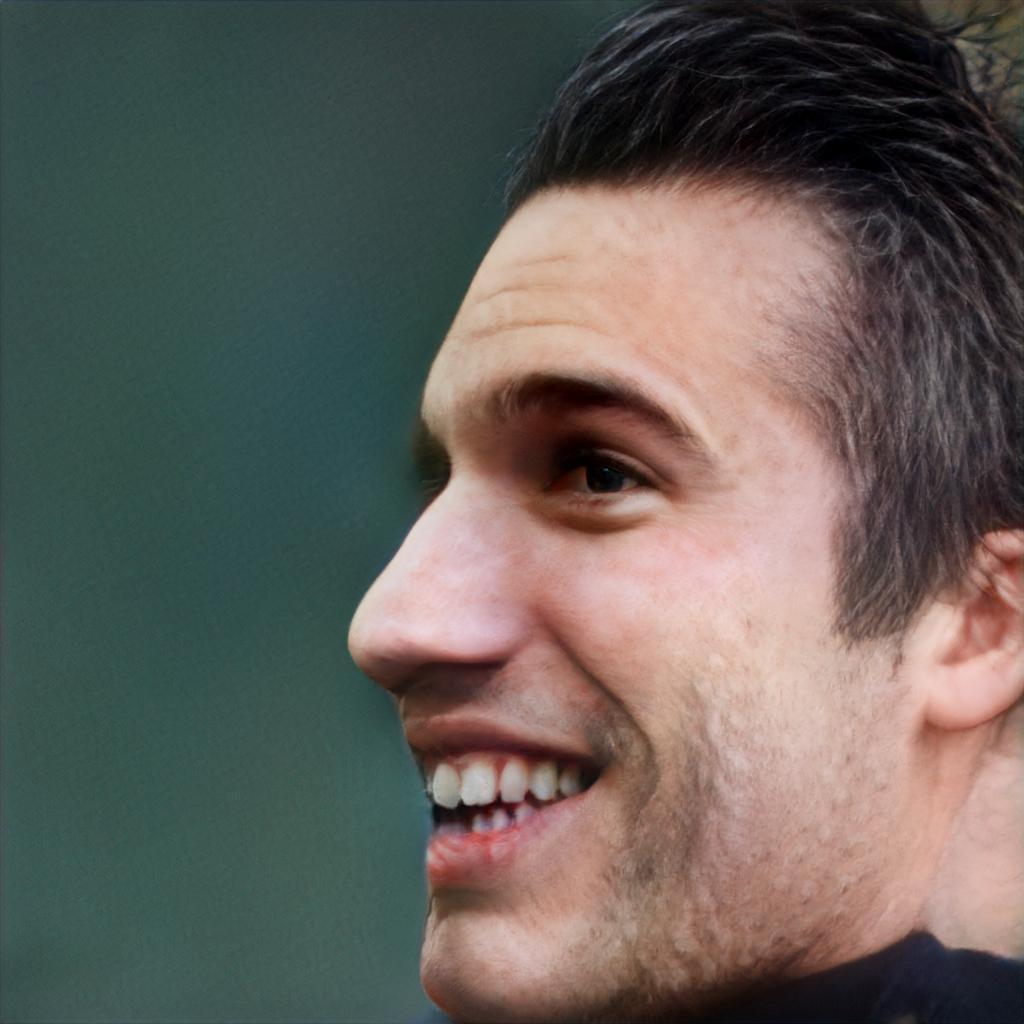}
        \tabularnewline

        \includegraphics[width=0.09\textwidth]{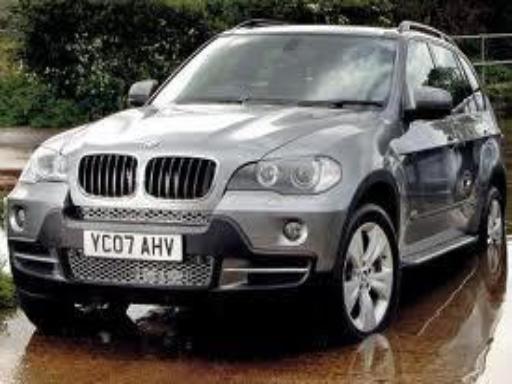} &
        \includegraphics[width=0.09\textwidth]{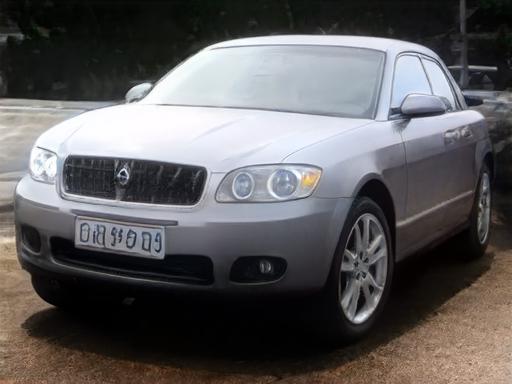} &
        \includegraphics[width=0.09\textwidth]{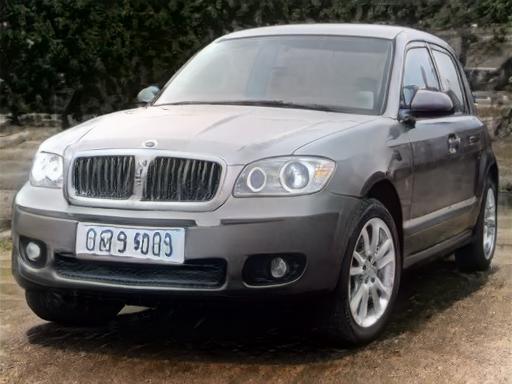} &
        \includegraphics[width=0.09\textwidth]{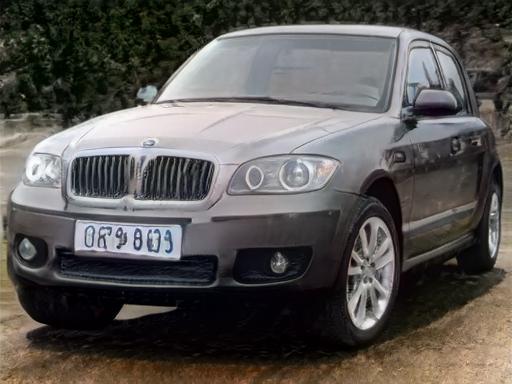} &
        \includegraphics[width=0.09\textwidth]{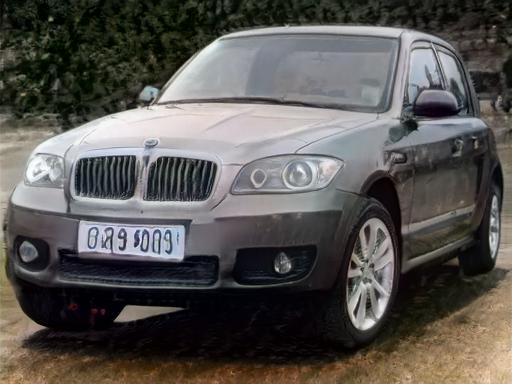} \\

        Input & \multicolumn{4}{l}{\quad $\text{ReStyle}_{pSp}$ Iterative Outputs $\longrightarrow$}\\

        \includegraphics[width=0.09\textwidth]{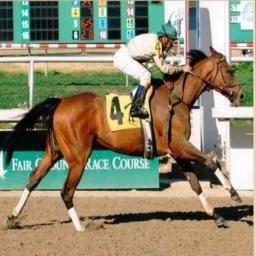} &
        \includegraphics[width=0.09\textwidth]{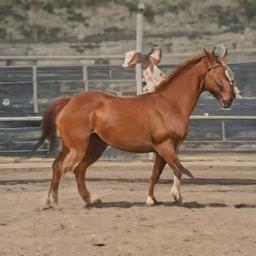} &
        \includegraphics[width=0.09\textwidth]{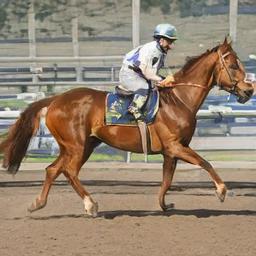} &
        \includegraphics[width=0.09\textwidth]{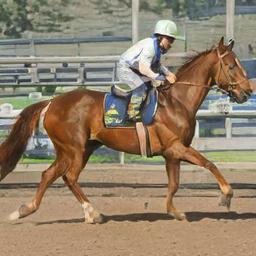} &
        \includegraphics[width=0.09\textwidth]{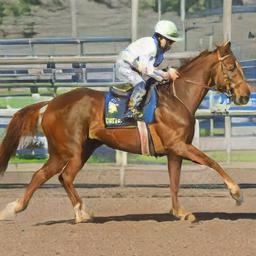}
        \tabularnewline
        
        \includegraphics[width=0.09\textwidth]{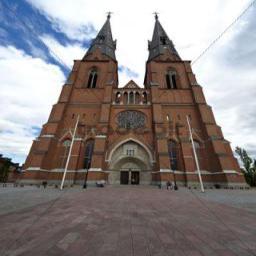} &
        \includegraphics[width=0.09\textwidth]{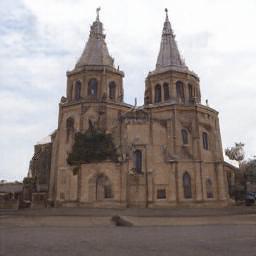} &
        \includegraphics[width=0.09\textwidth]{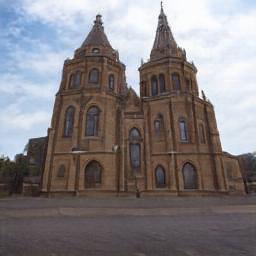} &
        \includegraphics[width=0.09\textwidth]{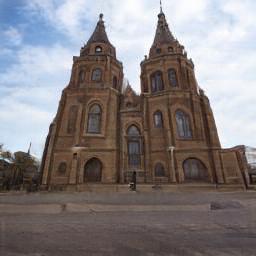} &
        \includegraphics[width=0.09\textwidth]{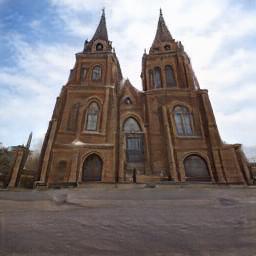} \\
        
        Input & \multicolumn{4}{l}{\quad $\text{ReStyle}_{e4e}$ Iterative Outputs $\longrightarrow$}

    \end{tabular}
    }
    \vspace{0.1cm}
    \caption{Given the input image on the left, we visualize the intermediate outputs of ReStyle applied over pSp~\cite{richardson2020encoding} and e4e~\cite{tov2021designing}.}
    \label{fig:visualize_steps}
\end{figure} 

\vspace{0.15cm}
\topic{\textit{ReStyle's iterative progress.}}
We now turn to Figure~\ref{fig:visualize_steps} and show how the reconstruction quality incrementally improves with each additional step of ReStyle. Specifically, observe how $\text{ReStyle}_{pSp}$ is able to gradually improve the reconstruction of the highly non-frontal input image in the top row. Similarly, notice how $\text{ReStyle}_{e4e}$ is able to iteratively refine the posture of the horse rider and capture the skewed structure of the church building. 

\subsection{Editability via Latent Space Manipulations}~\label{sec:editing}
Previous works \cite{zhu2020domain, zhu2020improved, tov2021designing, zhu2020improved} have discussed the importance of evaluating the editability of inversion methods. Here, we show that the editability achieved by ReStyle is comparable to that of the conventional encoders.
Since e4e is designed specifically for image manipulations, we choose to show that inversions obtained by combining e4e with ReStyle are still editable.
We show visual examples in Figure \ref{fig:editing_comparison}. Compared to e4e, ReStyle is able to better reconstruct the input while still allowing for realistic edits. Notably, observe the more plausible edits over ReStyle's inversions compared to those obtained via optimization. For example, observe the artifacts in the front bumpers of the car edits when applied over the optimization-based inversions.

\subsection{Encoder Bootstrapping}~\label{sec:bootstrapping}
Finally, we explore a new concept, which we call \textit{encoder bootstrapping.} To motivate this idea, let us consider the image toonification task in which we would like to translate real face images into their toonified, or animated, version. 
Pinkney \etal~\cite{pinkney2020resolution} propose solving this image-to-image task by projecting each real input image to its closest toon image in the latent space of a toon StyleGAN obtained via fine-tuning the FFHQ StyleGAN generator.
In a similar sense, ReStyle can be applied over pSp to solve this task. 
Here, ReStyle is initialized with the average toon latent code and its corresponding image. Then, $N$ steps are performed to translate the image to its toonified version.

\begin{figure}
    \centering
    \setlength{\belowcaptionskip}{-10pt}
    \includegraphics[width=0.425\textwidth]{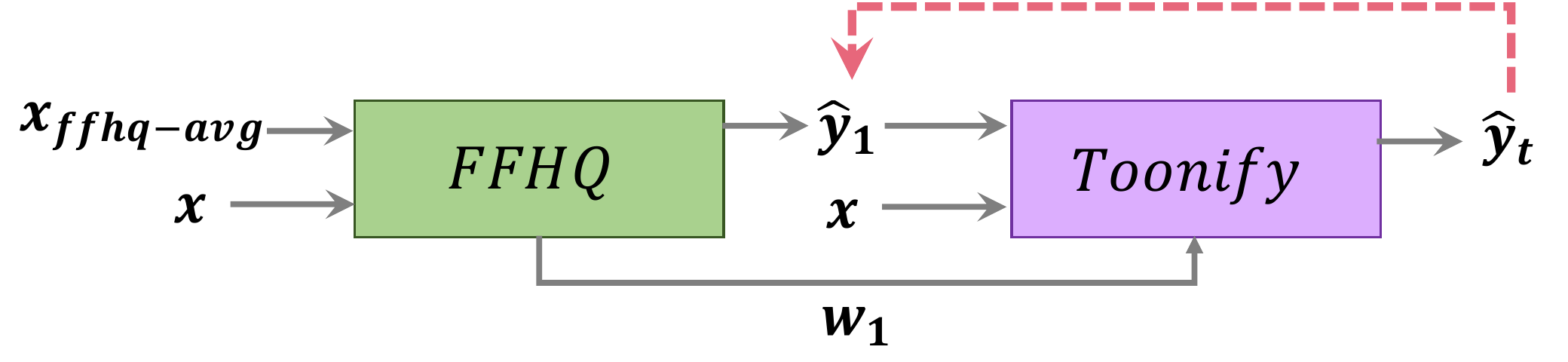}
    \vspace{0.05cm}
    \caption{\textit{Encoder bootstrapping overview.}}
    \vspace{-0.15cm}
    \label{fig:encoder_bootstrap_diagram}
\end{figure}

With encoder bootstrapping, we take a slightly different approach. Rather than initializing the iterative process using the average toon image, we first pass the given real image to an encoder tasked with embedding real images into the latent space of a StyleGAN trained on FFHQ. Doing so will result in an inverted code $\textbf{w}_1$ and reconstructed image $\hat{\textbf{y}}_1$. This inverted code and reconstructed image are then taken to initialize the toonification translation using ReStyle. This idea is illustrated in Figure~\ref{fig:encoder_bootstrap_diagram}. Notice that this technique is possible thanks to the residual nature of ReStyle. By utilizing the FFHQ encoder to obtain a better initialization, we are able to more easily learn a proper residual for translating the input image while more faithfully preserving identity.

\begin{figure}
    \centering
    \setlength{\belowcaptionskip}{-5pt}
    \setlength{\tabcolsep}{1pt}
    {\small
    \begin{tabular}{c c c c c c}
    
        \includegraphics[width=0.0925\textwidth]{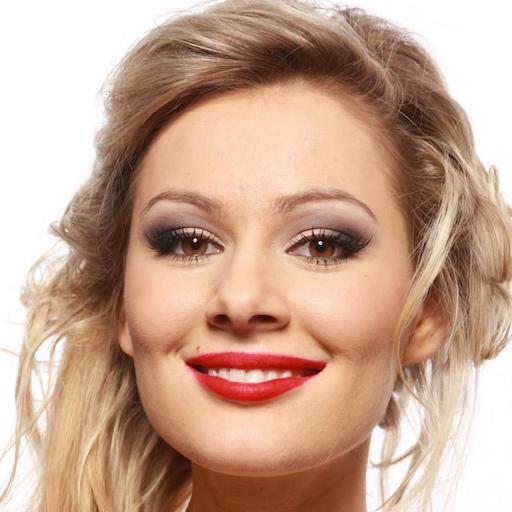} &
        \includegraphics[width=0.0925\textwidth]{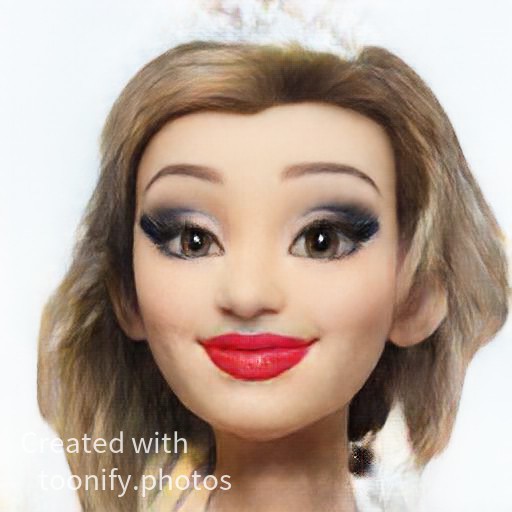} &
        \includegraphics[width=0.0925\textwidth]{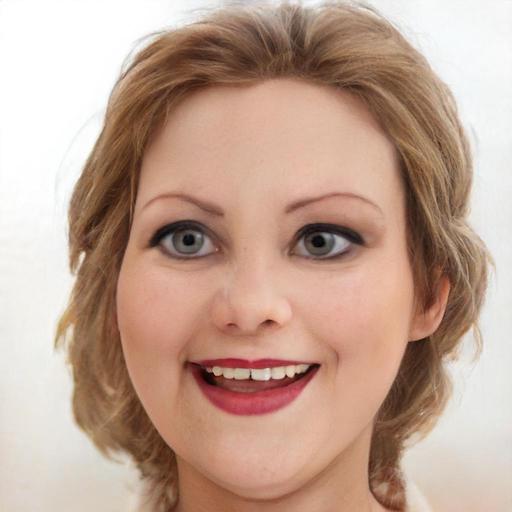} &
        \includegraphics[width=0.0925\textwidth]{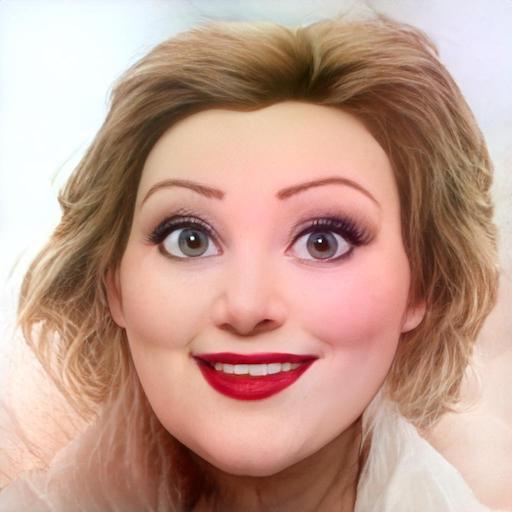}
        \tabularnewline
    
        \includegraphics[width=0.0925\textwidth]{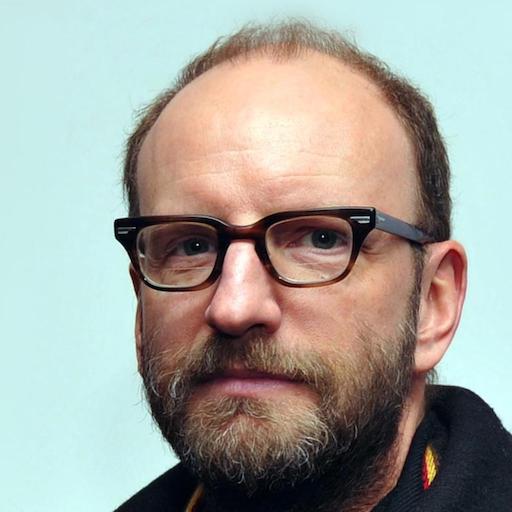} &
        \includegraphics[width=0.0925\textwidth]{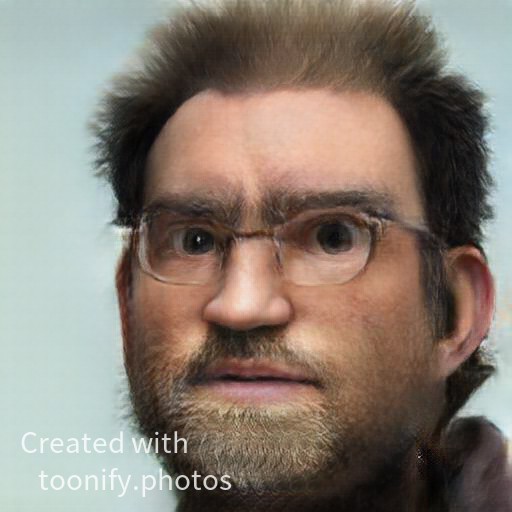} &
        \includegraphics[width=0.0925\textwidth]{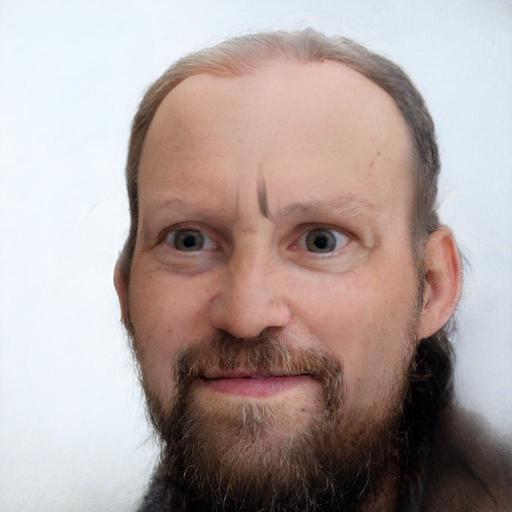} &
        \includegraphics[width=0.0925\textwidth]{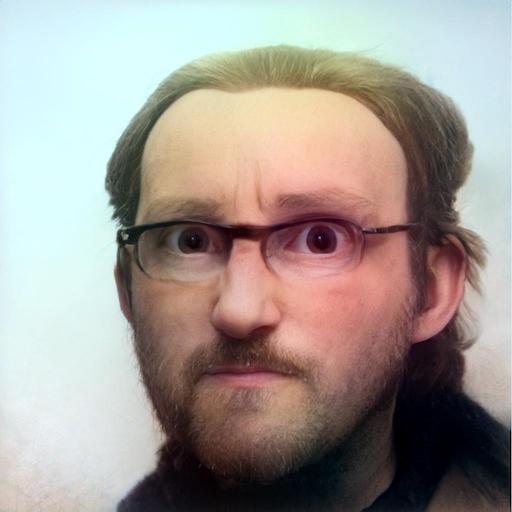}
        \tabularnewline

        \includegraphics[width=0.0925\textwidth]{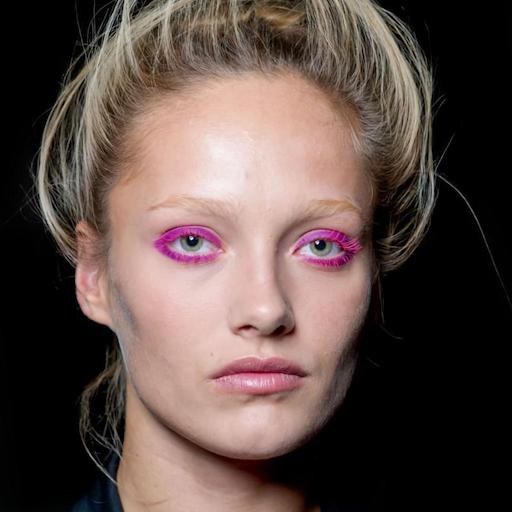} &
        \includegraphics[width=0.0925\textwidth]{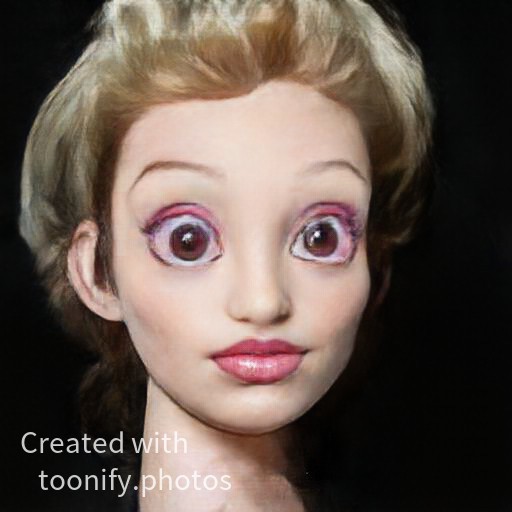} &
        \includegraphics[width=0.0925\textwidth]{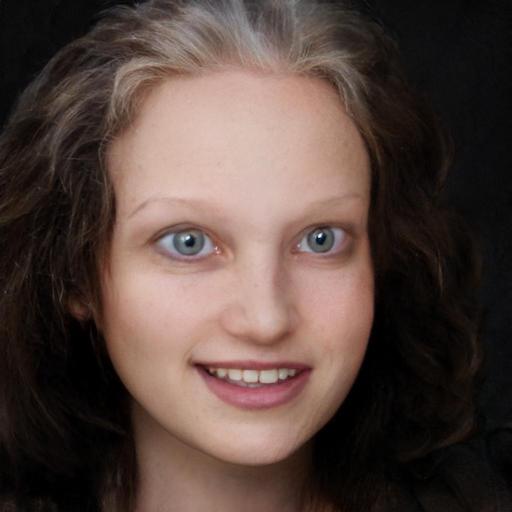} &
        \includegraphics[width=0.0925\textwidth]{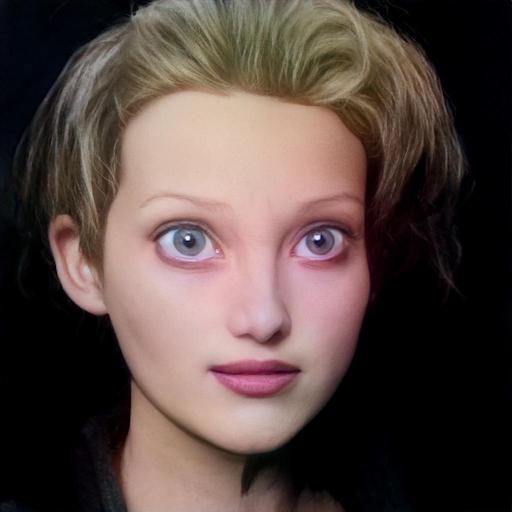}
        \tabularnewline
        
        Input & Toonify~\cite{pinkney2020resolution} & $\text{ReStyle}_{pSp}$ & $\text{ReStyle}_{pSp}^{BS}$
        
    \end{tabular}}
    \vspace{0.1cm}
    \caption{
    \textit{Toonify comparison}. Applying ReStyle with bootstrapping, denoted $\text{ReStyle}_{pSp}^{BS}$, is able to better preserve the identity characteristics of the input real image.}
    \label{fig:toonify_comparison}
\end{figure}

\begin{figure}
    \centering
    \setlength{\belowcaptionskip}{-20pt}
    \setlength{\tabcolsep}{1pt}
    {\small
    
    \begin{tabular}{c l}
    
        \includegraphics[width=0.0925\textwidth]{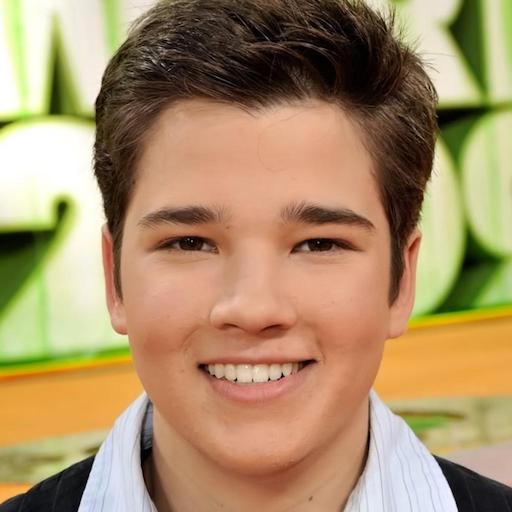} &
        \includegraphics[width=0.2775\textwidth]{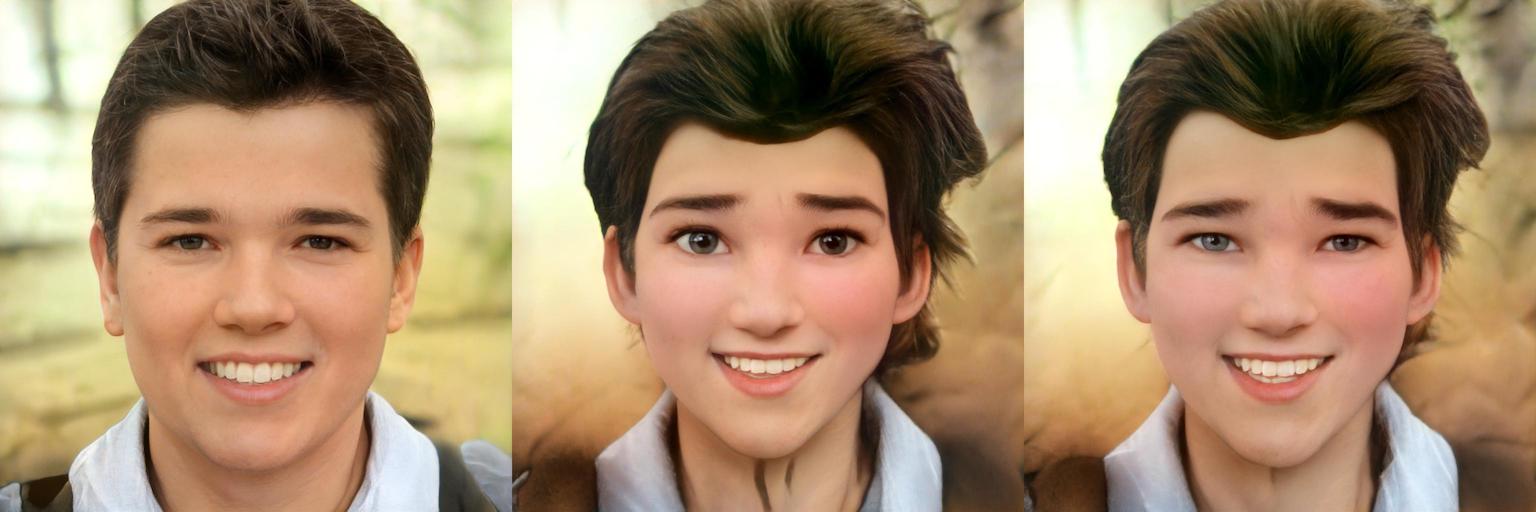}
        \tabularnewline

        \includegraphics[width=0.0925\textwidth]{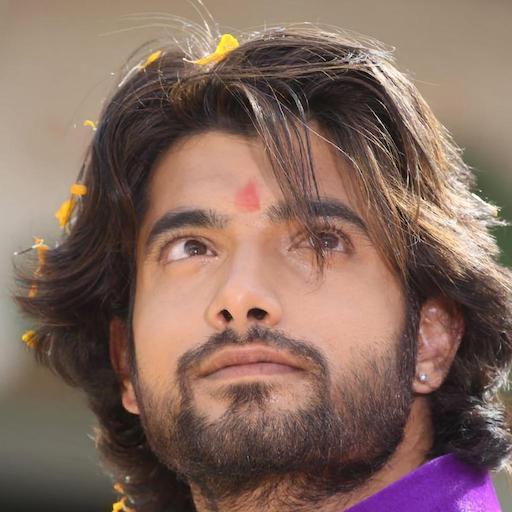} &
        \includegraphics[width=0.2775\textwidth]{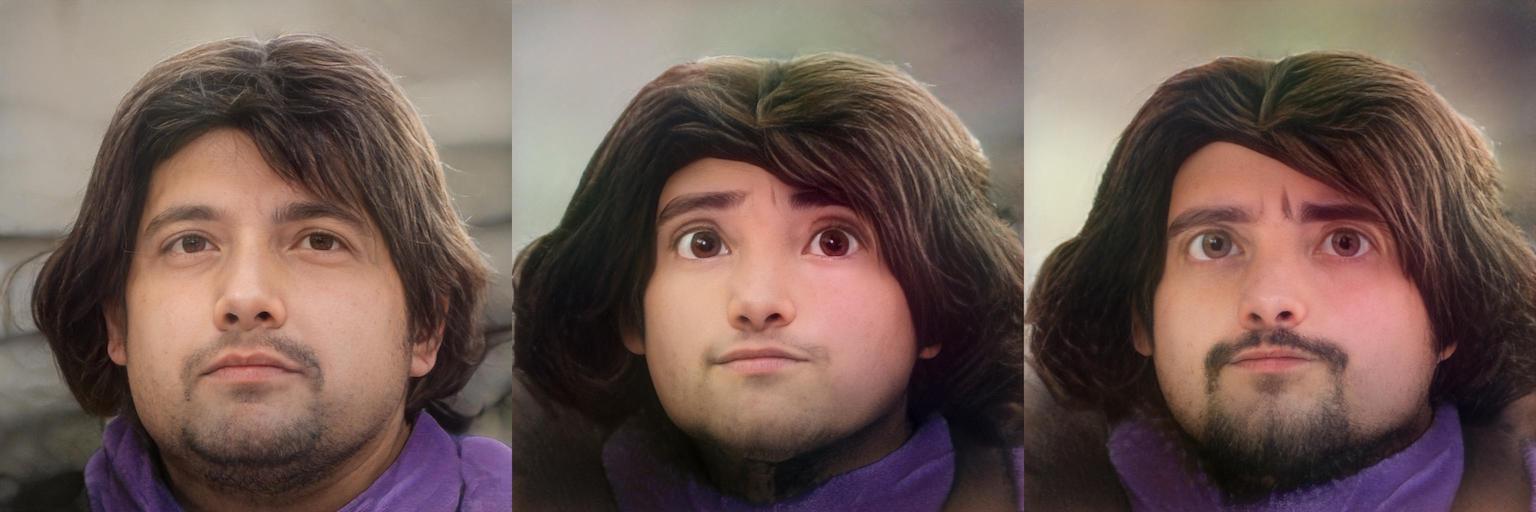}
        \tabularnewline
        
        Input & \quad Inverted \quad \quad  Iterative Outputs $\longrightarrow$ \\
        
    \end{tabular}
    }
    \vspace{0.05cm}
    \caption{For each input, we show the inverted image obtained after a single step of our $\text{ReStyle}_{pSp}$ FFHQ encoder followed by the iterative outputs of our $\text{ReStyle}_{pSp}$ toonify encoder.}
    \label{fig:toonify_mixing_steps}
\end{figure}

We compare several real-to-toon variants in Figure~\ref{fig:toonify_comparison}. Observe how bootstrapping the toonification process with the FFHQ code
results in translations able to better capture the input characteristics and toonify style. 
Observe the ability of the bootstrapped variant to better preserve make-up, eyeglasses, hairstyle, and expression. 
In Figure~\ref{fig:toonify_mixing_steps}, we visualize the inverted 
real image used to initialize the toonify encoder followed ReStyle's toonified outputs.

The bootstrapping technique is intriguing as it is not immediately clear why the code in the FFHQ latent space results in a meaningful code in the toonify space. We refer the reader to Appendix~\ref{sec:analysis_toonify} for further analysis.

\vspace{-0.5cm}
\section{Conclusions}
\vspace{-0.1cm}
In our work, we have focused on improving the inversion accuracy of encoders and presented a new scheme for training GAN encoders. Instead of predicting the inversion in one shot, we perform multiple forward passes, which more accurately and quickly converge to the target inversion. In a sense, this scheme allows the encoder to \textit{learn} how to efficiently guide its convergence to the desired inversion. 
Moreover, the encoder is trained on a larger, richer set of images consisting not only of the original dataset itself, but also the intermediate reconstructions. 
We also explored pairing the ReStyle scheme with a bootstrapping technique for the image toonification task.
We view this bootstrapping idea and the resulting transformations to be intriguing and may further open the door for additional tasks, leveraging the nature of our residual-based, iterative scheme.

\section*{Acknowledgements}
We would like to thank Elad Richardson, Omer Tov, Rinon Gal, Ron Mokady, and Yotam Nitzan for their early feedback and discussions. We would also like to thank the reviewers for their insightful remarks and constructive comments. This work was supported in part by the Israel Science Foundation under Grant No. 2366/16 and Grant No. 2492/20.

{\small
\bibliographystyle{ieee_fullname}
\bibliography{egbib}
}

\clearpage
\appendix
\appendixpage
In this appendix, we provide additional details and analysis to complement those provided in the main manuscript. Along with the additional details, we also perform an ablation study to validate our design choices and provide a large gallery of comparisons and results at full resolution obtained using the proposed ReStyle scheme.

\section{Additional Details}
\subsection{The ReStyle Encoder Architecture}~\label{sec:fpn_arch}
\begin{figure}
    \centering
    \setlength{\belowcaptionskip}{-2.5pt}
    \includegraphics[width=0.425\textwidth]{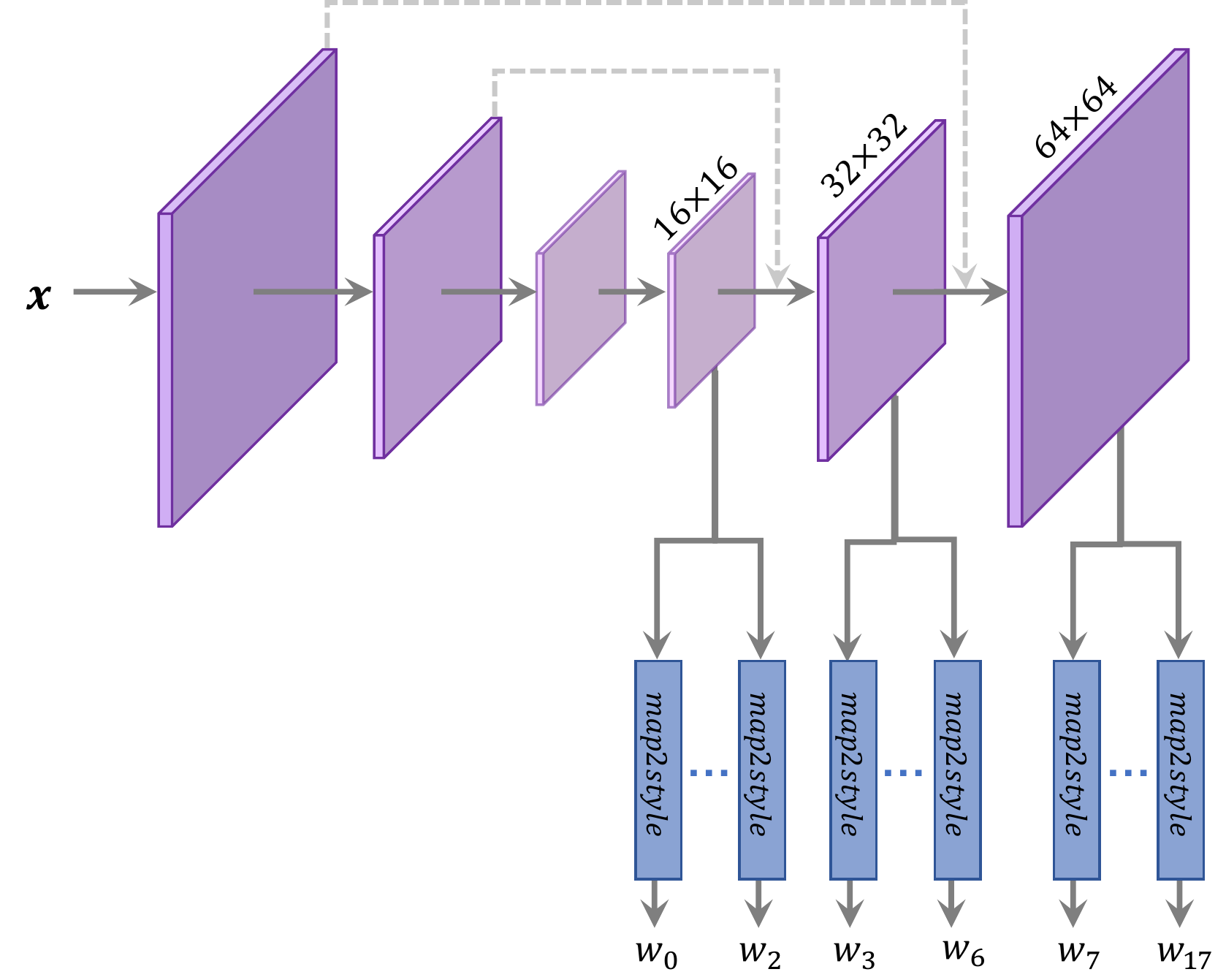}
    \vspace{0.1cm}
    \caption{Visualization of the original FPN architecture used in Richardson \etal \cite{richardson2020encoding} and Tov \etal \cite{tov2021designing}. A visualization of our simplified encoder architecture is provided in the main paper.}
    \label{fig:fpn_architecture}
\end{figure}

We begin by providing additional details regarding the ReStyle encoder architecture presented in Section~\ref{sec:encoder_arch}. Recall that our simplified architecture is derived from the architecture used in Richardson \etal.~\cite{richardson2020encoding}. There, the authors employ an FPN-based architecture for encoding real images into the StyleGAN latent space. Specifically, the encoder extracts the style input vectors using three intermediate feature maps of spatial resolutions $64\times64$ (for inputs $0-2$), $32\times32$ (for inputs $3-6$), and $16\times16$ (for inputs $7-17$). Each style vector is extracted from their corresponding feature map using a \textit{map2style} block, which is a small convolutional network containing a series of 2-strided convolutions with LeakyReLU activations. This FPN-based architecture is illustrated in Figure~\ref{fig:fpn_architecture}.

With ReStyle, we take a simpler approach. Instead of extracting the style vectors from three intermediate feature maps along the encoder, each style input is extracted from the final $16\times16$ feature map and a map2style block, as illustrated in Figure 3 in the main paper. An ablation study comparing these two architectures is provided below in Appendix~\ref{ablation_study}.

\subsection{Datasets}
Here, we provide additional details regarding the datasets used in each of the evaluated domains.

\vspace{0.25cm}
\topic{\textit{Human Faces.}}
For the human facial domain, we use all $70,000$ images from the FFHQ~\cite{karras2019style} dataset for training the ReStyle encoders. For evaluations, we use all $2,824$ test images from the CelebA-HQ~\cite{liu2015deep, karras2017progressive} dataset using the official train-test split.

\vspace{0.25cm}
\topic{\textit{Cars.}}
For the cars domain, we use the Stanford Cars~\cite{KrauseStarkDengFei-Fei_3DRR2013} dataset with training performed using the $8,144$ training images. For our evaluations, due to the large test set ($8,041$ images), we randomly select $1,000$ images from the test set with all metrics computed using the selected subset.

\vspace{0.25cm}
\topic{\textit{AFHQ Wild.}}
Here, training and evaluations are performed on $4,738$ and $500$ images, respectively, taken from the official AFHQ~\cite{choi2020stargan} Wild dataset. 

\vspace{0.25cm}
\topic{\textit{Horses.}}
From the LSUN~\cite{yu2016lsun} Horse dataset, we randomly select $10,000$ to be used for training and $2,215$ images used for evaluations. 

\vspace{0.25cm}
\topic{\textit{Churches.}}
For the churches domain we use all $126,227$ training images and $300$ testing images from the official LSUN~\cite{yu2016lsun} Church dataset.

\subsection{Baselines}~\label{sec:baselines}
In our evaluations in Sections~\ref{sec:comparison} and ~\ref{sec:editing}, we compare ReStyle with various encoder-based inversion methods. Below, we provide additional details on the baselines evaluated in the main paper.

\vspace{0.25cm}
\topic{\textit{IDInvert.}}
We compared our ReStyle approach to the IDInvert encoder from Zhu \etal ~\cite{zhu2020domain} on the human facial and cars domains. For the human facial domain, we use the official pre-trained model, which employs a StyleGAN1~\cite{karras2019style} generator. For the cars domain, we re-trained IDInvert using an input resolution of $512\times384$ and the official StyleGAN2 generator. Note, due to the long training time required by IDInvert (over three weeks on two NVIDIA P40 GPUs), we chose to train IDInvert only on the cars domain.

\vspace{0.25cm}
\topic{\textit{pSp.}}
For the human facial domain, we use the official pre-trained model from Richardson \etal ~\cite{richardson2020encoding}. For all other domains, we trained pSp using StyleGAN2 generators and default pSp hyper-parameters. During training, we also incorporated the MOCO-based similarity loss from ~\cite{tov2021designing} which was shown to improve reconstruction quality.

\vspace{0.25cm}
\topic{\textit{e4e.}}
For our comparison with Tov \etal ~\cite{tov2021designing}, we trained e4e on the AFHQ~\cite{choi2020stargan} Wild dataset using the official implementation and default e4e hyper-parameters. All other domains were evaluated using official pre-trained models.

\begin{figure}
    \centering
    \setlength{\belowcaptionskip}{-2.5pt}
    \setlength{\tabcolsep}{1pt}
    
    {\small 
    \begin{tabular}{c c l}

        \includegraphics[width=0.09\textwidth]{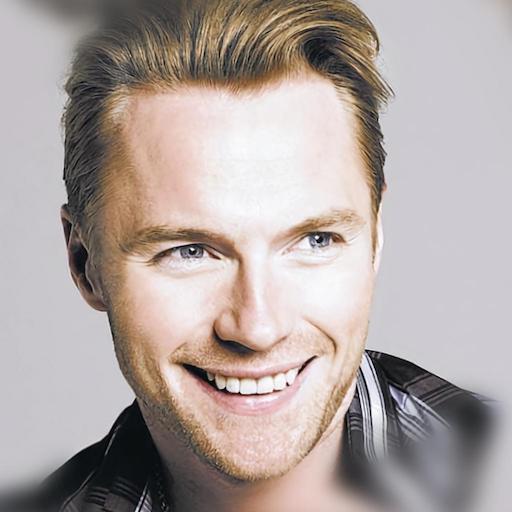} &
        \raisebox{0.20in}{\rotatebox[origin=t]{90}{Outputs}} &
        \includegraphics[width=0.36\textwidth]{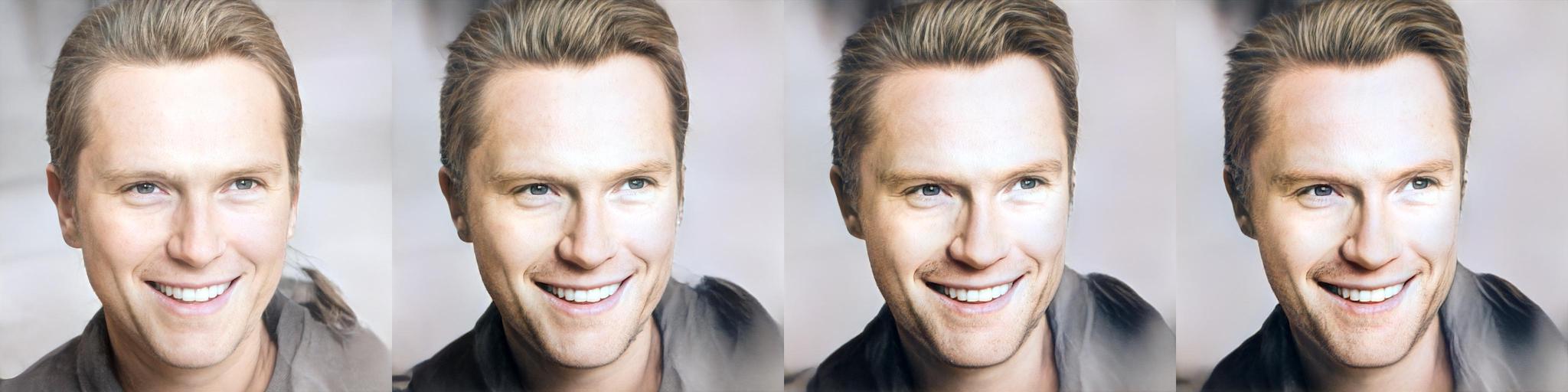}
        \tabularnewline
        
        &
        \raisebox{0.220in}{\rotatebox[origin=t]{90}{Heatmap}} &
        \includegraphics[width=0.36\textwidth]{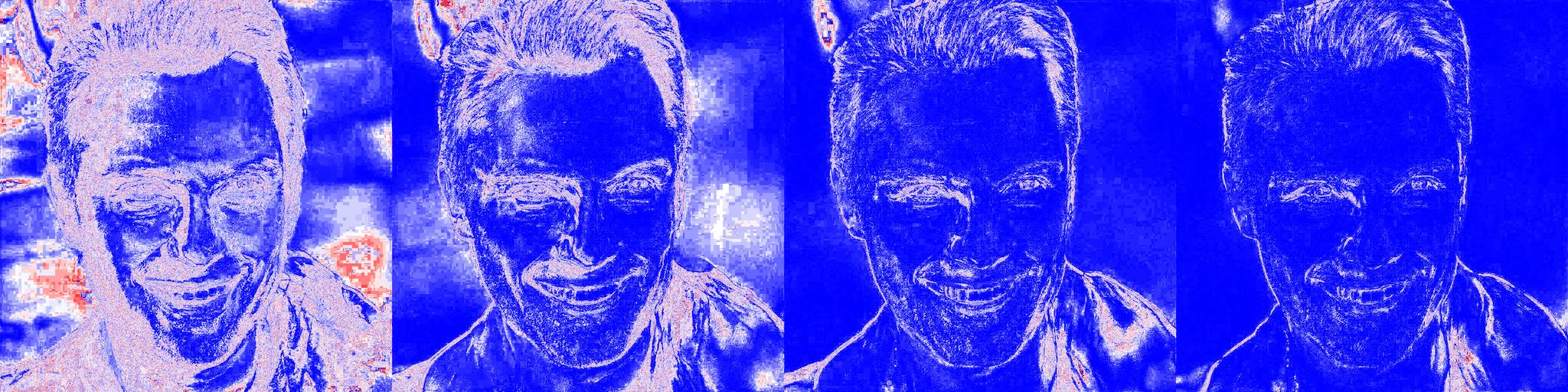}
        \tabularnewline
        
        \includegraphics[width=0.09\textwidth]{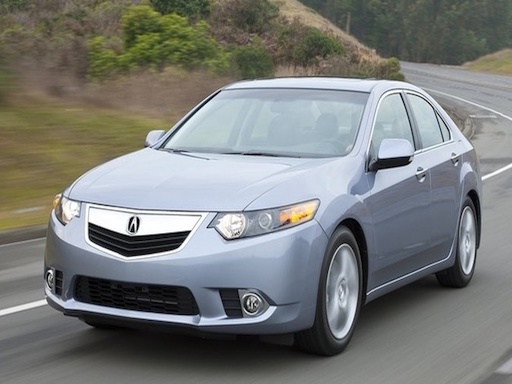} &
        \raisebox{0.15in}{\rotatebox[origin=t]{90}{Outputs}} &
        \includegraphics[width=0.36\textwidth]{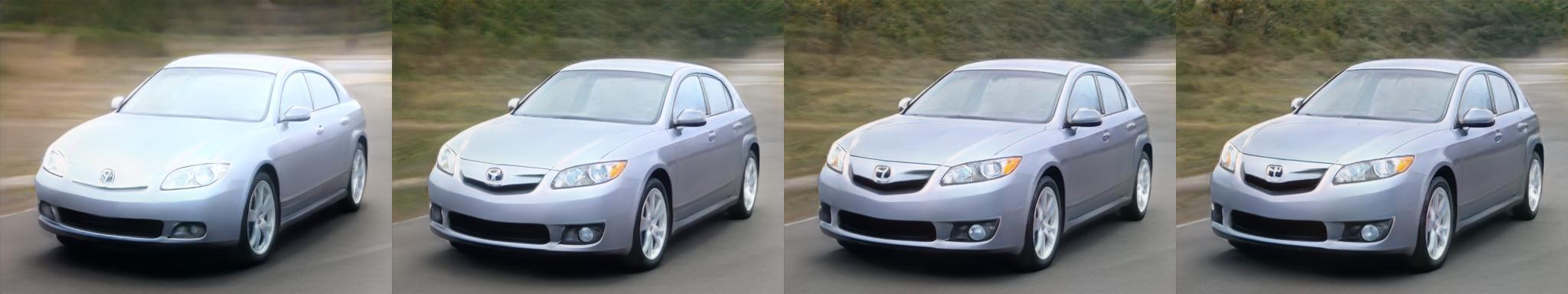}
        \tabularnewline
        
        &
        \raisebox{0.15in}{\rotatebox[origin=t]{90}{Heatmap}} &
        \includegraphics[width=0.36\textwidth]{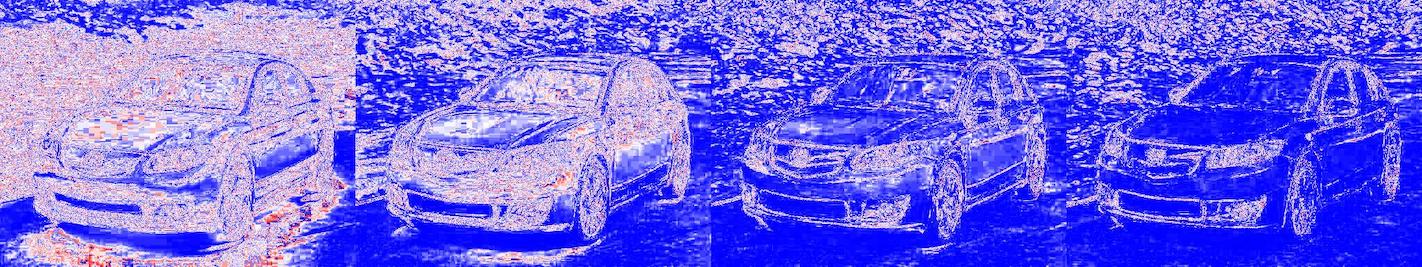}
        \tabularnewline
        
        \includegraphics[width=0.09\textwidth]{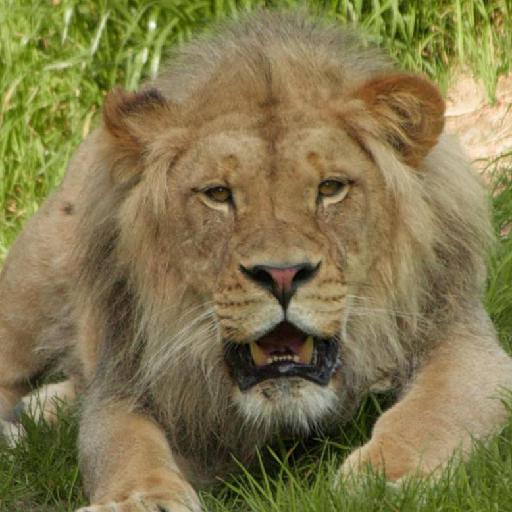} &
        \raisebox{0.220in}{\rotatebox[origin=t]{90}{Outputs}} &
        \includegraphics[width=0.36\textwidth]{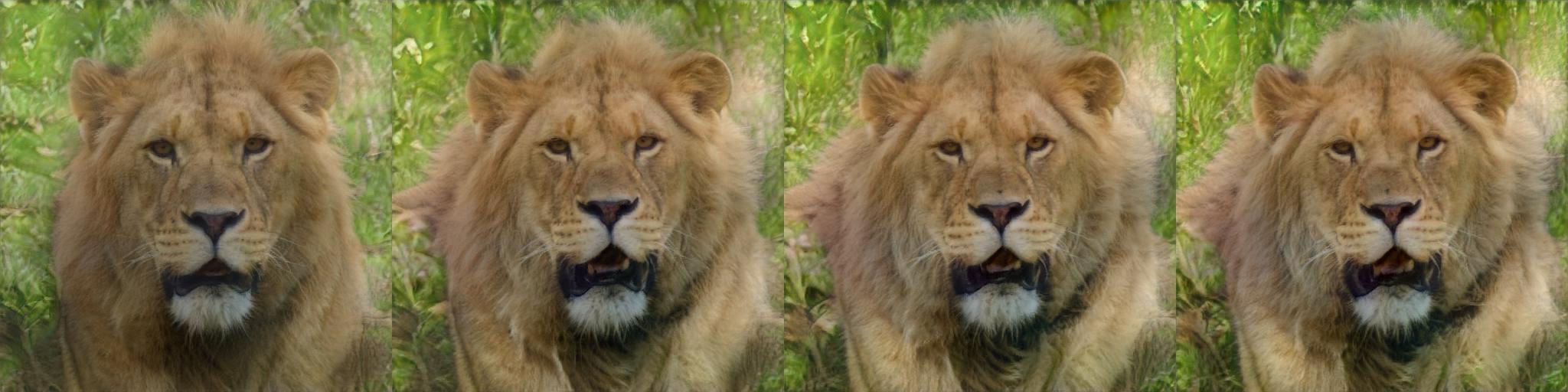}
        \tabularnewline
        
        &
        \raisebox{0.220in}{\rotatebox[origin=t]{90}{Heatmap}} &
        \includegraphics[width=0.36\textwidth]{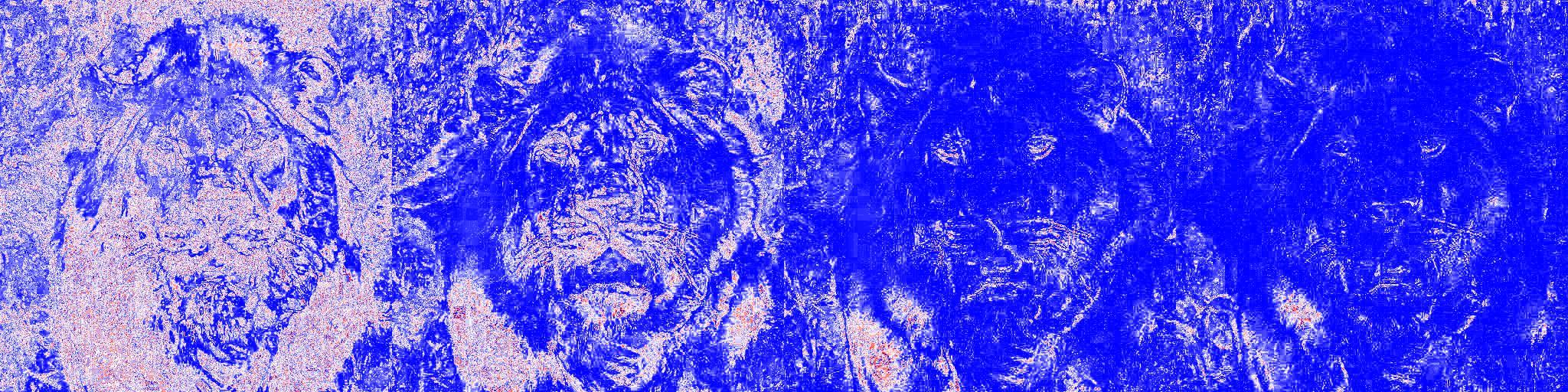}
        \tabularnewline

        Input & & {}\quad Iterative Outputs $\longrightarrow$

    \end{tabular}
    }
    \vspace{0.1cm}
    \caption{\textit{Visualizing the iterative refinement.} For each image we visualize the iterative outputs produced by ReStyle. Below each output, we show a heatmap illustrating which image regions were altered the most at the corresponding step. Note, all heatmaps are normalized globally with respect to each other. \textcolor{red}{Red} regions indicate a large change with \textcolor{blue}{blue} representing small changes in the pixel-space.}
    \label{fig:per_image_diffs}
\end{figure}

\section{ReStyle Analysis}~\label{sec:additional_analysis}
In this section, we provide additional analyses to complement those performed in the main paper (Section~\ref{sec:analysis}). 

\vspace{0.25cm}
\topic{\textit{Where's the focus? (Part II)}}
In Section~\ref{sec:analysis} (Figures 6 and 7), we showed which image regions change the most at each inference step and showed how the magnitude of change decreases over time. There, the resulting Figures were obtained by averaging over all $2,000+$ test images. To complement these Figures, we can examine this behavior while observing each image \textit{independently} rather than averaging over all images. Consider Figure~\ref{fig:per_image_diffs}. There, we illustrate the intermediate outputs of $\text{ReStyle}_{pSp}$ alongside the normalized heat-maps where \textcolor{red}{red} denotes a large pixel change and \textcolor{blue}{blue} denotes a small pixel change. As shown, in the early iterations, ReStyle focuses on adjusting global features such as the head pose or the car shape while in subsequent iterations finer details are adjusted.   

\vspace{0.25cm}
\topic{\textit{How many steps are needed?}}
The ReStyle analysis performed in the main paper and above explore ReStyle's behavior in the image space. Here, we explore the behavior of ReStyle in the latent space. Specifically, we analyze (i) which latent entries change the most across the inference steps and (ii) how many steps are needed for convergence. To do so, we focus on the cars domain and perform the following process. Consider some iteration $t$ and some latent entry $\textbf{w}_l$ where $l\in[1,k]$ and $k$ is the number of style inputs of the generator. To compute how much the values of $\textbf{w}_l$ change between iterations $t-1$ and $t$ we compute the squared difference between the latent entries averaged across all test samples. 

\begin{figure}
    \centering
    \includegraphics[width=0.5\textwidth]{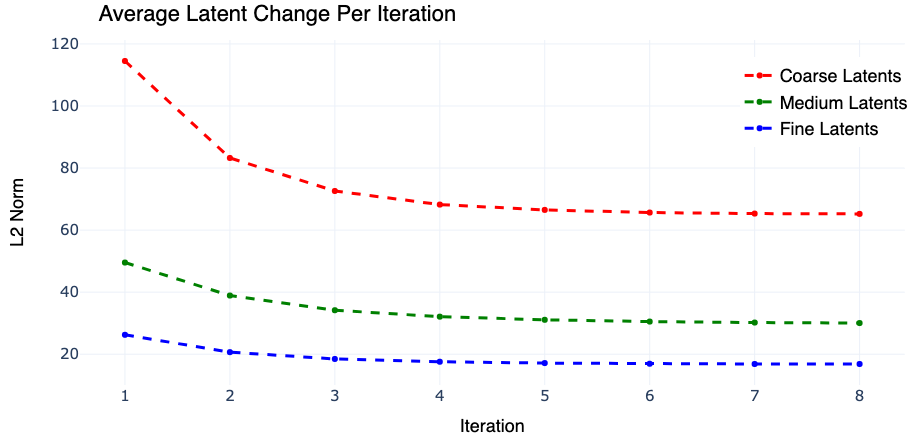}
    \caption{\textit{Which latents change the most?} We plot the average magnitude of change in each group of latent inputs per step during inference. As shown, ReStyle focuses on adjusting the coarse and medium-level inputs and converges after a few steps.}
    \label{fig:avg_latent_change}
\end{figure}

That is,
\begin{equation}
    \textbf{d}_{l,t} = \frac{1}{N} {\sum_{i=1}^N{ \left ( \textbf{w}_{l,t}^{(i)} - \textbf{w}_{l,t-1}^{(i)} \right ) ^2 }},
\end{equation}
\begin{equation}
    v_{l,t} = ||\textbf{d}_{l,t}||_2
\end{equation}
where $\textbf{w}_{l,t}^{(i)}$ is the $l$-th latent entry of the $i$-th sample obtained in iteration $t$ and $v_{l,t}$ is the $L_2$ norm of the average difference $d_{l,t}$. Having computed the average change of each of the $k$ latent entries across all test samples, we group the entries into the coarse, medium, and fine inputs as defined by StyleGAN and compute the average change of each group.

In Figure~\ref{fig:avg_latent_change}, we plot the change along each step for each of the three groups. As can be seen, the coarse input group attains the most change in the early iterations as the encoder focuses on refining the background and pose of the inversions. Conversely, the fine styles undergo the least change and converge after only a few steps, indicating that refining these aspects (e.g., the color) of the output image may be easier for the encoder. Overall, it can be seen that a few number of steps are needed for the encoder to converge to its final inversion prediction. Another interpretation of the above phenomenon is that the learned residuals decrease at every inference step, as is desired.

\begin{figure*}
    \centering
    \setlength{\belowcaptionskip}{-5.5pt}
    \begin{minipage}{.5\textwidth}
        \setlength{\tabcolsep}{1pt}
        \centering
            {\small 
            \begin{tabular}{c c l}
                \raisebox{0.175in}{\rotatebox[origin=t]{90}{$\text{pSp}_{naive}$}} &
                    \includegraphics[width=0.16\textwidth]{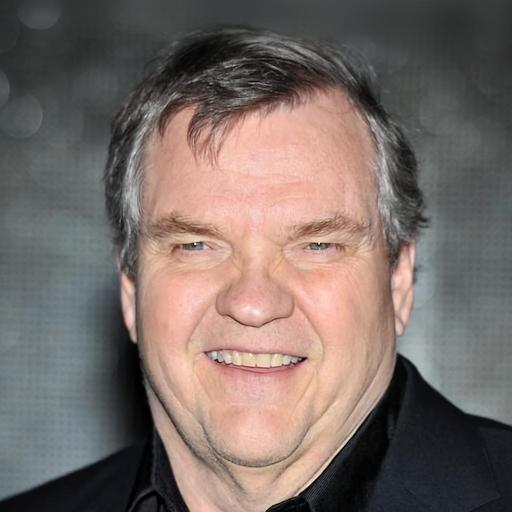} &
                    \includegraphics[width=0.80\textwidth]{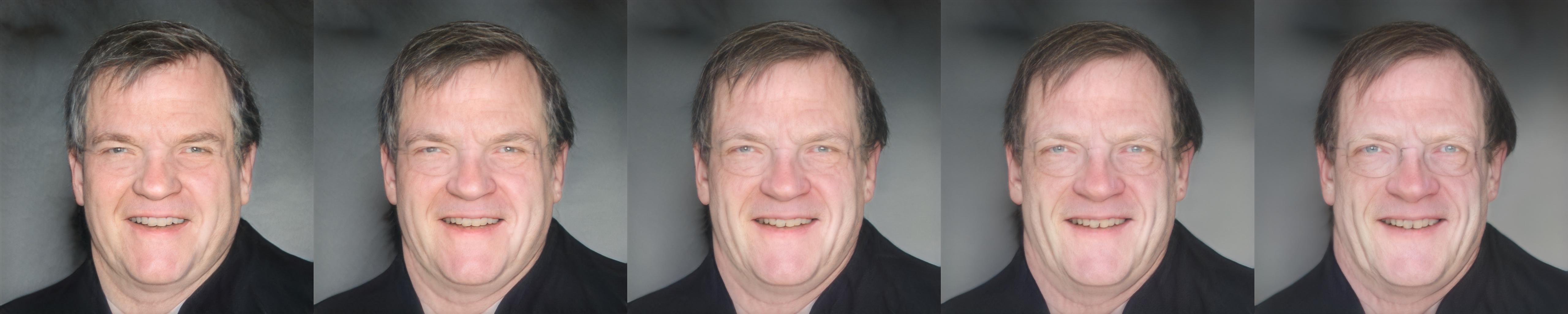} \\
                \raisebox{0.175in}{\rotatebox[origin=t]{90}{$\text{ReStyle}_{pSp}$}} & 
                    \includegraphics[width=0.16\textwidth]{images/appendix/naive_iteration/comparison/1005.jpg} &
                    \includegraphics[width=0.80\textwidth]{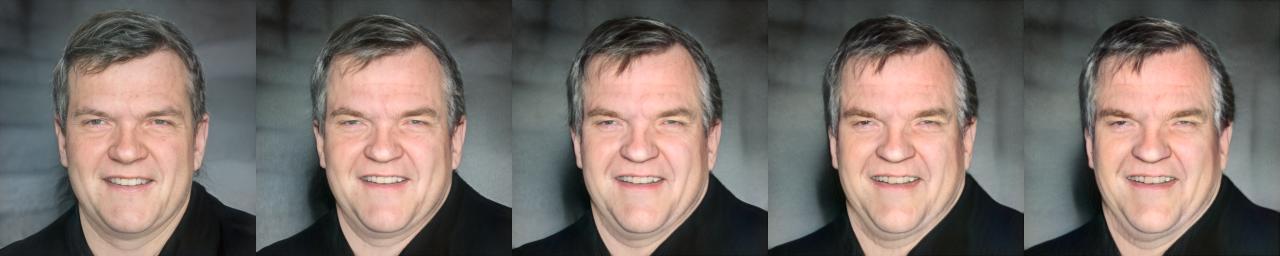} \\
            
                \raisebox{0.175in}{\rotatebox[origin=t]{90}{$\text{e4e}_{naive}$}} &
                    \includegraphics[width=0.16\textwidth]{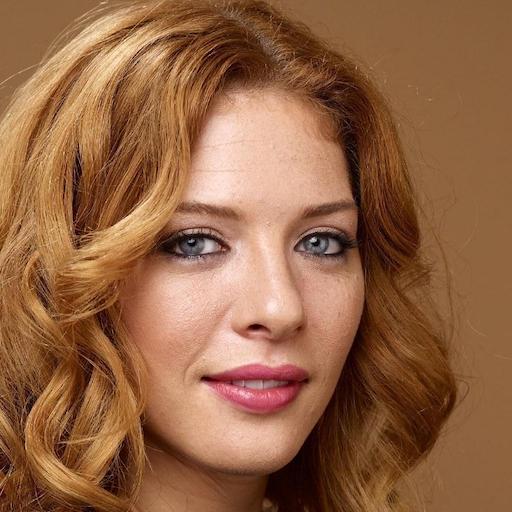} &
                    \includegraphics[width=0.80\textwidth]{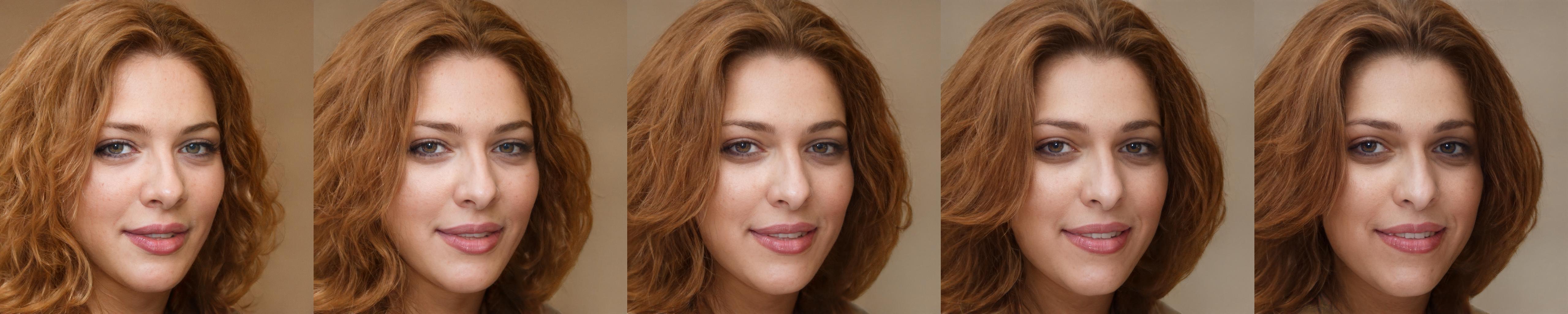} \\
                \raisebox{0.175in}{\rotatebox[origin=t]{90}{$\text{ReStyle}_{e4e}$}} & 
                    \includegraphics[width=0.16\textwidth]{images/appendix/naive_iteration/comparison/120.jpg} &
                    \includegraphics[width=0.80\textwidth]{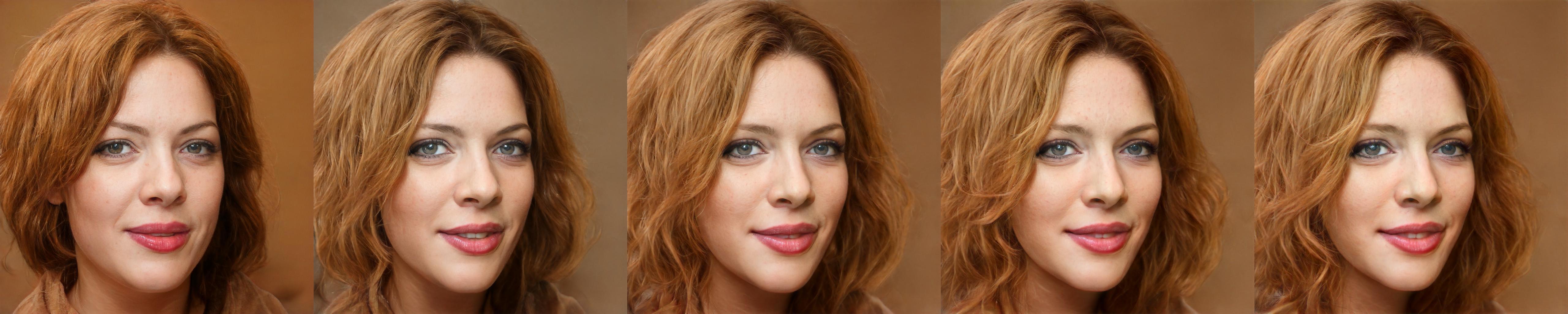} \\
                
            & Input & {}\quad Iterative Outputs $\longrightarrow$ 

            \end{tabular}
            }
        \label{fig:naive_iteration_faces}
    \end{minipage}%
    \begin{minipage}{.5\textwidth}
        \setlength{\tabcolsep}{1pt}
        \centering
            {\small
            \begin{tabular}{c c l}

                \raisebox{0.175in}{\rotatebox[origin=t]{90}{$\text{pSp}_{naive}$}} &
                    \includegraphics[height=0.16\textwidth]{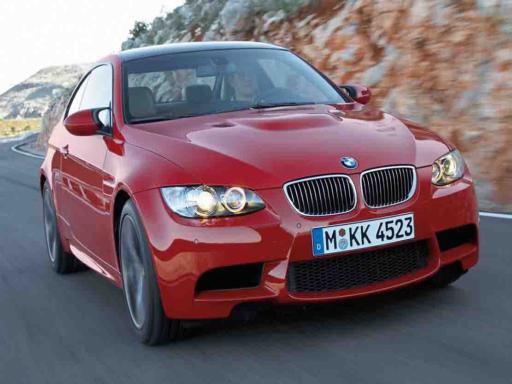} &
                    \includegraphics[width=0.64\textwidth]{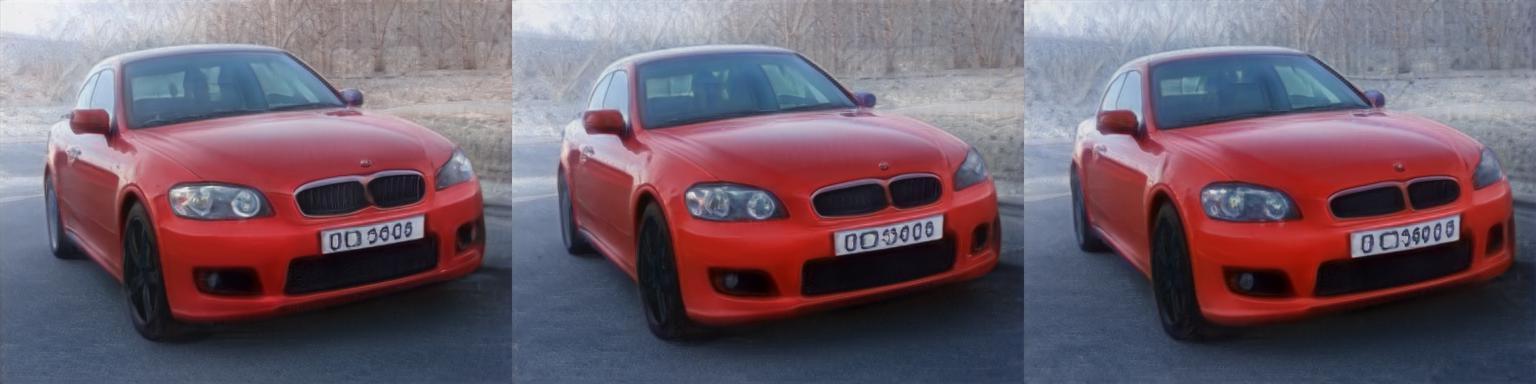} \\
                \raisebox{0.175in}{\rotatebox[origin=t]{90}{$\text{ReStyle}_{pSp}$}} & 
                    \includegraphics[height=0.16\textwidth]{images/appendix/naive_iteration/comparison/00103.jpg} &
                    \includegraphics[width=0.64\textwidth]{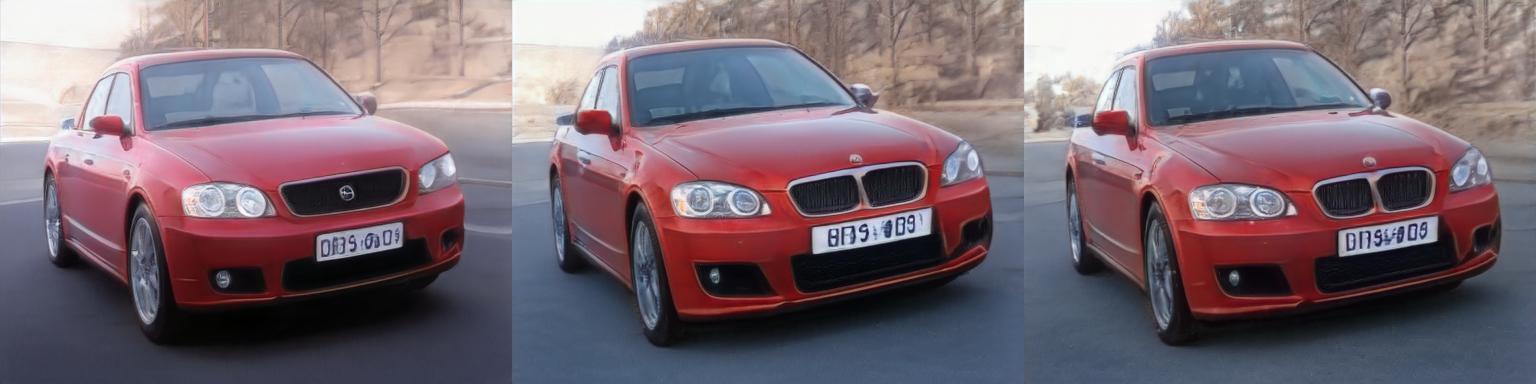} \\
           
                \raisebox{0.175in}{\rotatebox[origin=t]{90}{$\text{e4e}_{naive}$}} &
                    \includegraphics[height=0.16\textwidth]{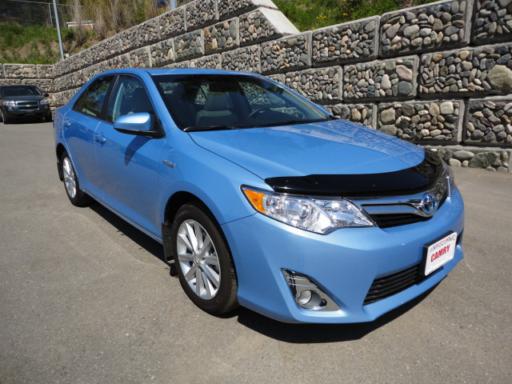} &
                    \includegraphics[width=0.64\textwidth]{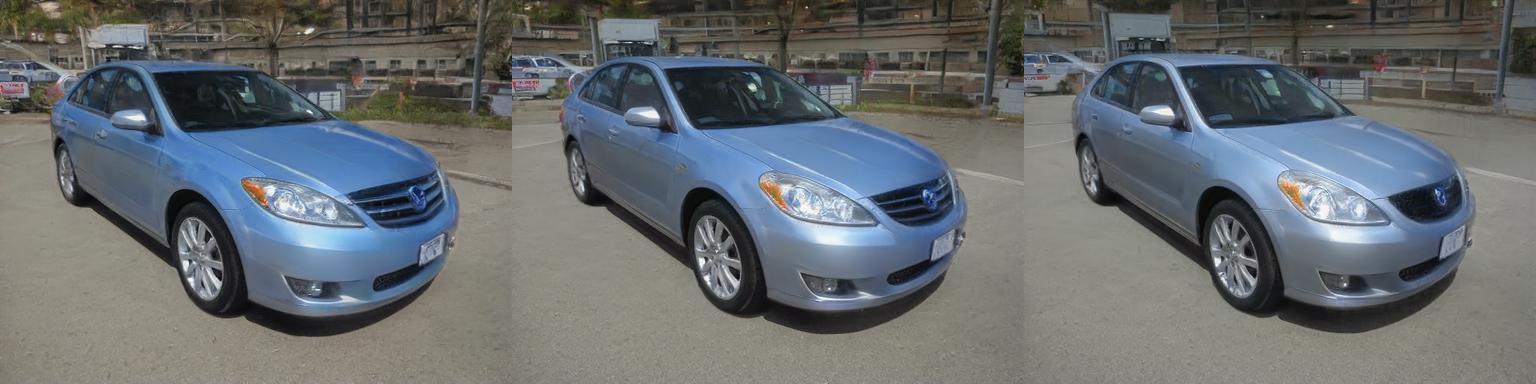} \\
                \raisebox{0.175in}{\rotatebox[origin=t]{90}{$\text{ReStyle}_{e4e}$}} & 
                    \includegraphics[height=0.16\textwidth]{images/appendix/naive_iteration/comparison/00004.jpg} &
                    \includegraphics[width=0.64\textwidth]{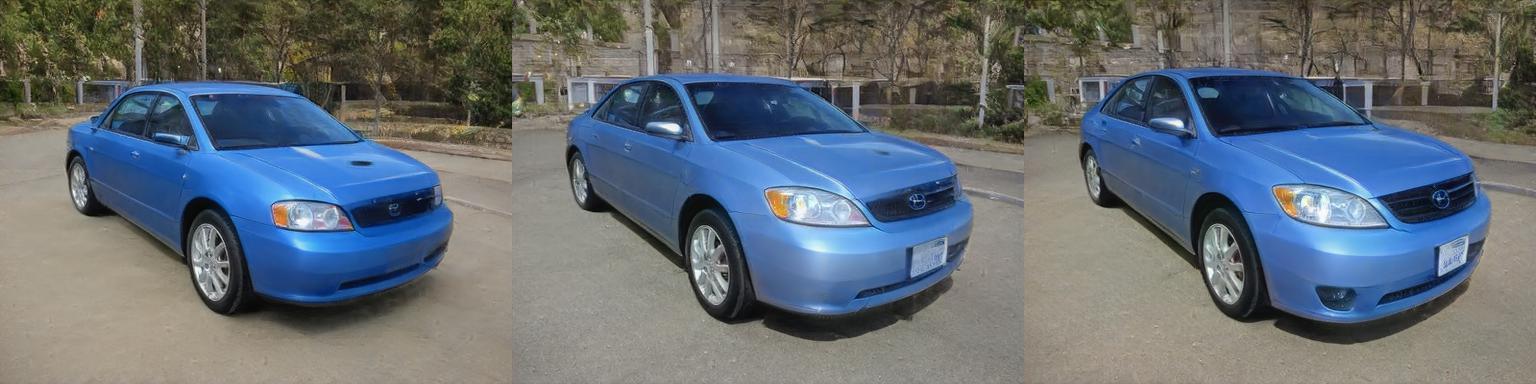} \\

            & Input & {}\quad Iterative Outputs $\longrightarrow$
                
            \end{tabular}
            }
        \label{fig:naive_iteration_cars}
    \end{minipage}
    \vspace{0.1cm}
    \caption{\textit{Ablation study.} 
    Using a pre-trained pSp or e4e encoder and simply feeding the output image multiple times leads to deteriorating reconstructions. Conversely, our dedicated ReStyle scheme incrementally improves its reconstruction outputs with each additional step.}
    \label{fig:naive_iteration}
\end{figure*}

\section{Ablation Study}~\label{ablation_study}
In this section, we validate our design choices for our ReStyle training scheme and encoder architecture.

\vspace{0.1cm}
\topic{\textit{The Iterative Training Scheme.}}
We begin by showing that a dedicated iterative training scheme is truly needed. A natural first attempt for creating an iterative inversion scheme is simply using a pre-trained conventional encoder and passing the output image back as input multiple times. 
Notice that there are key differences between the above formulation and the ReStyle formulation. First, in the ReStyle scheme, we pass both the input and current output to the encoder. Second, a ReStyle encoder is trained explicitly to output a \textit{residual} with respect to the previous latent at each step.

In Figure~\ref{fig:naive_iteration} we provide a comparison on the human facial and cars domains using both a pSp encoder and e4e encoder trained using the ReStyle formulation and naive formulation presented above. As can be seen, at each additional step, the naive formulation moves away from the reconstruction of the input image. This shows that our dedicated iterative, residual-based training scheme is indeed needed over readily-available encoders.

\begin{table}
    \setlength{\tabcolsep}{2.5pt}
    \centering
    \begin{tabular}{c l c c c}
    \toprule
    Domain & Method & $\downarrow$ LPIPS & $\downarrow$ MSE & $\downarrow$ Runtime \\
    \midrule
    \multirow{2}{*}{\begin{tabular}{@{}c@{}}Faces \\ ($1024$)\end{tabular}}
    & $\text{ReStyle}_{fpn}$ & 0.03 & 0.14 & 0.538 \\
    & $\text{ReStyle}_{simple}$ & 0.03 & \textbf{0.13} & \textbf{0.451} \\
    \midrule 
    \multirow{2}{*}{\begin{tabular}{@{}c@{}}Cars \\ ($512$)\end{tabular}}
    & $\text{ReStyle}_{fpn}$ & 0.08 & 0.26 & 0.411 \\
    & $\text{ReStyle}_{simple}$ & \textbf{0.07} & \textbf{0.25} & \textbf{0.361} \\
    \midrule 
    \multirow{2}{*}{\begin{tabular}{@{}c@{}}Wild \\ ($512$)\end{tabular}}
    & $\text{ReStyle}_{fpn}$ & 0.06 & 0.23 & 0.413 \\
    & $\text{ReStyle}_{simple}$ & 0.06 & \textbf{0.21} &  \textbf{0.363} \\
    \midrule 
    \multirow{2}{*}{\begin{tabular}{@{}c@{}}Churches \\ ($256$)\end{tabular}}
    & $\text{ReStyle}_{fpn}$ & 0.09 & 0.28 & 0.355 \\
    & $\text{ReStyle}_{simple}$ & 0.09 & \textbf{0.26} & \textbf{0.298} \\
    \midrule 
    \multirow{2}{*}{\begin{tabular}{@{}c@{}}Horses \\ ($256$)\end{tabular}}
    & $\text{ReStyle}_{fpn}$  & 0.09 & 0.32 & 0.355 \\
    & $\text{ReStyle}_{simple}$ & 0.09 & \textbf{0.31} & \textbf{0.298} \\
    \midrule 
    \end{tabular}
    \caption{Ablation study on applying ReStyle to pSp using our simpler encoder architecture variants compared to the original FPN-based architecture. All results are computed on the final reconstructions after $5$ steps. As shown, our simplified architecture is comparable to the FPN-variant with a reduced inference time.}
    \vspace{-0.75em}
\label{tb:ablation_fpn}
\end{table}

\vspace{0.1cm}
\topic{\textit{The ReStyle Encoder Architecture.}}
Here, we show that our simplified encoder architectural design choice leads to a negligible difference in reconstruction quality while reducing the inference time compared to the FPN-based architecture from Richardson \etal ~\cite{richardson2020encoding}. Table~\ref{tb:ablation_fpn} summarizes the reconstruction results across $5$ domains. For each domain and encoder, we perform $5$ steps during inference and display quantitative results computed on the final output. As shown, the two variants attain nearly identical reconstruction quality in terms of both $L_2$ and LPIPS similarity. However, our simplified architecture is more than $10\%$ faster.

\section{Analyzing the Toonify Latent Space}~\label{sec:analysis_toonify}
In Section~\ref{sec:bootstrapping}, we explored a new \textit{encoder bootstrapping} technique for performing image toonification~\cite{pinkney2020resolution}. Rather than initializing ReStyle using the average latent code and corresponding toon image, the iterative process is initialized by first inverting the real image into the FFHQ StyleGAN latent space. The inverted code and reconstructed image are then used to initialize the ReStyle toonify encoder. Thanks to the improved initialization from the initial inversion, the toonify encoder is able to learn a more faithful translation of the real image into its corresponding toon image. However, it is not trivial to assume that the real inverted code in the FFHQ latent space corresponds to the same identity characteristics in the toonify latent space. 

\begin{figure}
    \centering
    \setlength{\belowcaptionskip}{-2.5pt}
    \setlength{\tabcolsep}{1pt}
    \begin{tabular}{c c c c c}

        \raisebox{0.275in}{\rotatebox[origin=t]{90}{FFHQ}} & 
        \includegraphics[width=0.11\textwidth]{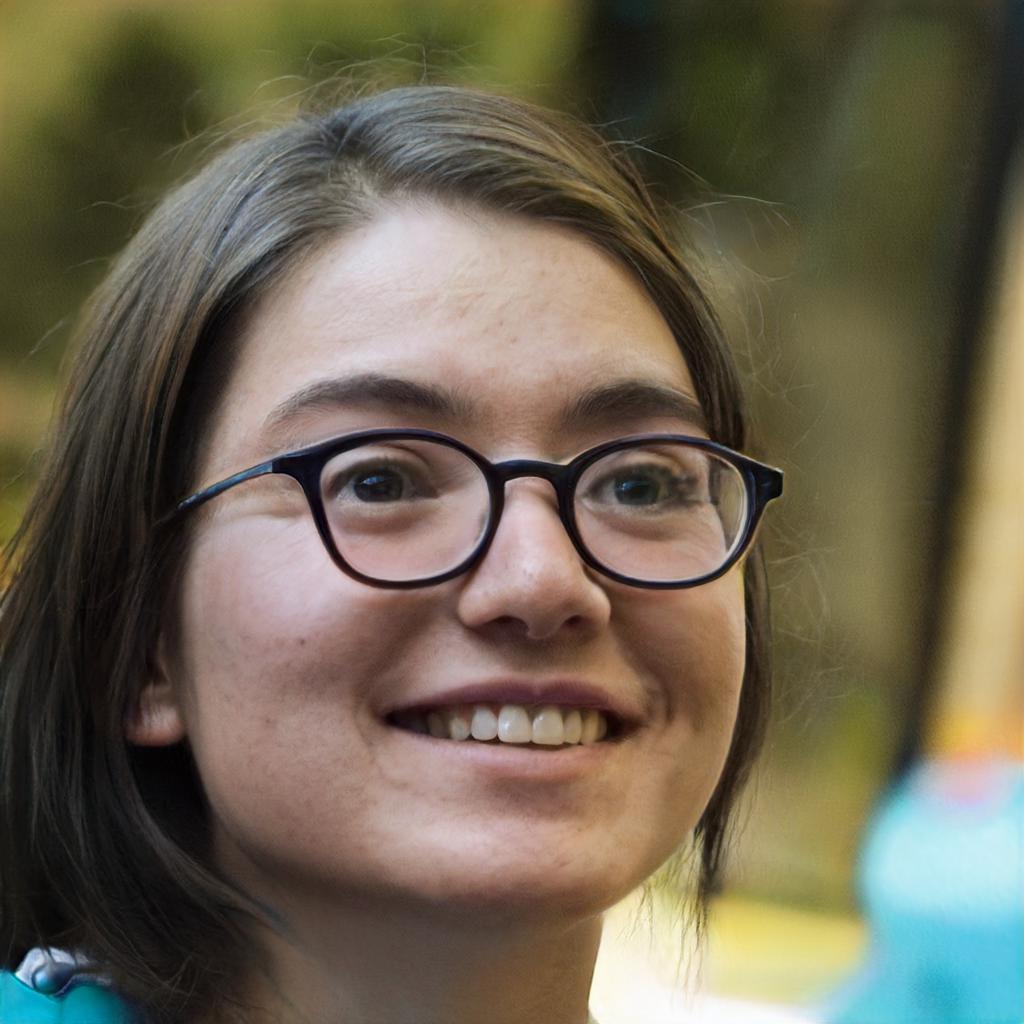} &
        \includegraphics[width=0.11\textwidth]{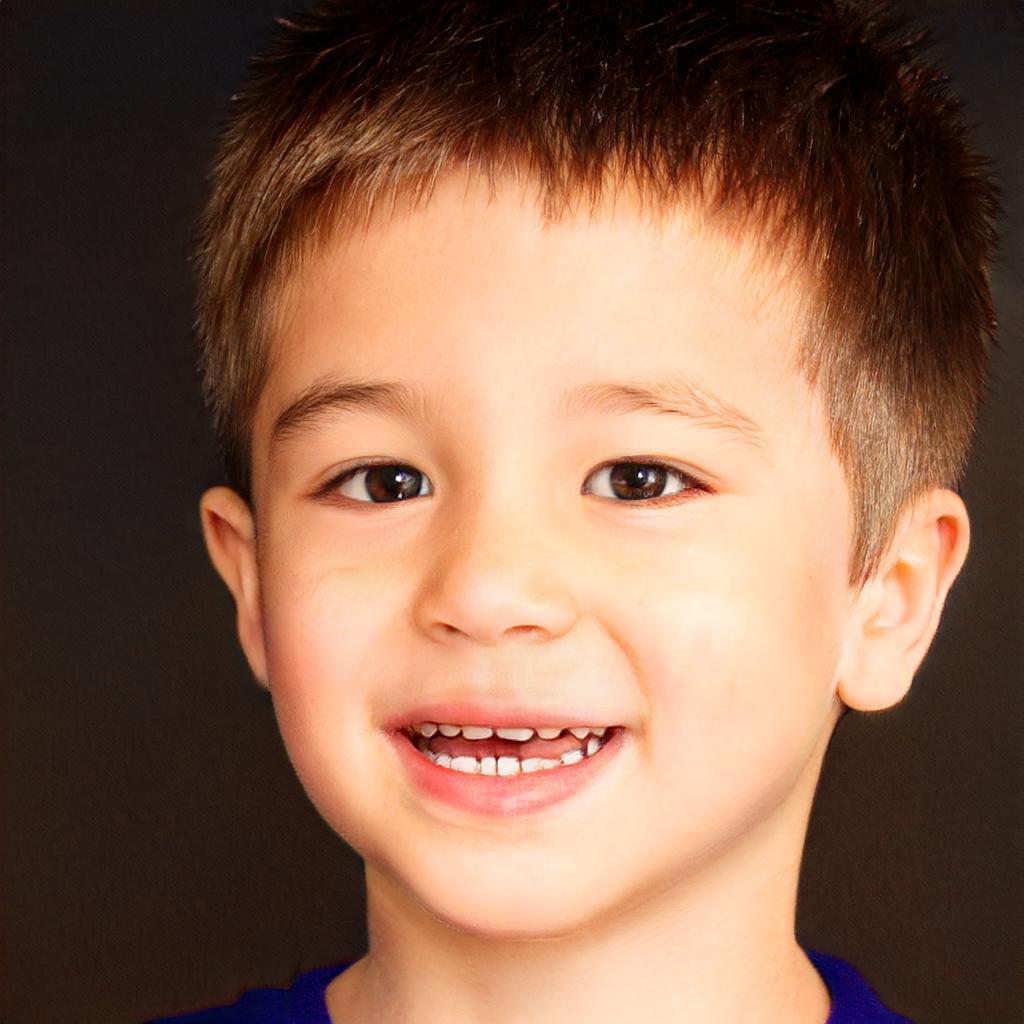} &
        \includegraphics[width=0.11\textwidth]{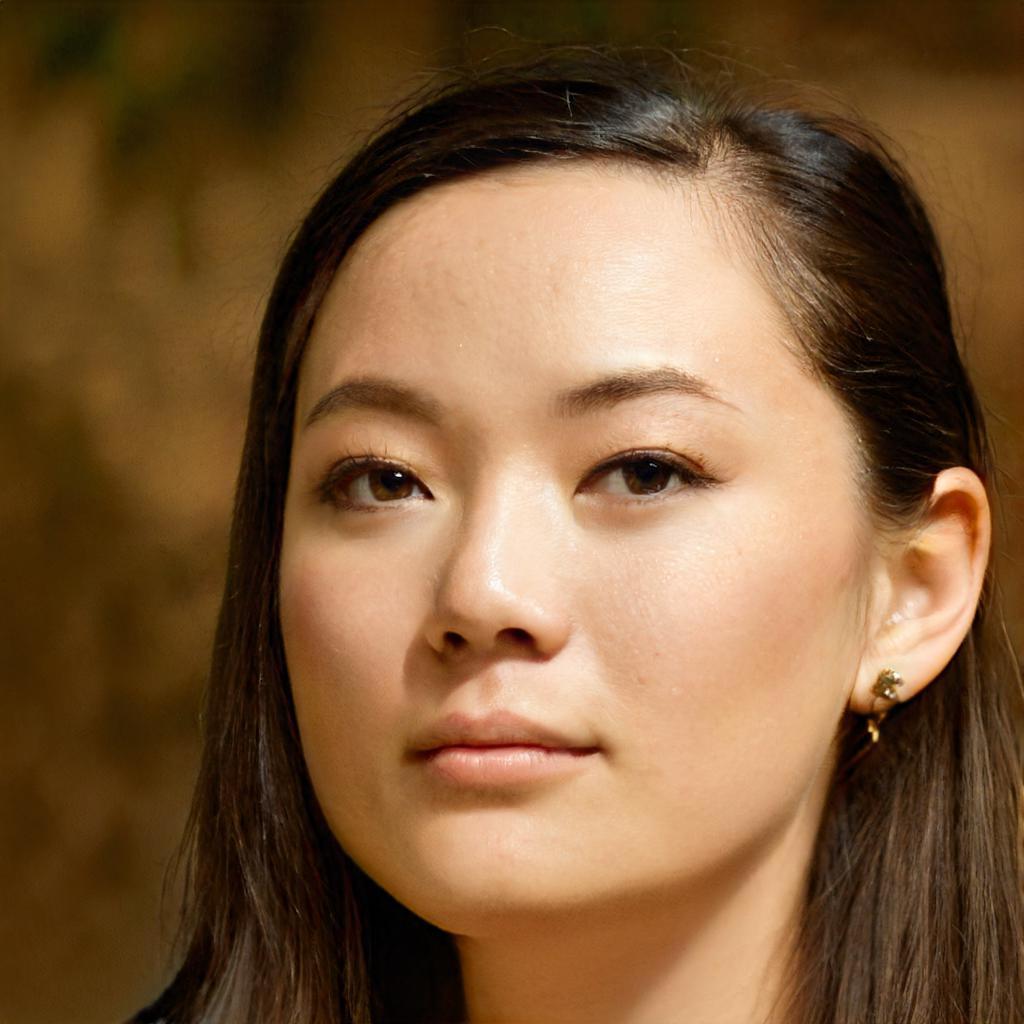} &
        \includegraphics[width=0.11\textwidth]{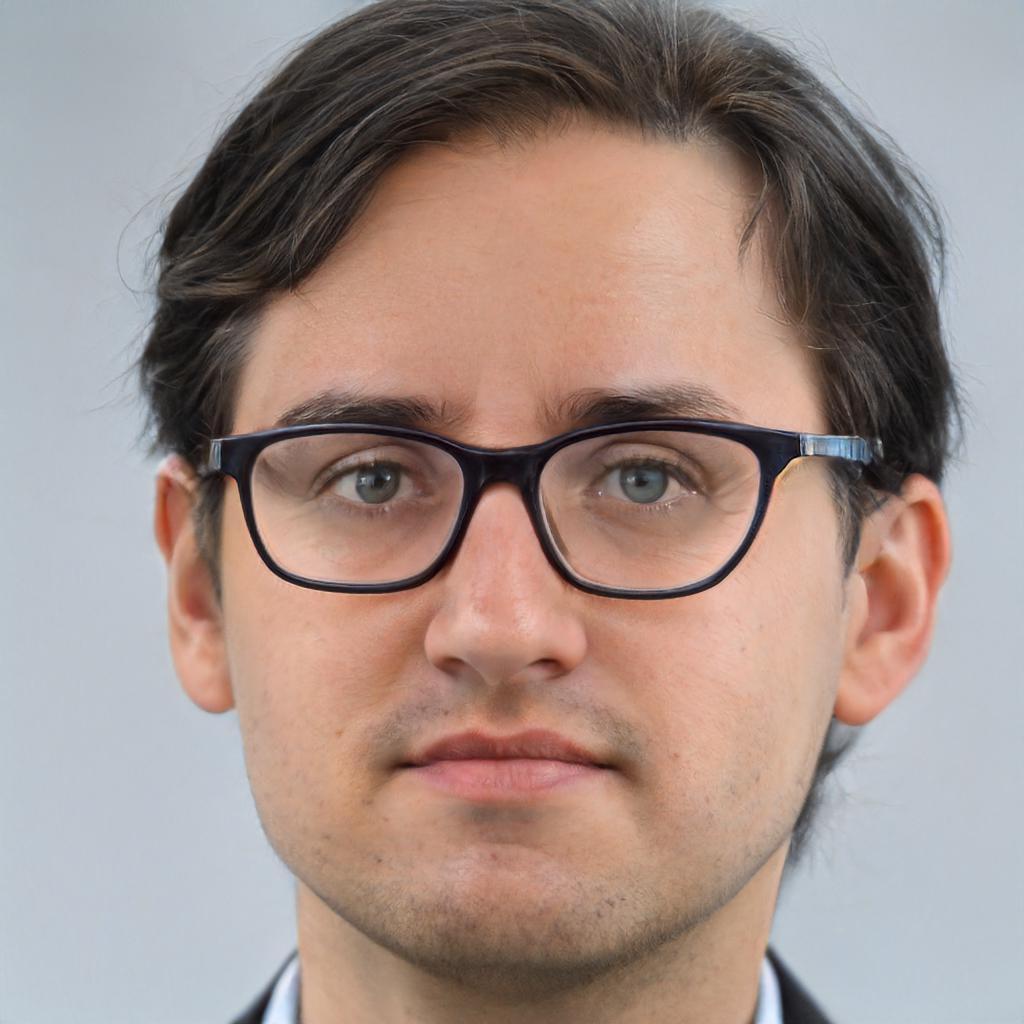} \\
        
        \raisebox{0.275in}{\rotatebox[origin=t]{90}{Toonify}} & 
        \includegraphics[width=0.11\textwidth]{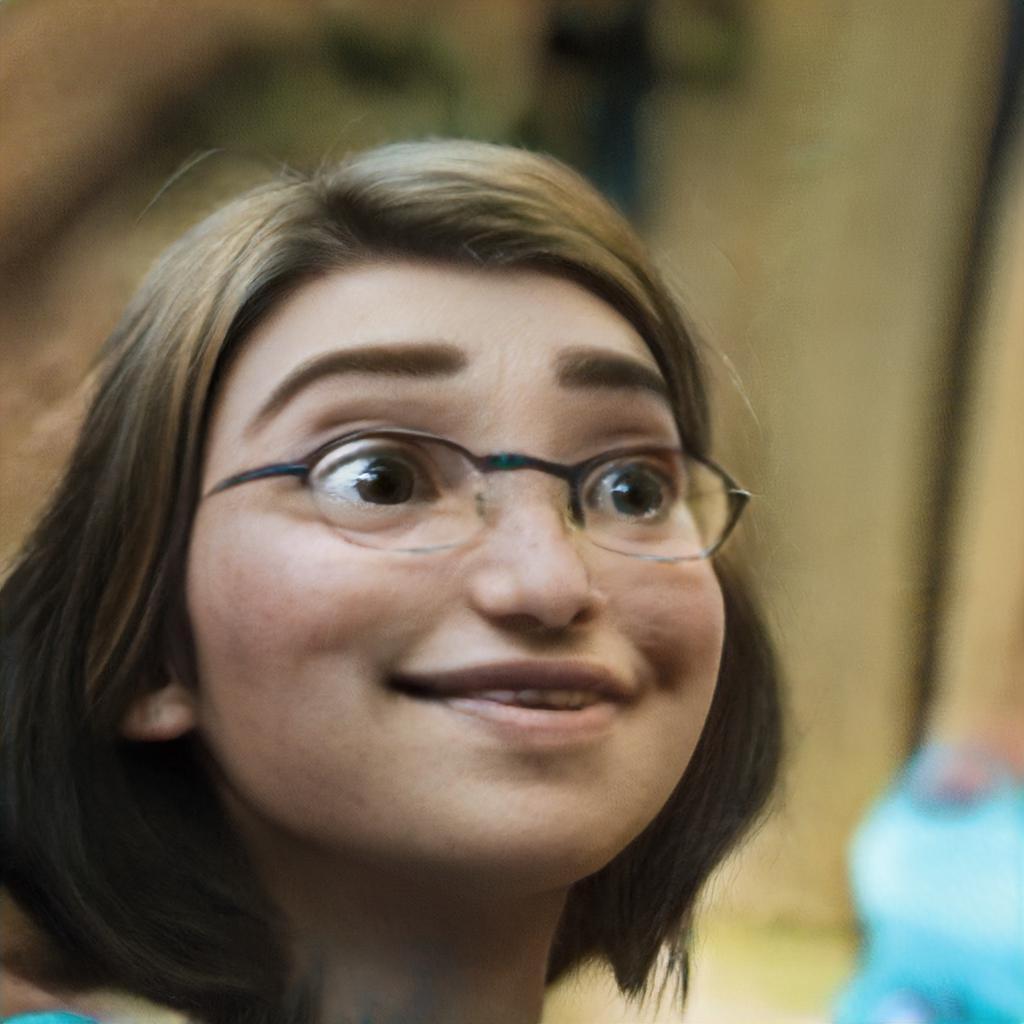} &
        \includegraphics[width=0.11\textwidth]{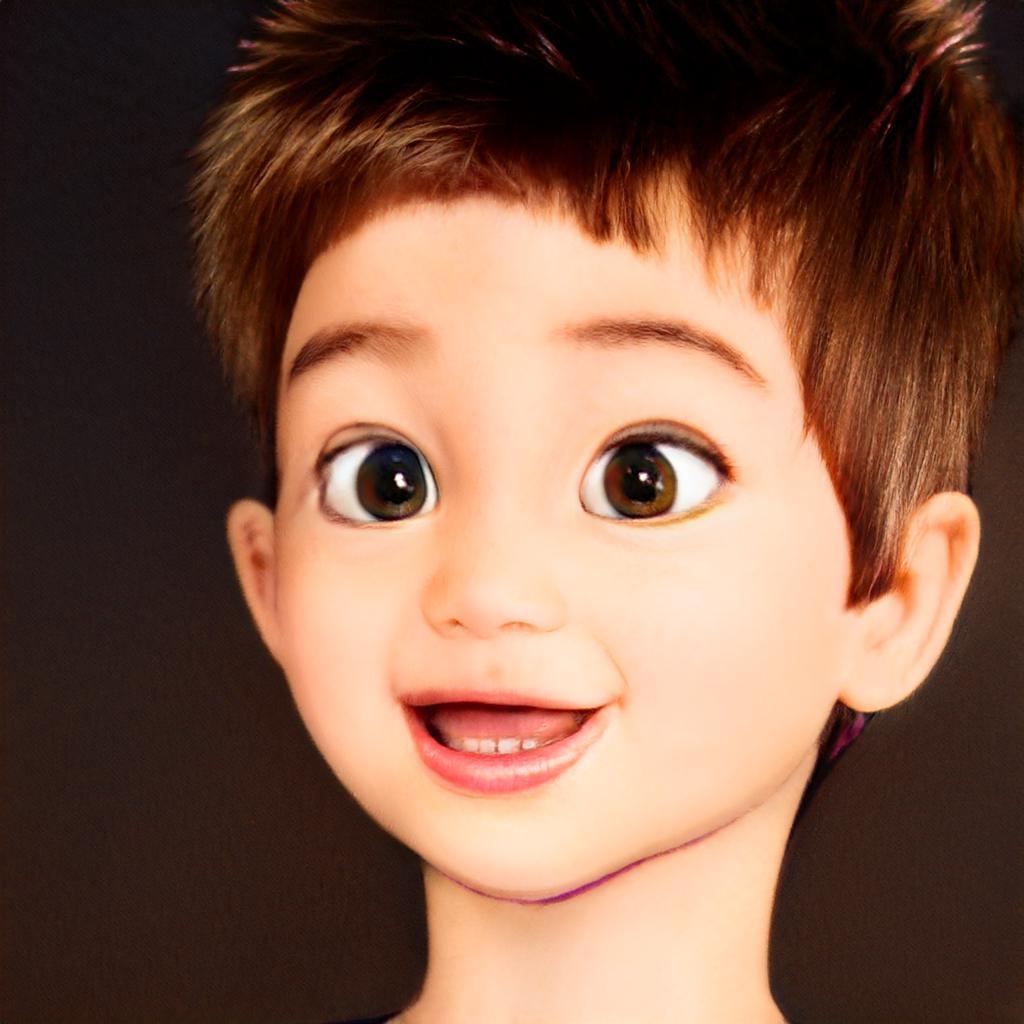} &
        \includegraphics[width=0.11\textwidth]{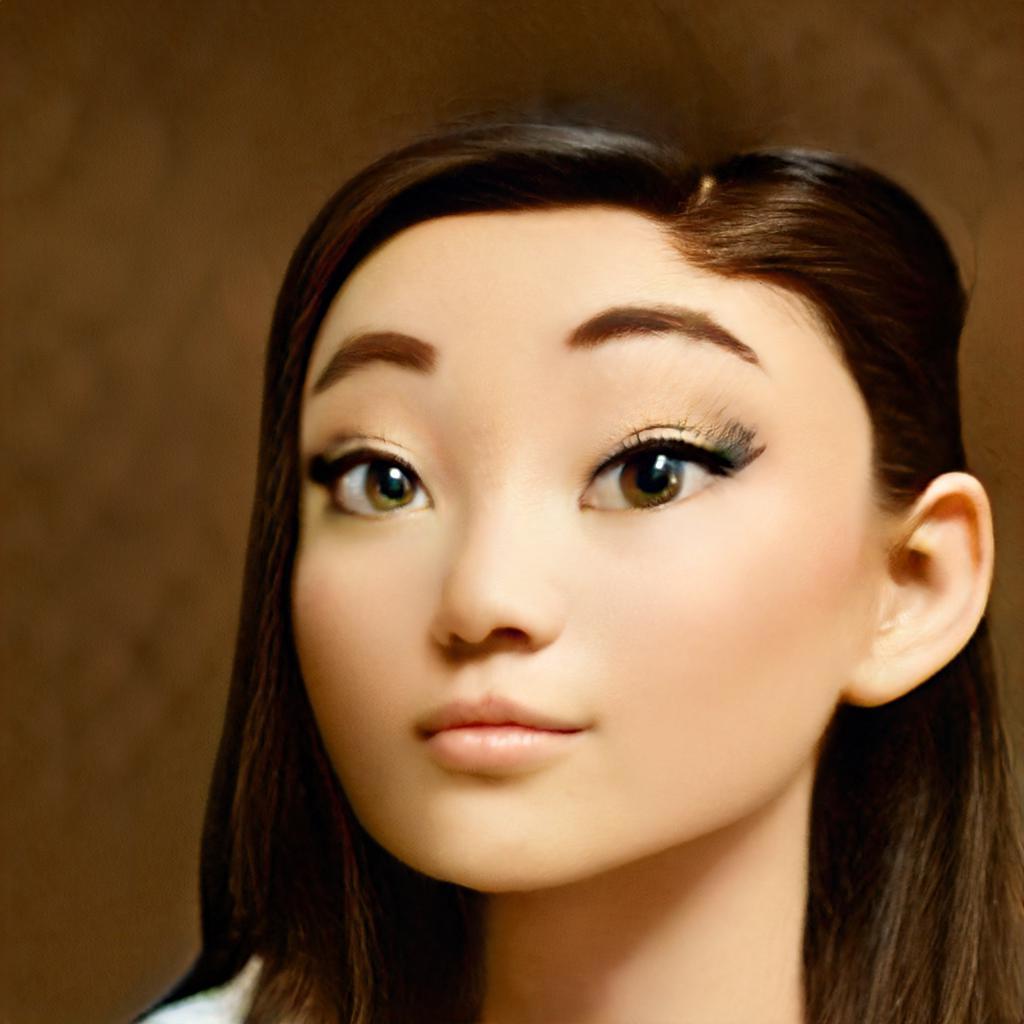} &
        \includegraphics[width=0.11\textwidth]{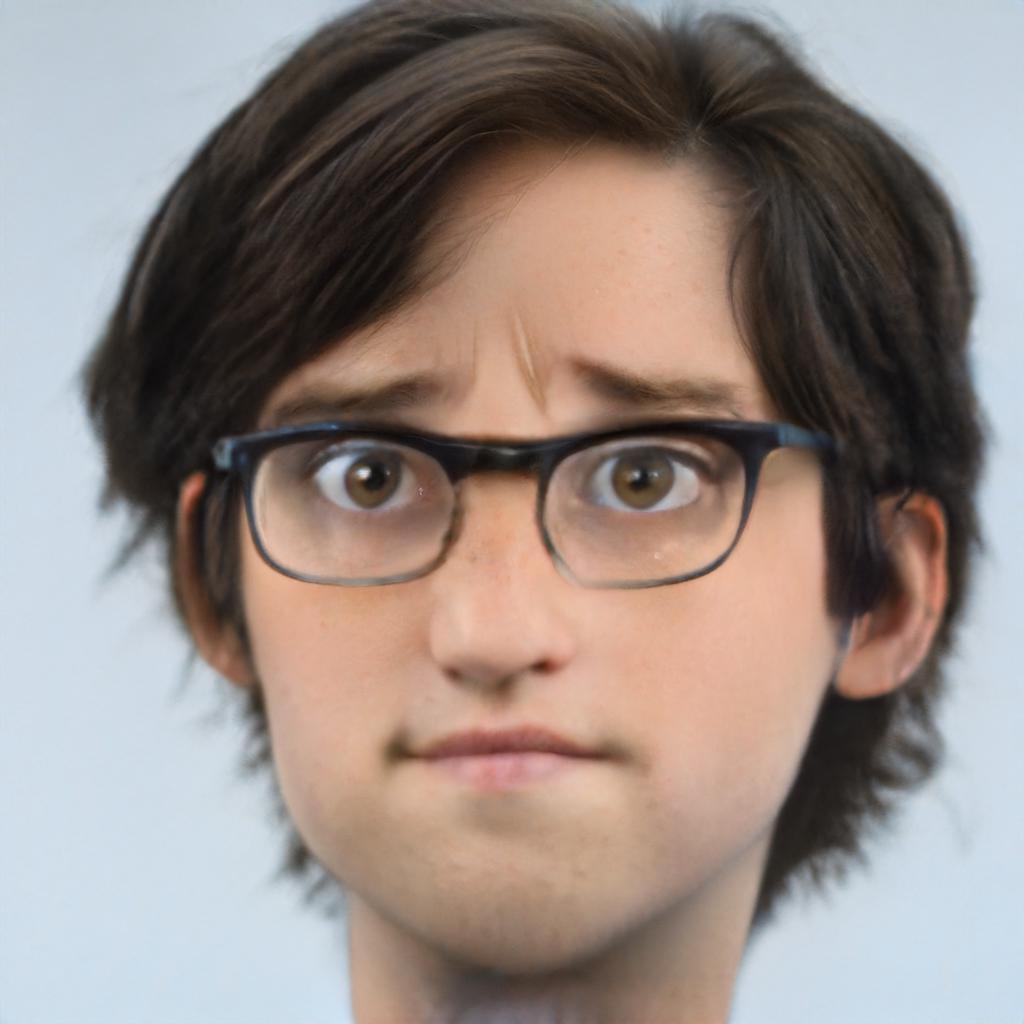}
        
    \end{tabular}
    \vspace{0.05cm}
    \caption{Synthesized images generated by passing the same randomly sampled latent code throug the FFHQ StyleGAN generator and Toonify StyleGAN generator. As shown, the two latent spaces are well-aligned.}
    \label{fig:ffhq_toon_random_latents}
\end{figure}

To show that the above is in fact true we randomly sample $\textbf{w}$ vectors from the FFHQ latent space. We then pass the same $\textbf{w}$ vector to both the FFHQ StyleGAN generator and toonify StyleGAN generator to see if the synthesized images are semantically similar. We provide several examples in Figure~\ref{fig:ffhq_toon_random_latents}. As shown, the same latent code produces semantically similar images in both latent spaces, indicating that the two latent spaces are indeed well-aligned. As a result of the above, we gain valuable insights as to the effectiveness of the encoder bootstrapping technique in the task of image toonification.

\section{Quantitative Results}~\label{sec:quant_comparison}
In Section~\ref{sec:comparison}, we quantitatively compared various inversion methods by constructing \textit{quality-time graphs} which allowed us to visualize how reconstruction quality changes with respect to inference time. In Figures~\ref{fig:quantitative_comparison} and ~\ref{fig:quantitative_comparison_2} we provide the quality-time graphs for both the $L_2$ and LPIPS loss metrics across all five domains. For the human facial domain we additionally measure the identity similarity of the reconstructed images by using the Curricularface~\cite{huang2020curricularface} method for facial recognition.

\section{Additional Results}~\label{sec:additional_results}
The remainder of this document contains additional comparisons and results, as follows:
\begin{enumerate}
    \item Figures~\ref{fig:faces_comparison_psp} and \ref{fig:faces_comparison_e4e} contain additional comparisons on the human facial domain. Additionally, Figure~\ref{fig:faces_comparison_challenging_poses} provides a comparison on more challenging input poses and expressions.
    \item Figures~\ref{fig:cars_comparison_psp} and \ref{fig:cars_comparison_e4e} contain additional comparisons on the cars domain.
    \item Figures~\ref{fig:churches_comparison_psp} and \ref{fig:churches_comparison_e4e} contain additional comparisons on the churches domain.
    \item Figure~\ref{fig:wild_comparison_psp} contains additional comparisons on the wild animals domain.
    \item Figure~\ref{fig:horses_comparison_e4e} contains additional comparisons on the horses domain.
    \item Figure~\ref{fig:idinvert_comparison} contains comparisons to the IDInvert encoder from Zhu \etal~\cite{zhu2020domain} on the human facial and cars domains.
    \item Figure~\ref{fig:faces_steps_visualization} shows iterative outputs generated by ReStyle applied over pSp~\cite{richardson2020encoding} on the human facial domain.
    \item Figure~\ref{fig:cars_steps_visualization} shows iterative outputs generated by ReStyle applied over e4e~\cite{tov2021designing} on the cars domain.
    \item Figure~\ref{fig:churches_steps_visualization} shows iterative outputs generated by ReStyle applied over pSp~\cite{tov2021designing} on the churches domain.    
    \item Figure~\ref{fig:horses_steps_visualization} shows iterative outputs generated by ReStyle applied over e4e~\cite{tov2021designing} on the horses domain.
    \item Figures~\ref{fig:faces_editing} contains editing comparisons with the optimization technique from Karras \etal~\cite{karras2020analyzing} on the human facial domain obtained using InterFaceGAN~\cite{shen2020interpreting}. 
    \item Figures~\ref{fig:cars_editing} contains editing comparisons with the optimization technique from Karras \etal~\cite{karras2020analyzing} on the cars domain obtained using GANSpace~\cite{harkonen2020ganspace}. 
    \item Figures~\ref{fig:horses_editing} contains editing comparisons with the optimization technique from Karras \etal~\cite{karras2020analyzing} on the horses domain obtained using SeFa~\cite{shen2020closedform}. 
    \item Figures~\ref{fig:encoder_mixing} and \ref{fig:encoder_mixing_2} contain additional results on the image toonification task using the encoder bootstrapping method presented in Section~\ref{sec:bootstrapping}.
    
\end{enumerate}
Note, all results are shown at full-resolution: $1024\times1024$ for human faces, $512\times512$ for cars and wild animal faces, and $256\times256$ for churches and horses.

\begin{figure*}
    \centering
    \setlength{\belowcaptionskip}{-2.5pt}
    \setlength{\tabcolsep}{1pt}
    \begin{tabular}{c c c}
        \includegraphics[width=0.33\textwidth]{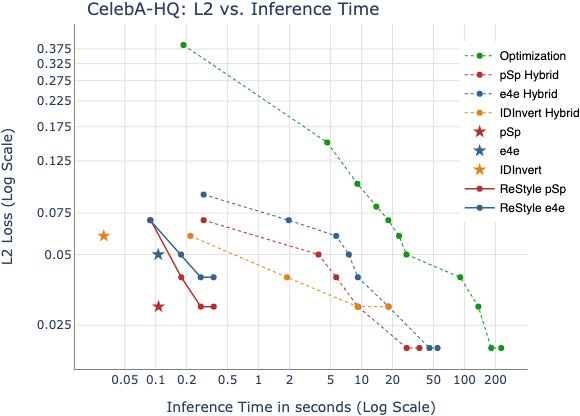} &
        \hspace{0.05cm}
        \includegraphics[width=0.33\textwidth]{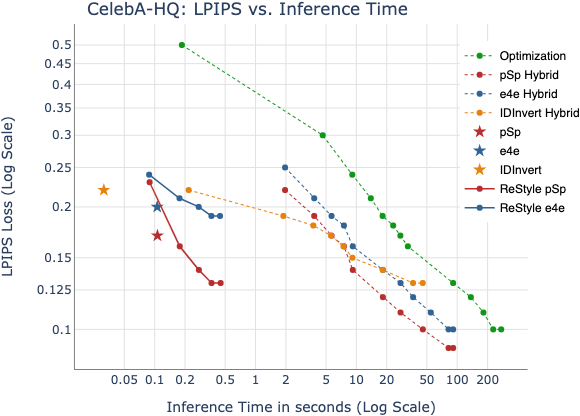} &       
        \hspace{0.05cm}
        \includegraphics[width=0.33\textwidth]{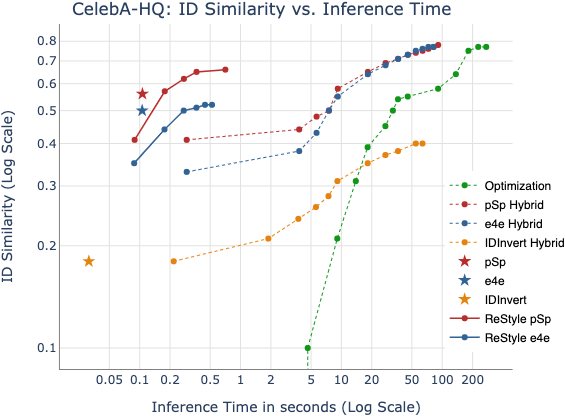}
    \end{tabular}
    
    \vspace{0.5cm}
    
    \begin{tabular}{c c}
        \includegraphics[width=0.33\textwidth]{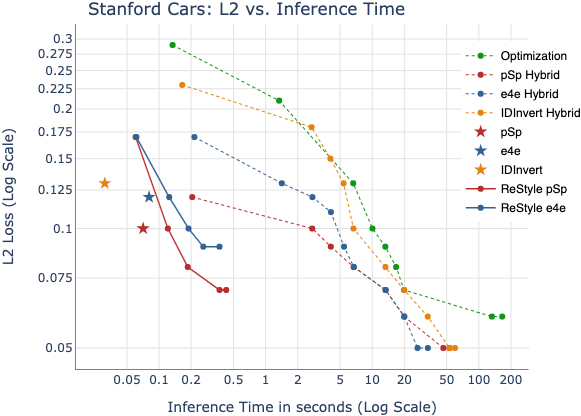} &
        \hspace{0.05cm}
        \includegraphics[width=0.33\textwidth]{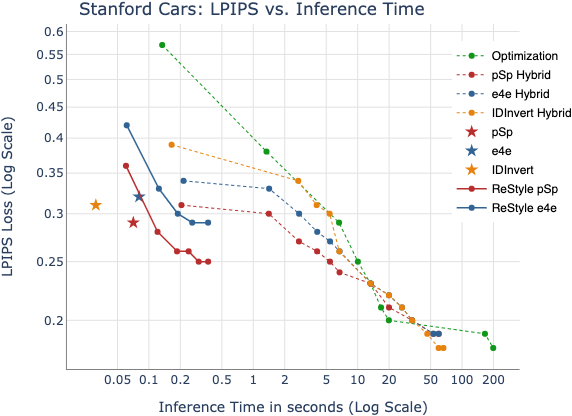}
    \end{tabular}
    
    \vspace{0.5cm}
    
    \begin{tabular}{c c}
        \includegraphics[width=0.33\textwidth]{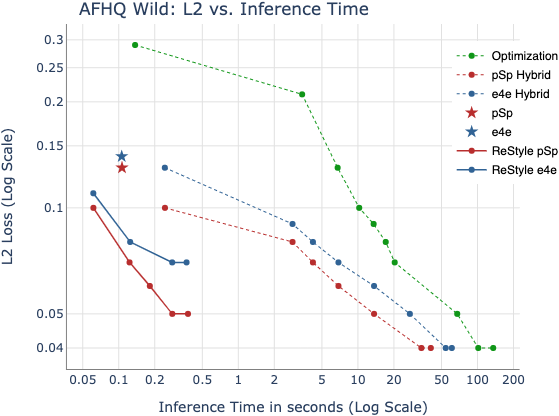} &
        \hspace{0.05cm}
        \includegraphics[width=0.33\textwidth]{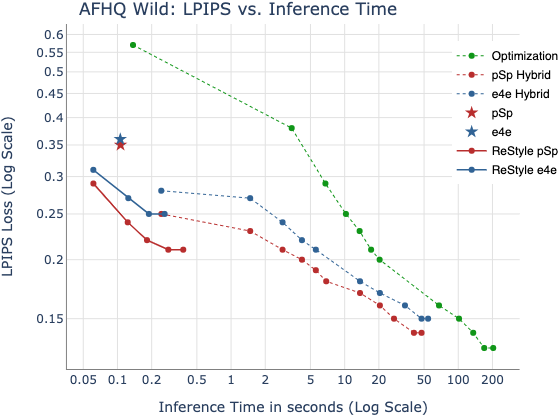}
    \end{tabular}

    \vspace{0.5cm}
    \caption{\textit{Quantitative comparison.} We compare ReStyle with current state-of-the-art optimization-based and encoder-based methods 
    by analyzing reconstruction via multiple evaluation metrics while measuring each method's run-time during inference.
    Each encoder-based method is represented using a $\star$ symbol. The corresponding hybrid method is marked using a dashed line of the same color with the ReStyle applied over the base method shown using a solid line of the same color.
    Optimization results are shown using a dashed \textcolor{ForestGreen}{green} line.
    Methods based on pSp are shown in \textcolor{red}{red} with methods based on e4e shown in \textcolor{blue}{blue}. Finally, results obtained using IDInvert~\cite{zhu2020domain} are shown in \textcolor{orange}{orange}.  
    Note that both axes are shown in log-scale. 
    }
    \label{fig:quantitative_comparison}
\end{figure*}
\begin{figure*}
    \centering
    \setlength{\belowcaptionskip}{-2.5pt}
    \setlength{\tabcolsep}{1pt}
    
\begin{tabular}{c c}
        \includegraphics[width=0.33\textwidth]{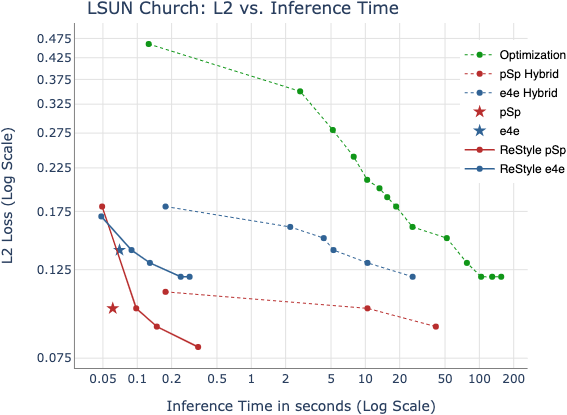} &
        \hspace{0.05cm}
        \includegraphics[width=0.33\textwidth]{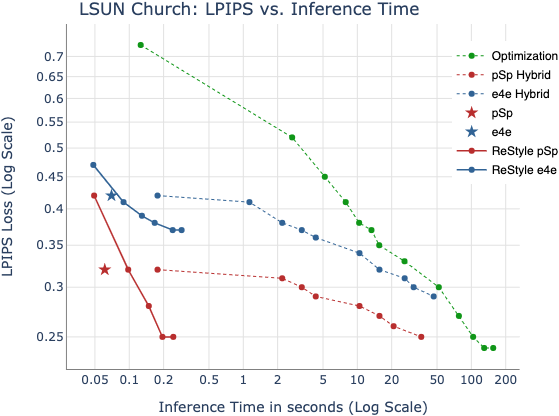}
    \end{tabular}
    
    \vspace{0.5cm}
    
    \begin{tabular}{c c}
        \includegraphics[width=0.33\textwidth]{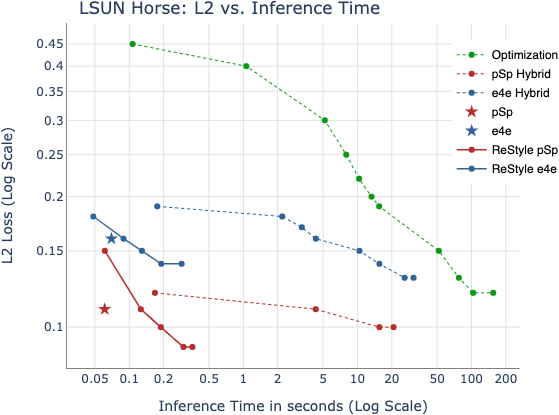} &
        \hspace{0.05cm}
        \includegraphics[width=0.33\textwidth]{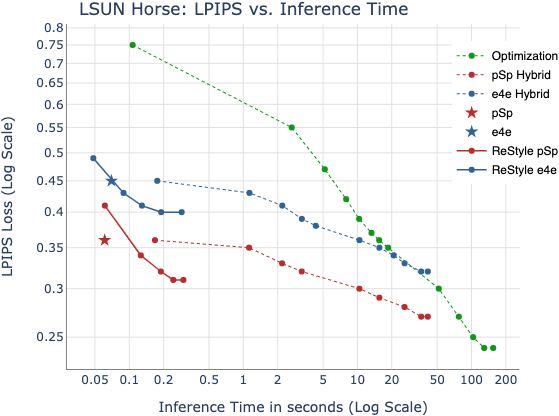}
    \end{tabular}
    
    \vspace{0.5cm}
    \caption{\textit{Quantitative comparison.} Same setting as Figure~\ref{fig:quantitative_comparison} for the LSUN~\cite{yu2016lsun} Church and LSUN Horse domains. 
    }
    \label{fig:quantitative_comparison_2}
\end{figure*}

\begin{figure*}
    \centering
    \setlength{\belowcaptionskip}{-6pt}
    \setlength{\tabcolsep}{1pt}
    {\small
    \begin{tabular}{c c c c c}

        Input & Optimization & Hybrid & pSp & $\text{ReStyle}_{pSp}$ \\

        \includegraphics[width=0.14\textwidth]{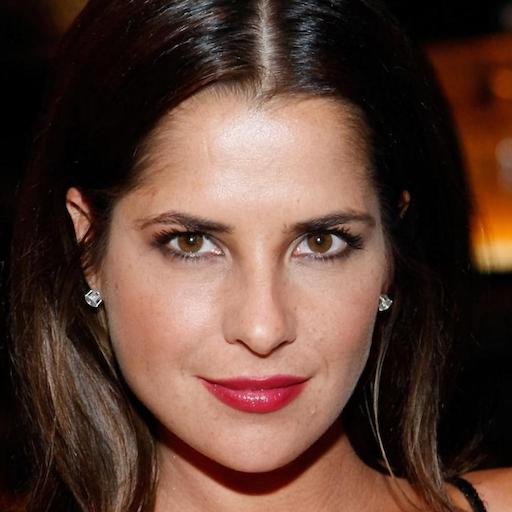} &
        \includegraphics[width=0.14\textwidth]{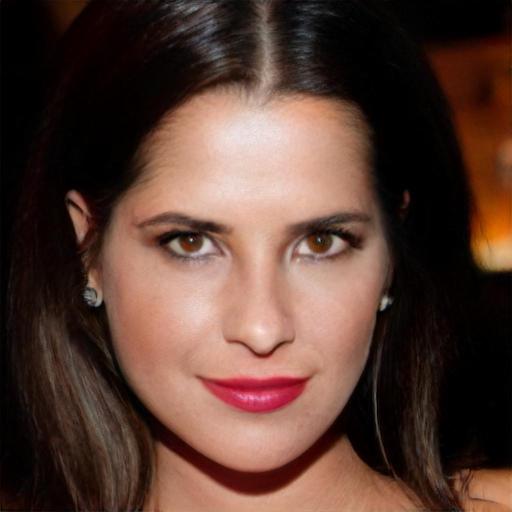} &
        \includegraphics[width=0.14\textwidth]{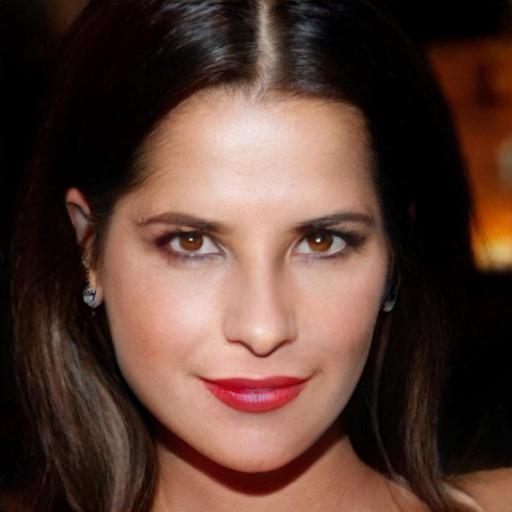} &
        \includegraphics[width=0.14\textwidth]{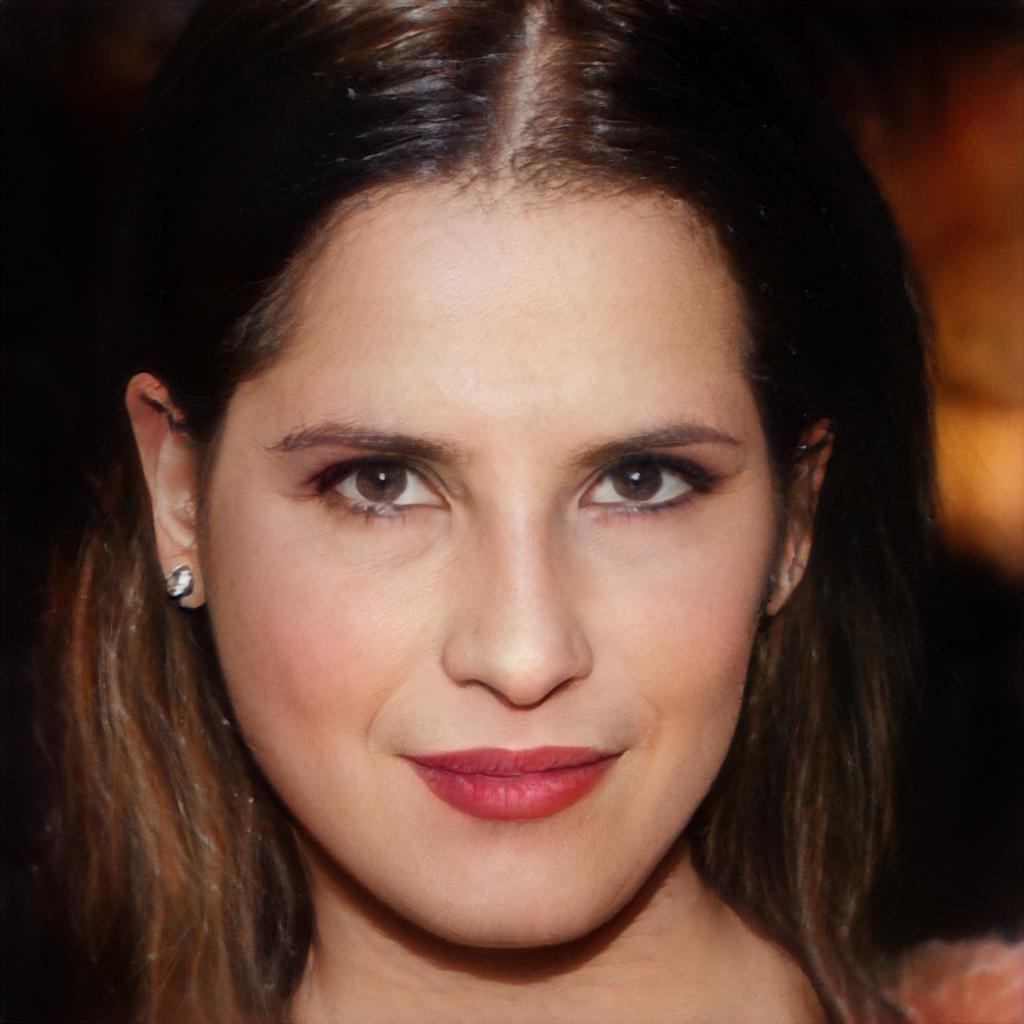} &
        \includegraphics[width=0.14\textwidth]{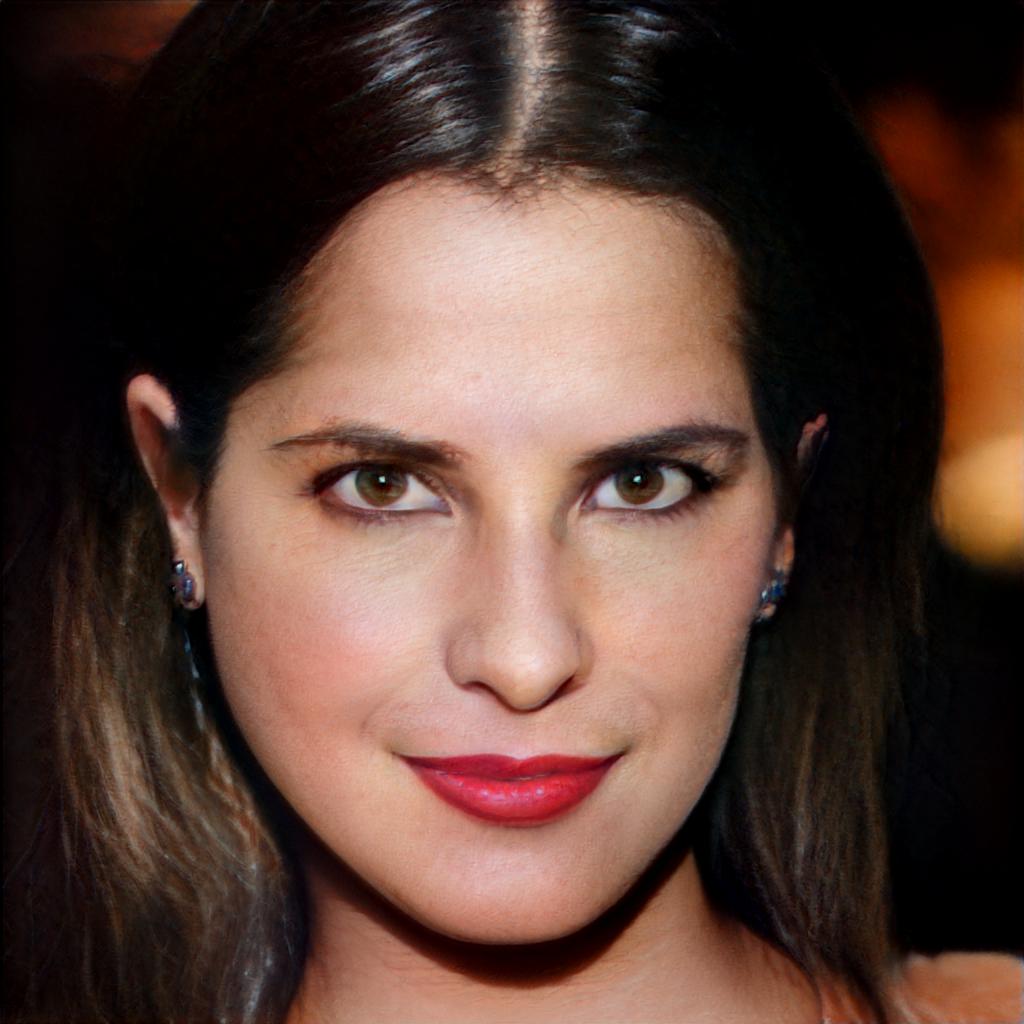} \\

        \includegraphics[width=0.14\textwidth]{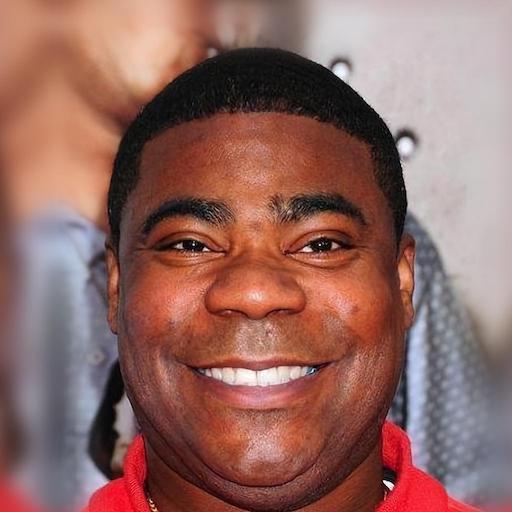} &
        \includegraphics[width=0.14\textwidth]{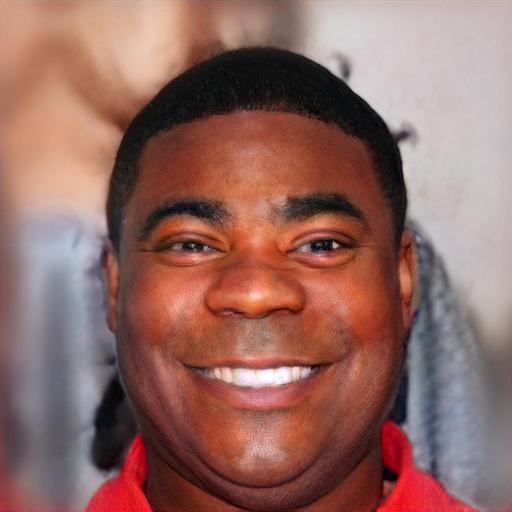} &
        \includegraphics[width=0.14\textwidth]{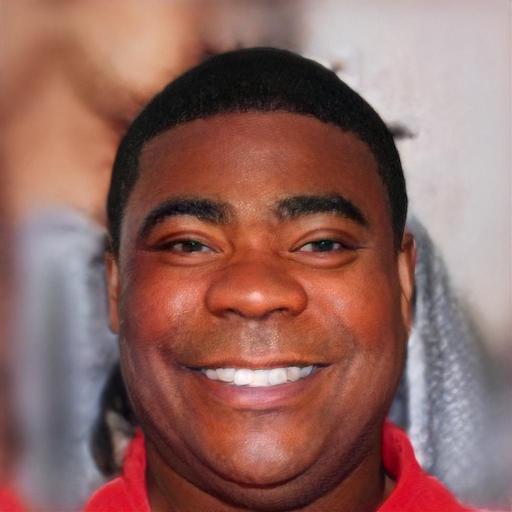} &
        \includegraphics[width=0.14\textwidth]{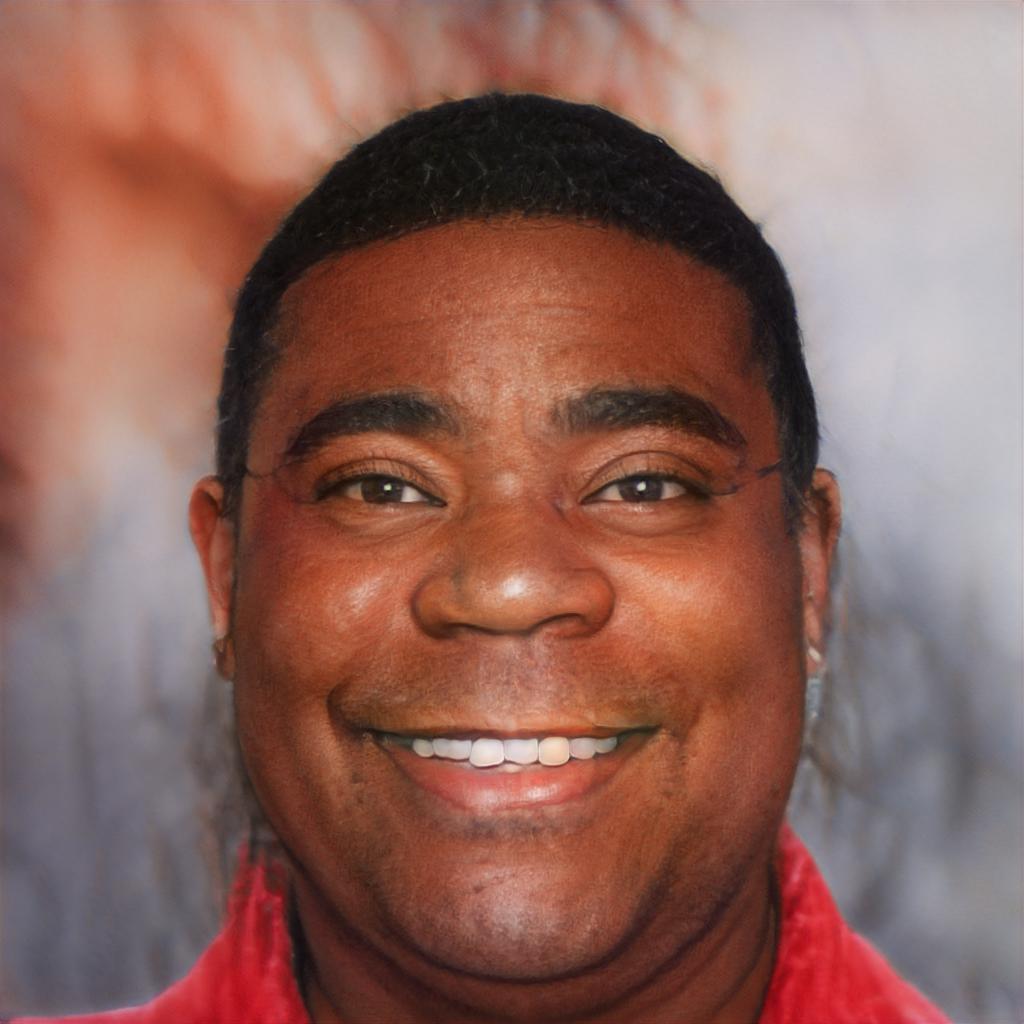} &
        \includegraphics[width=0.14\textwidth]{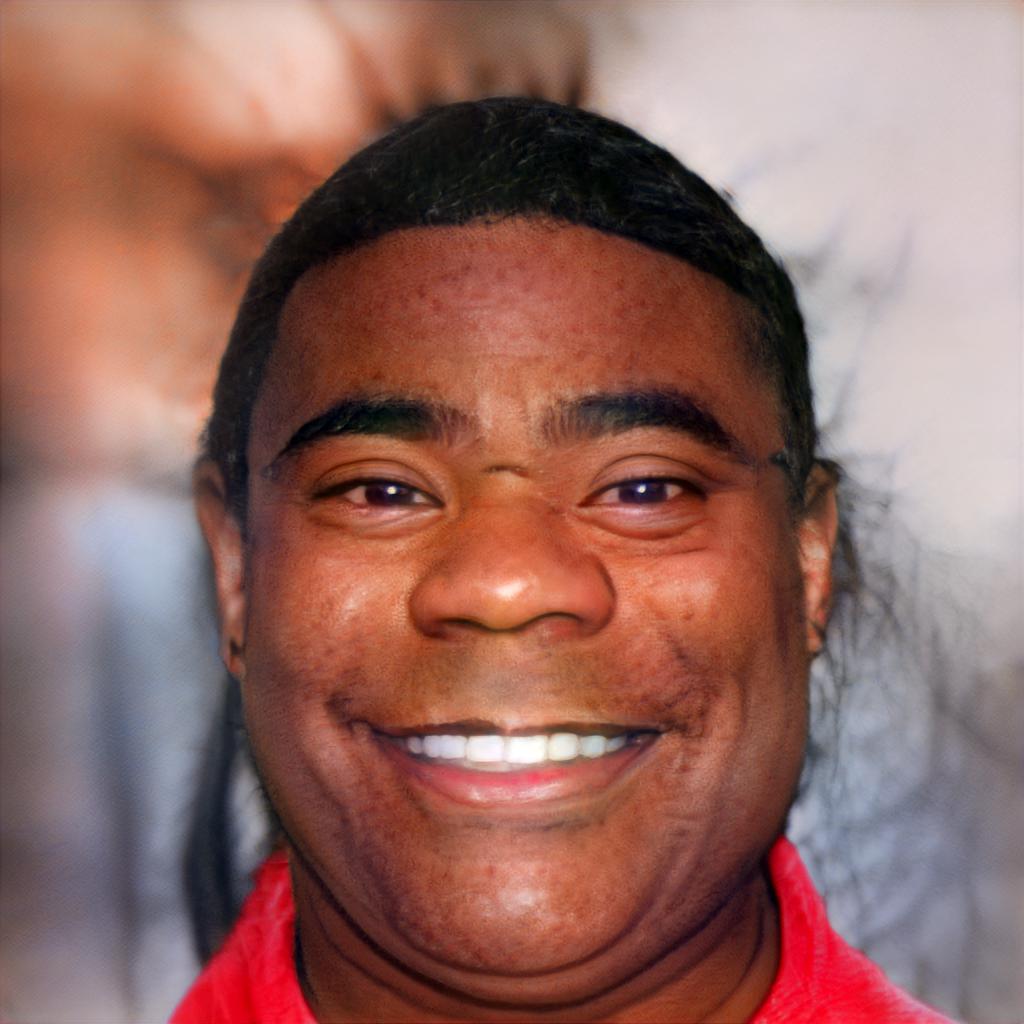} \\
        
        \includegraphics[width=0.14\textwidth]{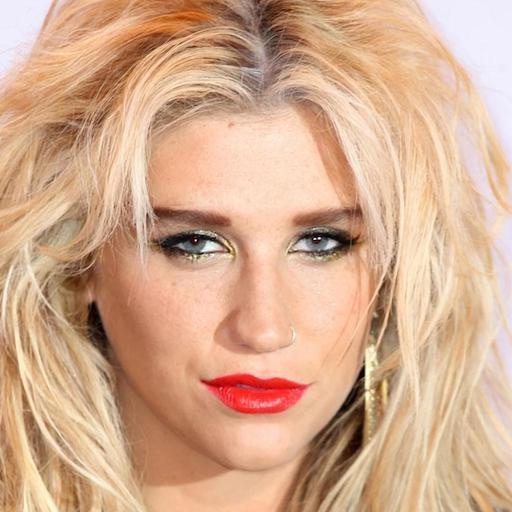} &
        \includegraphics[width=0.14\textwidth]{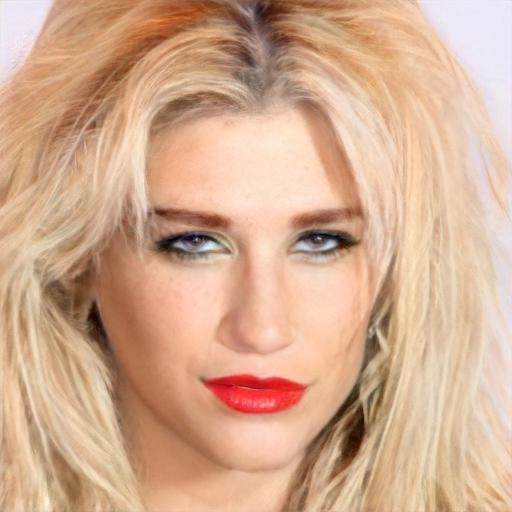} &
        \includegraphics[width=0.14\textwidth]{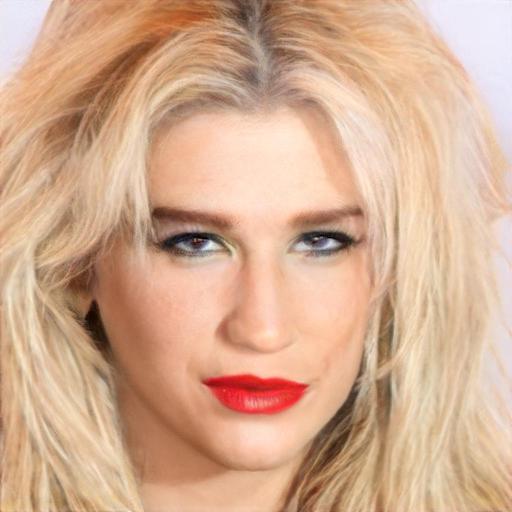} &
        \includegraphics[width=0.14\textwidth]{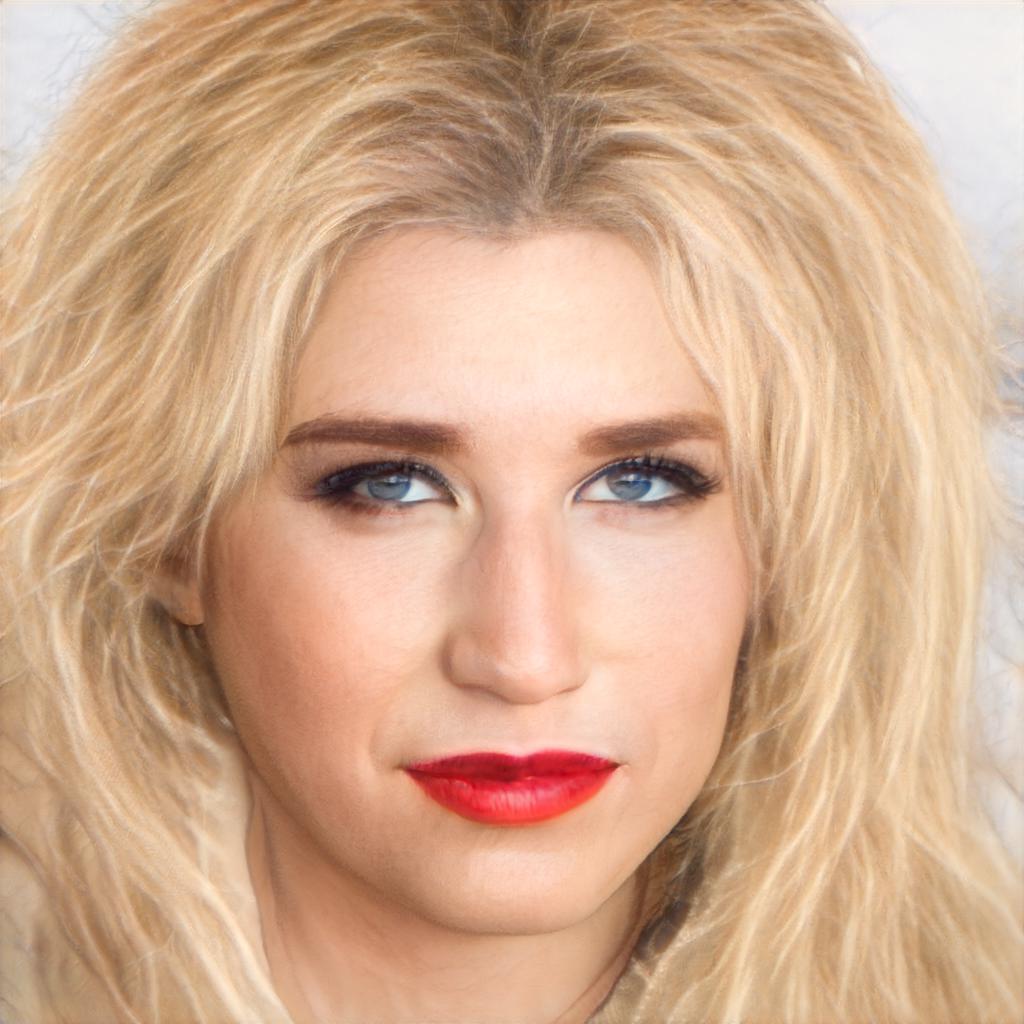} &
        \includegraphics[width=0.14\textwidth]{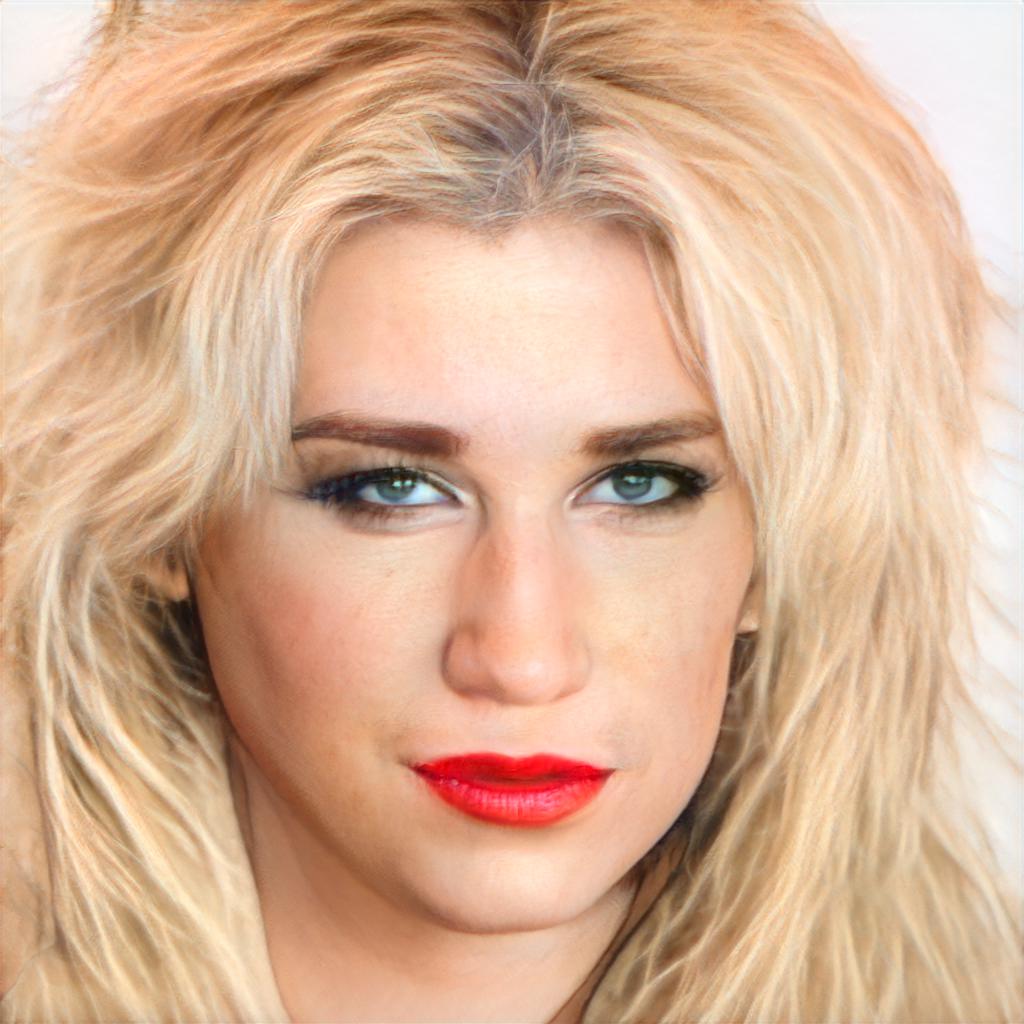} \\
\end{tabular}
    }
    \vspace{0.35cm}
    \caption{Additional qualitative comparisons on the CelebA-HQ~\cite{liu2015deep, karras2017progressive} test set. Here, we apply ReStyle over the pSp encoder~\cite{richardson2020encoding}.}
    \label{fig:faces_comparison_psp}
\end{figure*}
\begin{figure*}
    \centering
    \setlength{\belowcaptionskip}{-6pt}
    \setlength{\tabcolsep}{1pt}
    {\small
    \begin{tabular}{c c c c c}

        Input & Optimization & Hybrid & e4e & $\text{ReStyle}_{e4e}$ \\

        \includegraphics[width=0.14\textwidth]{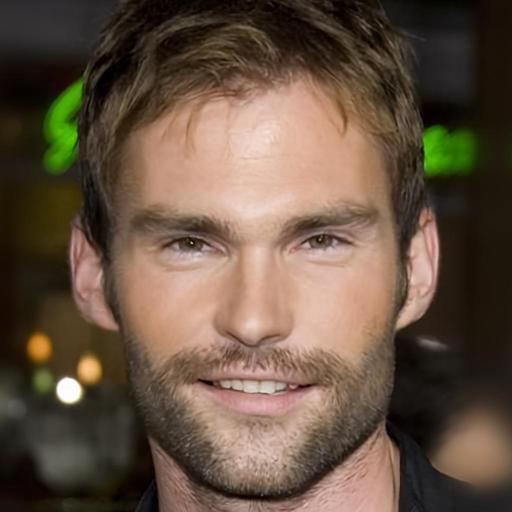} &
        \includegraphics[width=0.14\textwidth]{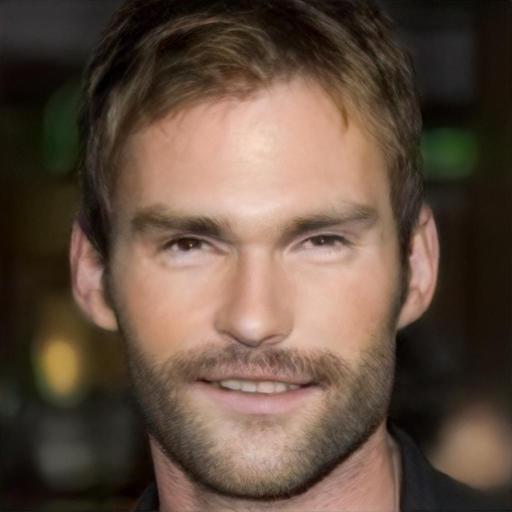} &
        \includegraphics[width=0.14\textwidth]{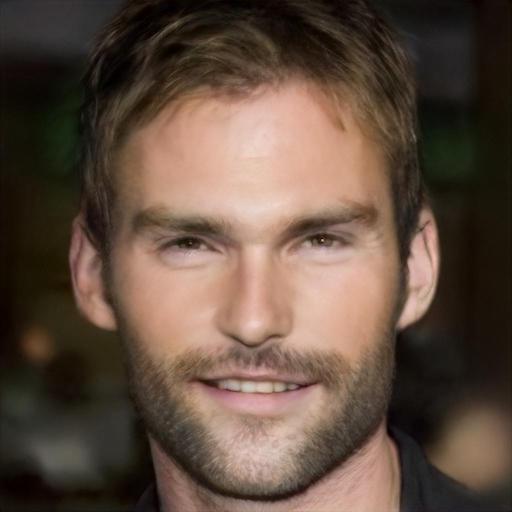} &
        \includegraphics[width=0.14\textwidth]{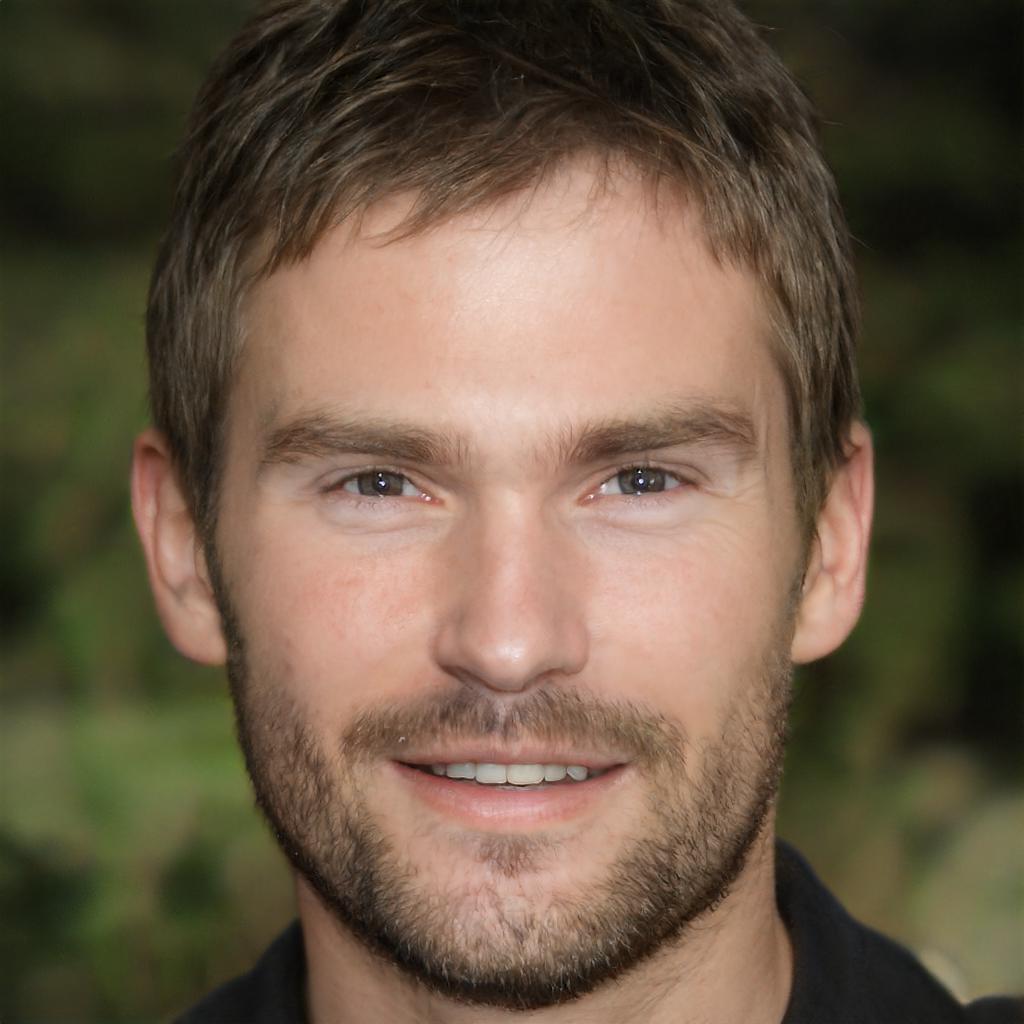} &
        \includegraphics[width=0.14\textwidth]{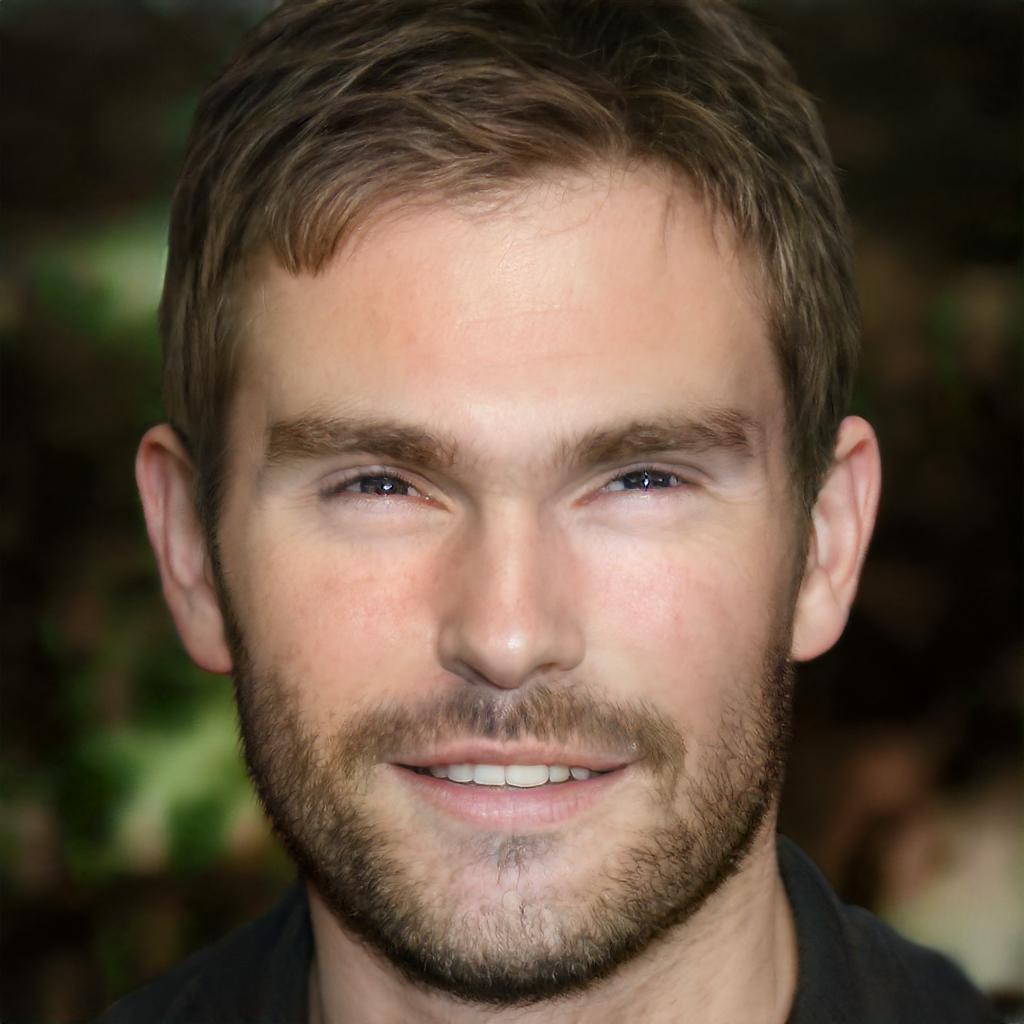} \\

        \includegraphics[width=0.14\textwidth]{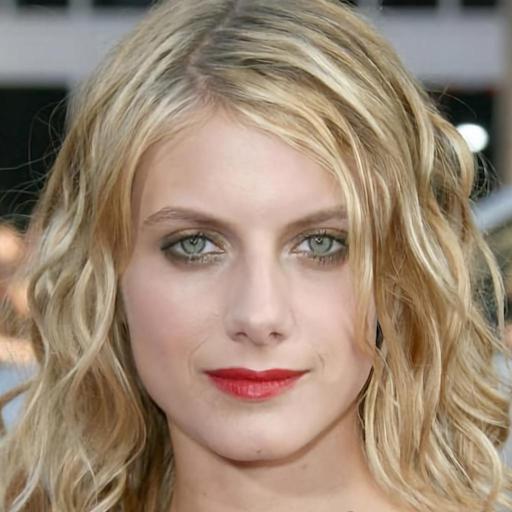} &
        \includegraphics[width=0.14\textwidth]{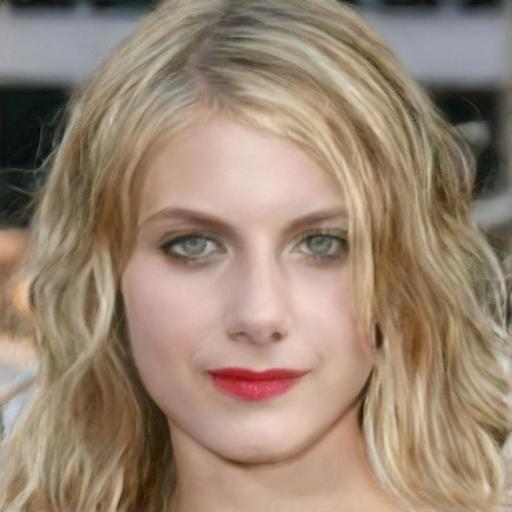} &
        \includegraphics[width=0.14\textwidth]{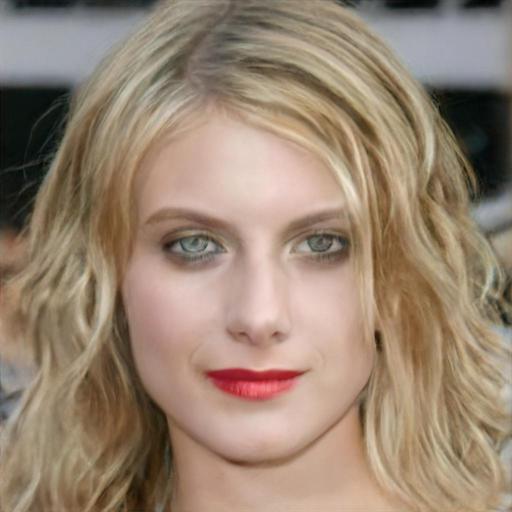} &
        \includegraphics[width=0.14\textwidth]{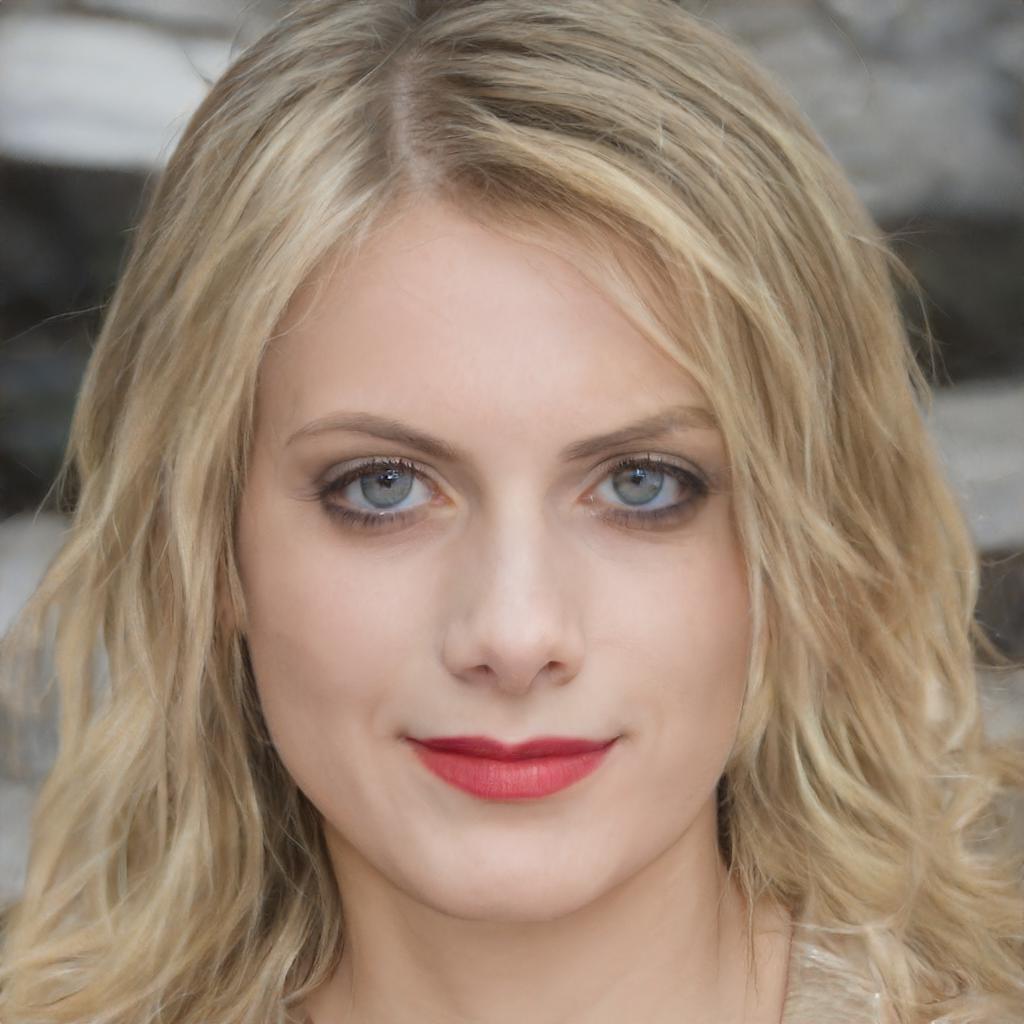} &
        \includegraphics[width=0.14\textwidth]{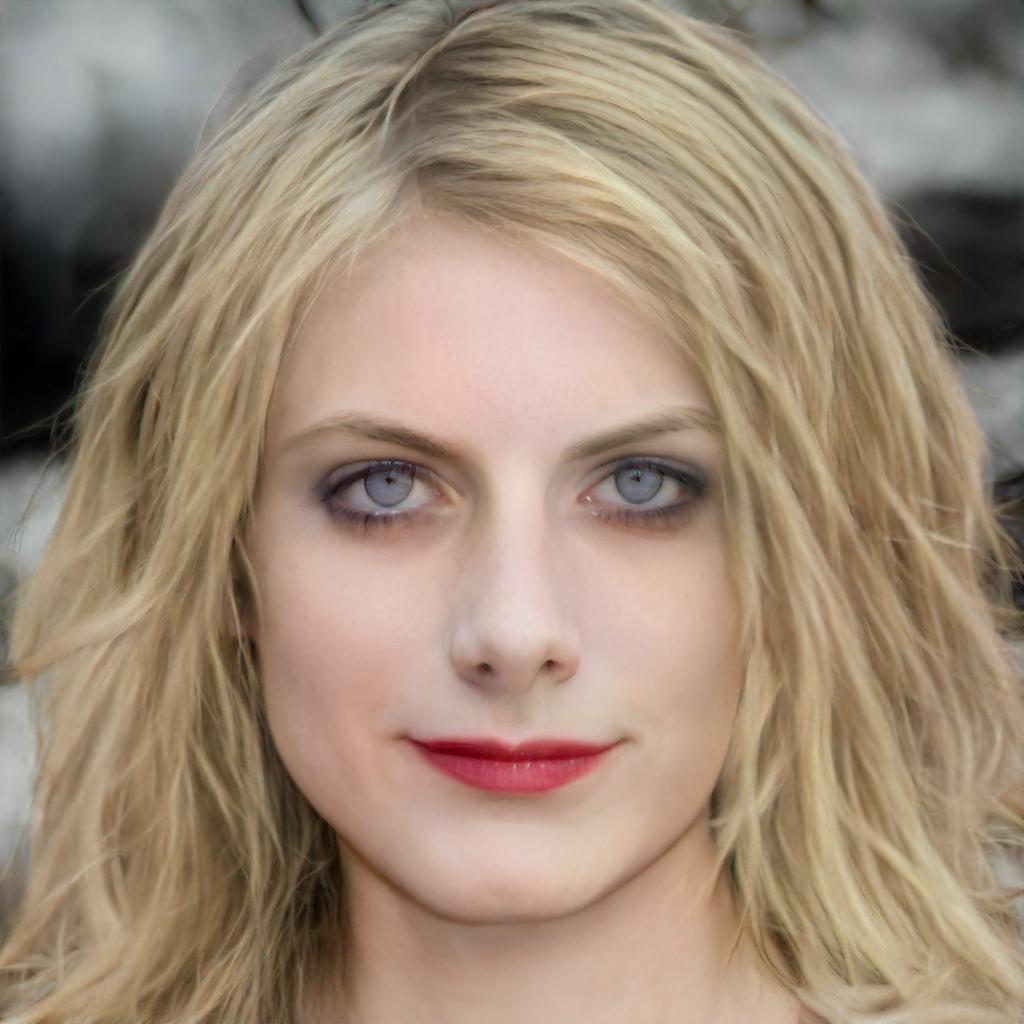} \\
        
        \includegraphics[width=0.14\textwidth]{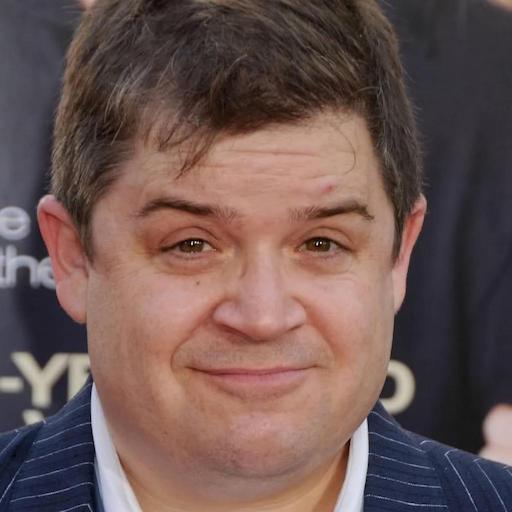} &
        \includegraphics[width=0.14\textwidth]{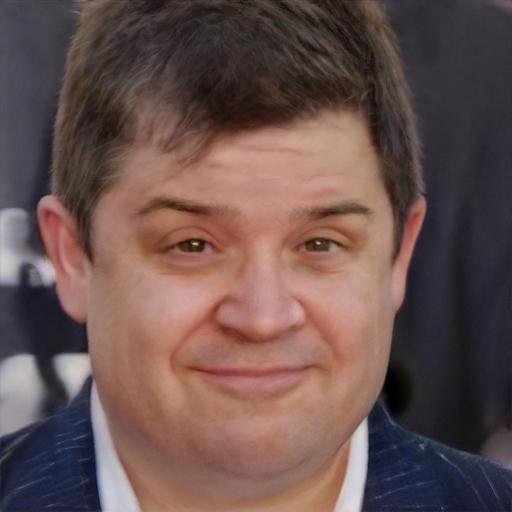} &
        \includegraphics[width=0.14\textwidth]{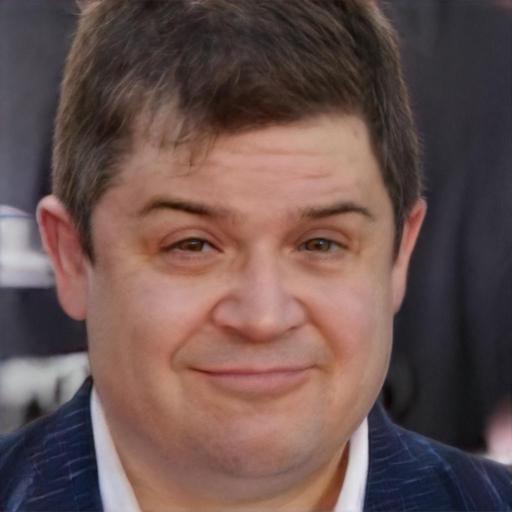} &
        \includegraphics[width=0.14\textwidth]{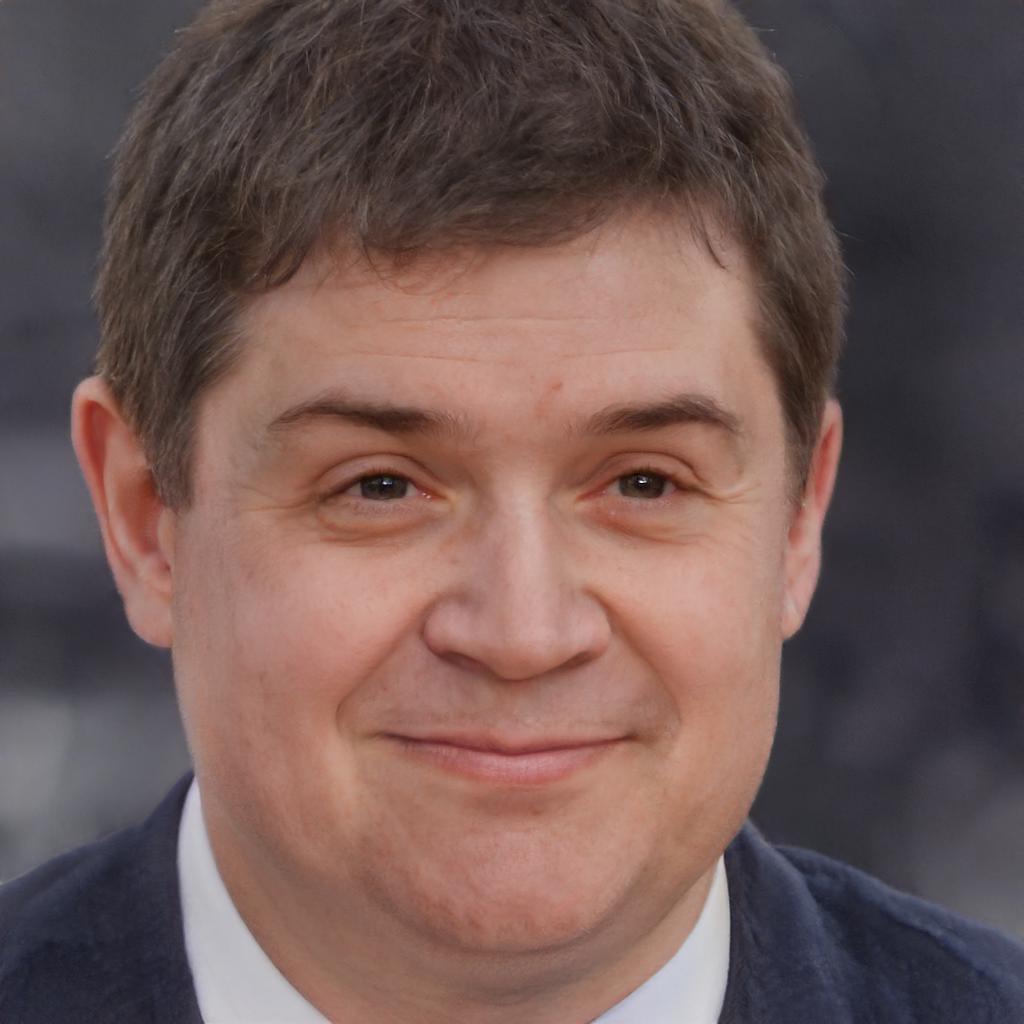} &
        \includegraphics[width=0.14\textwidth]{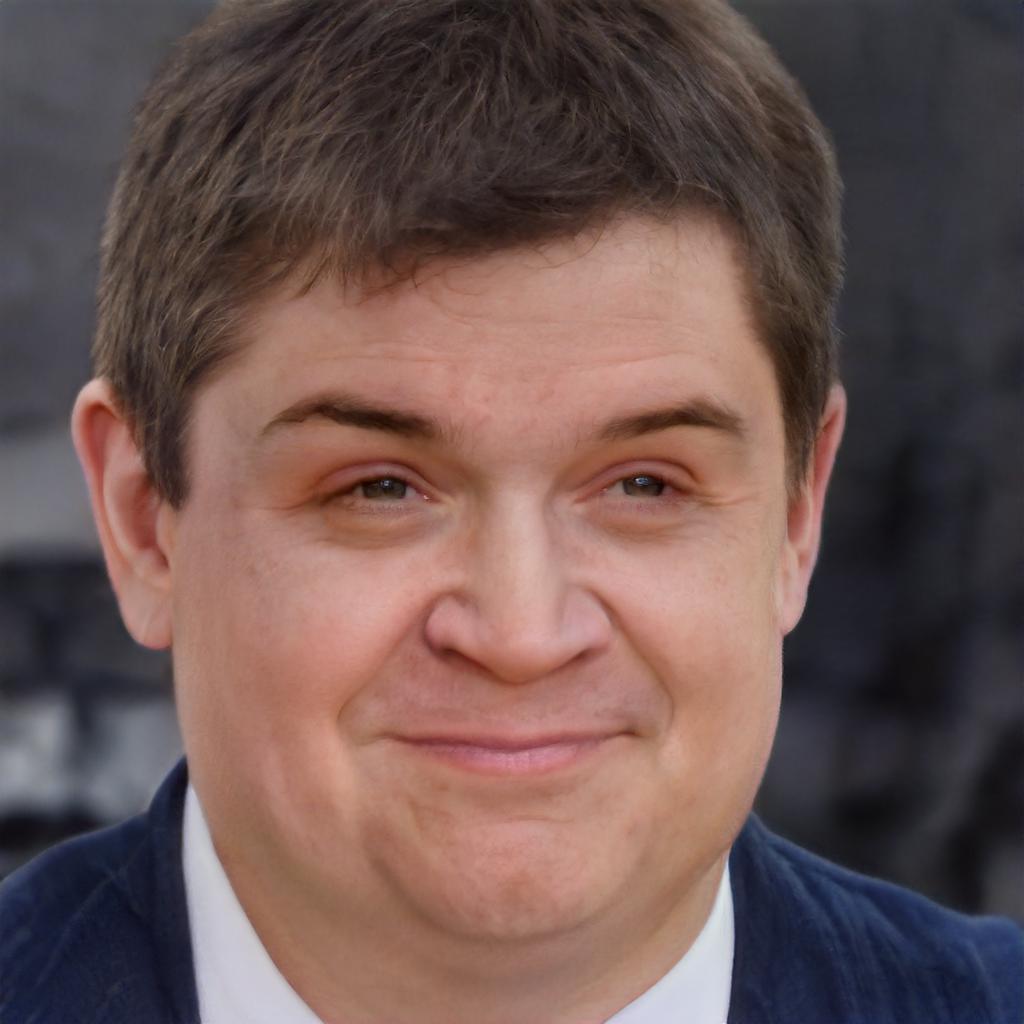} \\
\end{tabular}
    }
    \vspace{0.35cm}
    \caption{Additional qualitative comparisons on the CelebA-HQ~\cite{liu2015deep, karras2017progressive} test set. Here, we apply ReStyle over the e4e encoder~\cite{tov2021designing}.}    \label{fig:faces_comparison_e4e}
\end{figure*}
\begin{figure*}
    \centering
    \setlength{\belowcaptionskip}{-6pt}
    \setlength{\tabcolsep}{1pt}
    {\small
    \begin{tabular}{c c c c c}

        Input & Optimization & Hybrid & pSp & $\text{ReStyle}_{pSp}$ \\

        \includegraphics[width=0.14\textwidth]{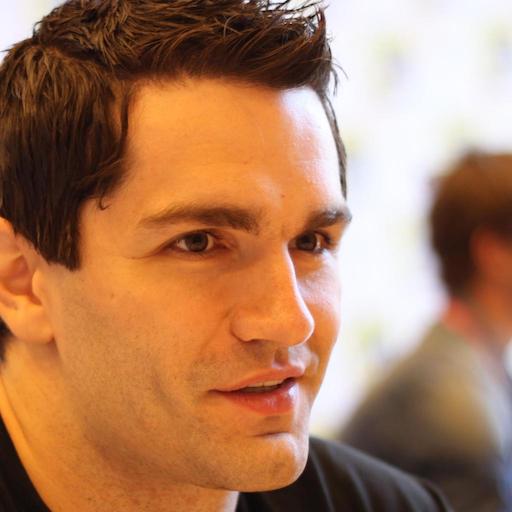} &
        \includegraphics[width=0.14\textwidth]{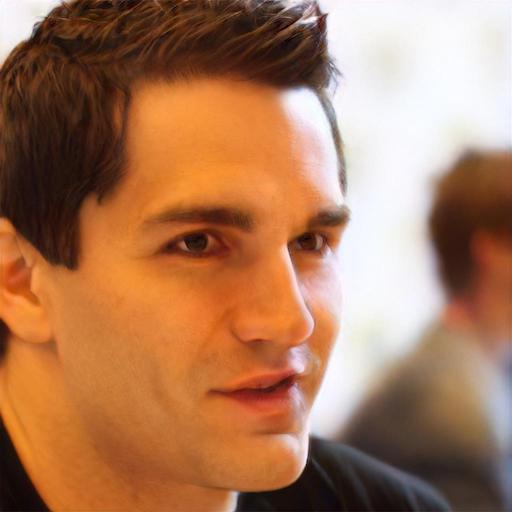} &
        \includegraphics[width=0.14\textwidth]{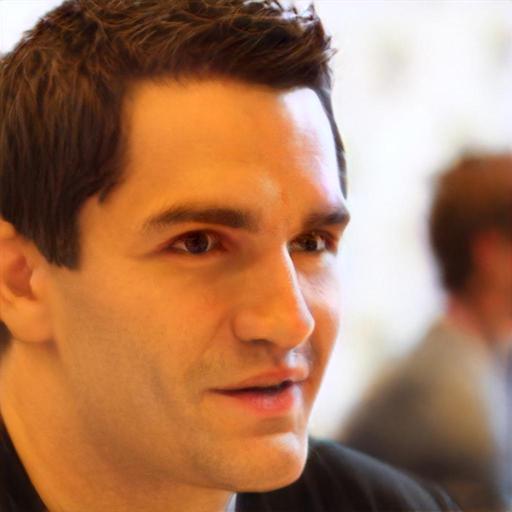} &
        \includegraphics[width=0.14\textwidth]{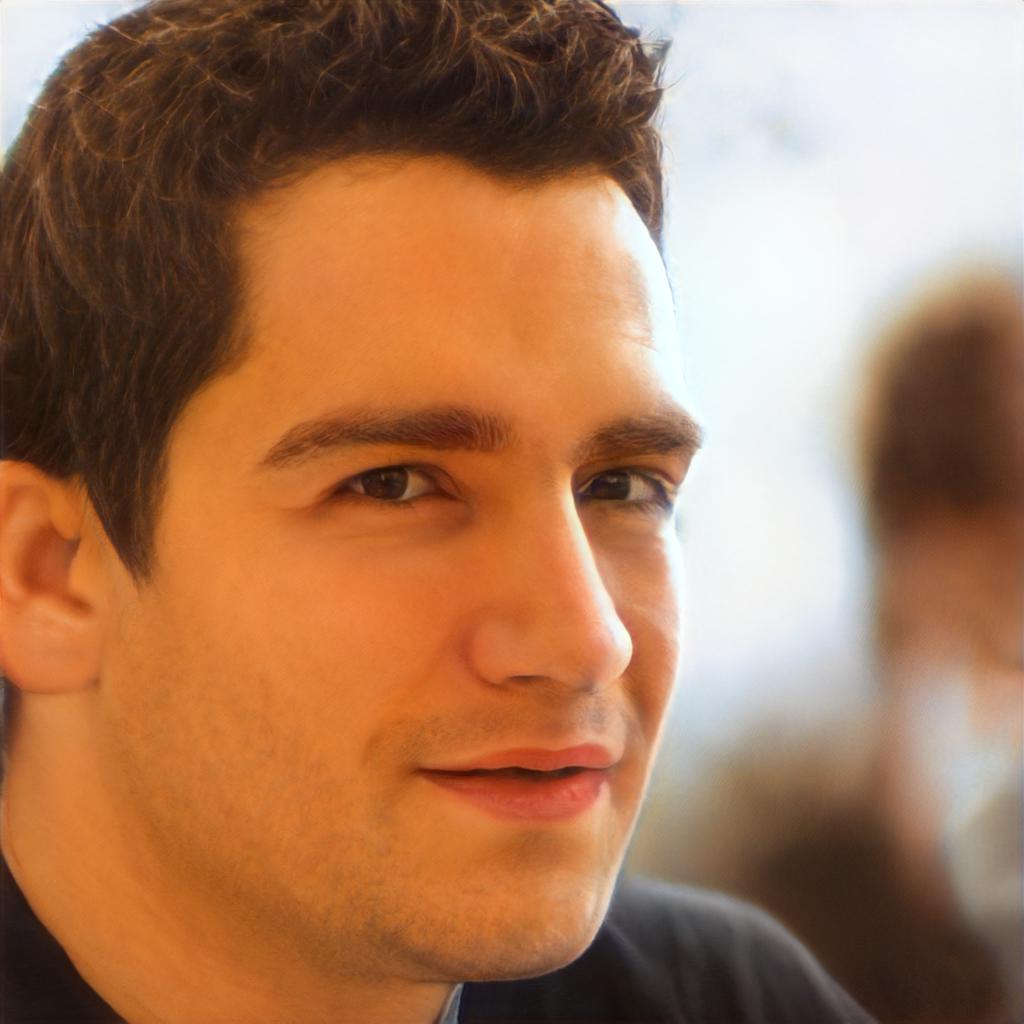} &
        \includegraphics[width=0.14\textwidth]{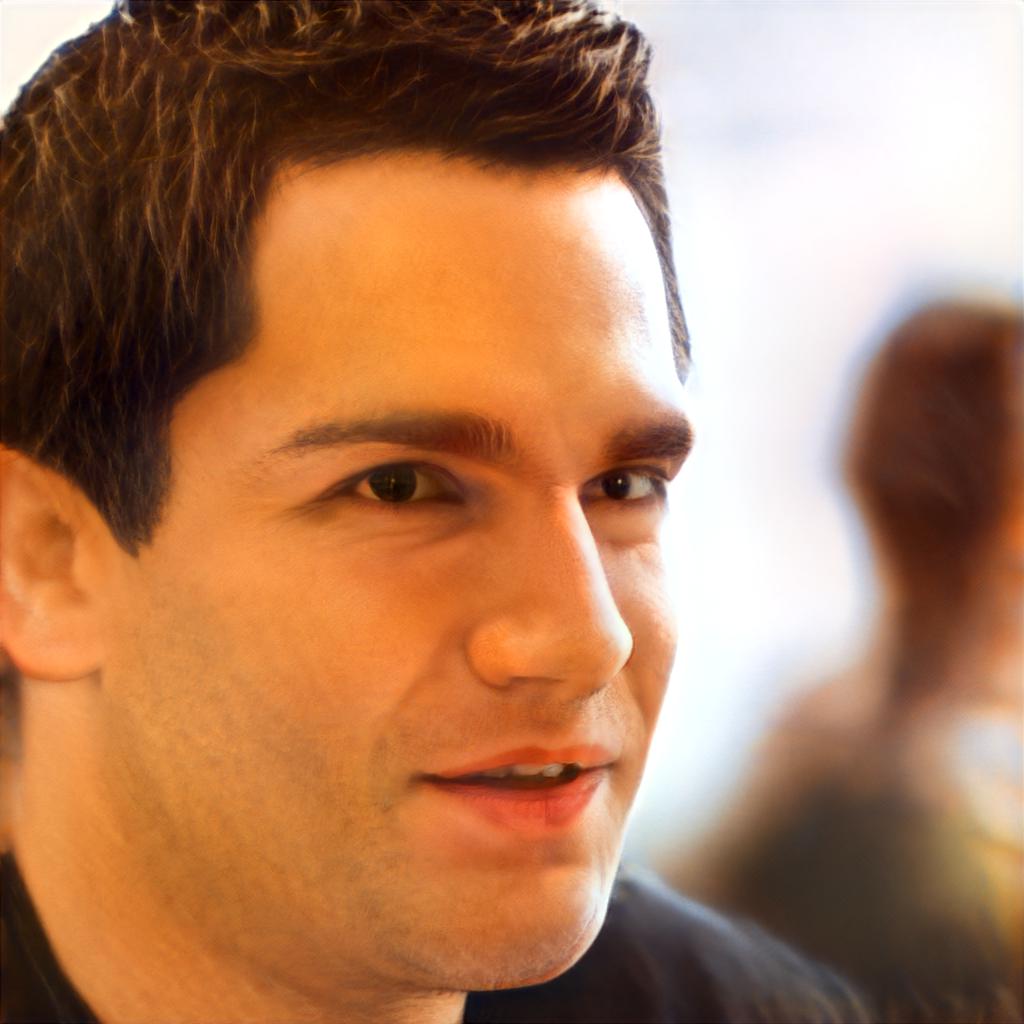} \\
        
        \includegraphics[width=0.14\textwidth]{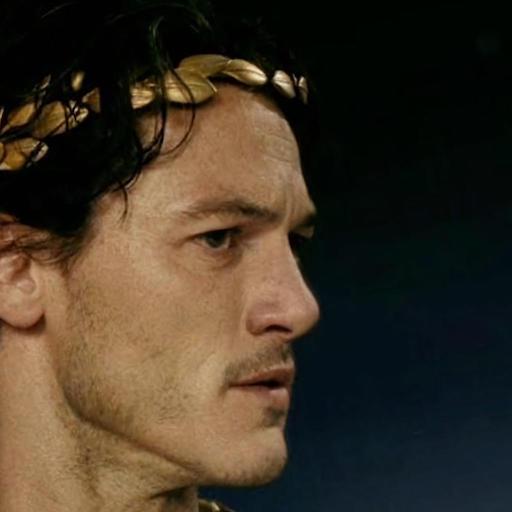} &
        \includegraphics[width=0.14\textwidth]{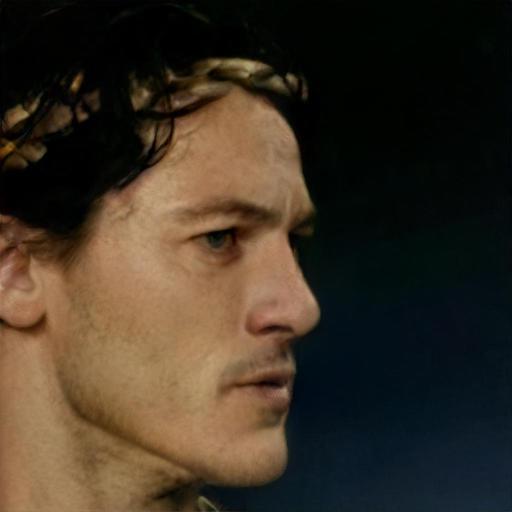} &
        \includegraphics[width=0.14\textwidth]{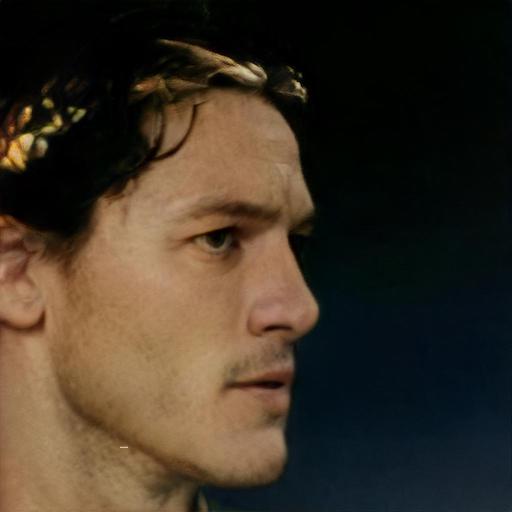} &
        \includegraphics[width=0.14\textwidth]{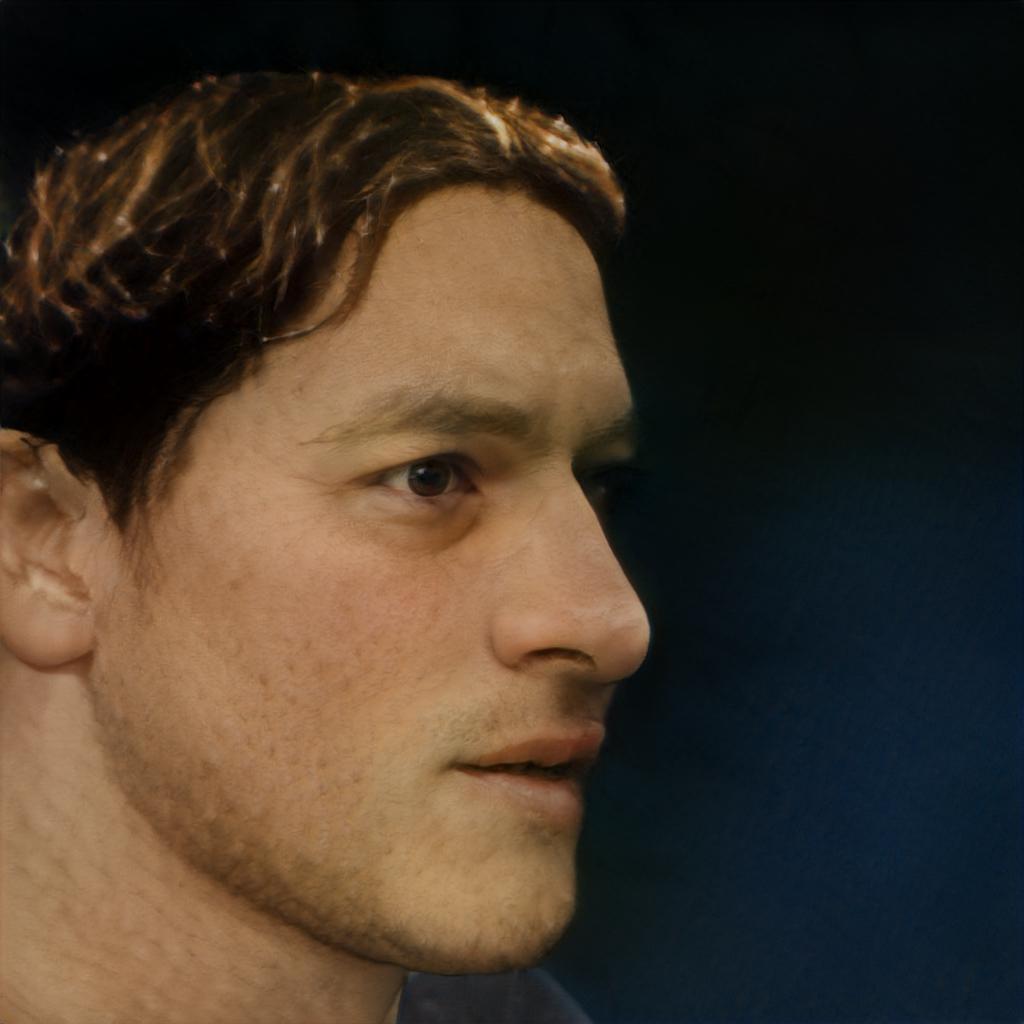} &
        \includegraphics[width=0.14\textwidth]{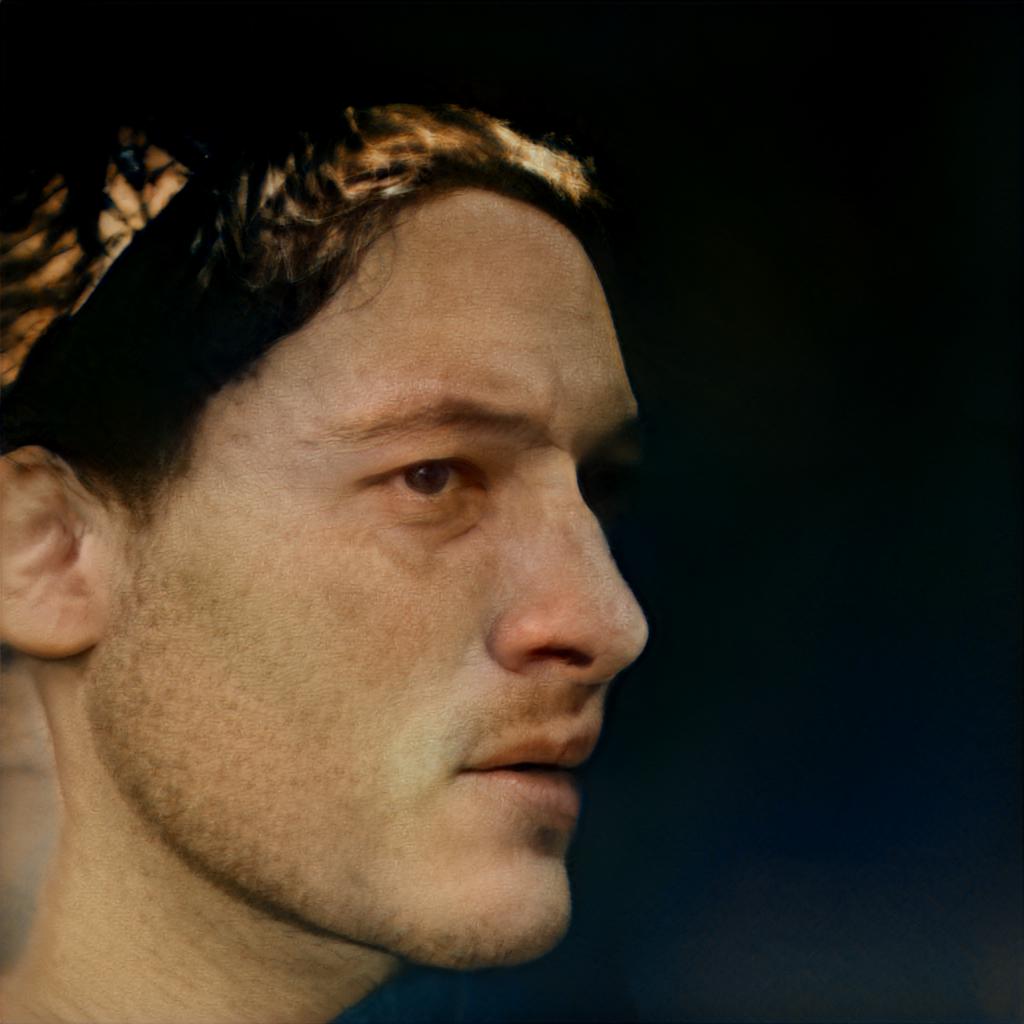} \\
        
        \includegraphics[width=0.14\textwidth]{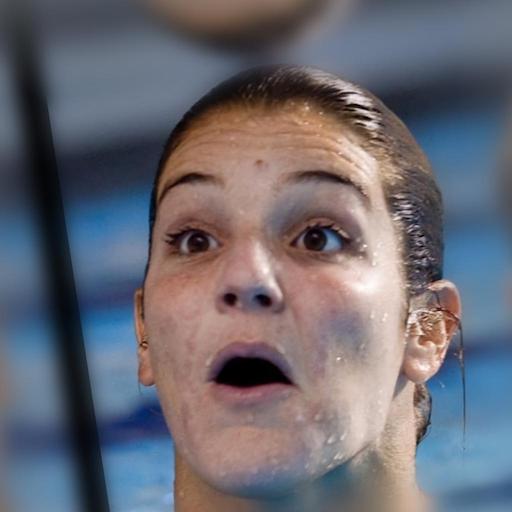} &
        \includegraphics[width=0.14\textwidth]{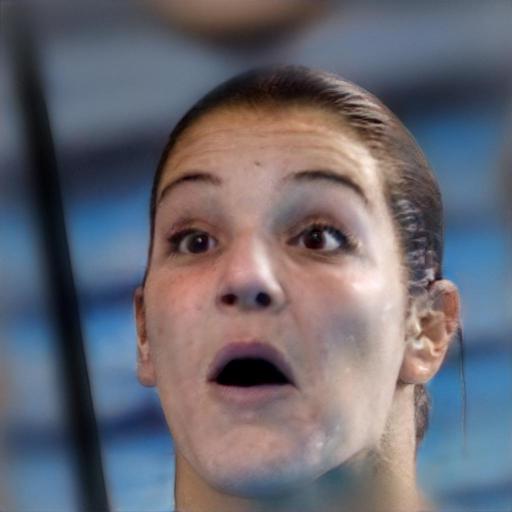} &
        \includegraphics[width=0.14\textwidth]{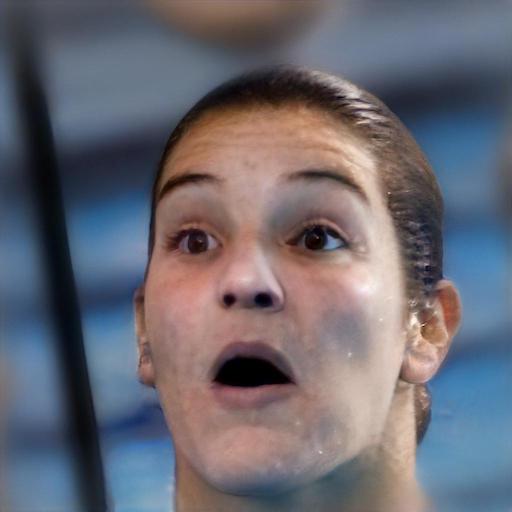} &
        \includegraphics[width=0.14\textwidth]{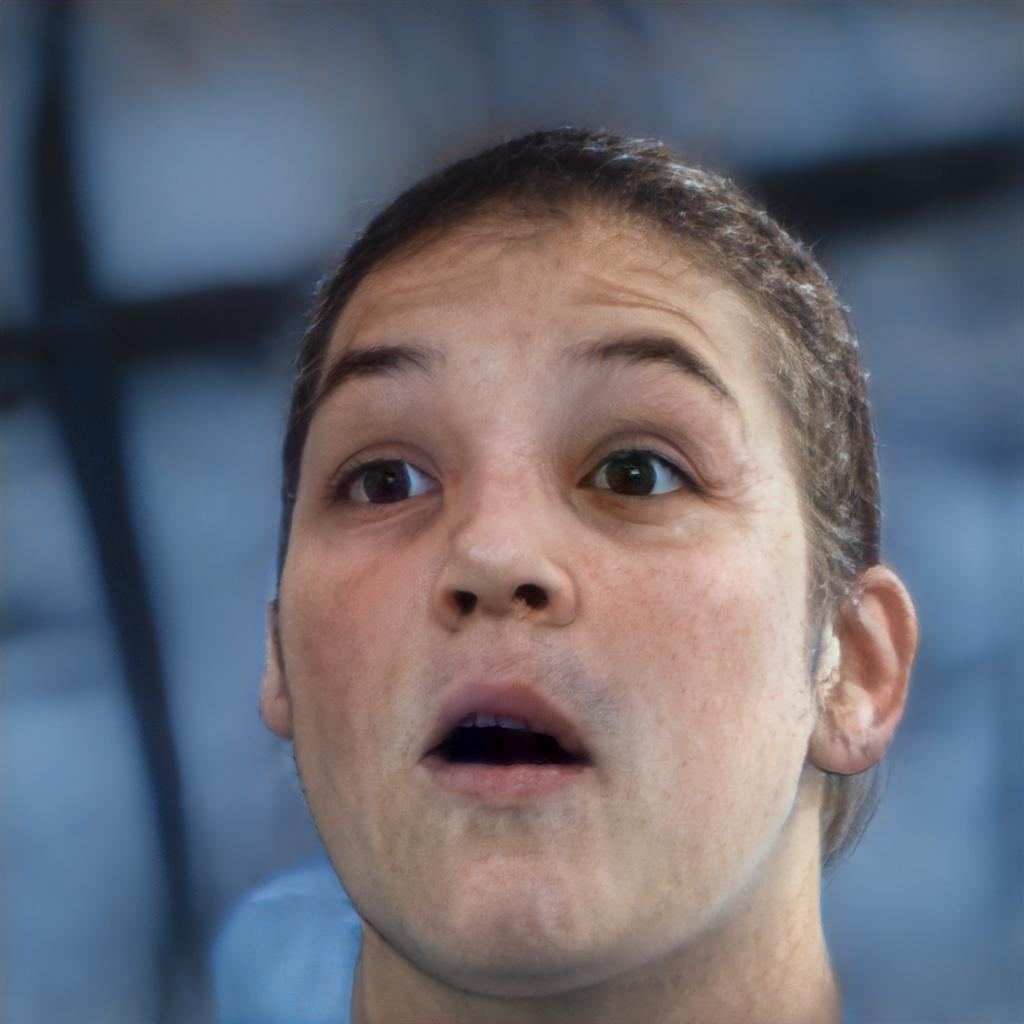} &
        \includegraphics[width=0.14\textwidth]{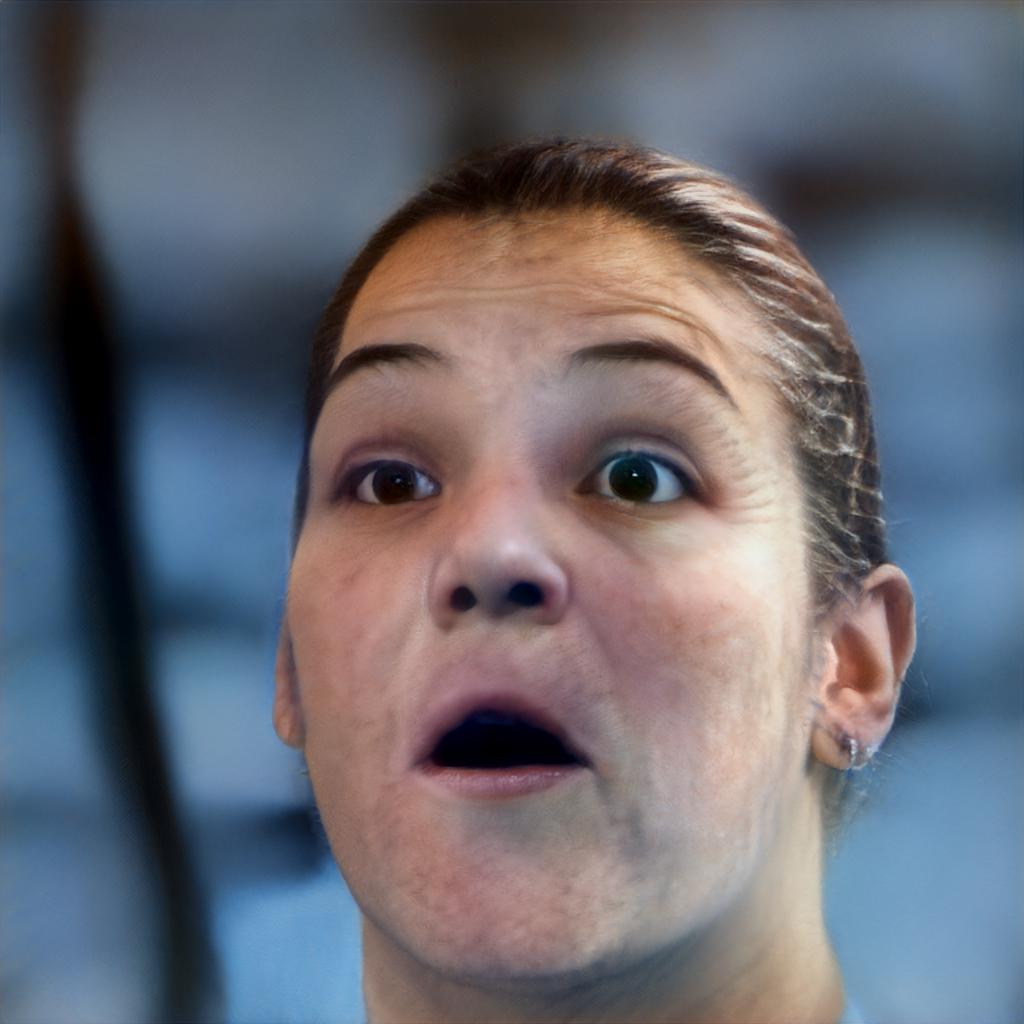} \\
\end{tabular}
    }
    \vspace{0.35cm}
    \caption{Additional qualitative comparisons on the CelebA-HQ~\cite{liu2015deep, karras2017progressive} test set on more challenging input poses and expressions. Here, we apply ReStyle over the pSp encoder~\cite{richardson2020encoding}.}       \label{fig:faces_comparison_challenging_poses}
\end{figure*}
\begin{figure*}
    \centering
    \setlength{\belowcaptionskip}{-6pt}
    \setlength{\tabcolsep}{1pt}
    {\small
    \begin{tabular}{c c c c c}

        Input & Optimization & Hybrid & pSp & $\text{ReStyle}_{pSp}$ \\

        \includegraphics[width=0.15\textwidth]{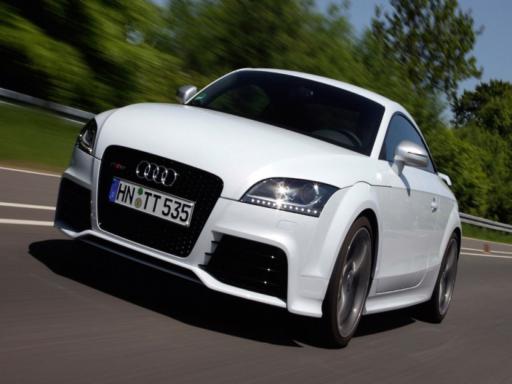} &
        \includegraphics[width=0.15\textwidth]{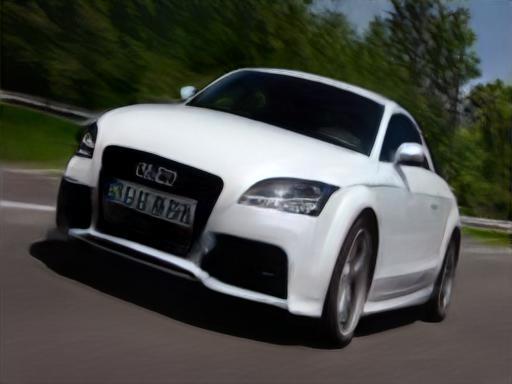} &
        \includegraphics[width=0.15\textwidth]{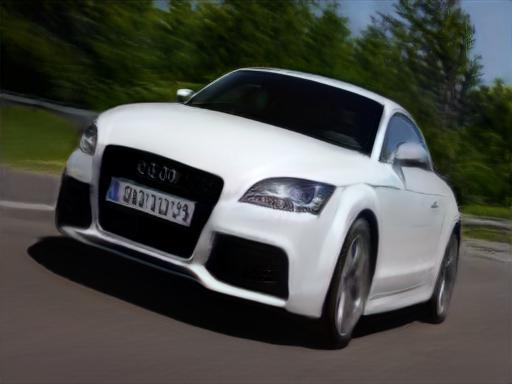} &
        \includegraphics[width=0.15\textwidth]{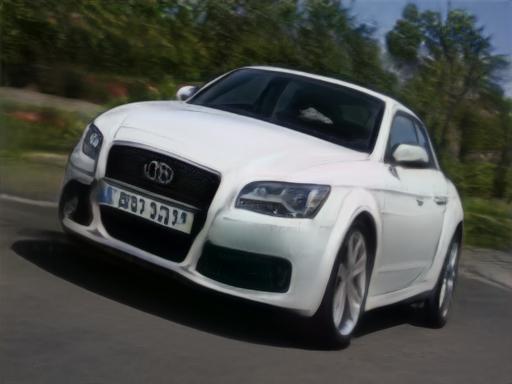} &
        \includegraphics[width=0.15\textwidth]{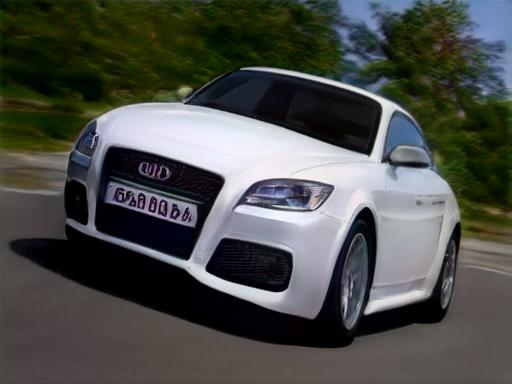} \\

        \includegraphics[width=0.15\textwidth]{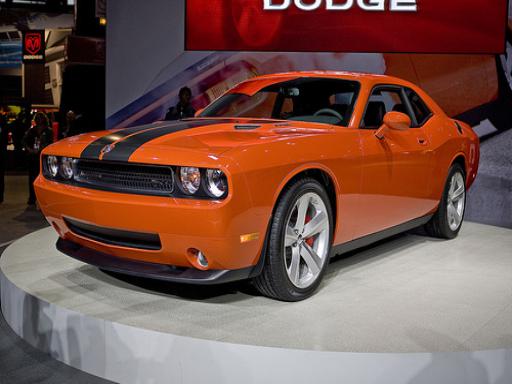} &
        \includegraphics[width=0.15\textwidth]{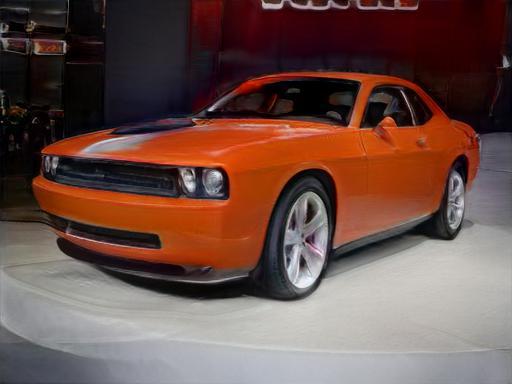} &
        \includegraphics[width=0.15\textwidth]{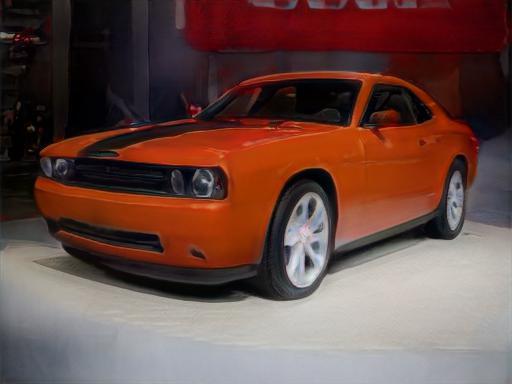} &
        \includegraphics[width=0.15\textwidth]{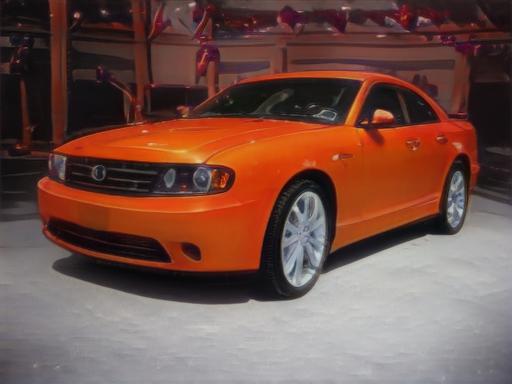} &
        \includegraphics[width=0.15\textwidth]{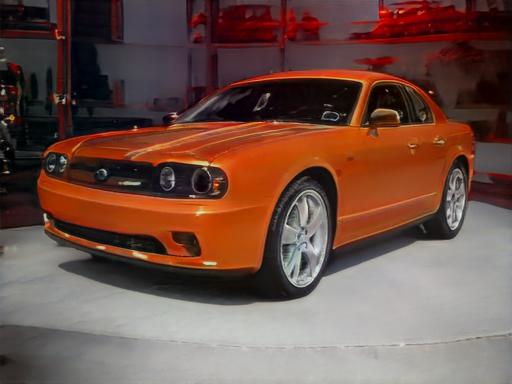} \\
        
        \includegraphics[width=0.15\textwidth]{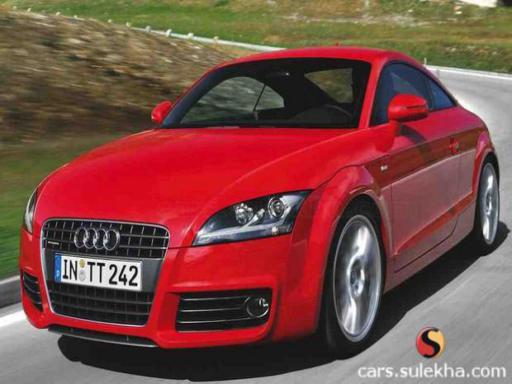} &
        \includegraphics[width=0.15\textwidth]{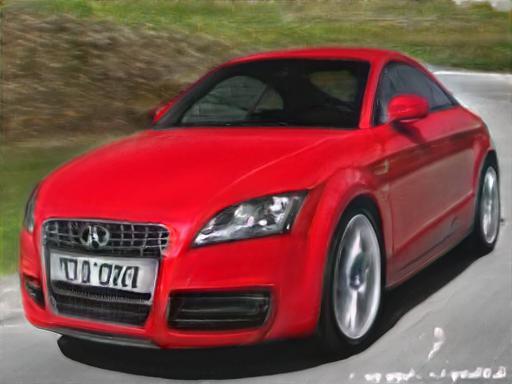} &
        \includegraphics[width=0.15\textwidth]{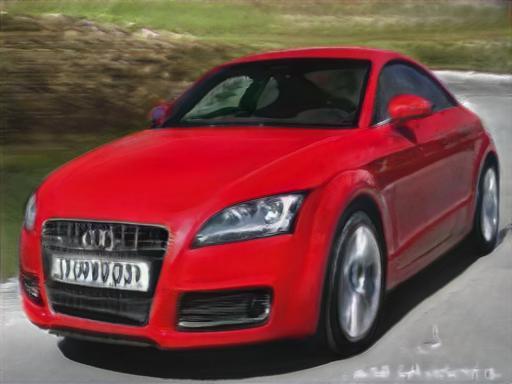} &
        \includegraphics[width=0.15\textwidth]{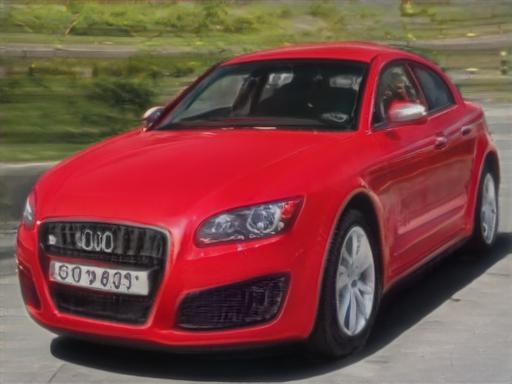} &
        \includegraphics[width=0.15\textwidth]{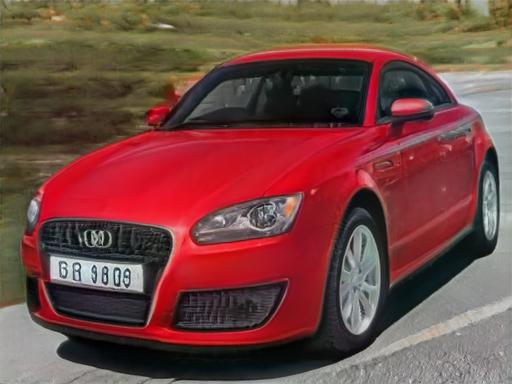} \\
\end{tabular}
    }
    \vspace{0.45cm}
    \centering
    \caption{Additional qualitative comparisons on the Stanford Cars~\cite{KrauseStarkDengFei-Fei_3DRR2013} test set. Here, we apply ReStyle over the pSp encoder~\cite{richardson2020encoding}.}    \label{fig:cars_comparison_psp}
\end{figure*}
\begin{figure*}
    \centering
    \setlength{\belowcaptionskip}{-6pt}
    \setlength{\tabcolsep}{1pt}
    {\small
    \begin{tabular}{c c c c c}

        Input & Optimization & Hybrid & e4e & $\text{ReStyle}_{e4e}$ \\

        \includegraphics[width=0.15\textwidth]{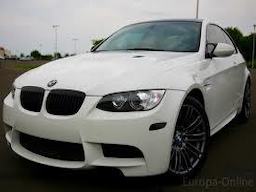} &
        \includegraphics[width=0.15\textwidth]{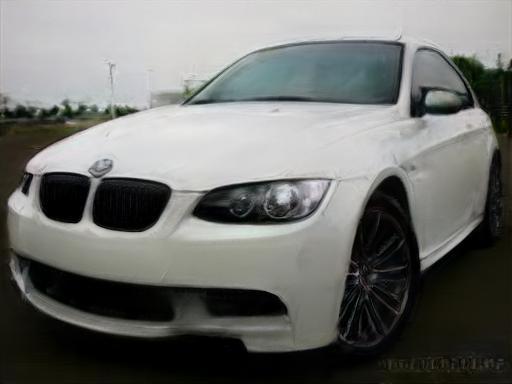} &
        \includegraphics[width=0.15\textwidth]{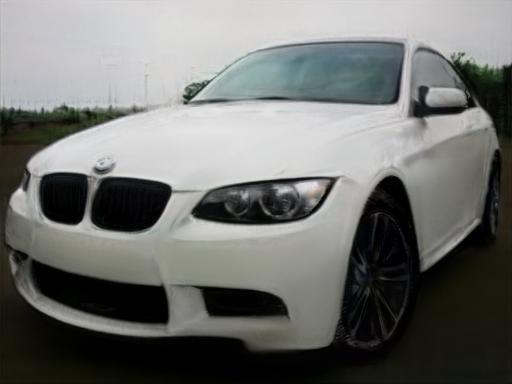} &
        \includegraphics[width=0.15\textwidth]{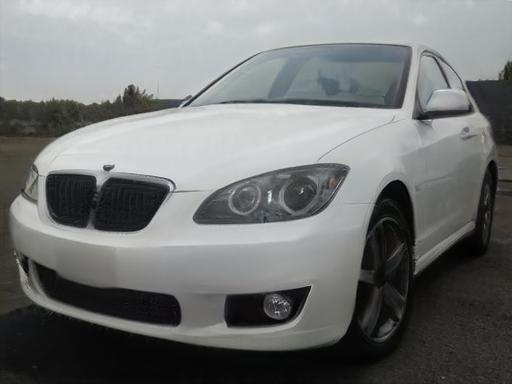} &
        \includegraphics[width=0.15\textwidth]{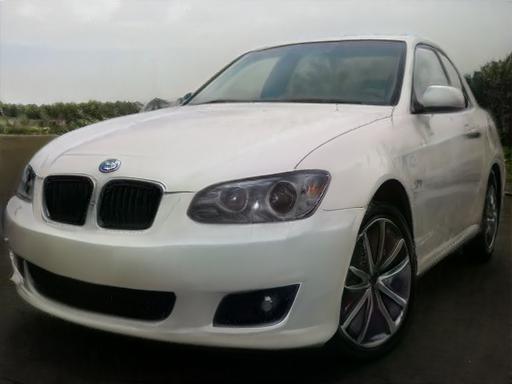} \\

        \includegraphics[width=0.15\textwidth]{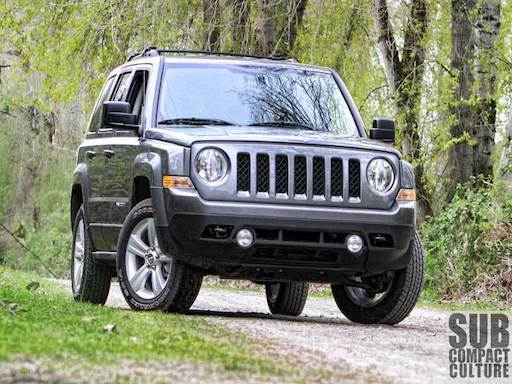} &
        \includegraphics[width=0.15\textwidth]{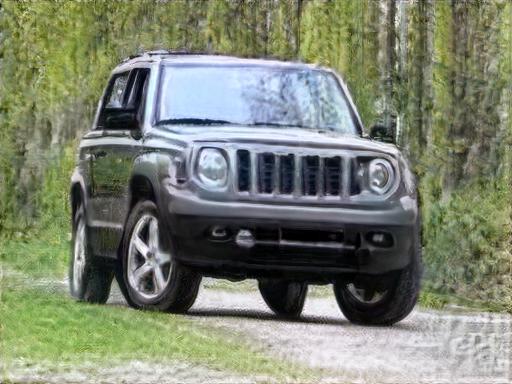} &
        \includegraphics[width=0.15\textwidth]{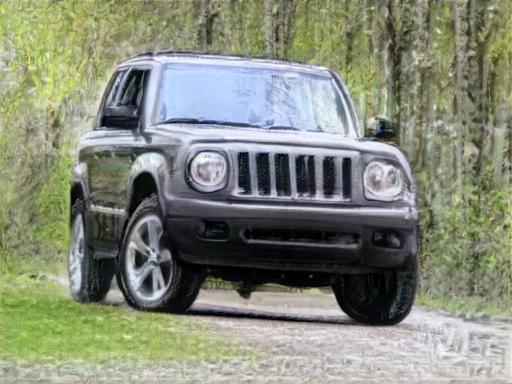} &
        \includegraphics[width=0.15\textwidth]{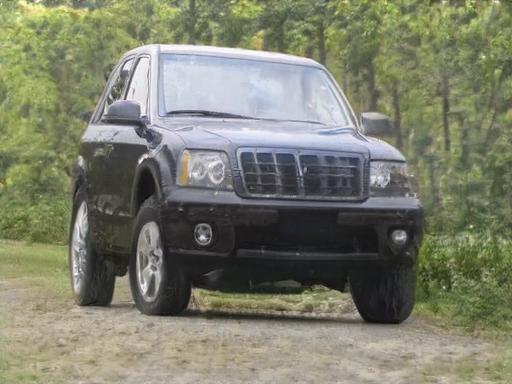} &
        \includegraphics[width=0.15\textwidth]{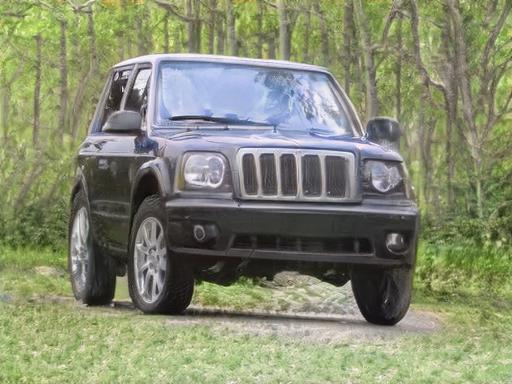} \\
        
        \includegraphics[width=0.15\textwidth]{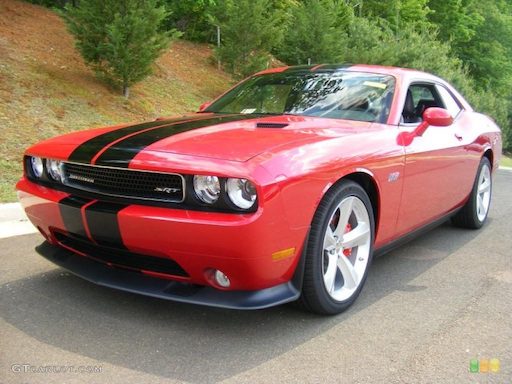} &
        \includegraphics[width=0.15\textwidth]{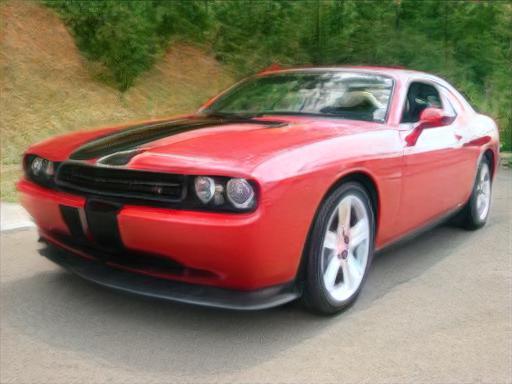} &
        \includegraphics[width=0.15\textwidth]{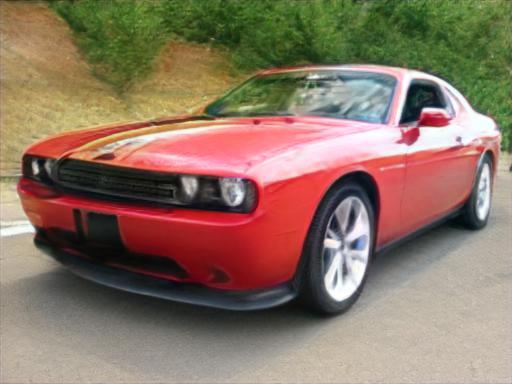} &
        \includegraphics[width=0.15\textwidth]{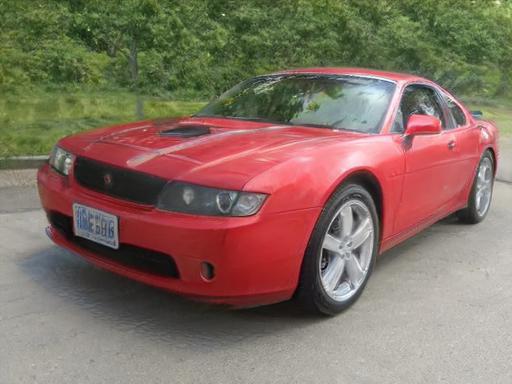} &
        \includegraphics[width=0.15\textwidth]{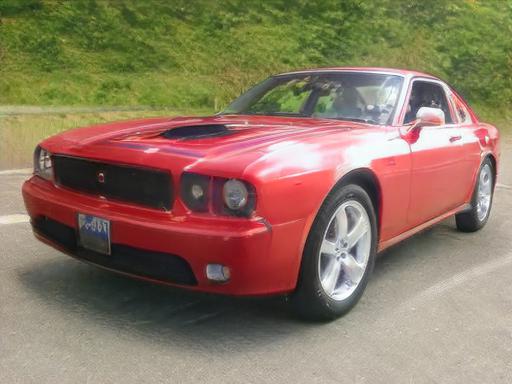} \\
\end{tabular}
    }
    \vspace{0.45cm}
    \caption{Additional qualitative comparisons on the Stanford Cars~\cite{KrauseStarkDengFei-Fei_3DRR2013} test set. Here, we apply ReStyle over the e4e encoder~\cite{tov2021designing}.}   
    \label{fig:cars_comparison_e4e}
\end{figure*}
\begin{figure*}
    \centering
    \setlength{\belowcaptionskip}{-6pt}
    \setlength{\tabcolsep}{1pt}
    {\small
    \begin{tabular}{c c c c c}

        Input & Optimization & Hybrid & pSp & $\text{ReStyle}_{pSp}$ \\

        \includegraphics[width=0.14\textwidth]{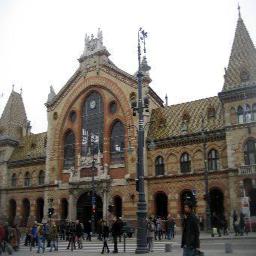} &
        \includegraphics[width=0.14\textwidth]{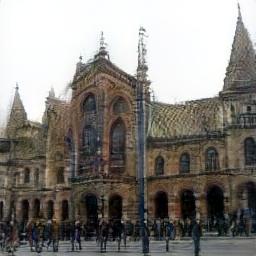} &
        \includegraphics[width=0.14\textwidth]{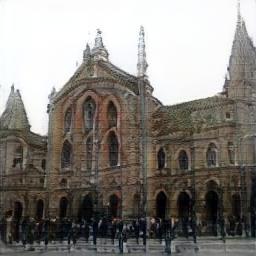} &
        \includegraphics[width=0.14\textwidth]{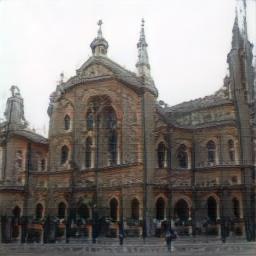} &
        \includegraphics[width=0.14\textwidth]{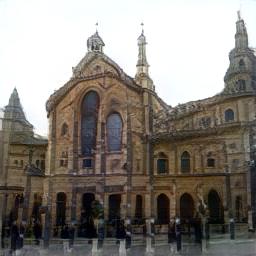} \\

        \includegraphics[width=0.14\textwidth]{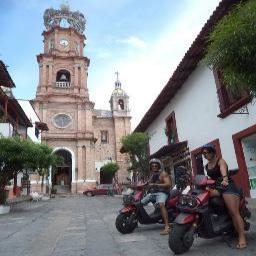} &
        \includegraphics[width=0.14\textwidth]{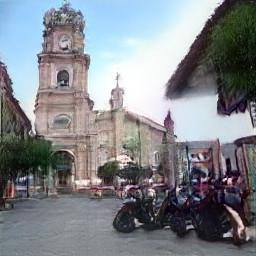} &
        \includegraphics[width=0.14\textwidth]{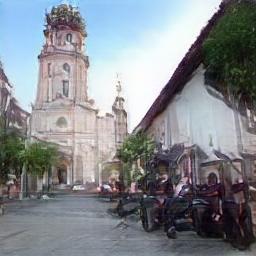} &
        \includegraphics[width=0.14\textwidth]{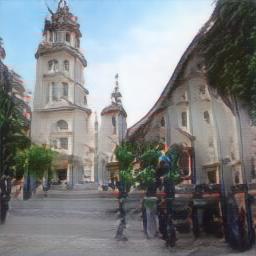} &
        \includegraphics[width=0.14\textwidth]{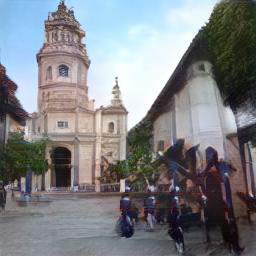} \\
\end{tabular}
    }
    \vspace{0.45cm}
    \caption{Additional qualitative comparisons on the LSUN~\cite{yu2016lsun} Church test set. Here, we apply ReStyle over the pSp encoder~\cite{richardson2020encoding}.}
    \label{fig:churches_comparison_psp}
\end{figure*}
\begin{figure*}
    \centering
    \setlength{\belowcaptionskip}{-6pt}
    \setlength{\tabcolsep}{1pt}
    {\small
    \begin{tabular}{c c c c c}

        Input & Optimization & Hybrid & e4e & $\text{ReStyle}_{e4e}$ \\

        \includegraphics[width=0.14\textwidth]{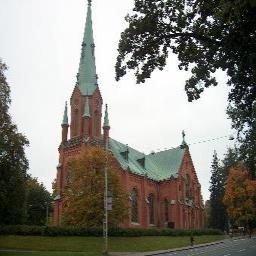} &
        \includegraphics[width=0.14\textwidth]{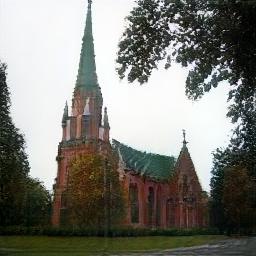} &
        \includegraphics[width=0.14\textwidth]{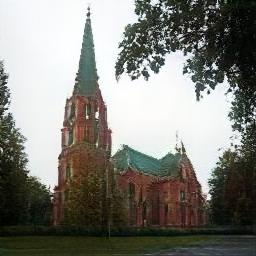} &
        \includegraphics[width=0.14\textwidth]{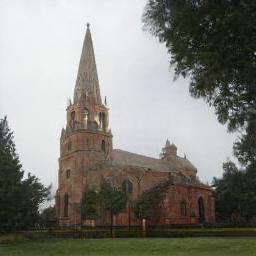} &
        \includegraphics[width=0.14\textwidth]{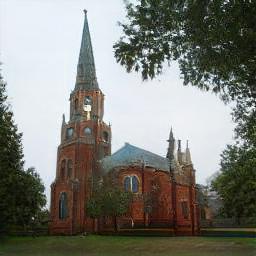} \\

        \includegraphics[width=0.14\textwidth]{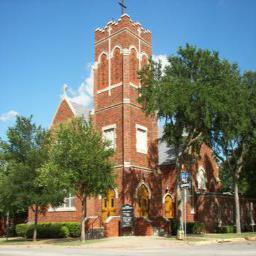} &
        \includegraphics[width=0.14\textwidth]{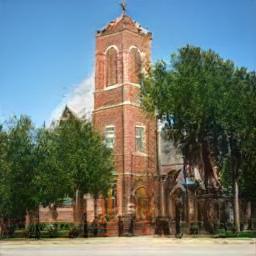} &
        \includegraphics[width=0.14\textwidth]{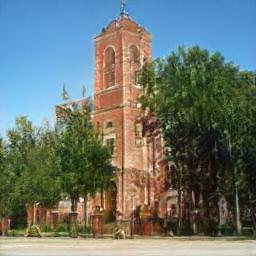} &
        \includegraphics[width=0.14\textwidth]{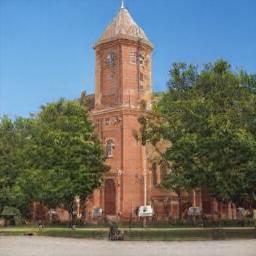} &
        \includegraphics[width=0.14\textwidth]{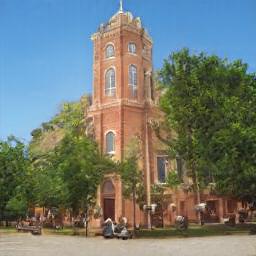} \\
\end{tabular}
    }
    \vspace{0.5cm}
    \centering
    \caption{Additional qualitative comparisons on the LSUN~\cite{yu2016lsun} Church test set. Here, we apply ReStyle over the e4e encoder~\cite{tov2021designing}.}    
    \label{fig:churches_comparison_e4e}
\end{figure*}
\begin{figure*}
    \centering
    \setlength{\belowcaptionskip}{-6pt}
    \setlength{\tabcolsep}{1pt}
    {\small
    \begin{tabular}{c c c c c}

        Input & Optimization & Hybrid & pSp & $\text{ReStyle}_{pSp}$ \\

        \includegraphics[width=0.14\textwidth]{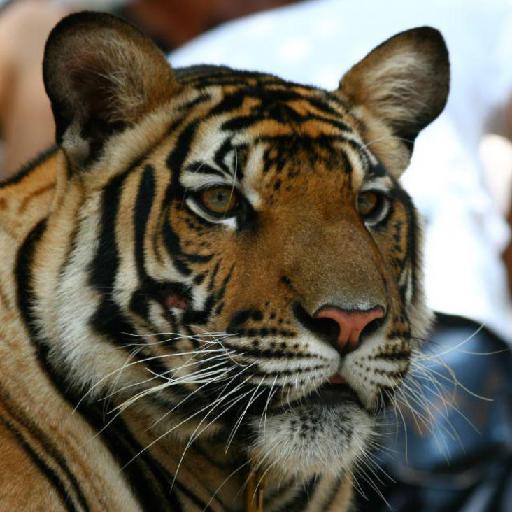} &
        \includegraphics[width=0.14\textwidth]{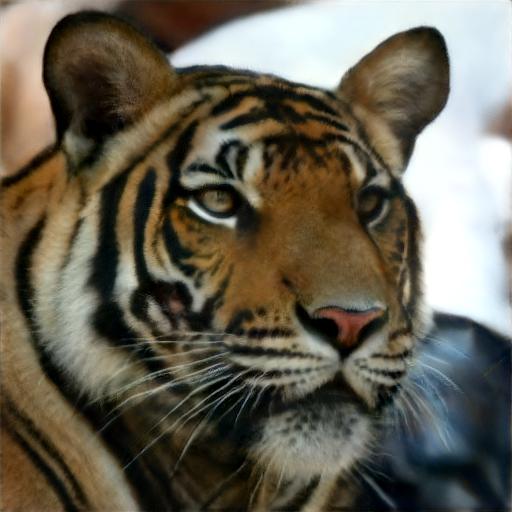} &
        \includegraphics[width=0.14\textwidth]{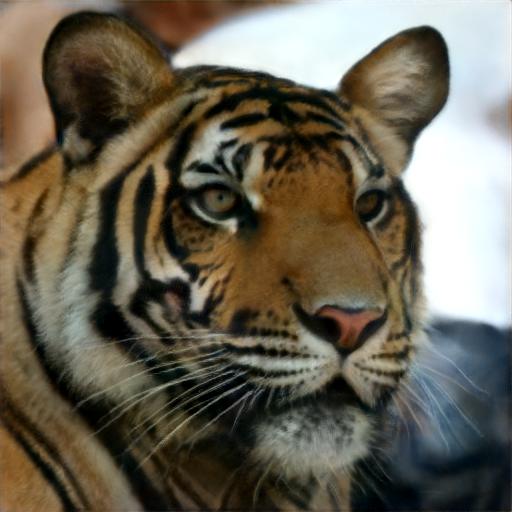} &
        \includegraphics[width=0.14\textwidth]{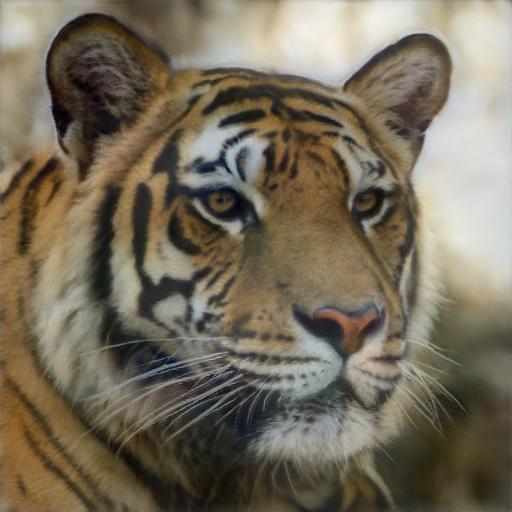} &
        \includegraphics[width=0.14\textwidth]{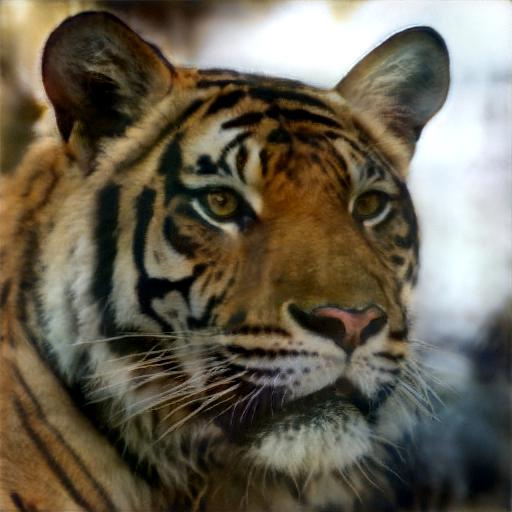} \\

        \includegraphics[width=0.14\textwidth]{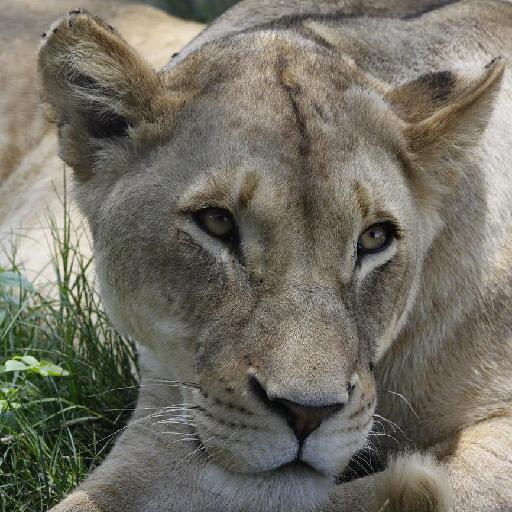} &
        \includegraphics[width=0.14\textwidth]{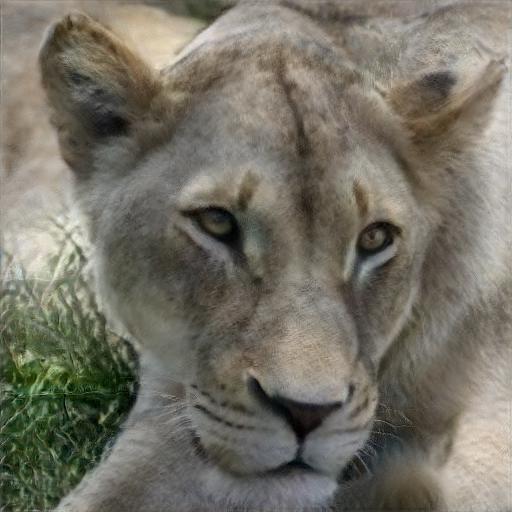} &
        \includegraphics[width=0.14\textwidth]{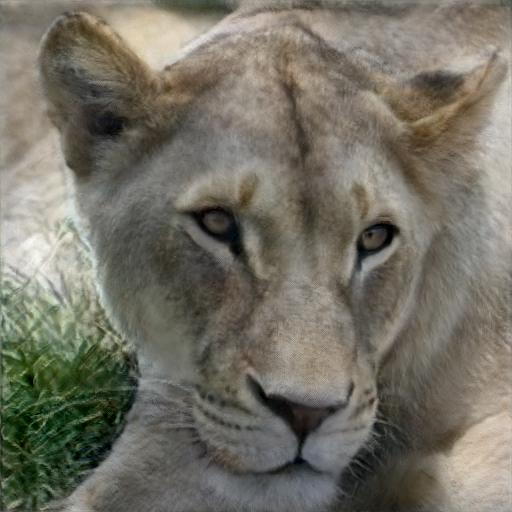} &
        \includegraphics[width=0.14\textwidth]{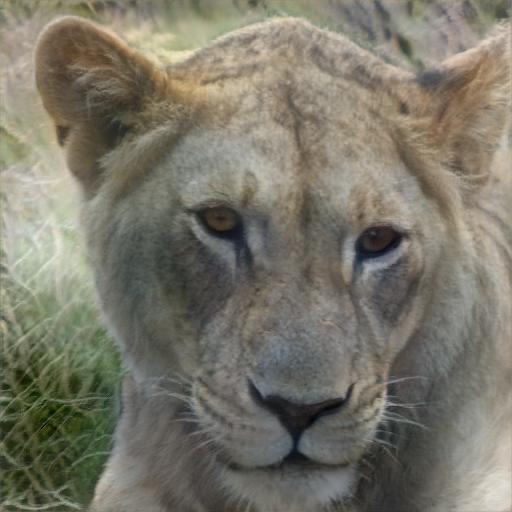} &
        \includegraphics[width=0.14\textwidth]{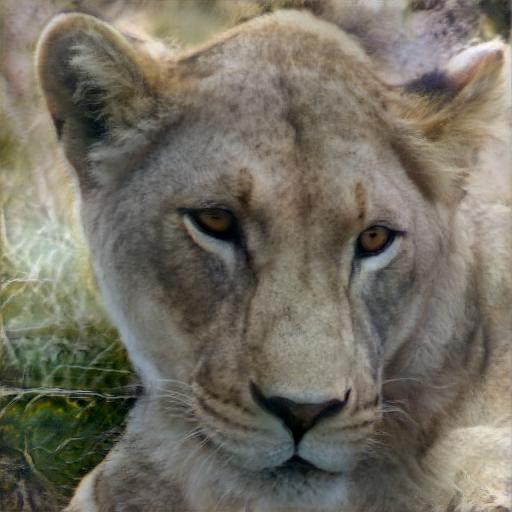} \\
        
        \includegraphics[width=0.14\textwidth]{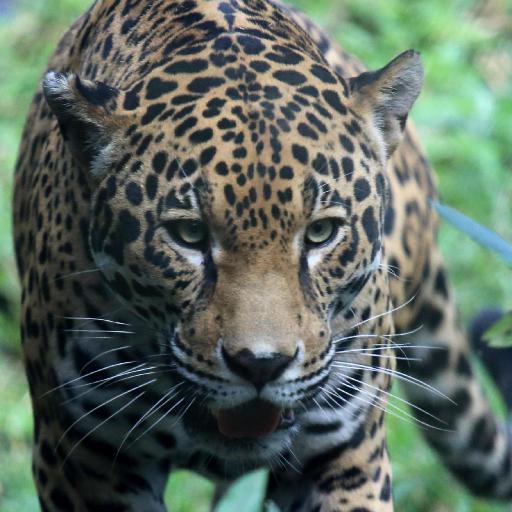} &
        \includegraphics[width=0.14\textwidth]{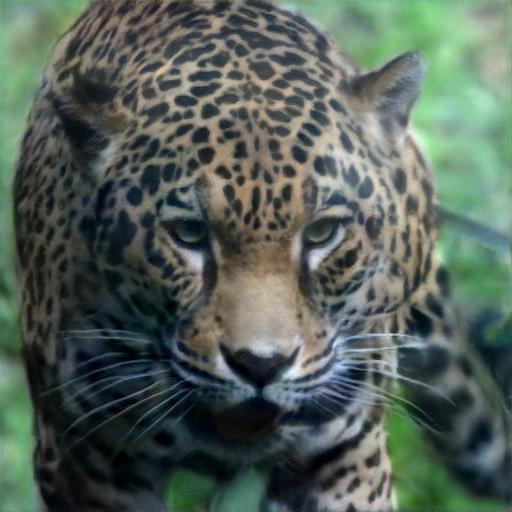} &
        \includegraphics[width=0.14\textwidth]{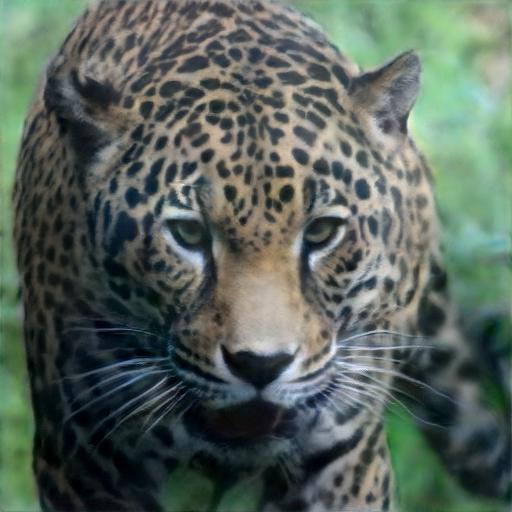} &
        \includegraphics[width=0.14\textwidth]{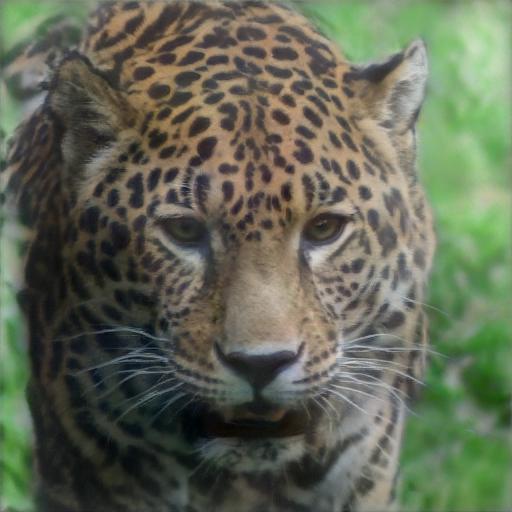} &
        \includegraphics[width=0.14\textwidth]{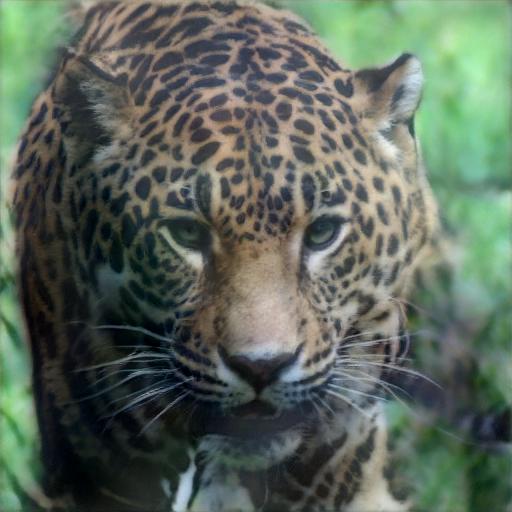} \\
\end{tabular}
    }
    \vspace{0.45cm}
    \caption{Additional qualitative comparisons on the AFHQ~\cite{choi2020stargan} Wild test set. Here, we apply ReStyle over the pSp encoder~\cite{richardson2020encoding}.}
    \label{fig:wild_comparison_psp}
\end{figure*}
\begin{figure*}
    \centering
    \setlength{\belowcaptionskip}{-6pt}
    \setlength{\tabcolsep}{1pt}
    {\small
    \begin{tabular}{c c c c c}

        Input & Optimization & Hybrid & e4e & $\text{ReStyle}_{e4e}$ \\

        \includegraphics[width=0.14\textwidth]{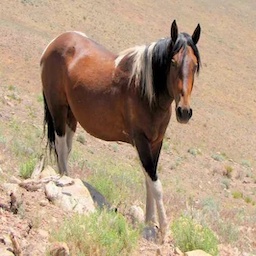} &
        \includegraphics[width=0.14\textwidth]{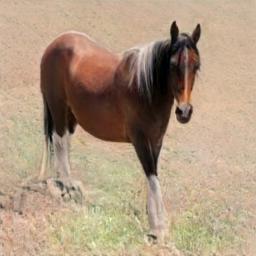} &
        \includegraphics[width=0.14\textwidth]{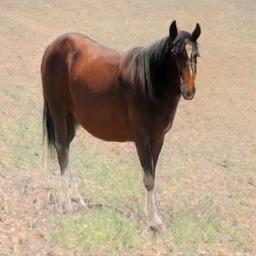} &
        \includegraphics[width=0.14\textwidth]{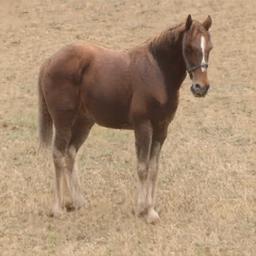} &
        \includegraphics[width=0.14\textwidth]{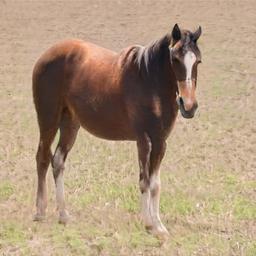} \\

        \includegraphics[width=0.14\textwidth]{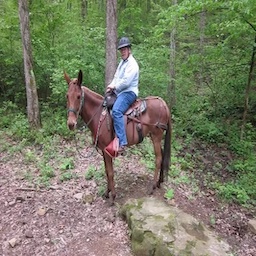} &
        \includegraphics[width=0.14\textwidth]{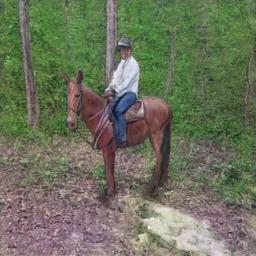} &
        \includegraphics[width=0.14\textwidth]{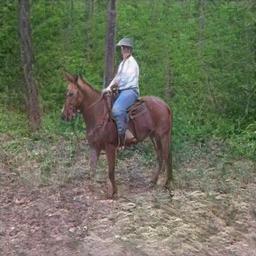} &
        \includegraphics[width=0.14\textwidth]{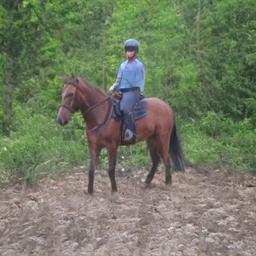} &
        \includegraphics[width=0.14\textwidth]{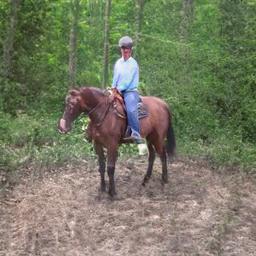} \\
        
        \includegraphics[width=0.14\textwidth]{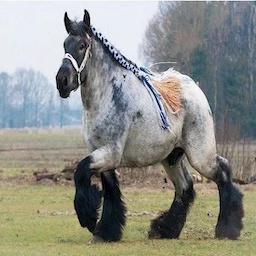} &
        \includegraphics[width=0.14\textwidth]{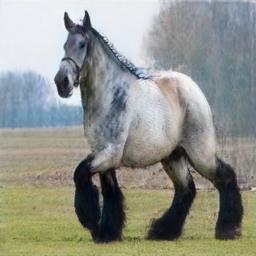} &
        \includegraphics[width=0.14\textwidth]{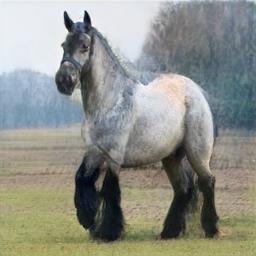} &
        \includegraphics[width=0.14\textwidth]{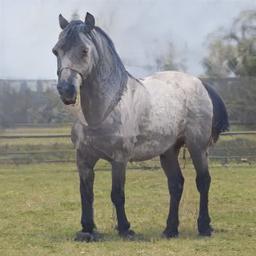} &
        \includegraphics[width=0.14\textwidth]{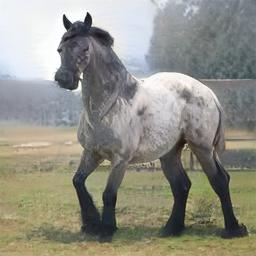} \\
\end{tabular}
    }
    \vspace{0.45cm}
    \caption{Additional qualitative comparisons on the LSUN~\cite{yu2016lsun} Horse test set. Here, we apply ReStyle over the e4e encoder~\cite{tov2021designing}.}
    \label{fig:horses_comparison_e4e}
\end{figure*}
\begin{figure*}
    \centering
    \setlength{\belowcaptionskip}{-6pt}
    \setlength{\tabcolsep}{1pt}
    {\small
    \begin{tabular}{c c c c c c}

        \includegraphics[width=0.14\textwidth]{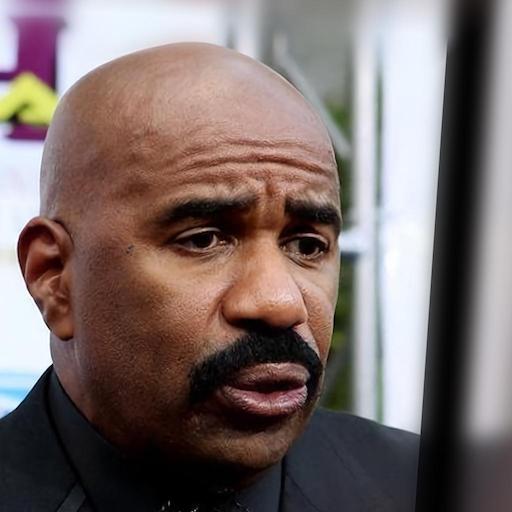} &
        \includegraphics[width=0.14\textwidth]{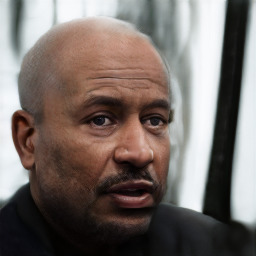} &
        \includegraphics[width=0.14\textwidth]{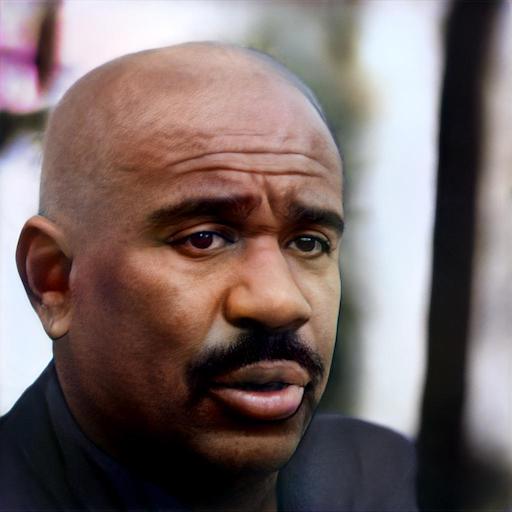} &
        \hspace{0.5cm}
        \includegraphics[width=0.14\textwidth]{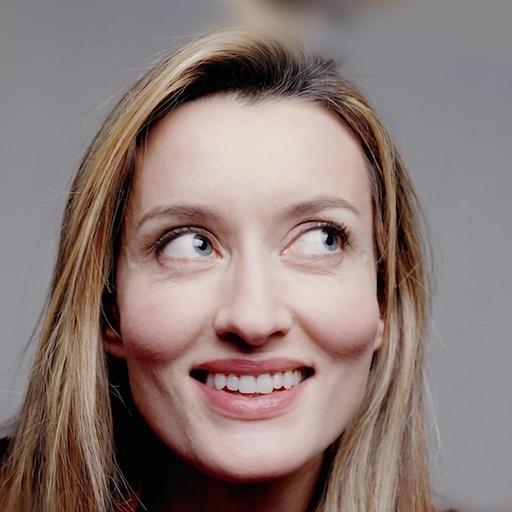} &
        \includegraphics[width=0.14\textwidth]{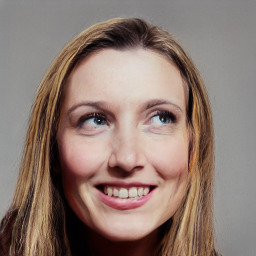} &
        \includegraphics[width=0.14\textwidth]{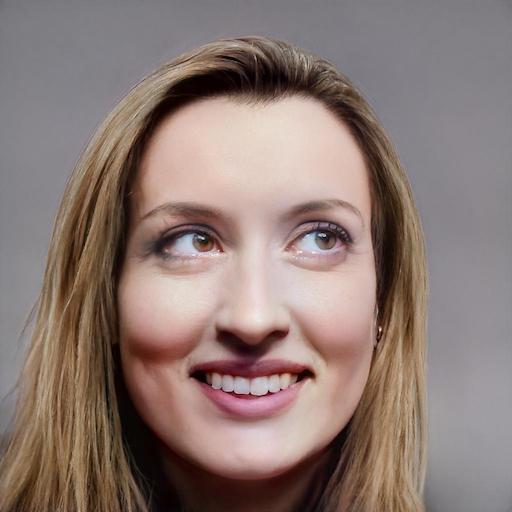} \\
        
        \includegraphics[width=0.14\textwidth]{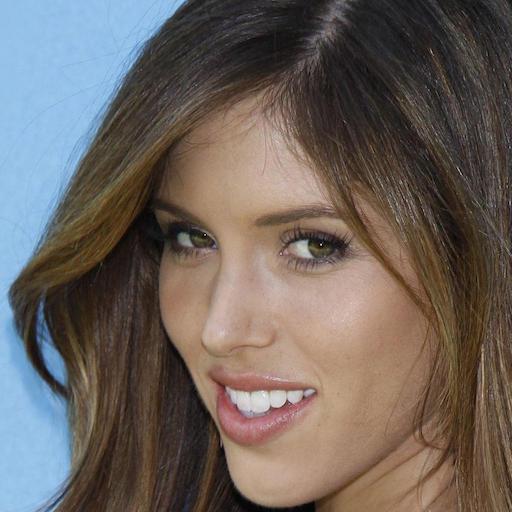} &
        \includegraphics[width=0.14\textwidth]{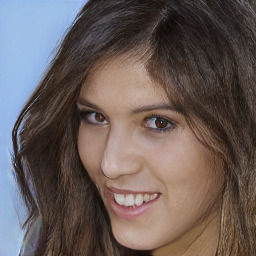} &
        \includegraphics[width=0.14\textwidth]{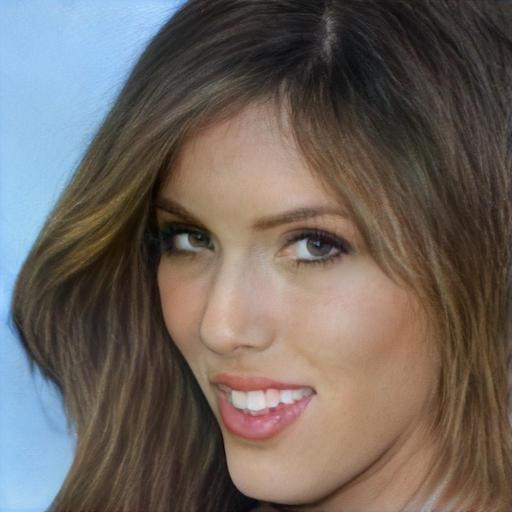} &
        \hspace{0.5cm}
        \includegraphics[width=0.14\textwidth]{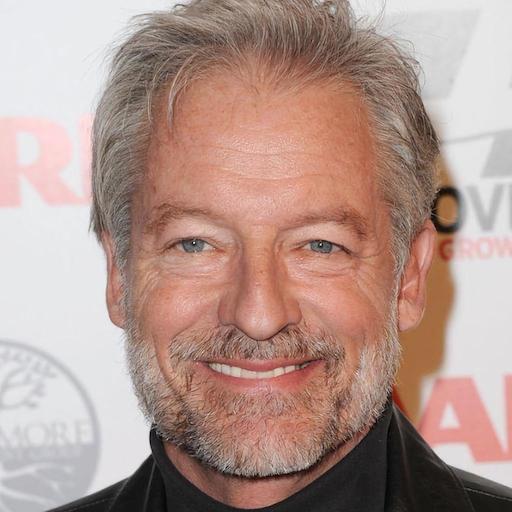} &
        \includegraphics[width=0.14\textwidth]{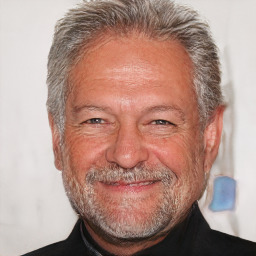} &
        \includegraphics[width=0.14\textwidth]{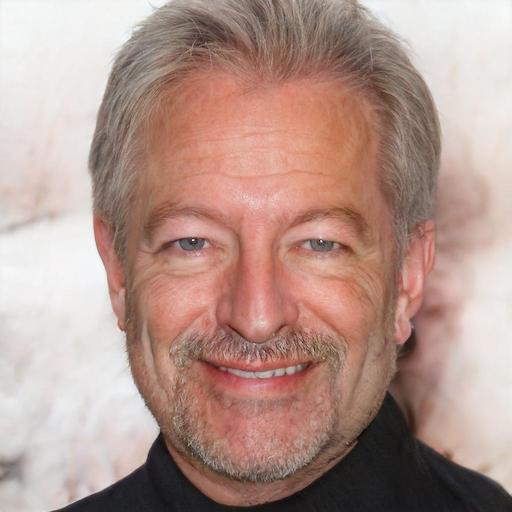} \\
\end{tabular}
    }
    \vspace{0.45cm}
    \caption{Qualitative comparisons with the IDInvert encoder from Zhu \etal~\cite{zhu2020domain} on the
    CelebA-HQ~\cite{liu2015deep, karras2017progressive} and Stanford Cars~\cite{KrauseStarkDengFei-Fei_3DRR2013} test sets. Here, we show results with ReStyle applied over both the pSp and e4e encoders.}
    \label{fig:idinvert_comparison}
\end{figure*}

\begin{figure*}
    \centering
    \setlength{\belowcaptionskip}{-6pt}
    \setlength{\tabcolsep}{1pt}
    {\small
    \begin{tabular}{c l}

        Input & \qquad Iterative Outputs $\longrightarrow$ \\

        \includegraphics[width=0.14\textwidth]{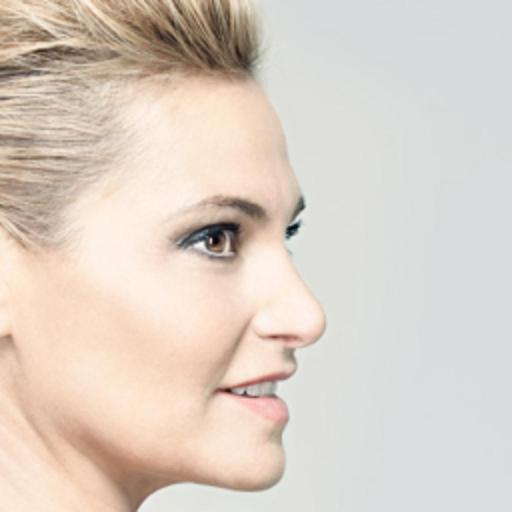} &
        \includegraphics[width=0.7\textwidth]{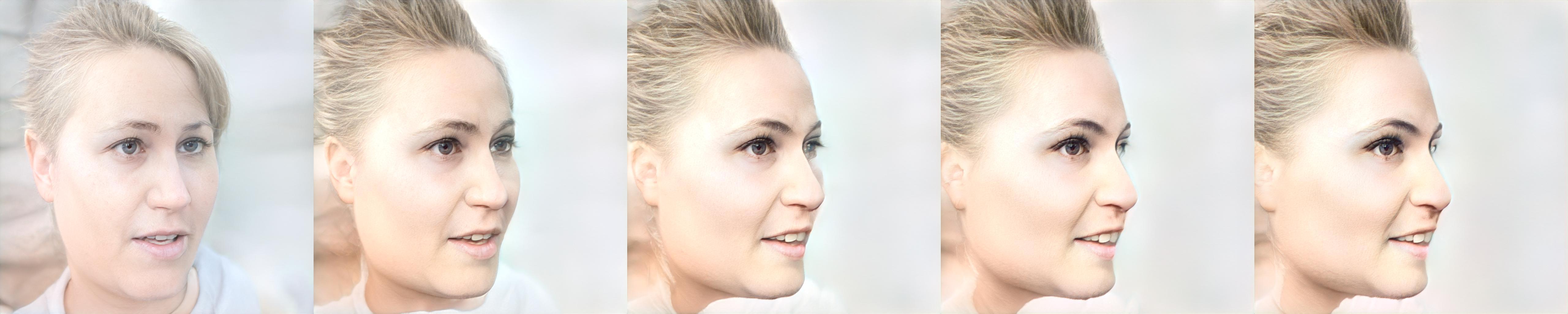} \\

        \includegraphics[width=0.14\textwidth]{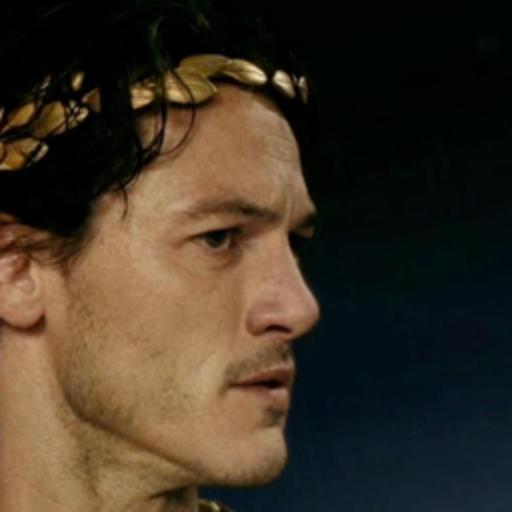} &
        \includegraphics[width=0.7\textwidth]{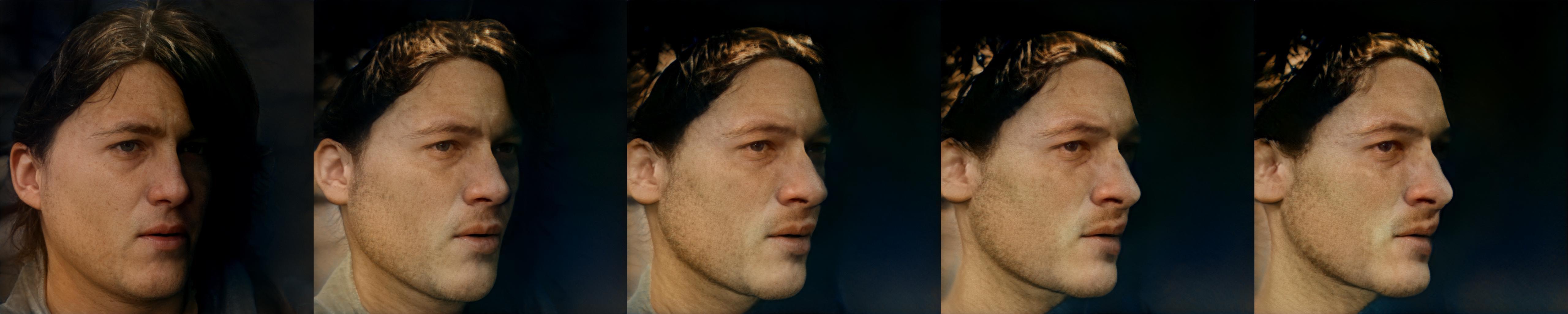} \\
        
        \includegraphics[width=0.14\textwidth]{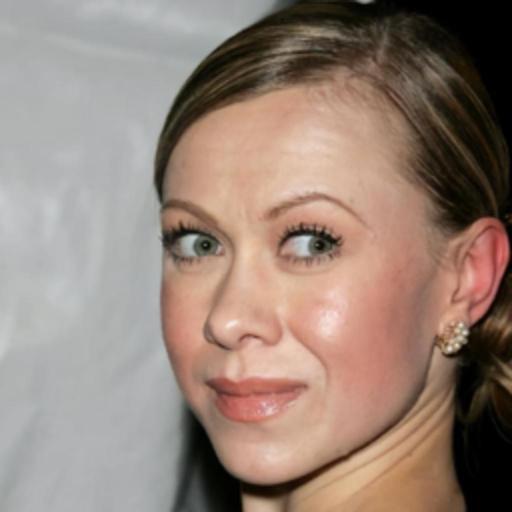} &
        \includegraphics[width=0.7\textwidth]{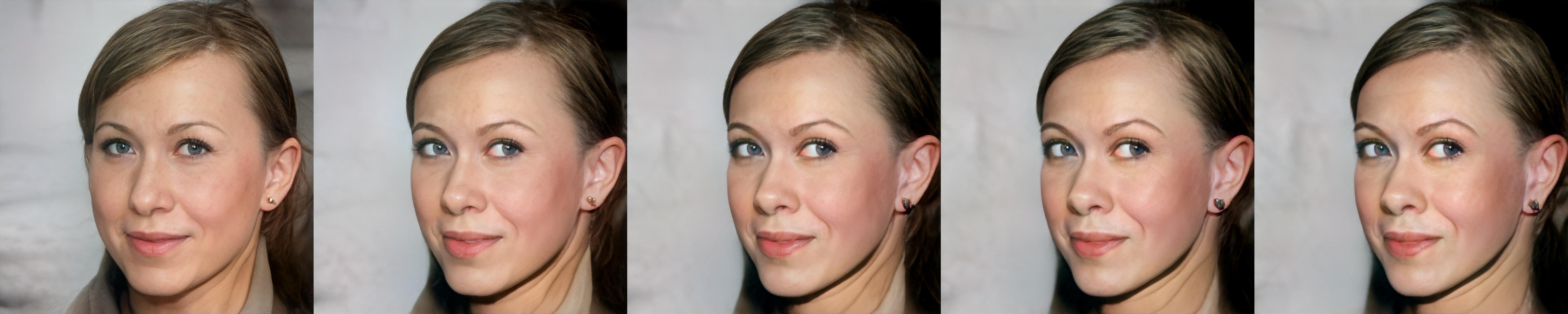} \\
        
        \includegraphics[width=0.14\textwidth]{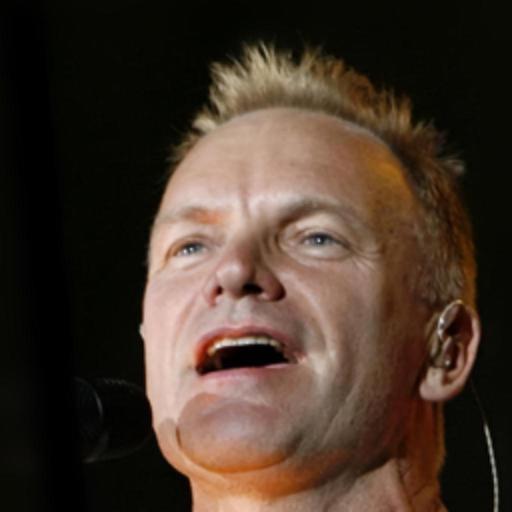} &
        \includegraphics[width=0.7\textwidth]{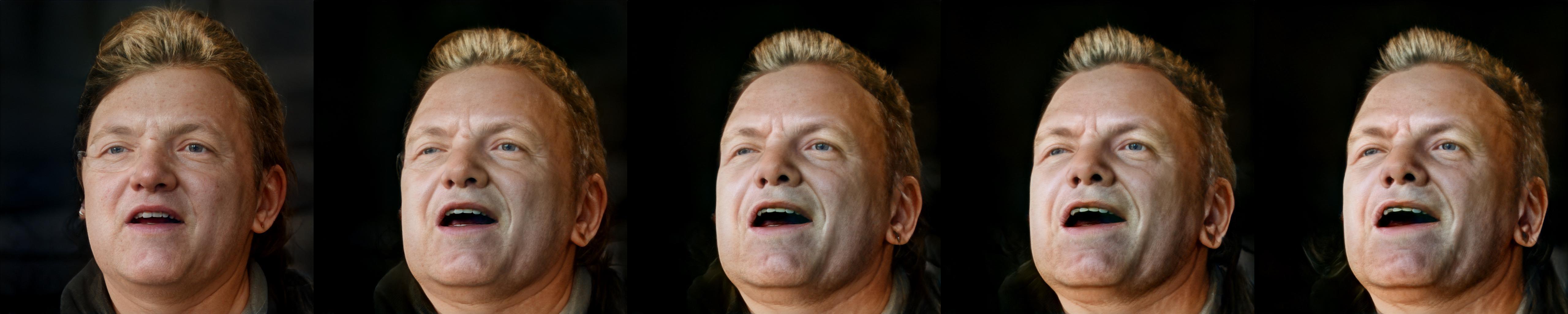} \\

        \includegraphics[width=0.14\textwidth]{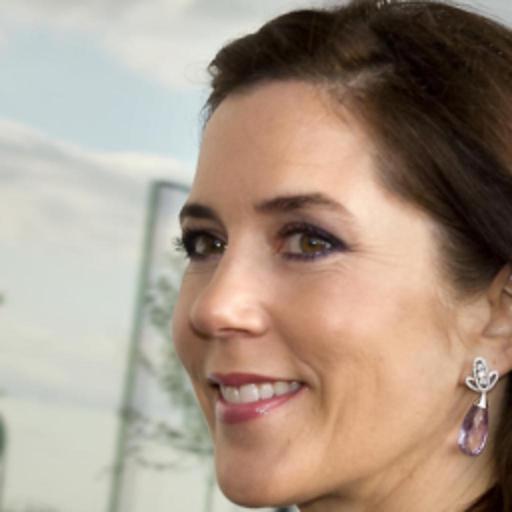} &
        \includegraphics[width=0.7\textwidth]{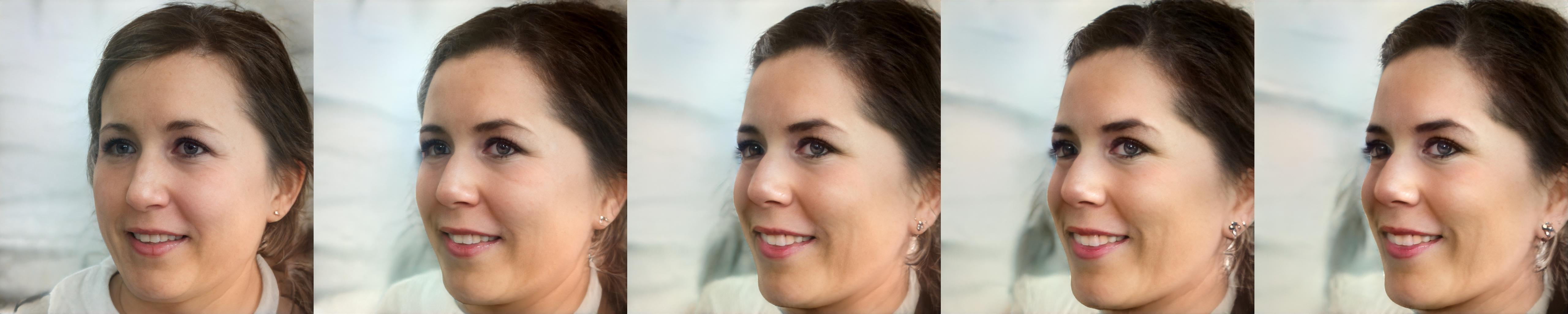} \\
    
        \includegraphics[width=0.14\textwidth]{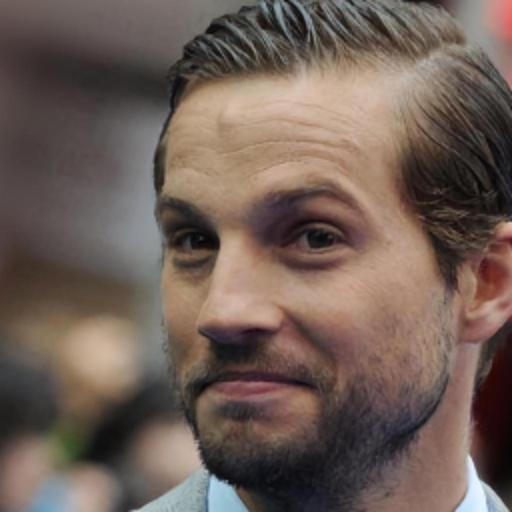} &
        \includegraphics[width=0.7\textwidth]{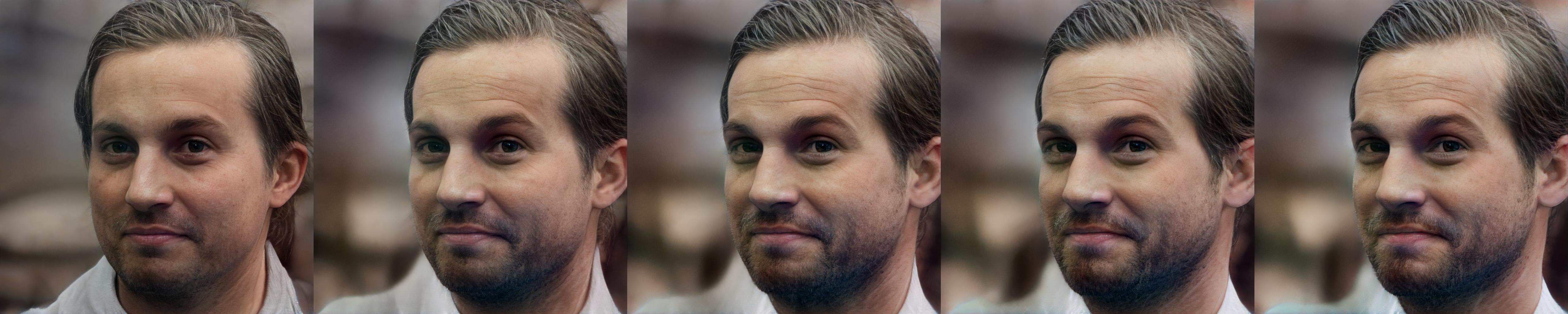} \\
        
        \includegraphics[width=0.14\textwidth]{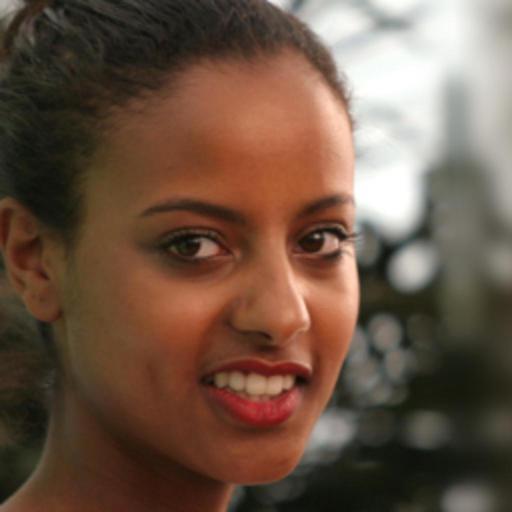} &
        \includegraphics[width=0.7\textwidth]{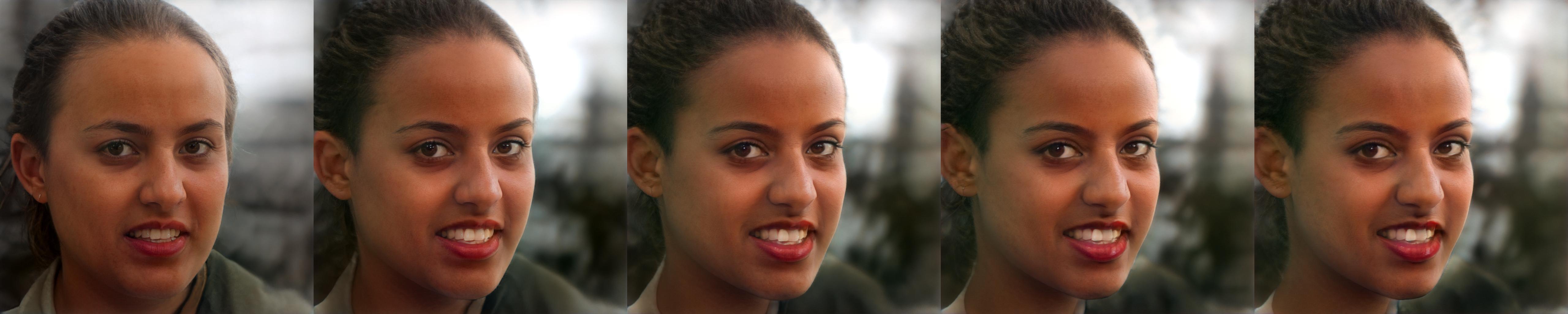} \\
        
        \includegraphics[width=0.14\textwidth]{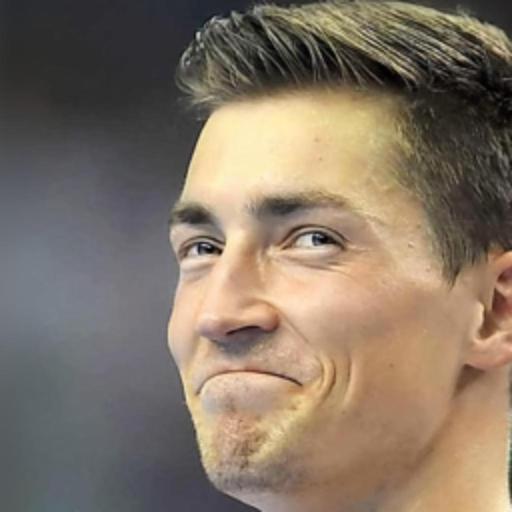} &
        \includegraphics[width=0.7\textwidth]{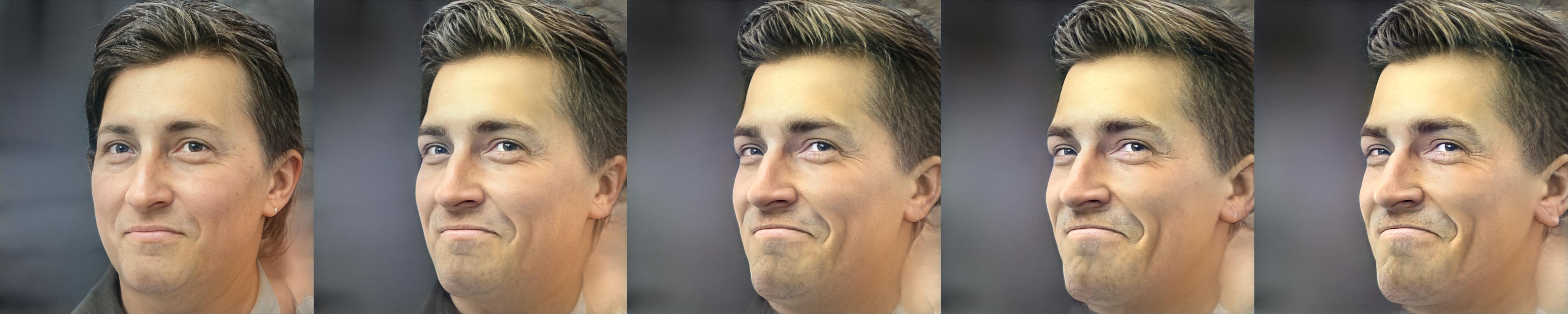} \\

        Input & \qquad Iterative Outputs $\longrightarrow$
    
    \end{tabular}
    }
    \vspace{0.35cm}
    \caption{Iterative outputs generated by ReStyle applied over pSp on the human facial domain.}
    \label{fig:faces_steps_visualization}
\end{figure*}
\begin{figure*}
    \centering
    \setlength{\belowcaptionskip}{-6pt}
    \setlength{\tabcolsep}{1pt}
    {\small
    \begin{tabular}{c l}

        Input & \qquad Iterative Outputs $\longrightarrow$ \\

        \includegraphics[width=0.14\textwidth]{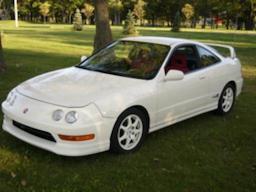} &
        \includegraphics[width=0.7\textwidth]{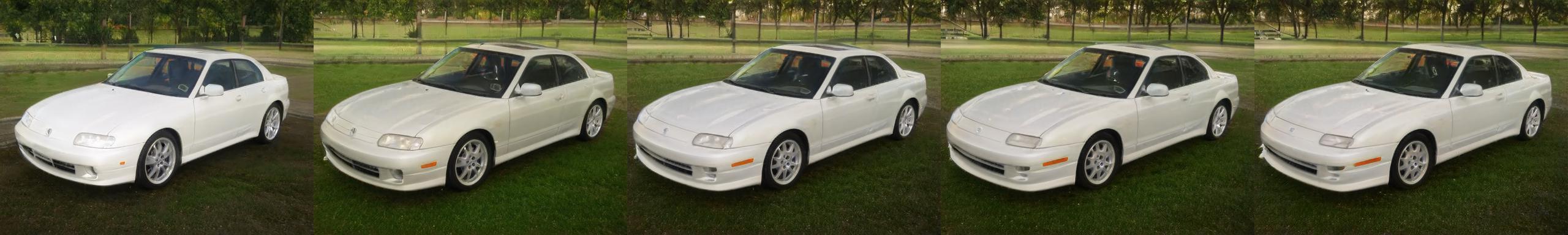} \\

        \includegraphics[width=0.14\textwidth]{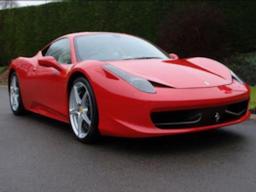} &
        \includegraphics[width=0.7\textwidth]{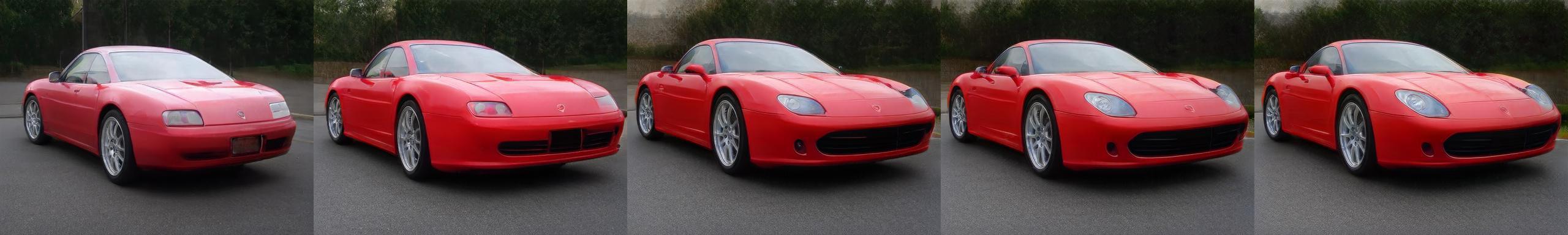} \\
        
        \includegraphics[width=0.14\textwidth]{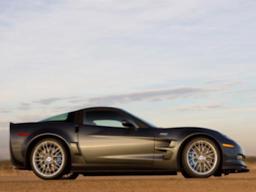} &
        \includegraphics[width=0.7\textwidth]{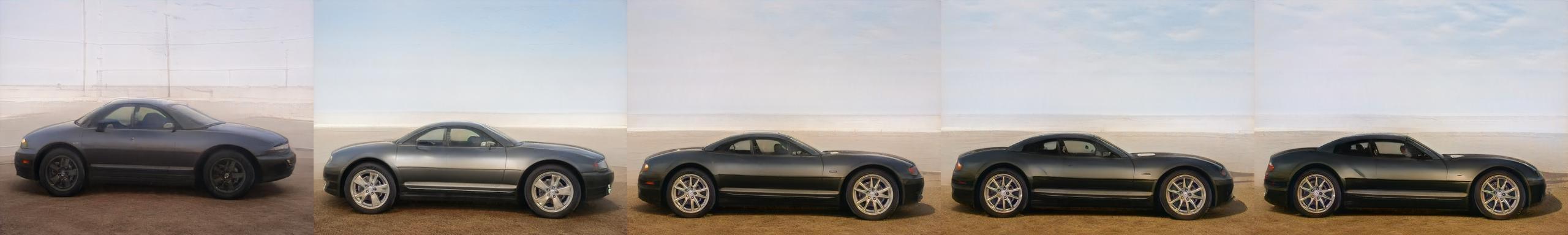} \\

        \includegraphics[width=0.14\textwidth]{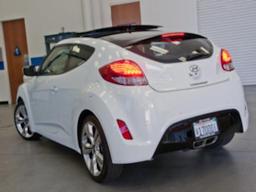} &
        \includegraphics[width=0.7\textwidth]{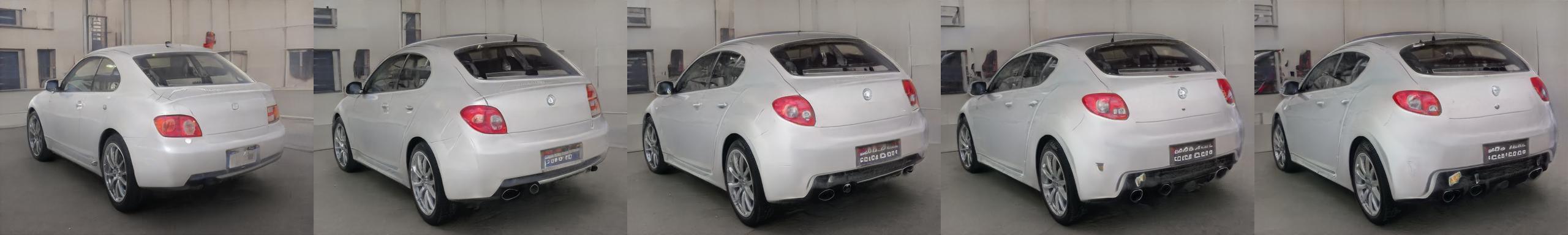} \\

        \includegraphics[width=0.14\textwidth]{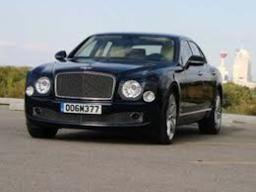} &
        \includegraphics[width=0.7\textwidth]{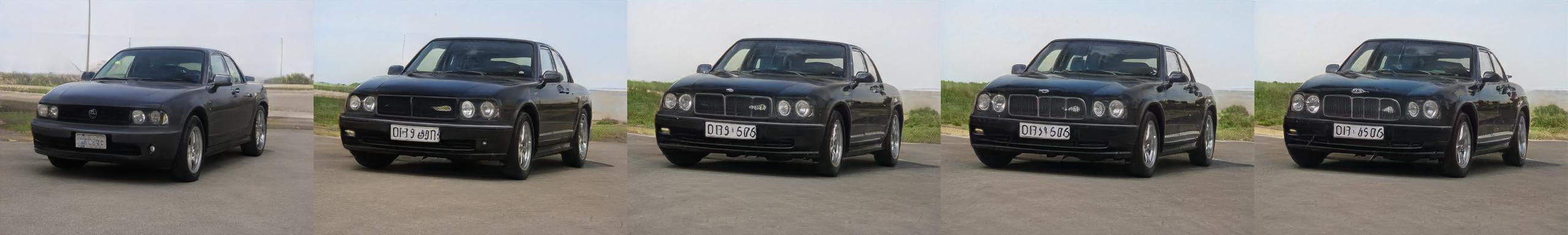} \\

        \includegraphics[width=0.14\textwidth]{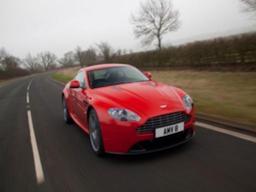} &
        \includegraphics[width=0.7\textwidth]{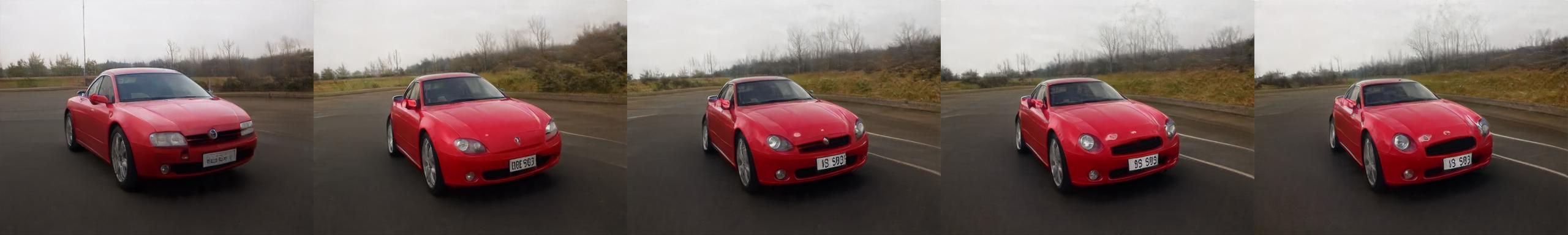} \\ 

        \includegraphics[width=0.14\textwidth]{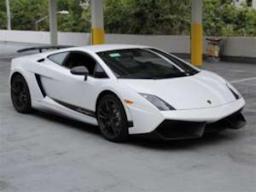} &
        \includegraphics[width=0.7\textwidth]{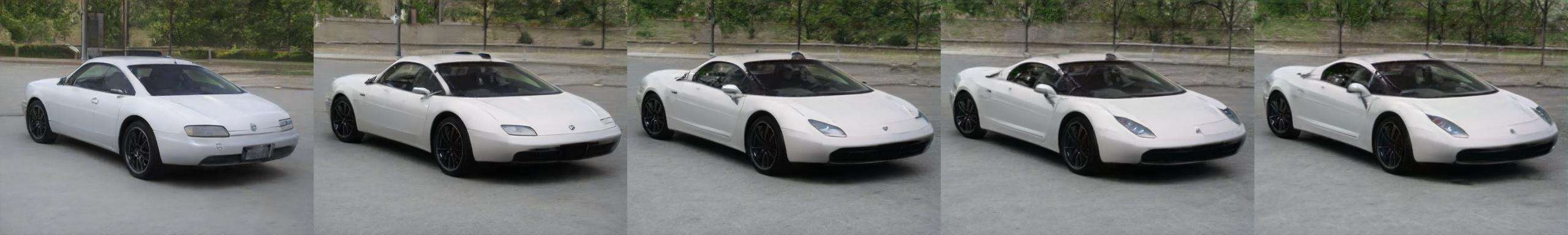} \\
\end{tabular}
    }
    \vspace{0.35cm}
    \caption{Iterative outputs generated by ReStyle applied over e4e on the cars domain.}
    \label{fig:cars_steps_visualization}
\end{figure*}
\begin{figure*}
    \centering
    \setlength{\belowcaptionskip}{-6pt}
    \setlength{\tabcolsep}{1pt}
    {\small
    \begin{tabular}{c l}

        Input & \qquad Iterative Outputs $\longrightarrow$ \\

        \includegraphics[width=0.14\textwidth]{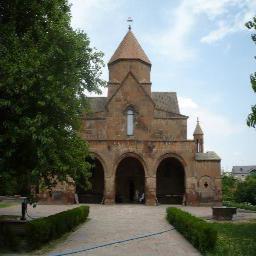} &
        \includegraphics[width=0.7\textwidth]{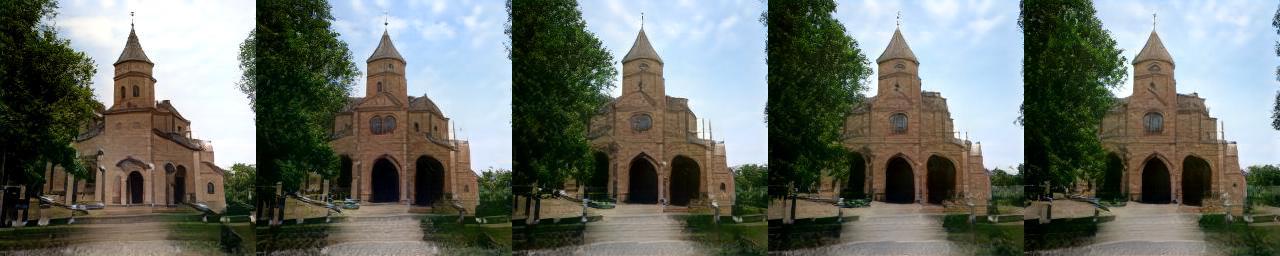} \\

        \includegraphics[width=0.14\textwidth]{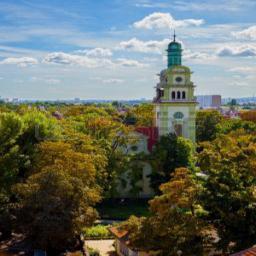} &
        \includegraphics[width=0.7\textwidth]{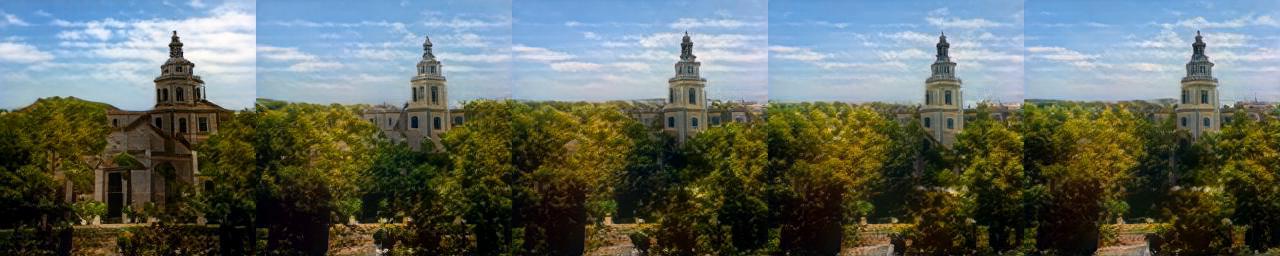} \\
        
        \includegraphics[width=0.14\textwidth]{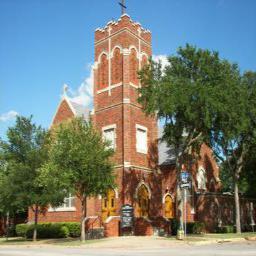} &
        \includegraphics[width=0.7\textwidth]{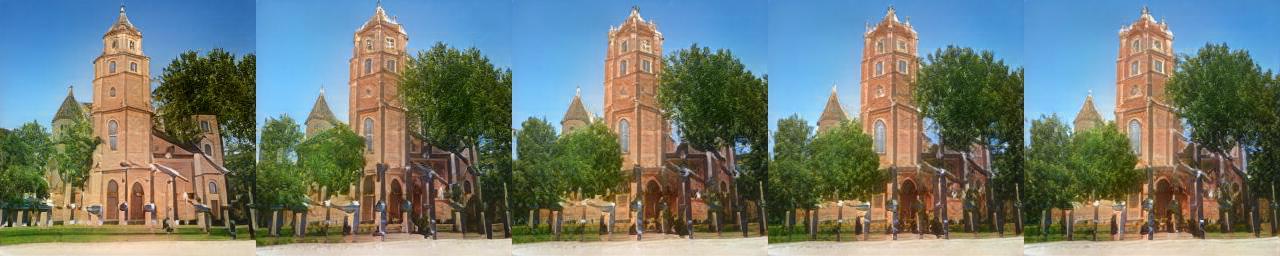} \\

        \includegraphics[width=0.14\textwidth]{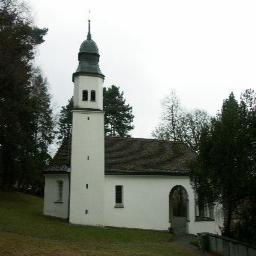} &
        \includegraphics[width=0.7\textwidth]{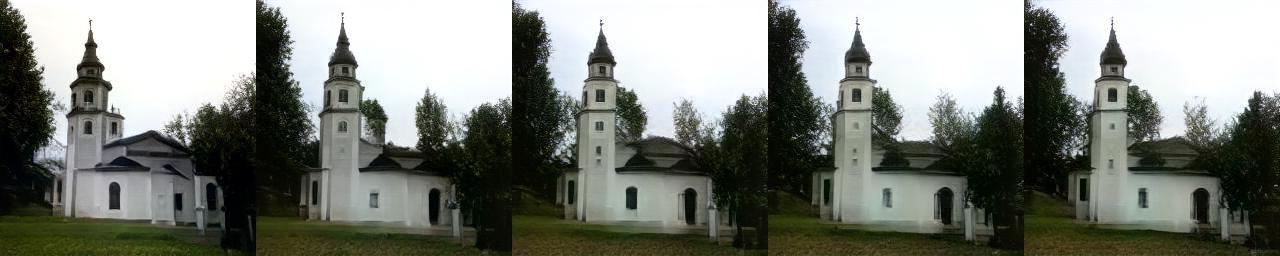} \\

        \includegraphics[width=0.14\textwidth]{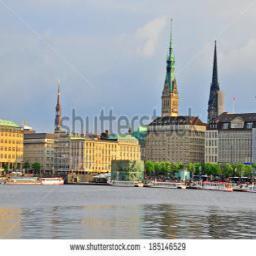} &
        \includegraphics[width=0.7\textwidth]{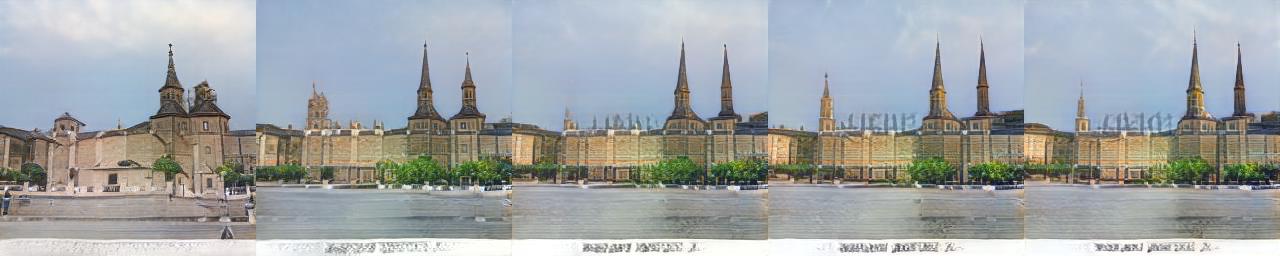} \\

        \includegraphics[width=0.14\textwidth]{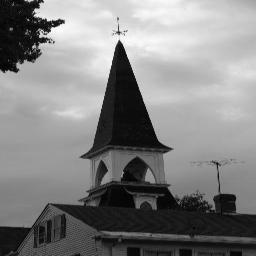} &
        \includegraphics[width=0.7\textwidth]{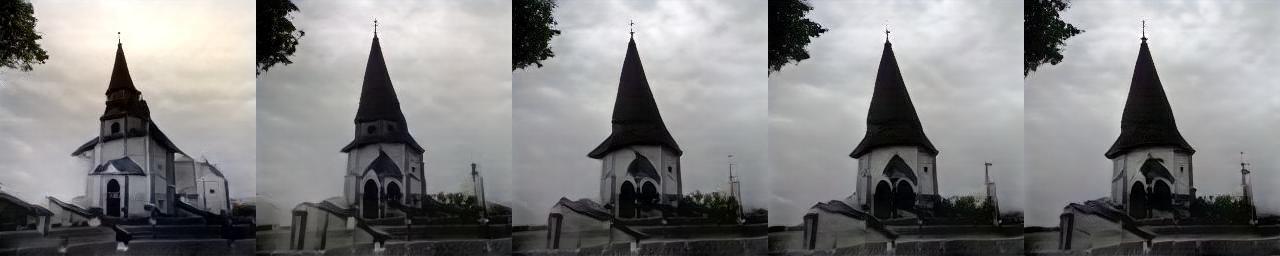} \\
\end{tabular}
    }
    \vspace{0.35cm}
    \caption{Iterative outputs generated by ReStyle applied over pSp on the churches domain.}
    \label{fig:churches_steps_visualization}
\end{figure*}
\begin{figure*}
    \centering
    \setlength{\belowcaptionskip}{-6pt}
    \setlength{\tabcolsep}{1pt}
    {\small
    \begin{tabular}{c l}

        Input & \qquad Iterative Outputs $\longrightarrow$ \\

        \includegraphics[width=0.14\textwidth]{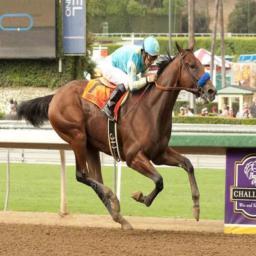} &
        \includegraphics[width=0.7\textwidth]{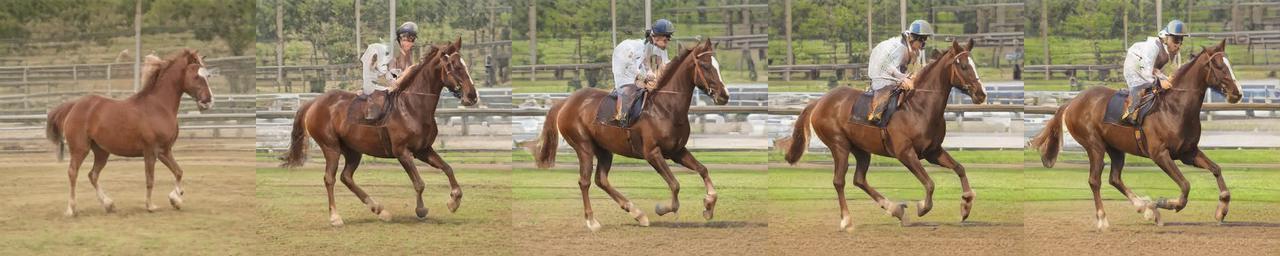} \\

        \includegraphics[width=0.14\textwidth]{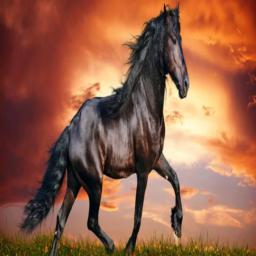} &
        \includegraphics[width=0.7\textwidth]{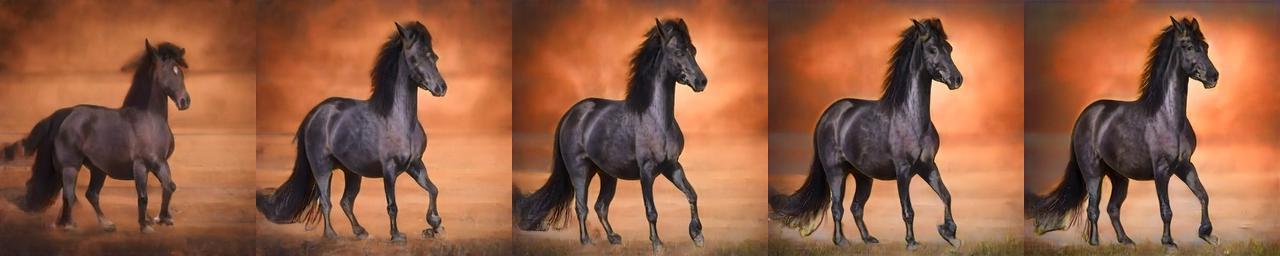} \\
        
        \includegraphics[width=0.14\textwidth]{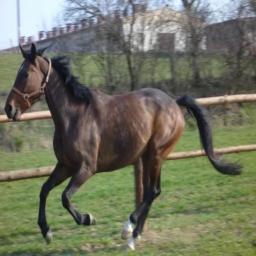} &
        \includegraphics[width=0.7\textwidth]{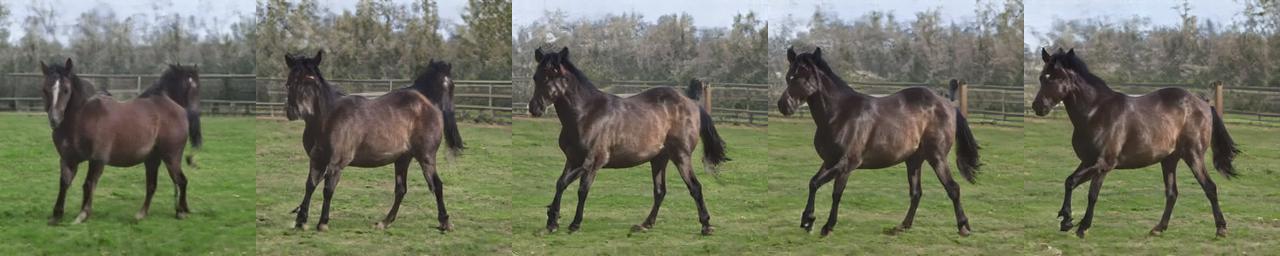} \\

        \includegraphics[width=0.14\textwidth]{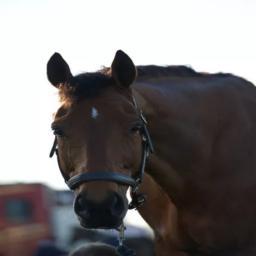} &
        \includegraphics[width=0.7\textwidth]{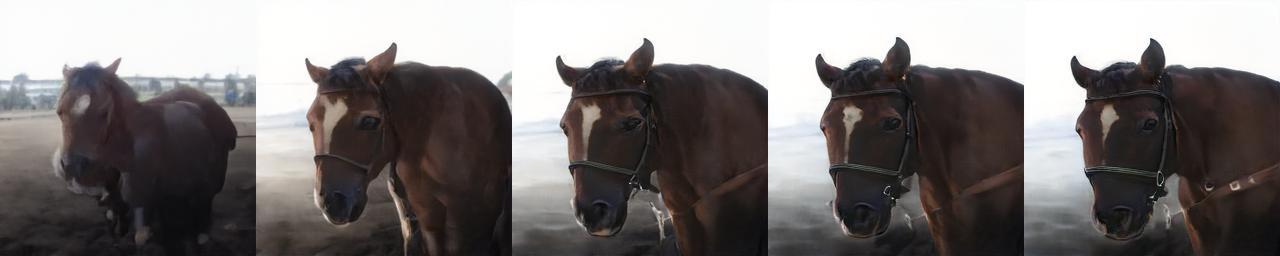} \\

        \includegraphics[width=0.14\textwidth]{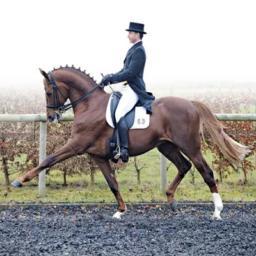} &
        \includegraphics[width=0.7\textwidth]{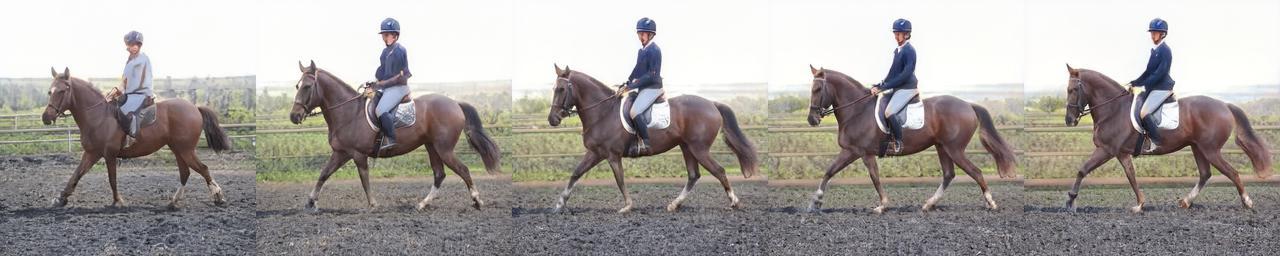} \\

        \includegraphics[width=0.14\textwidth]{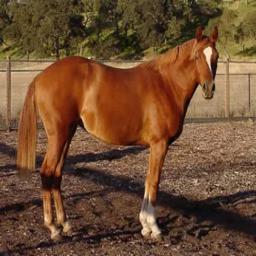} &
        \includegraphics[width=0.7\textwidth]{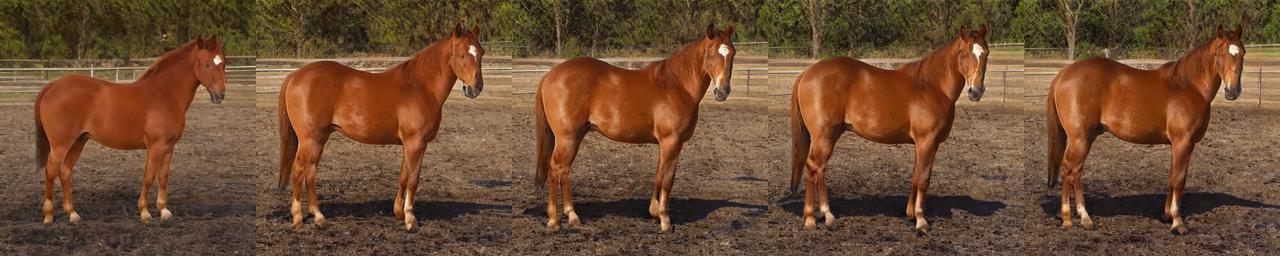} \\ 

        \includegraphics[width=0.14\textwidth]{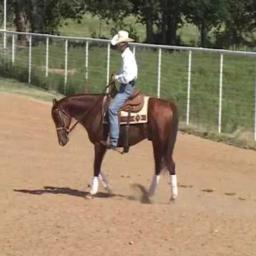} &
        \includegraphics[width=0.7\textwidth]{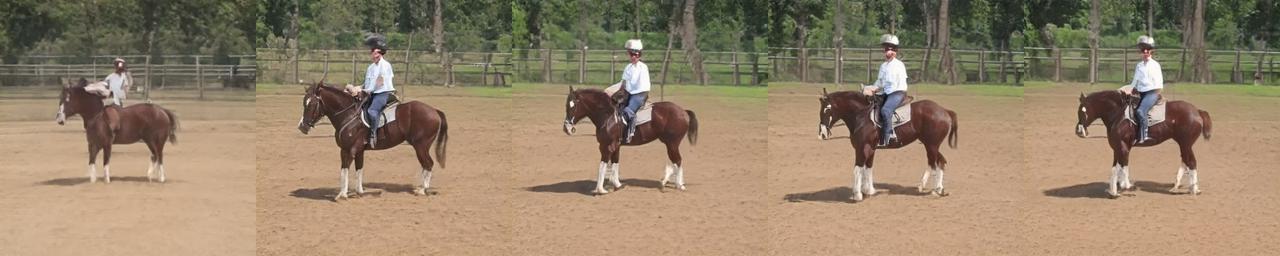} \\
    
        \includegraphics[width=0.14\textwidth]{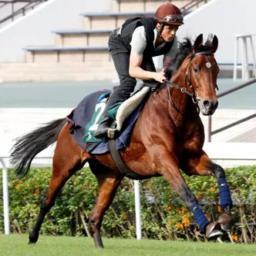} &
        \includegraphics[width=0.7\textwidth]{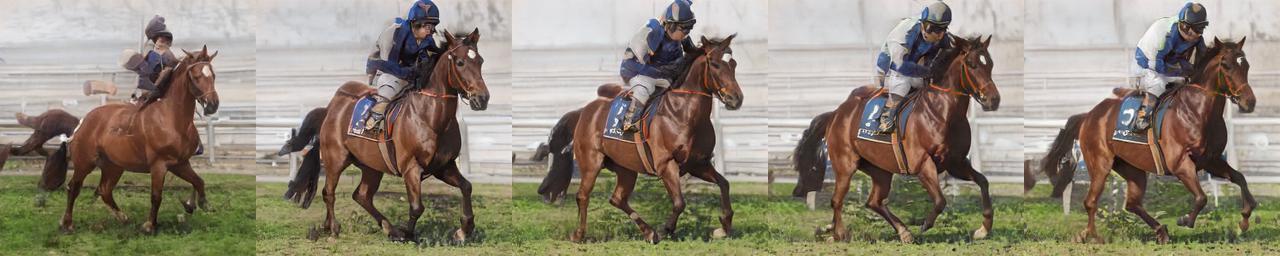} \\

        Input & \qquad Iterative Outputs $\longrightarrow$
    
    \end{tabular}
    }
    \vspace{0.35cm}
    \caption{Iterative outputs generated by ReStyle applied over e4e on the horses domain.}
    \label{fig:horses_steps_visualization}
\end{figure*}

\begin{figure*}
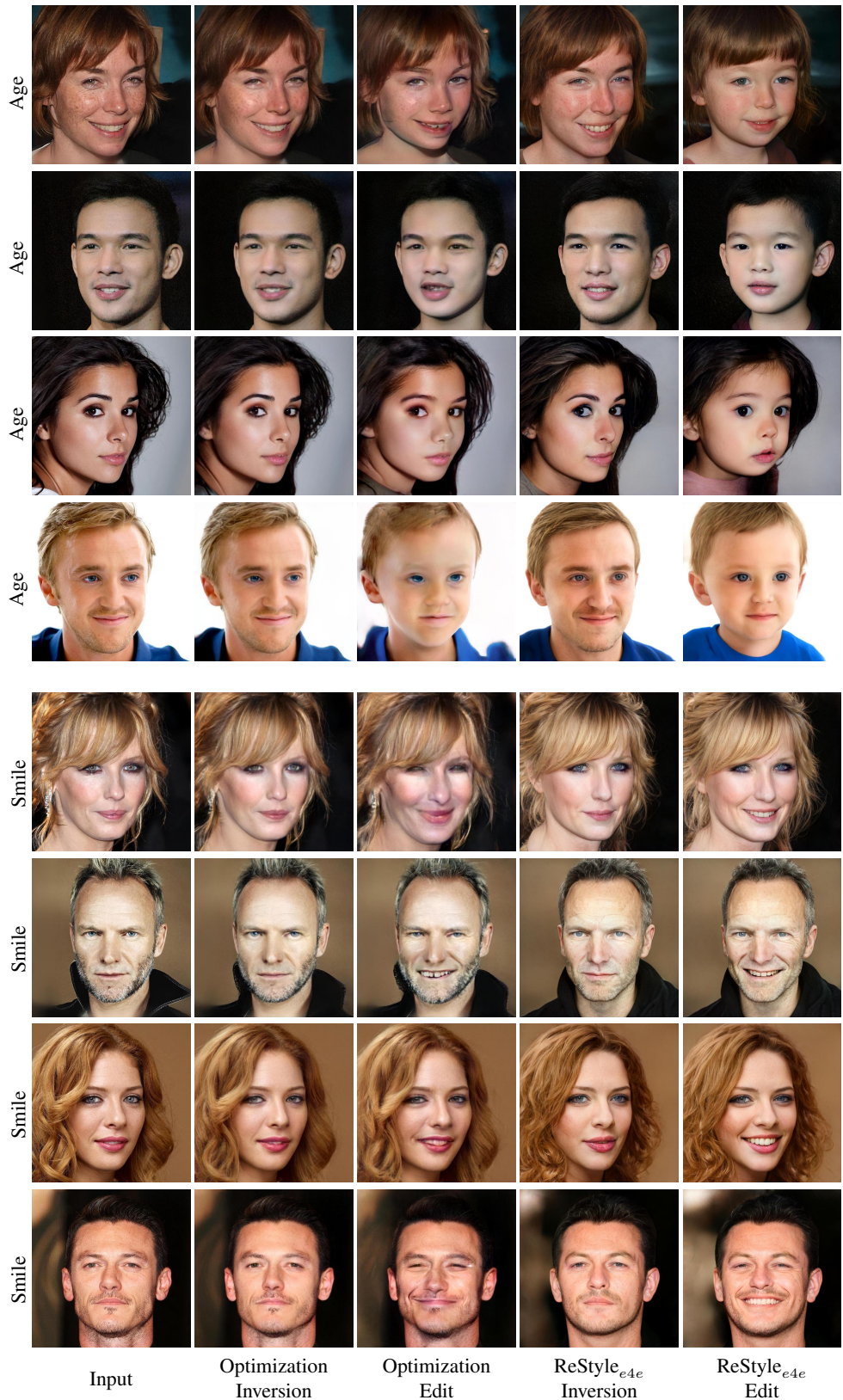

    \centering
    \setlength{\belowcaptionskip}{-6pt}
    \setlength{\tabcolsep}{1pt}
    {\small
    %
    }
    \vspace{0.35cm}
    \caption{Editing comparisons with the optimization technique from Karras \etal~\cite{karras2020analyzing} on the human facial domain obtained using InterFaceGAN~\cite{shen2020interpreting}. Here ReStyle is applied over the e4e encoder. We perform two edits: an age edit and smile edit.}
    \label{fig:faces_editing}
\end{figure*}
\begin{figure*}
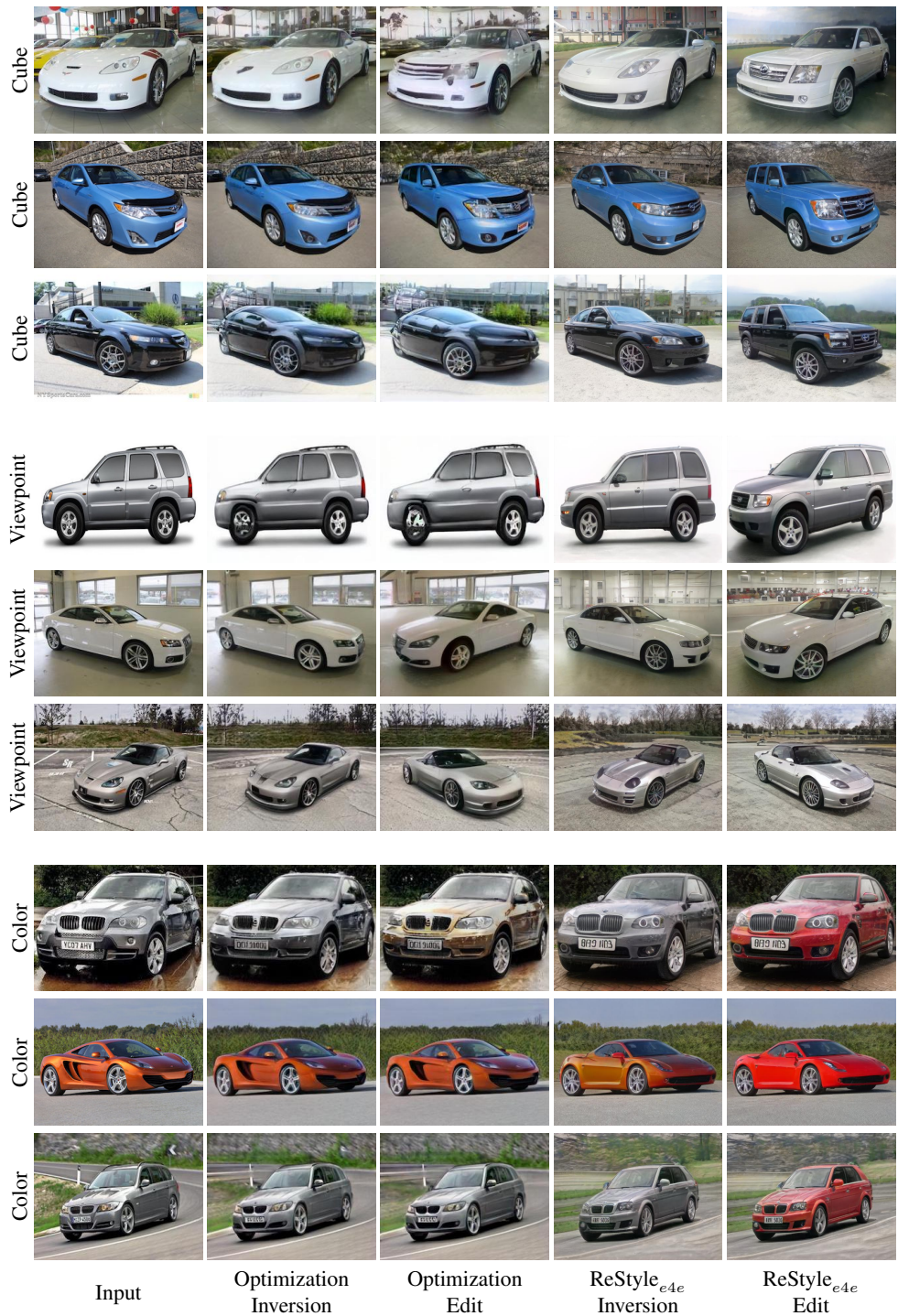

    \centering
    \setlength{\belowcaptionskip}{-6pt}
    \setlength{\tabcolsep}{1pt}
    {\small
    %
    }
    \vspace{0.35cm}
    \caption{Editing comparisons with the optimization technique from Karras \etal~\cite{karras2020analyzing} on the cars domain obtained using GANSpace~\cite{harkonen2020ganspace}. Here ReStyle is applied over the e4e encoder. We perform three edits: a cube-shape edit, viewpoint edit, and color edit.}
    \label{fig:cars_editing}
\end{figure*}
\begin{figure*}
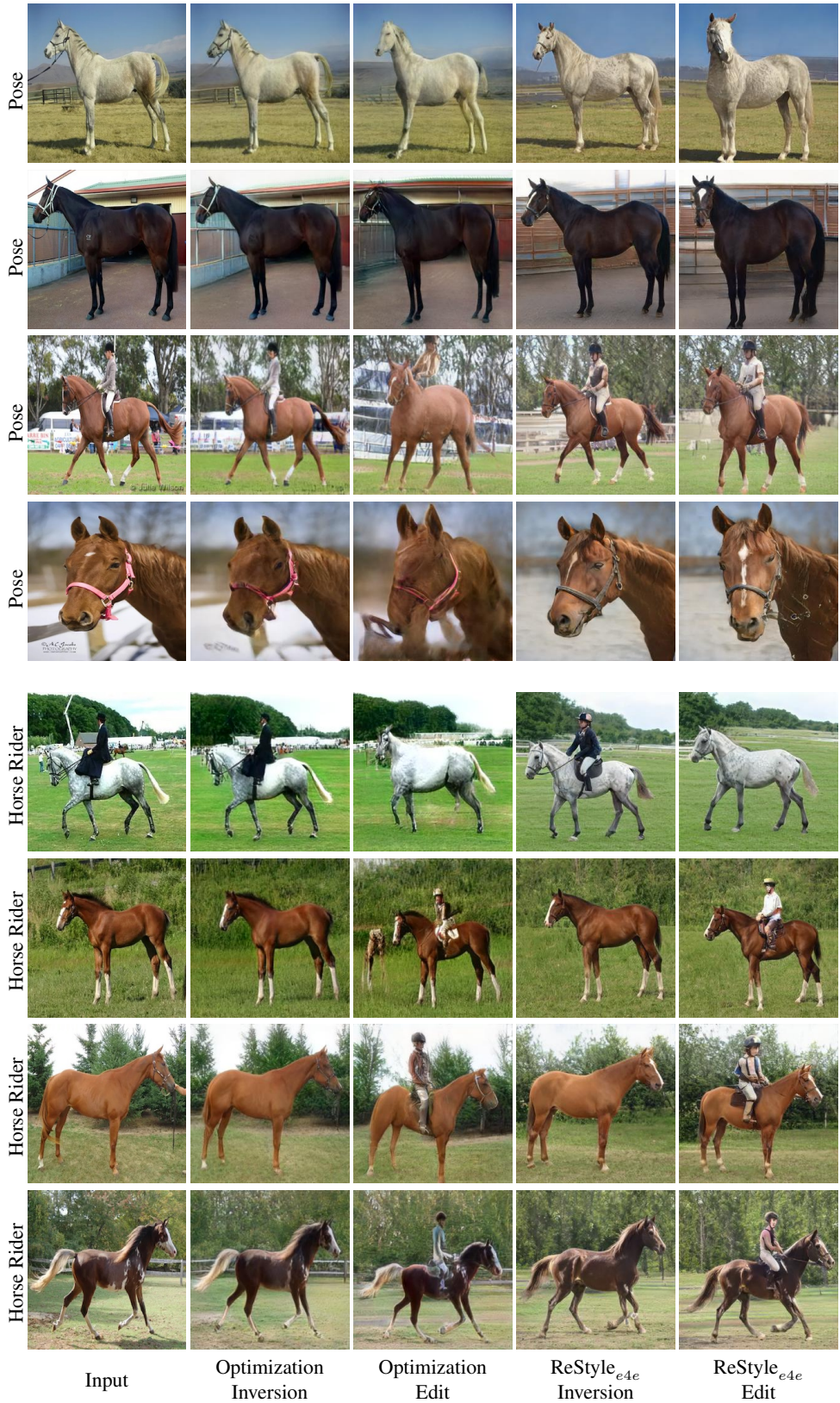

    \centering
    \setlength{\belowcaptionskip}{-6pt}
    \setlength{\tabcolsep}{1pt}
    {\small
    %
    }
    \vspace{0.35cm}
    \caption{Editing comparisons with the optimization technique from Karras \etal~\cite{karras2020analyzing} on the horses domain obtained using SeFa~\cite{shen2020interpreting}. Here ReStyle is applied over the e4e encoder. We perform two edits: a pose edit and an edit to add/remove a horse rider.}
    \label{fig:horses_editing}
\end{figure*}

\begin{figure*}
    \centering
    \setlength{\belowcaptionskip}{-2.5pt}
    \setlength{\tabcolsep}{1pt}
    \begin{tabular}{c c c}

        \includegraphics[width=0.16\textwidth]{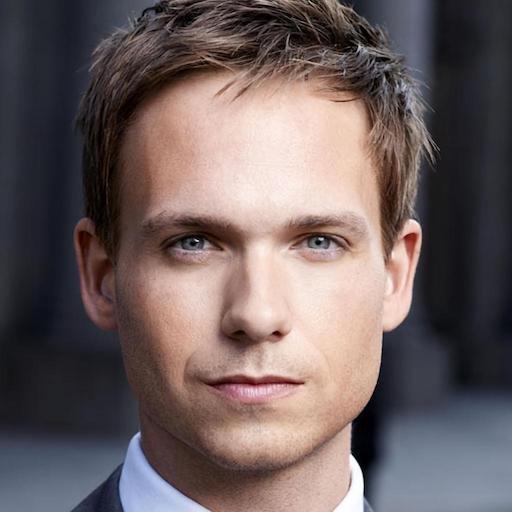} &
        \includegraphics[width=0.80\textwidth]{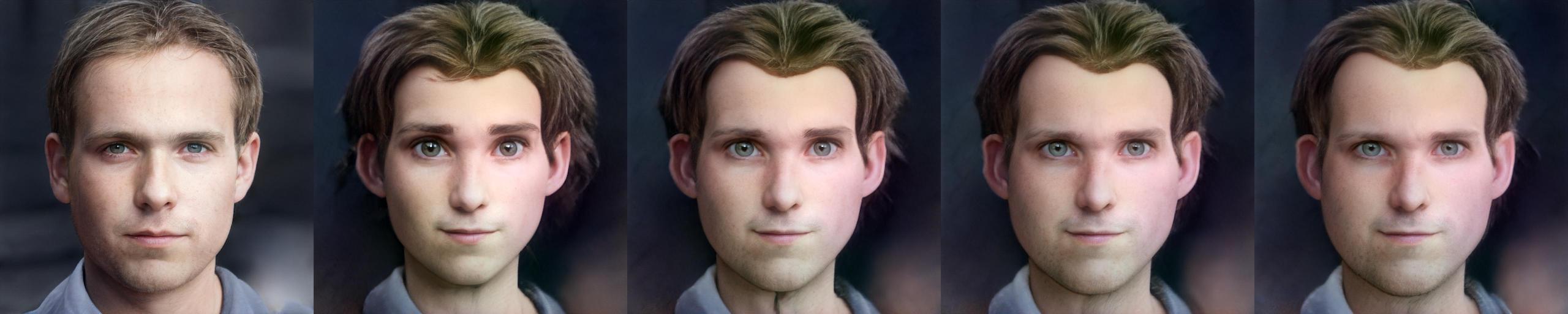}
        \tabularnewline

        \includegraphics[width=0.16\textwidth]{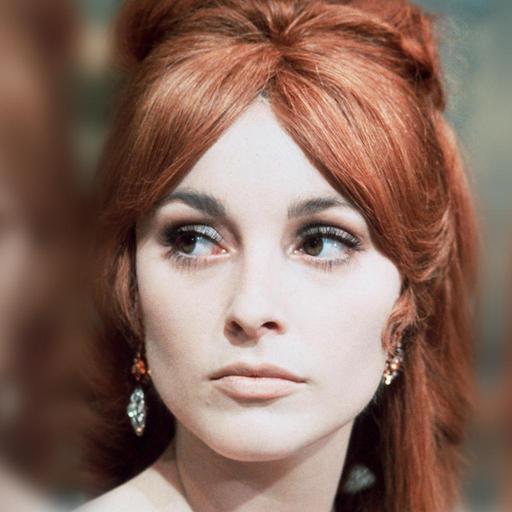} &
        \includegraphics[width=0.80\textwidth]{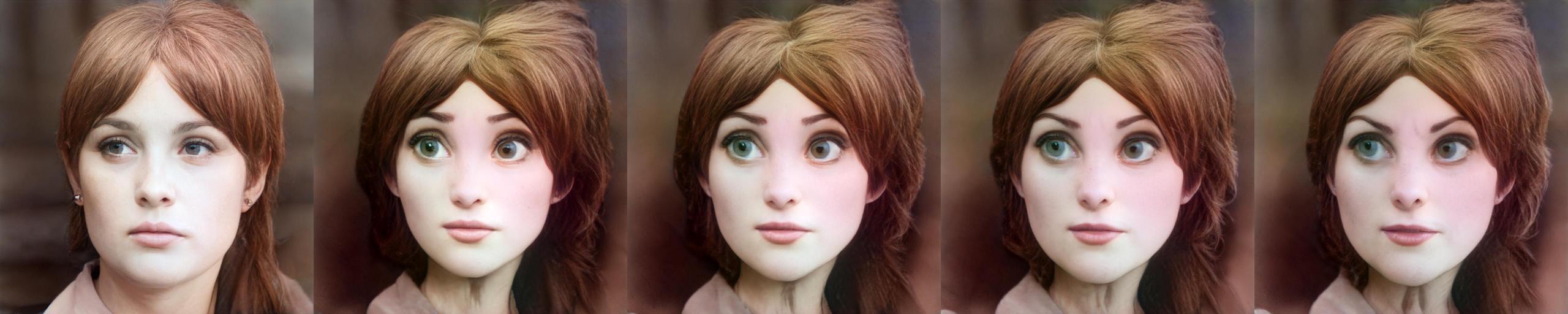}
        \tabularnewline
        
        \includegraphics[width=0.16\textwidth]{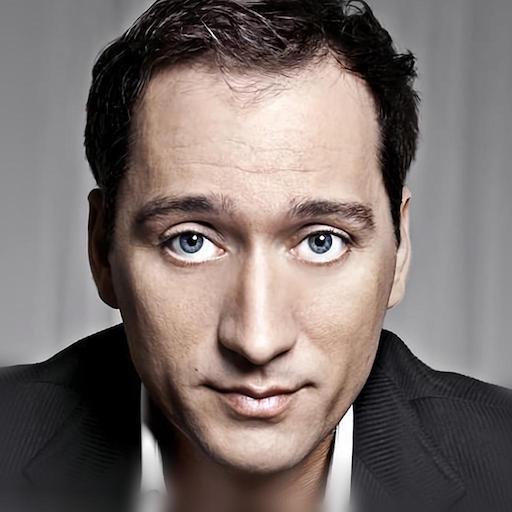} &
        \includegraphics[width=0.80\textwidth]{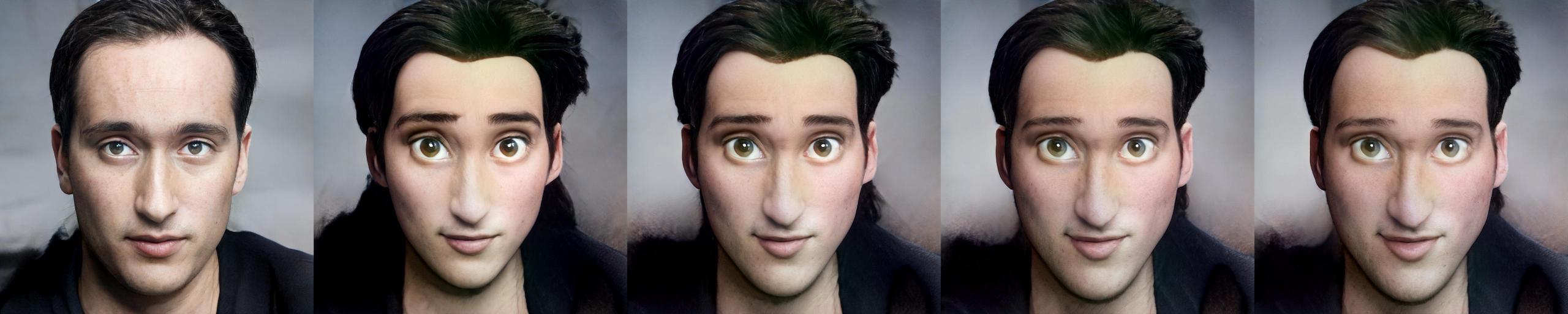}
        \tabularnewline

        \includegraphics[width=0.16\textwidth]{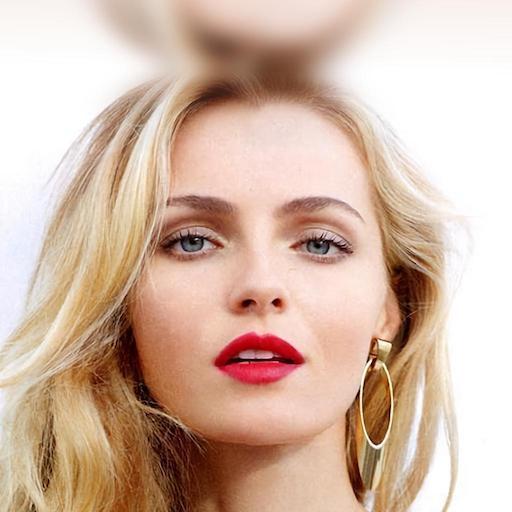} &
        \includegraphics[width=0.80\textwidth]{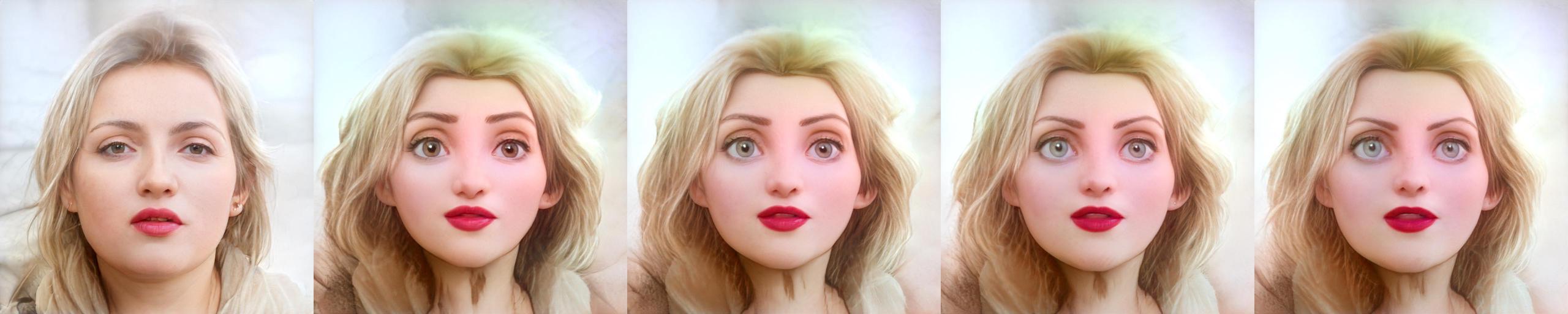}
        \tabularnewline
        
        \includegraphics[width=0.16\textwidth]{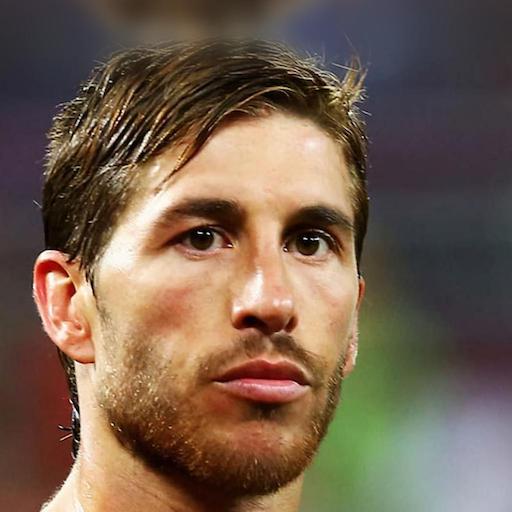} &
        \includegraphics[width=0.80\textwidth]{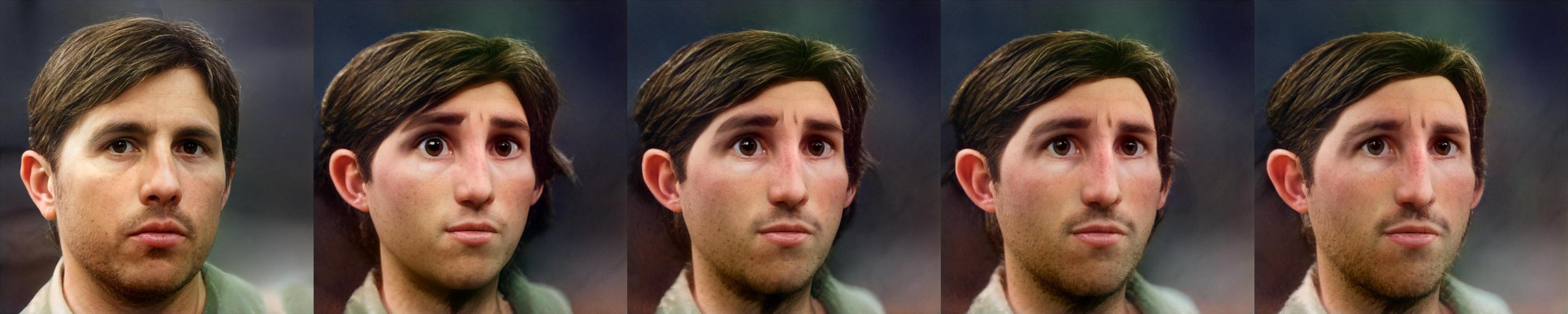}
        \tabularnewline

        \includegraphics[width=0.16\textwidth]{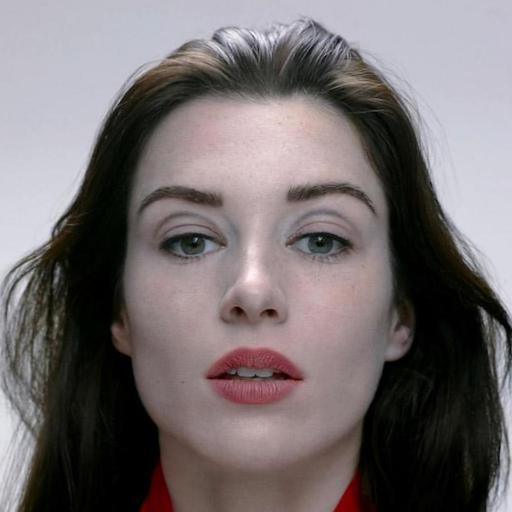} &
        \includegraphics[width=0.80\textwidth]{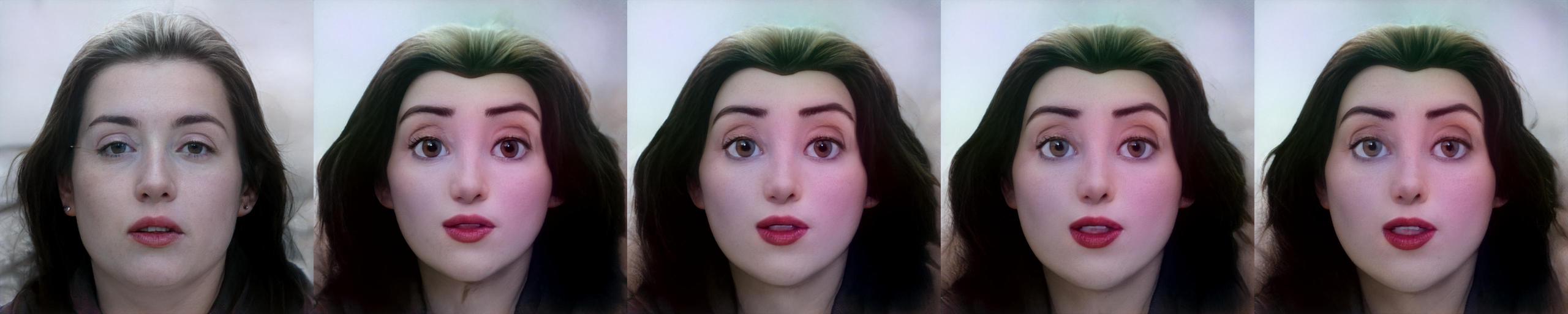}
        \tabularnewline
        
        \includegraphics[width=0.16\textwidth]{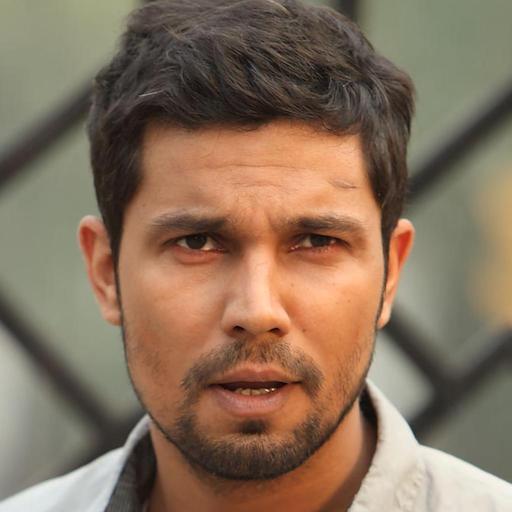} &
        \includegraphics[width=0.80\textwidth]{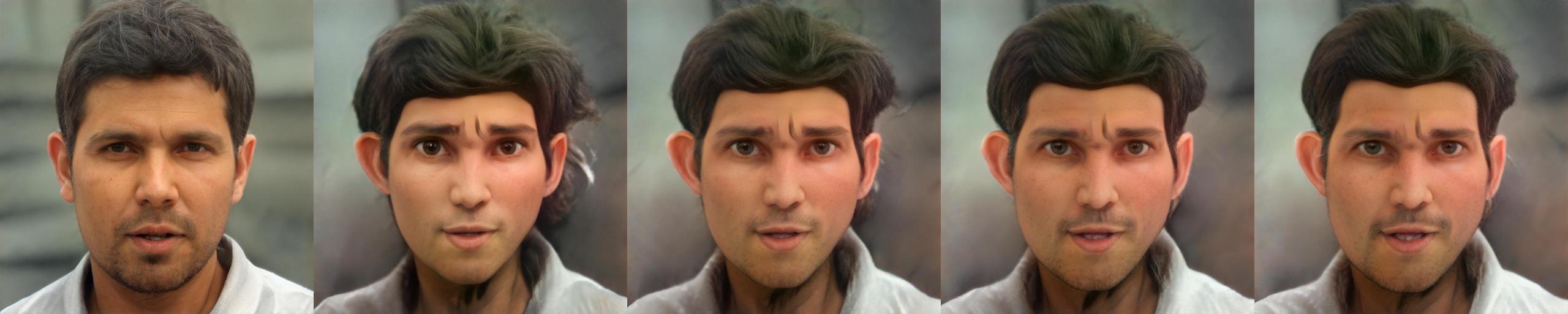}
        \tabularnewline

        Input & Inverted \qquad \qquad Iterative Outputs $\longrightarrow$ \qquad \qquad \qquad \qquad \qquad \qquad \qquad \qquad \qquad \qquad

    \end{tabular}
    \vspace{0.45cm}
    \caption{Additional results on the image toonification task using ReStyle with our encoder bootstrapping technique. For each input, we show the inverted image obtained after a single step of our $\text{ReStyle}_{pSp}$ FFHQ encoder followed by the iterative outputs of our $\text{ReStyle}_{pSp}$ toonify encoder.}
    \label{fig:encoder_mixing}
\end{figure*}
\begin{figure*}
    \centering
    \setlength{\belowcaptionskip}{-2.5pt}
    \setlength{\tabcolsep}{1pt}
    \begin{tabular}{c c c}

        \includegraphics[width=0.16\textwidth]{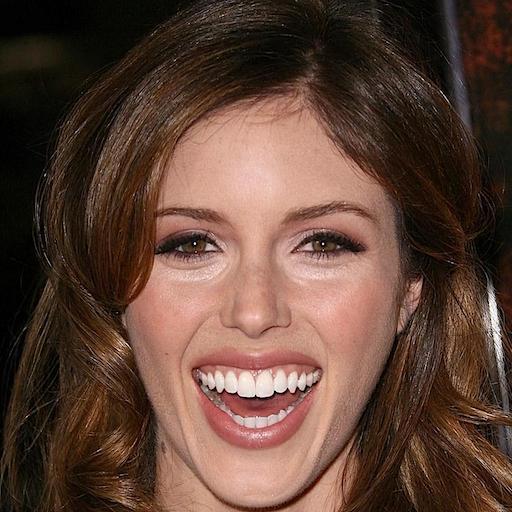} &
        \includegraphics[width=0.80\textwidth]{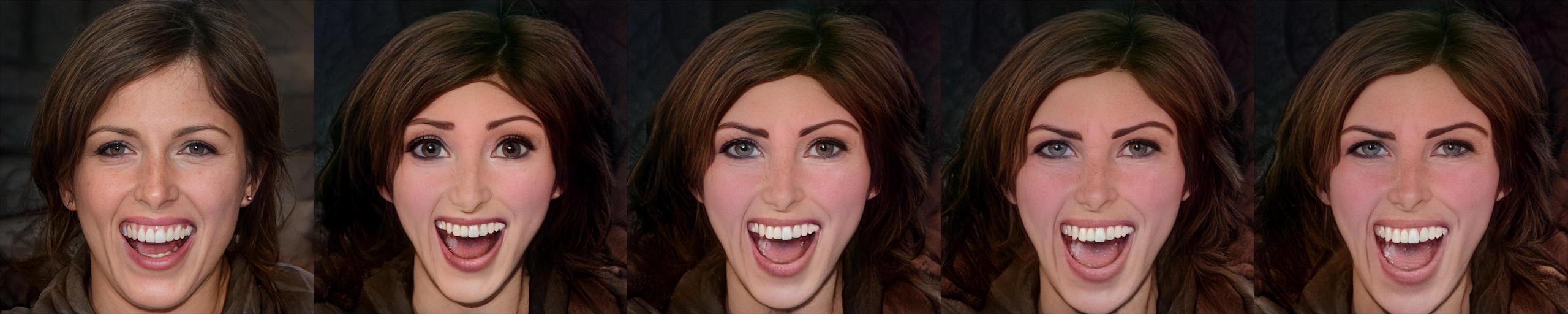}
        \tabularnewline

        \includegraphics[width=0.16\textwidth]{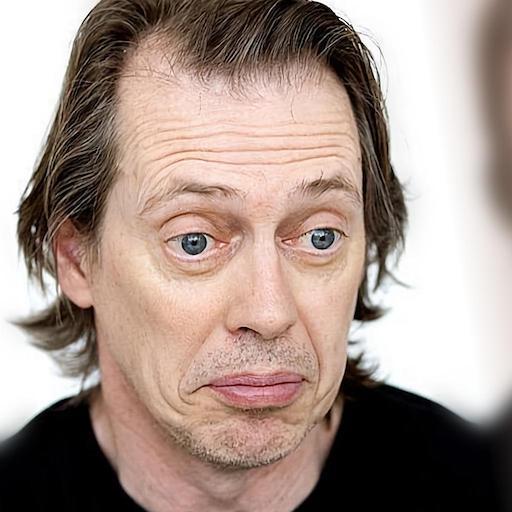} &
        \includegraphics[width=0.80\textwidth]{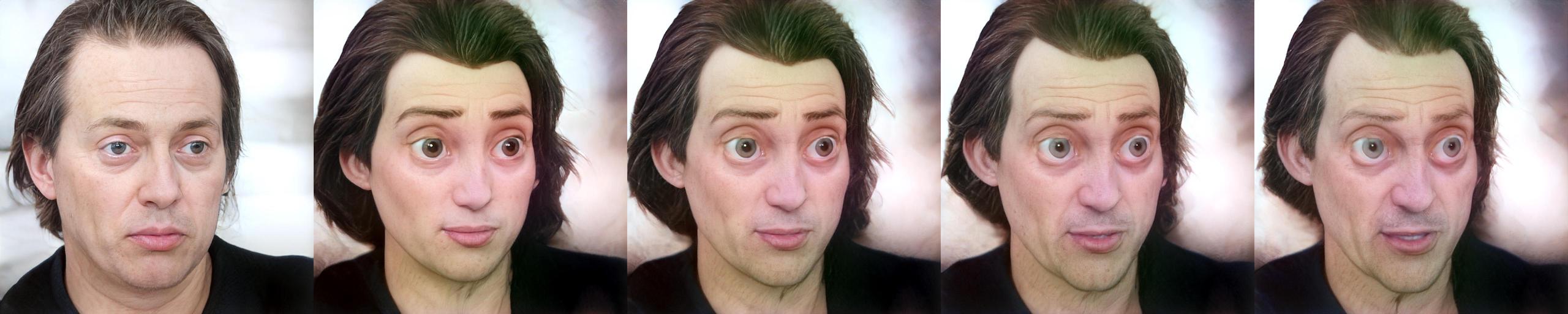}
        \tabularnewline
        
        \includegraphics[width=0.16\textwidth]{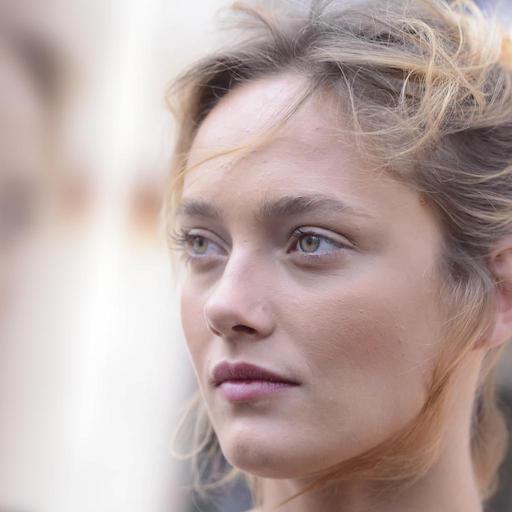} &
        \includegraphics[width=0.80\textwidth]{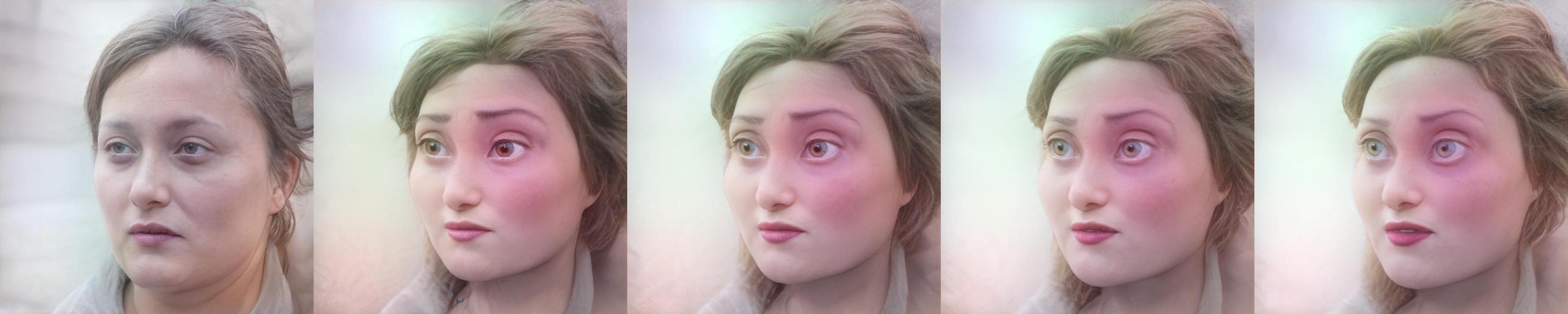}
        \tabularnewline

        \includegraphics[width=0.16\textwidth]{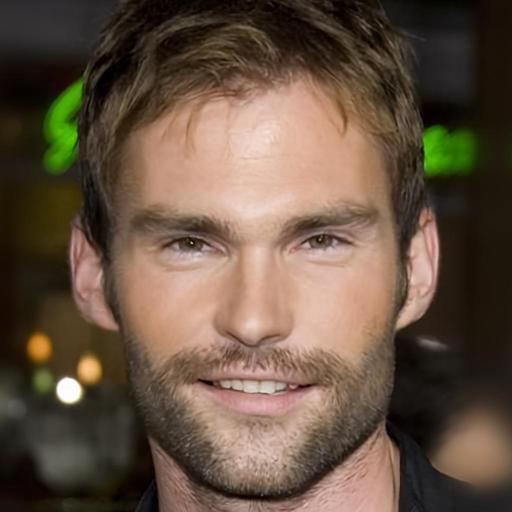} &
        \includegraphics[width=0.80\textwidth]{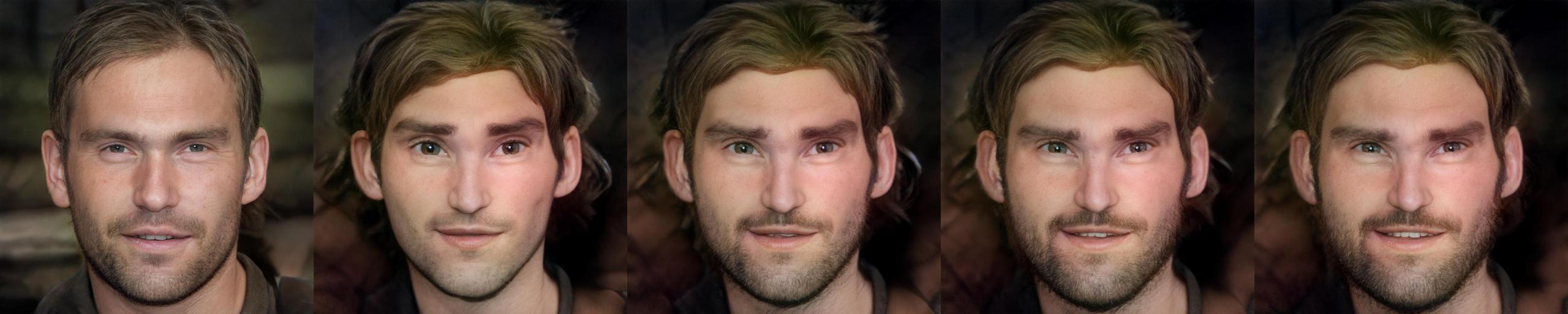}
        \tabularnewline
        
        \includegraphics[width=0.16\textwidth]{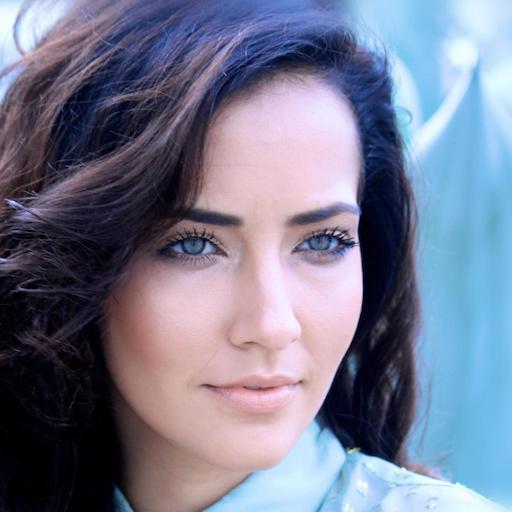} &
        \includegraphics[width=0.80\textwidth]{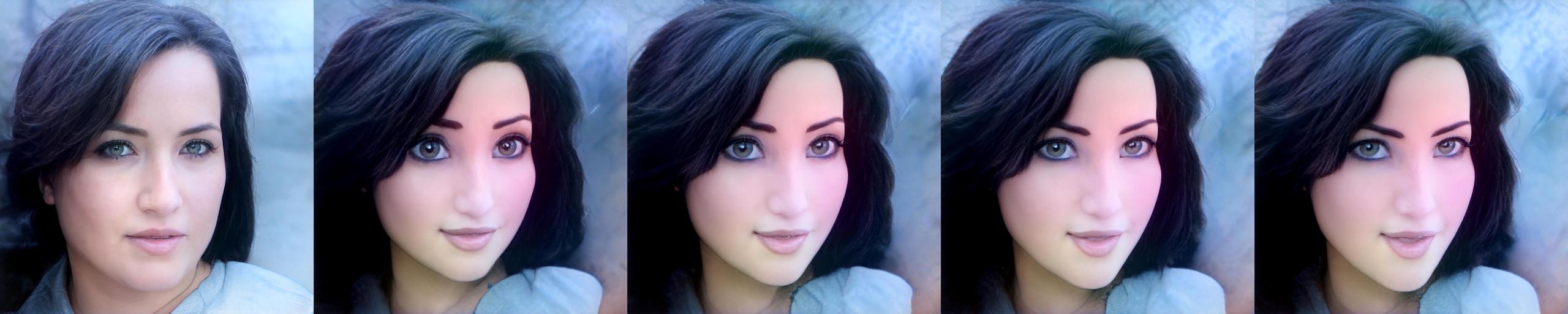}
        \tabularnewline

        \includegraphics[width=0.16\textwidth]{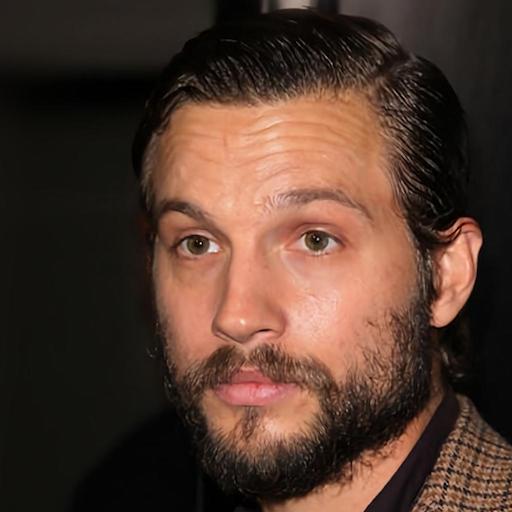} &
        \includegraphics[width=0.80\textwidth]{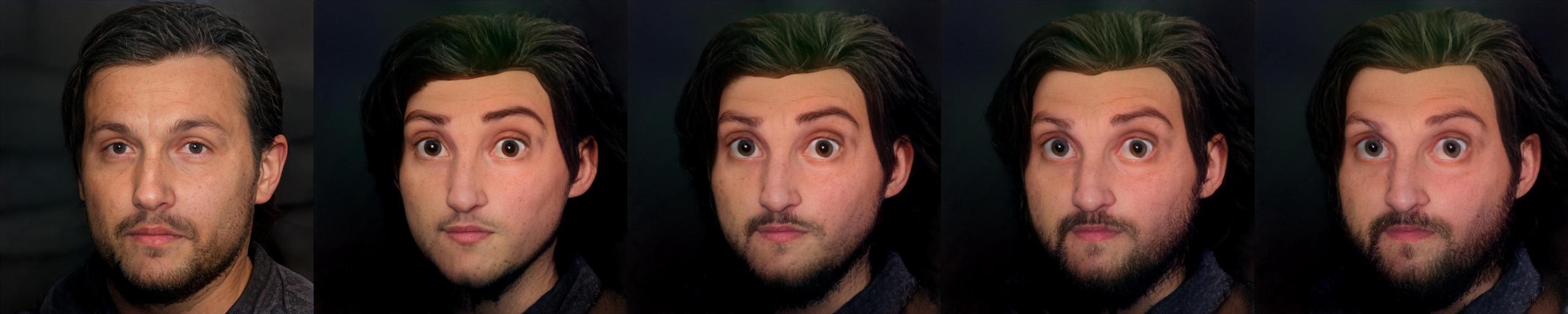}
        \tabularnewline
        
        \includegraphics[width=0.16\textwidth]{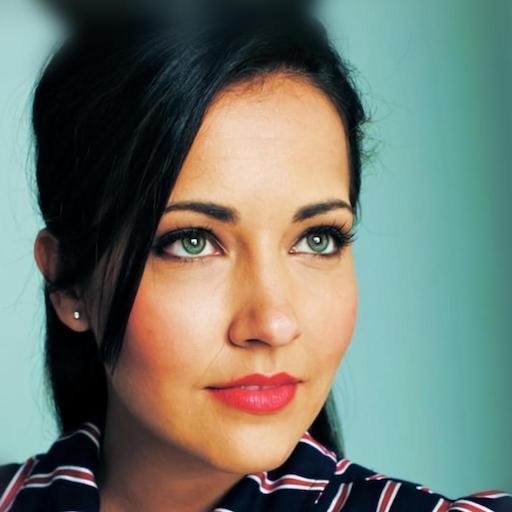} &
        \includegraphics[width=0.80\textwidth]{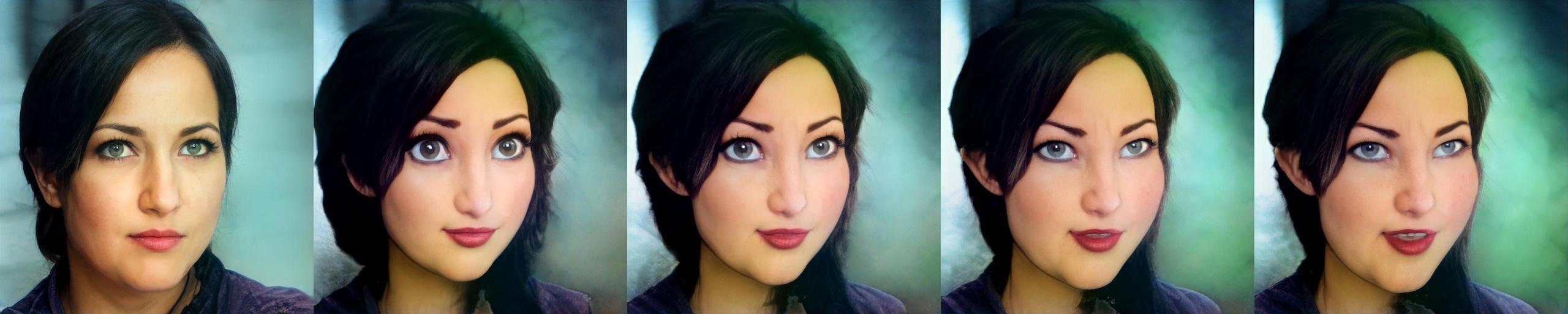}
        \tabularnewline

        Input & Inverted \qquad \qquad Iterative Outputs $\longrightarrow$ \qquad \qquad \qquad \qquad \qquad \qquad \qquad \qquad \qquad \qquad

    \end{tabular}
    \vspace{0.45cm}
    \caption{Additional image toonification results obtained using the same setting as Figure~\ref{fig:encoder_mixing}.}
    \label{fig:encoder_mixing_2}
\end{figure*}

\end{document}